\documentclass{article}

\usepackage[accepted]{icml2024}

\usepackage[utf8]{inputenc} % allow utf-8 input
\usepackage[T1]{fontenc}    % use 8-bit T1 fonts
\usepackage{hyperref}       % hyperlinks
\usepackage{url}            % simple URL typesetting

\usepackage{booktabs}       % professional-quality tables
\usepackage{amsfonts}       % blackboard math symbols
\usepackage{nicefrac}       % compact symbols for 1/2, etc.
\usepackage{microtype}      % microtypography
\usepackage{xcolor}         % colors

\usepackage{rmclarke}
\usepackage{algorithm}
\usepackage{algorithmic}
\usepackage{multirow}
\usepackage{graphicx}
\usepackage{makecell}
\usepackage{siunitx}
\usepackage{tabularx}
\usepackage{graphbox}
\usepackage{wrapfig}
\usepackage{subcaption}
\usepackage{enumitem}

\hypersetup{hidelinks}

\newcolumntype{U}{@{\,}>{\scriptsize}l}

\begin{document}

\twocolumn[
\icmltitle{Studying K-FAC Heuristics by Viewing Adam through a Second-Order Lens}



\icmlsetsymbol{equal}{*}

\begin{icmlauthorlist}
\icmlauthor{Ross M.\ Clarke}{cam}
\icmlauthor{José Miguel Hernández-Lobato}{cam}
\end{icmlauthorlist}

\icmlaffiliation{cam}{University of Cambridge}

\icmlcorrespondingauthor{Ross M.\ Clarke}{rmc78@cam.ac.uk}

 \icmlkeywords{ICML, Machine Learning, Optimization for Machine Learning, K-FAC, Adam, Levenberg-Marquardt Damping, Second-Order Optimization}

\vskip 0.3in
]



\printAffiliationsAndNotice{}  % leave blank if no need to mention equal contribution

\begin{abstract}
    Research into optimisation for deep learning is characterised by a tension between the computational efficiency of first-order, gradient-based methods (such as SGD and Adam) and the theoretical efficiency of second-order, curvature-based methods (such as quasi-Newton methods and K-FAC). Noting that second-order methods often only function effectively with the addition of stabilising heuristics (such as Levenberg-Marquardt damping), we ask how much these (as opposed to the second-order curvature model) contribute to second-order algorithms' performance. We thus study \emph{AdamQLR}: an optimiser combining damping and learning rate selection techniques from K-FAC \citep{martens_optimizing_2015} with the update directions proposed by Adam, inspired by considering Adam through a second-order lens. We evaluate AdamQLR on a range of regression and classification tasks at various scales and hyperparameter tuning methodologies, concluding K-FAC's adaptive heuristics are of variable standalone general effectiveness, and finding an \emph{untuned} AdamQLR setting can achieve comparable performance vs runtime to \emph{tuned} benchmarks.
    %and exhibiting superior generalisation to K-FAC on larger datasets.
\end{abstract}

\section{Introduction}
At the heart of any machine learning model is an optimisation problem, and at the heart of any training procedure is an optimisation algorithm. Most frequently seen in the literature are \emph{first-order} optimisers such as SGD, Adam \citep{kingma_adam_2015} and their variants, but exploratory studies have also been performed on \emph{second-order} algorithms such as quasi-Newton methods and K-FAC \citep{martens_optimizing_2015}. Broadly speaking, second-order algorithms aim to secure more rapid convergence to an optimal value of the objective function by making more principled individual updates, which in turn are more computationally costly than those employed by first-order methods. Combined with a generally more complicated implementation, first-order methods are still preferable to second-order approaches for most practitioners \citep{anil_scalable_2021}.

In part, this is a stability issue --- by virtue of taking larger individual steps, second-order optimisers carry an increased risk of significantly worsening the objective value if their approximate understanding of curvature in objective space is a poor representation of the true space. Most second-order approaches thus \emph{depend} on additional heuristics (such as curvature damping) for their viability. Heuristics commonly seen in first-order methods, such as weight decay or momentum applied to SGD, improve an already effective optimiser; by contrast, second-order methods' heuristics are \emph{essential} components, without which the optimiser will perform unstably or ineffectively --- a point we demonstrate in Appendix~\ref{sec:KFACAblations}. It is then natural to ask to what extent these heuristics are responsible for the documented benefits of second-order optimisers, and whether they might similarly improve first-order techniques.

In this paper, we study a damped automatic learning rate strategy, derived by applying K-FAC's damping and learning rate selection techniques to Adam. The resulting algorithm --- an optimiser whose default hyperparameters compare favourably with tuned baselines --- allows us to investigate the impact of K-FAC's heuristics. After reviewing related work in Section~\ref{sec:RelatedWork}, we present our study algorithm in Section~\ref{sec:AdamQLR}. We then justify our claims by experiment in Section~\ref{sec:Experiments} before Section~\ref{sec:Conclusion} concludes. Our main contributions are as follows:
\begin{itemize}
    \item We argue K-FAC's adaptive heuristics are of variable general empirical effectiveness
    \item We present a novel use of damped, second-order approximate learning rate selection in Adam
    \item We show our untuned study method often performs similarly to methods using tuned hyperparameters, exhibiting robustness to hyperparameters
\end{itemize}

\section{Related Work}
\label{sec:RelatedWork}

First-order methods form the bread and butter of modern machine learning, with SGD and Adam \citep{kingma_adam_2015} being most frequently seen. Adam belongs to a class of \emph{adaptive} first-order methods, which apply some kind of normalisation transformation to the observed gradients; other examples include Adagrad \citep{mcmahan_adaptive_2010,duchi_adaptive_2011} and RMSprop \citep{tieleman_neural_2012}.  \citet{balles_dissecting_2018} demonstrate that Adam essentially scales gradient signs by their variance. \citet{zhang_noisy_2018} show that Adam can be seen as a form of natural gradient mean field variational inference, whose mode-fitting behaviour is known to underestimate variance, corresponding to overestimating curvature in an optimisation task (see e.g.\ Figure 1.3 in \citet{turner_two_2011}). \citet{zhang_which_2019} use a noisy quadratic model to argue for the benefits of exponential moving averages and other components found in Adam. These methods achieve computational efficiency by using diagonal approximations or heuristics to understand curvature in the space, so ignore useful information which second-order methods incorporate.

Optimisers employing second-order derivative information are seen more often in the optimisation literature than in practical machine learning projects. The family of \emph{quasi-Newton} methods \citep{nocedal_numerical_2006} is inspired by the appearance of the Hessian matrix in a Taylor series truncated at quadratic order; this matrix characterises curvature in the model parameters. \citet{martens_deep_2010} use the Hessian-vector product trick \citep{pearlmutter_fast_1994} to work implicitly with the exact Hessian. Other work modifies the Hessian to avoid degeneracies --- a particular concern in saddle point-dense high-dimensional spaces \citep{pascanu_revisiting_2014, dauphin_identifying_2014} --- or introduces spatial and temporal averaging of a diagonal approximation to the Hessian \citep{yao_adahessian_2021}. Although not explicitly using second derivatives, SHAMPOO \citep{gupta_shampoo_2018} learns a factorised set of preconditioned matrices. However, in the non-convex, non-quadratic spaces of ML, the unaltered Hessian may be badly misleading, leading to diverging losses.

Where the system is viewed as a probabilistic model, an alternative curvature characterisation is the Fisher information matrix, which gives rise to the natural gradient family of methods \citep{amari_natural_1998}. Unlike the Hessian, the Fisher matrix characterises curvature in KL-divergence space between the predicted and ground truth probability distributions. Factorized Natural Gradient \citep{grosse_scaling_2015} approximates the Fisher using a Gaussian graphical model, while the Kronecker-Factored Approximate Curvature (K-FAC) method (\citet{martens_optimizing_2015} after an idea by \citet{heskes_natural_2000}) imposes a block-diagonal approximation to the Fisher and represents each block by a Kronecker product. Extensions to K-FAC include EKFAC \citep{george_fast_2018}, which learns the approximate Fisher in an eigenvalue-aligned basis, and a mini-block Fisher variant \citet{bahamou_mini-block_2023} targets the empirical Fisher by asserting that every diagonal block is itself block-diagonal. TNT \citep{ren_tensor_2021} exploits the Kronecker-factored covariance of the assumed tensor normal-distributed sample gradients to approximate the exact probabilistic Fisher matrix, while K-BFGS \citep{goldfarb_practical_2020} applies a similar factorisation strategy to the Hessian matrix, retaining theoretical guarantees from the classical BFGS optimiser \citep{broyden_convergence_1970, fletcher_new_1970, goldfarb_family_1970, shanno_conditioning_1970}. Although K-FAC can be applied in distributed settings, this is somewhat complex \citep{osawa_large-scale_2019}, and Fisher curvature expressions must be calculated anew for each different network architecture block.

Another line of work aims to accelerate first-order methods by dynamically adapting the learning rate to match the local optimisation dynamics. Originally this was predominantly done by imposing fixed learning rate schedules \citep{darken_note_1990,  li_exponential_2019, xu_learning_2019, loshchilov_sgdr_2017, smith_dont_2018}, but recent developments involve more dynamic adaptations by hypergradients \citep{franceschi_forward_2017, micaelli_non-greedy_2020, donini_marthe_2020, lorraine_optimizing_2020, clarke_scalable_2022}, online Bayesian optimisation \citep{jin_autolrs_2023}, or explicitly constructing an optimisation framework around the unique characteristics of deep neural networks \citep{bernstein_automatic_2023}. \citet{zhang_which_2019} and \citet{kwatra_autolr_2023} adopt a similar quadratic model methodology to our work, but the latter compute a finite-difference approximation to this model rather than using the exact curvature information as we do, and introduces additional hyperparameters controlling an exploration/exploitation trade-off.  \citet{niu_ml-bfgs_2023} uses a parallel approach to ours to incorporate momentum into L-BFGS \citep{liu_limited_1989}. These methods generally suffer an increased cost over simpler strategies, whether to discover a schedule, compute hypergradients or perform de-facto inline hyperparameter optimisation, in turn requiring a substantial validation dataset to be held aside.
%\citep{bernstein_automatic_2023} 
%\citep{kwatra_autolr_2023}
%\citep{jin_autolrs_2023}
%\citep{balles_dissecting_2018}
%\citep{zhang_which_2019}
%\citep{iyer_lrtuner_2021}
%\citep{zhang_noisy_2018}
%NOTE: Checked Slack back to 1st March 2023

% \vspace{-0.2cm}
\section{AdamQLR}
\label{sec:AdamQLR}
% \vspace{-0.2cm}

We consider the minimisation of $f(\vec{\theta})$, representing the loss function of some network parameterised by $\vec{\theta}$.

% \vspace{-0.2cm}

\subsection{First- and Second-Order Methods}
Many optimisation algorithms in ML take the form $\vec{\theta}_t \gets \vec{\theta}_{t-1} - \alpha \vec{u}(\vec{g}_t)$, where $\alpha$ is a learning rate and $\vec{u}$ some update function. This $\vec{u}$ may depend on an internal state and the current gradient $\vec{g}_t$, but not on any higher derivative. As is conventional, we call such algorithms \emph{first-order} optimisers. By contrast, \emph{second-order} optimisers take the form $\vec{\theta}_t \gets \vec{\theta}_{t-1} - \vec{C}^{-1} \vec{u}(\vec{g}_t)$, where $\vec{C}$ is some curvature matrix (often a damped Hessian, Fisher or Gauss-Newton matrix).

First-order methods broadly provide computational efficiency at the inconvenience of manually selecting $\alpha$, while second-order methods suffer a large computational cost to dynamically select an implicit $\alpha$ and improved update direction $\vec{d}$ using their more powerful objective models. However, a slew of `adaptive' first-order optimisers (such as Adam \citep{kingma_adam_2015} and relations) blur this distinction by constructing stateful models of the objective, which can often be interpreted as approximating the curvature of $f(\vec{\theta})$.

Moreover, practical second-order methods for ML are necessarily approximate, as the curvature $\vec{C}$ is otherwise intractably large. Further engineering is then required to mitigate the impact of approximate curvature and the inevitable non-convexity of $f$. For example, in K-FAC, \citet{martens_optimizing_2015} convincingly argue for a particular Kronecker factorisation of a block-diagonal $\vec{C}$, but then augment it with a raft of corrections and adaptive heuristics (including multiple periodically-updated damping/factorised Tikhonov regularisation terms, momentum, weight decay, exponential moving averages of curvature statistics and approximate exchange of expectations and Kronecker products). Further, these additions are seemingly \emph{essential} ingredients of a working K-FAC implementation.

A natural question is then whether curvature information or engineering heuristics contribute more to K-FAC's success. In particular, we might ask if accepting first-order methods' inaccurate curvature models and applying second-order stability techniques would blend computational efficiency with optimisation accuracy. We propose to study this possibility by adapting Adam using techniques from K-FAC.

\begin{figure*}
\vspace*{-2.7ex}
\begin{minipage}{0.49\linewidth}
\begin{algorithm}[H]
    \caption{Adam \citep{kingma_adam_2015}}
    \label{alg:AdamTransformed}
    \begin{algorithmic}
        \STATE $\vec{m}_0, \vec{v}_0 \gets \vec{0}$
        \FOR{$t = 1, 2, \cdots$ until $\vec{\theta}$ converged}
            \STATE $\vec{g}_t \gets \nabla_\vec{\theta} f (\vec{\theta}_{t-1})$
            \STATE $\vec{m}_t \gets \beta_1 \vec{m}_{t-1} + (1-\beta_1) \vec{g}_t$
            \STATE $\vec{v}_t \gets \beta_2 \vec{v}_{t-1} + (1-\beta_2) (\vec{g}_t \odot \vec{g}_t)$
            \STATE $\widehat{\vec{m}}_t \gets \frac{\vec{m}_t}{1 - \beta_1^t}$
            \STATE $\widehat{\vec{v}}_t \gets \frac{\vec{v}_t}{1 - \beta_2^t}$
            \STATE $\vec{d}_t \gets \frac{\widehat{\vec{m}}_t}{\sqrt{\widehat{\vec{v}}_t} + \epsilon}$
            % \STATE \vspace{2.9ex}
            \STATE \phantom{Update learning rate $\alpha$ according to \eqref{eq:LearningRateUpdate}}
            \STATE \phantom{Update damping $\lambda$ according to \eqref{eq:DampingUpdate}}
            \STATE $\vec{\theta}_t \gets \vec{\theta}_{t-1} - \alpha \vec{d}_t$
        \ENDFOR
    \end{algorithmic}
\end{algorithm}
\end{minipage}
\hfill
\begin{minipage}{0.49\linewidth}
\begin{algorithm}[H]
    \caption{AdamQLR}
    \label{alg:AdamQLR}
    \begin{algorithmic}
        \STATE $\vec{m}_0, \vec{v}_0 \gets \vec{0}$
        \FOR{$t = 1, 2, \cdots$ until $\vec{\theta}$ converged}
            \STATE $\vec{g}_t \gets \nabla_\vec{\theta} f (\vec{\theta}_{t-1})$
            \STATE $\vec{m}_t \gets \beta_1 \vec{m}_{t-1} + (1-\beta_1) \vec{g}_t$
            \STATE $\vec{v}_t \gets \beta_2 \vec{v}_{t-1} + (1-\beta_2) (\vec{g}_t \odot \vec{g}_t)$
            \STATE $\widehat{\vec{m}}_t \gets \frac{\vec{m}_t}{1 - \beta_1^t}$
            \STATE $\widehat{\vec{v}}_t \gets \frac{\vec{v}_t}{1 - \beta_2^t}$
            \STATE $\vec{d}_t \gets \frac{\widehat{\vec{m}}_t}{\sqrt{\widehat{\vec{v}}_t} + \epsilon}$
            {\color{solarizedMagenta}
            \STATE Update learning rate $\alpha$ according to \eqref{eq:LearningRateUpdate}
            \STATE Update damping $\lambda$ according to \eqref{eq:DampingUpdate}
            }
            \STATE $\vec{\theta}_t \gets \vec{\theta}_{t-1} - \alpha \vec{d}_t$
        \ENDFOR
    \end{algorithmic}
\end{algorithm}
\end{minipage}
\end{figure*}

\subsection{Adam Revisited}

Algorithm~\ref{alg:AdamTransformed} restates the Adam optimisation algorithm from \citet{kingma_adam_2015} applied to $f$, with some minor notational changes. Our proposed algorithm derives from our anecdotal observation that Adam often makes good choices of update direction, which we notate by $\vec{d}_t = \frac{\widehat{\vec{m}}_t}{\sqrt{\widehat{\vec{v}}_t} + \epsilon}$.

As we detail in Appendix~\ref{sec:CurvatureMatrices}, Adam is known to carry a diagonal approximation to the empirical Fisher matrix in $\widehat{\vec{v}}_t$. Then, the $\frac{1}{\sqrt{\widehat{\vec{v}}_t} + \epsilon}$ term in Algorithm~\ref{alg:AdamTransformed} effectively performs a curvature transformation on the averaged gradient $\widehat{\vec{m}}_t$ before computing a more traditional gradient-based update for $\vec{\theta}$. There are widely-known limitations to using the empirical Fisher in place of the true Fisher information matrix \citep{kunstner_limitations_2019}, and the square root is motivated only by a desire to be ``conservative'' \citep{kingma_adam_2015}. Indeed, \citet{zhang_noisy_2018} show Adam is very similar to one construction of natural gradient mean-field variational inference, a technique which is known to prioritise locally fitting modes of the target probability distribution \citep{turner_two_2011}. The consequent underestimation of global variance corresponds to overestimating local curvature in optimisation, justifying \citet{kingma_adam_2015}'s preference for a conservative estimate. Nonetheless, this formulation invites us to view Adam through a second-order optimisation lens, and ask whether common second-order optimiser heuristics might bring similar benefits to Adam.

\subsection{Adopting Heuristics from K-FAC}

After defining its Kronecker-factored block diagonal approximation to the curvature matrix, K-FAC \citep{martens_optimizing_2015} includes three important stabilising heuristics: Levenberg-Marquardt damping, and learning rate and momentum selection according to a local second-order model. Since Adam already implements a momentum correction in $\widehat{\vec{m}}_t$, we consider only the first two techniques.

Levenberg-Marquardt damping \citep{levenberg_method_1944, marquardt_algorithm_1963, roweis_levenberg-marquardt_1996} replaces the curvature matrix $\vec{C}$ with the damped version $\vec{C} + \lambda \vec{I}$. Its interpretations include approximating a trust region, enforcing positive definiteness of $\vec{C}$, preventing large updates in low-curvature directions and interpolating between gradient descent and full Newton updates. In effect, it imposes a `minimum curvature' on the objective to avoid near-zero eigenvalues in $\vec{C}$.

Let $M(\vec{\theta})$ be an approximate second-order model around $\vec{\theta}_{t-1}$, defined by a truncated Taylor series:
\begin{multline}
    \label{eq:QuadraticModel}
    M(\vec{\theta}) =
    f(\vec{\theta}_{t-1})
    + (\vec{\theta} - \vec{\theta}_{t-1})\trans \vec{g}_t \\
    + \frac{1}{2} (\vec{\theta} - \vec{\theta}_{t-1}) \trans (\vec{C} + \lambda \vec{I}) (\vec{\theta} - \vec{\theta}_{t-1}).
\end{multline}
The damping parameter $\lambda$ is adapted by comparing the change in objective value predicted by the model $(M(\vec{\theta}_t) - M(\vec{\theta}_{t-1}))$ to the actual observed change $(f(\vec{\theta}_{t}) - f(\vec{\theta}_{t-1}))$. This adjustment quantifies the model's reliability by a reduction ratio $\rho$, incorporating stepping factors\footnote{In the most general form we allow separate decrease and increase factors, but in practice we will often choose $\omega_\text{dec} = \frac{1}{\omega_\text{inc}}$ for simplicity. We also require $0 < \omega_\text{dec} < 1 < \omega_\text{inc}$.} $\omega_\text{dec}, \omega_\text{inc}$:
\begin{equation}
    \label{eq:DampingUpdate}
    \rho = \frac{f(\vec{\theta}_{t}) - f(\vec{\theta}_{t-1})}{M(\vec{\theta}_t) - M(\vec{\theta}_{t-1})};
    \hfill
    \lambda \gets \begin{cases}
        \omega_\text{dec} \lambda & \text{if } \rho > \frac{3}{4} \\
        \lambda & \text{if } \frac{1}{4} \leq \rho \leq \frac{3}{4} \\
        \omega_\text{inc} \lambda & \text{if } \rho < \frac{1}{4}
    \end{cases}.
\end{equation}
We discuss this formulation further in Appendix~\ref{sec:ReductionRatioCommentary}.

After choosing an update direction $\vec{d}_t$, a learning rate $\alpha$ is selected according to the second-order model $M$. We minimise $M(\vec{\theta}_{t-1} - \alpha \vec{d}_t)$ with respect to $\alpha$, which yields
\begin{equation}
    \label{eq:LearningRateUpdate}
    \alpha = \frac{\vec{g}_t\trans \vec{d}_t}{\vec{d}_t\trans (\vec{C} + \lambda \vec{I}) \vec{d}_t}.
\end{equation}

A minor rearrangement shows the large matrix $\vec{C}$ only appears in products with vectors. The Jacobian-vector product trick \citep{pearlmutter_fast_1994}, efficient Fisher decompositions \citep{martens_optimizing_2015} and similar techniques compute these quantities using only one additional backward pass per product with $\vec{C}$. In practice, the information value of these calculations outweighs this cost.

\subsection{Extending Adam}

Incorporating K-FAC's damping and learning rate selection strategies into Adam yields Algorithm~\ref{alg:AdamQLR}, which is easily implementable as a wrapper around vanilla Adam. We name this family of algorithms \emph{AdamQLR}, where \emph{QLR} indicates an optimiser-agnostic quadratic-model learning rate selection logic, which could be broadly applied (e.g. to SGD). Appendix~\ref{sec:ConvergenceOfAdamQLR} gives a sketch of a convergence argument for this algorithm.

One remaining consideration is the choice of a curvature matrix $\vec{C}$. We use the (true) Fisher matrix throughout, inspired by its connection with Adam's $\widehat{\vec{v}}_t$ buffer (see Appendix~\ref{sec:AdamAndFisherMatrix}), its use at the heart of K-FAC and its positive semi-definite guarantee. In short, we tune the damping parameter $\lambda$ to create a trust region in which our quadratic approximation --- specified by the Fisher --- is accurate. Then, given the Adam descent direction and the selected $\lambda$, we choose the optimal step size within this trust region. We exploit Jacobian-vector products and the efficient Fisher decomposition described in \citet[Appendix~C]{martens_optimizing_2015}, which computes exact products without explicitly storing $\vec{C}$.

Finally, our experiments found AdamQLR's training stability to be most threatened by selecting an unreasonably large $\alpha$ for a particular iteration, causing a divergent parameter update. The problem worsens in larger models with more prevalent low-curvature regions of the space to induce very large update sizes. We found that larger batch sizes improved our curvature estimates, leading to better performance despite the higher cost of each forward pass.

Now, the only remaining hyperparameters are $\beta_1$, $\beta_2$ and $\epsilon$ (from Adam) and an initial damping value $\lambda_0$. As Adam's hyperparameters are commonly fixed at the default values suggested by \citet{kingma_adam_2015}, and we show $\lambda_0$ to be sufficiently insensitive that a default value can be recommended (Section~\ref{sec:SensitivityStudiesSummary}), we claim that AdamQLR is robust, even without explicit hyperparameter tuning. In particular, we have encapsulated the learning rate $\alpha$ --- arguably the most important hyperparameter to select in many optimisation algorithms. We justify this claim in Section~\ref{sec:Experiments}.

Compared to Adam, we suffer additional forward and backward passes to compute $M(\vec{\theta}_t)$ and $(\vec{C} + \lambda \vec{I}) \vec{d}_t$. These turn out not to impede empirical performance, though we note a careful implementation would amortise the former cost. Our only significant additional memory cost is storing the vector $(\vec{C} + \lambda \vec{I}) \vec{d}_t$, making our approximate memory footprint four times that of SGD (where Adam's is three times SGD).

\section{Experiments}
\label{sec:Experiments}

We examine the training and test performance of AdamQLR against Adam and K-FAC in a variety of settings:
\begin{description}[itemsep=-0.05ex]
    \item[\citet{rosenbrock_automatic_1960} Function] Let $a=1$ and $b=100$, then $f(x, y) = (a - x)^2 + b(y - x^2)^2$
    \item[UCI~Energy] \citep{tsanas_accurate_2012} on an MLP with one hidden layer of 50 units
    \item[UCI~Protein] \citep{rana_uci_2013} on an MLP with one hidden layer of 100 units
    \item[Fashion-MNIST] \citep{xiao_fashion-mnist_2017} on an MLP with one hidden layer of 50 units
    \item[SVHN] \citep{netzer_reading_2011} on a ResNet-18 \citep{he_deep_2016}
    \item[CIFAR-10] \citep{krizhevsky_learning_2009} on a ResNet-18 \citep{he_deep_2016}
\end{description}
We also perform a study on Penn Treebank in Appendix~\ref{sec:PennTreebank}. On UCI datasets we generate random splits using the same sizes as \citet{gal_dropout_2016} and use MSE loss; otherwise, we separate the standard test set, randomly choose \nicefrac{1}{6} (Fashion-MNIST and SVHN) or \nicefrac{1}{10} (CIFAR-10) of the remaining data to form a validation set, and use cross-entropy loss. Code for all our experiments is available at 
\url{https://github.com/rmclarke/AdamThroughASecondOrderLens}.
We compare (see Appendix~\ref{sec:ChoiceOfBaselines} for further notes):

\begin{description}[itemsep=-0.05ex]
    \item[SGD Minimal] Classical mini-batched stochastic gradient descent, with tuned learning rate
    \item[SGD Full] \emph{SGD Minimal} with additional tuned momentum and weight decay
    \item[Adam] \citep{kingma_adam_2015} with tuned learning rate and fixed defaults for other hyperparameters
    \item[Adam (Untuned)] \emph{Adam} \citep{kingma_adam_2015}, fixing batch sizes\footnote{We use a `typical' batch size for each setting: full-batch for UCI Energy, 100 for UCI Protein, 50 for Fashion-MNIST, 256 for SVHN and 128 for CIFAR-10.} and learning rate of $0.001$ 
    \item[K-FAC] \citep{martens_optimizing_2015, botev_kfac-jax_2022} with tuned initial damping
    \item[K-FAC (Untuned)] \emph{K-FAC} \citep{martens_optimizing_2015, botev_kfac-jax_2022} with initial damping set to a default $1.0$ and fixed batch size 3200
    \item[AdamQLR (Tuned)] Algorithm~\ref{alg:AdamQLR}, using Fisher curvature for $\vec{C}$. We tune initial damping and damping adjustment factors $\omega_\text{dec}, \omega_\text{inc}$
    \item[AdamQLR (Untuned)] \emph{AdamQLR} with fixed batch size 3\,200, initial damping 0.001 and $\omega_\text{dec} = \frac{1}{\omega_\text{inc}} = 0.5$ (justified by Section~\ref{sec:SensitivityStudiesSummary} and Appendix~\ref{sec:SensitivityExperiments})
\end{description}
Except for the Rosenbrock Function and \emph{(Untuned)} variants, we also tune a batch size over $\{50, 100, 200, 400, 800, 1\,600, 3\,200\}$. All hyperparameter tuning uses ASHA \citep{li_system_2020} over 200 random initialisations, targeting a fixed number of training epochs, subject to a maximum runtime of 15~minutes (only reached for CIFAR-10; see Appendix~\ref{sec:ASHATimeExperiments} for experiments using runtime as the primary constraint). We then perform 50 runs using each of the best hyperparameters found (measured by final validation loss), then plot the mean and standard deviation of the median trends of each of 50 bootstrap samples of the results. Following \citet{botev_kfac-jax_2022}, any damping is clipped to $\lambda \geq 10^{-8}$. Except for the Rosenbrock Function, we give a numerical comparison of the end-of-training statistics in Table~\ref{tab:EpochConstrainedFinalResults}.

In Appendix~\ref{sec:ASHATimeExperiments}, we present analogous results where the hyperparameters are tuned to minimise training or validation losses after a fixed runtime, with no epoch constraint. Appendix~\ref{sec:AblationExperiments} details additional ablation studies on Adam, AdamQLR and K-FAC.

\subsection{Rosenbrock Function}
\begin{figure}
    \centering
    \includegraphics[width=\linewidth]{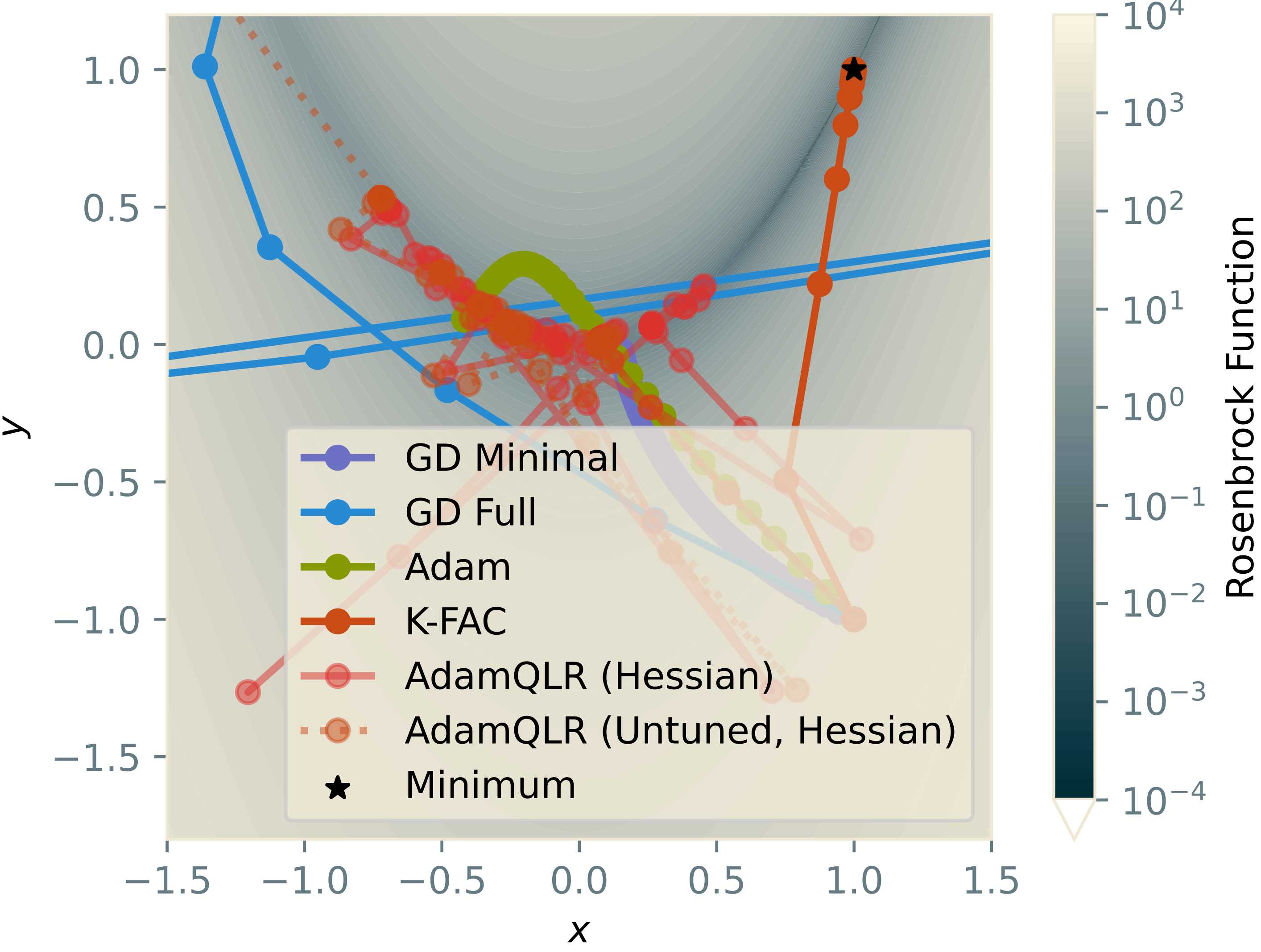}
    \caption{Optimisation trajectories over 200~steps from a fixed initial point on the Rosenbrock Function. Hyperparameter tuning used 200 standard-normal random initial points.}
    \label{fig:RosenbrockTrajectories}
    \vspace*{-2ex}
\end{figure}

The Rosenbrock Function \citep{rosenbrock_automatic_1960} provides a visualisable test bed for optimisation algorithms, containing non-linear correlations between its inputs and anisotropic curvature. We consider 200~optimisation steps, using $\mathcal{N}(\vec{0}, \vec{I})$-sampled initial $(x, y)$ values during hyperparameter tuning, and evaluate trajectories from the fixed starting point $(1, -1)$ in Figure~\ref{fig:RosenbrockTrajectories}. As there is no probabilistic model, we apply K-FAC to a least-squares formulation \citep[page 38]{brunet_contributions_2010}, and we use Hessian curvature in \emph{AdamQLR} for this experiment only. Additionally, we use gradient descent (\emph{GD}) in place of \emph{SGD}. Since there is no separate validation set, we tune hyperparameters on the objective function directly.

Here, \emph{GD Minimal} makes good initial progress into the central `valley', but its learning rate is too small to continue along the valley floor --- without regularisation, GD must select conservative step sizes to avoid diverging from poor initialisations. \emph{GD Full}'s hyperparameters cause it to bounce unstably around the optimisation space --- the algorithm is unable to recover from their overly aggressive settings. \emph{Adam}'s adaptive buffers allow it to target the valley more directly, eventually making slow progress along the valley floor, but it takes time to learn the new dynamics in the latter regime, initially `overshooting' the valley. \emph{K-FAC} demonstrates an impressive understanding of the optimisation space, making direct, rapid progress towards the optimum in spite of the challenging Euclidean curvature.

\emph{AdamQLR (Tuned)} demonstrates a combination of all these phenomena. It begins by updating in the same direction as \emph{Adam}, but with more aggressive step sizes, permitting faster initial progress. This confidence subsequently causes significant steps away from the central valley, but K-FAC's adaptive heuristics allow the algorithm to recover from these errors and return to the valley. Compared to \emph{Adam}, \emph{AdamQLR (Tuned)} performs much more exploration of the central valley, indicating the latter exploits its ability to recover from poor steps by accepting more aggressive updates. \emph{AdamQLR (Untuned)} follows a similar approach, though again with ill-suited hyperparameters for this toy problem. These initial results suggest K-FAC's heuristics do carry potential in their own right for first-order approaches.

\subsection{UCI Energy}
\begin{figure*}[p]
    \centering
    \newcommand{\rotatecaption}[1]{%
        \rotatebox[origin=c]{90}{\begin{minipage}{3cm}#1\end{minipage}} }
    \begin{tabularx}{0.82\linewidth}{p{2ex}XX}
        \rotatecaption{\subcaption{UCI Energy}\label{fig:UCIEnergyLosses}}
        & \includegraphics[align=c,width=\linewidth]{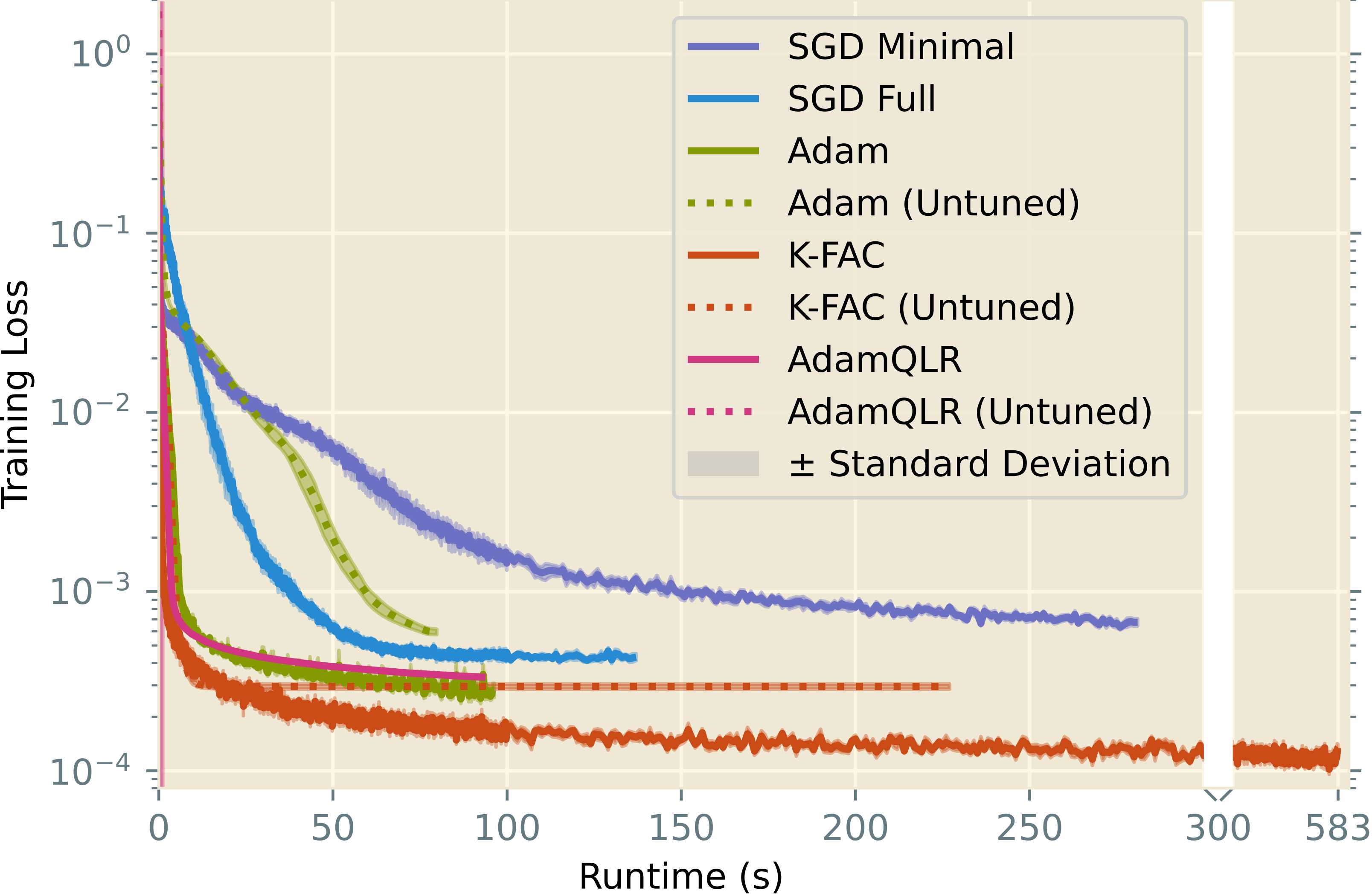}
        & \includegraphics[align=c,width=\linewidth]{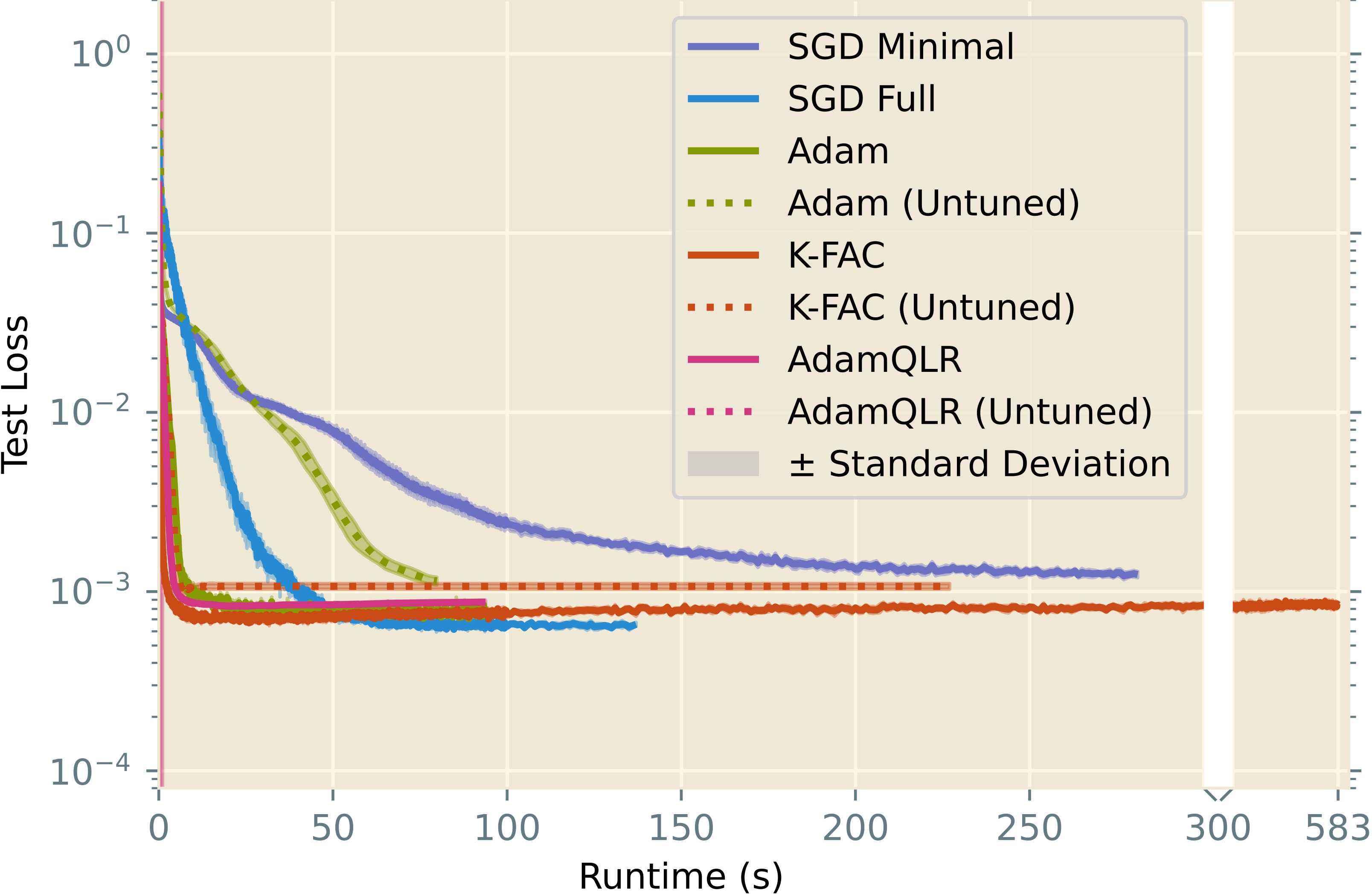} \vfill\\
        
        \rotatecaption{\subcaption{UCI Protein}\label{fig:UCIProteinLosses}}
        & \includegraphics[align=c,width=\linewidth]{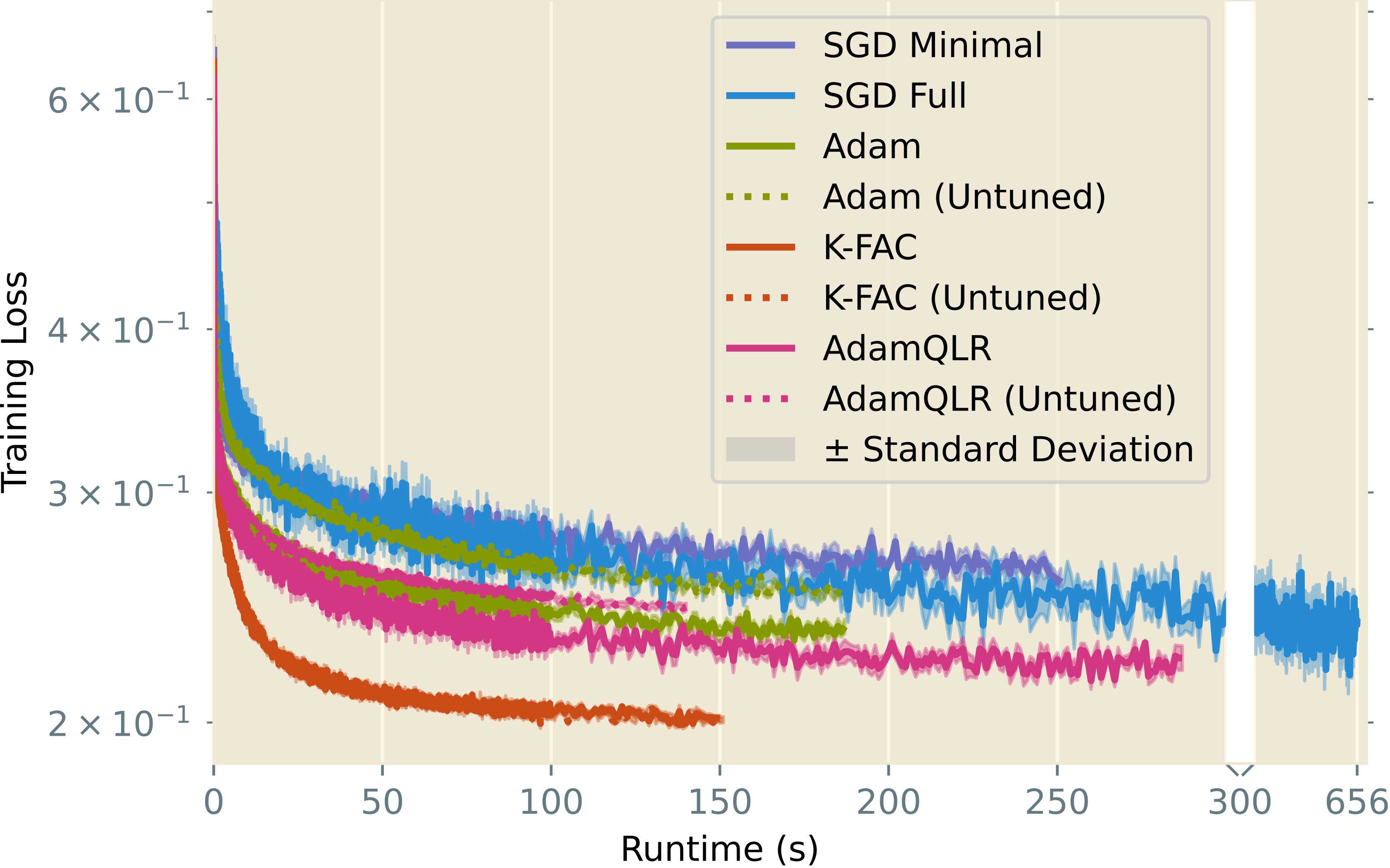}
        & \includegraphics[align=c,width=\linewidth]{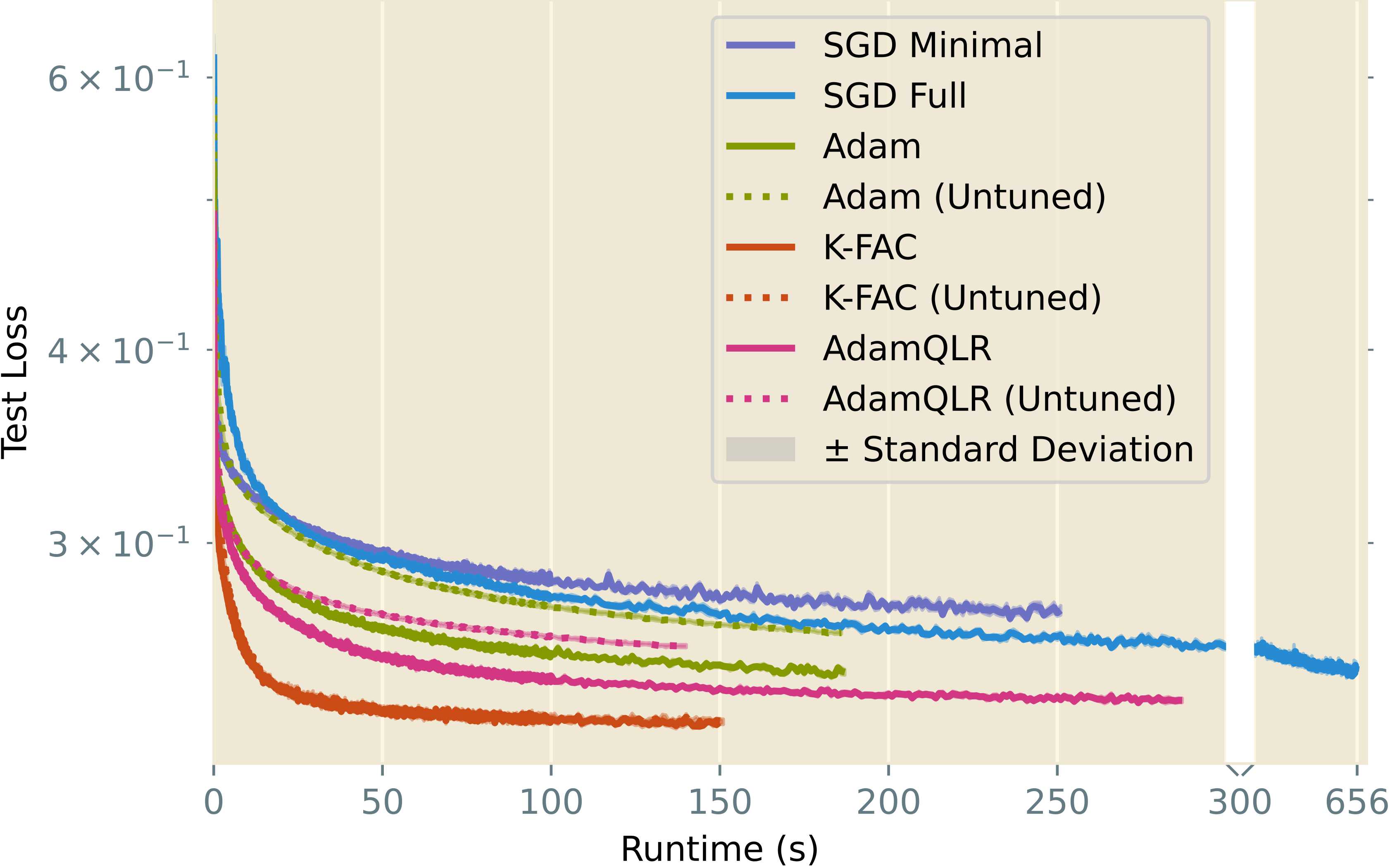} \vfill\\
        
        \rotatecaption{\subcaption{Fashion-MNIST}\label{fig:FashionMNISTAccuracies}}
        & \includegraphics[align=c,width=\linewidth]{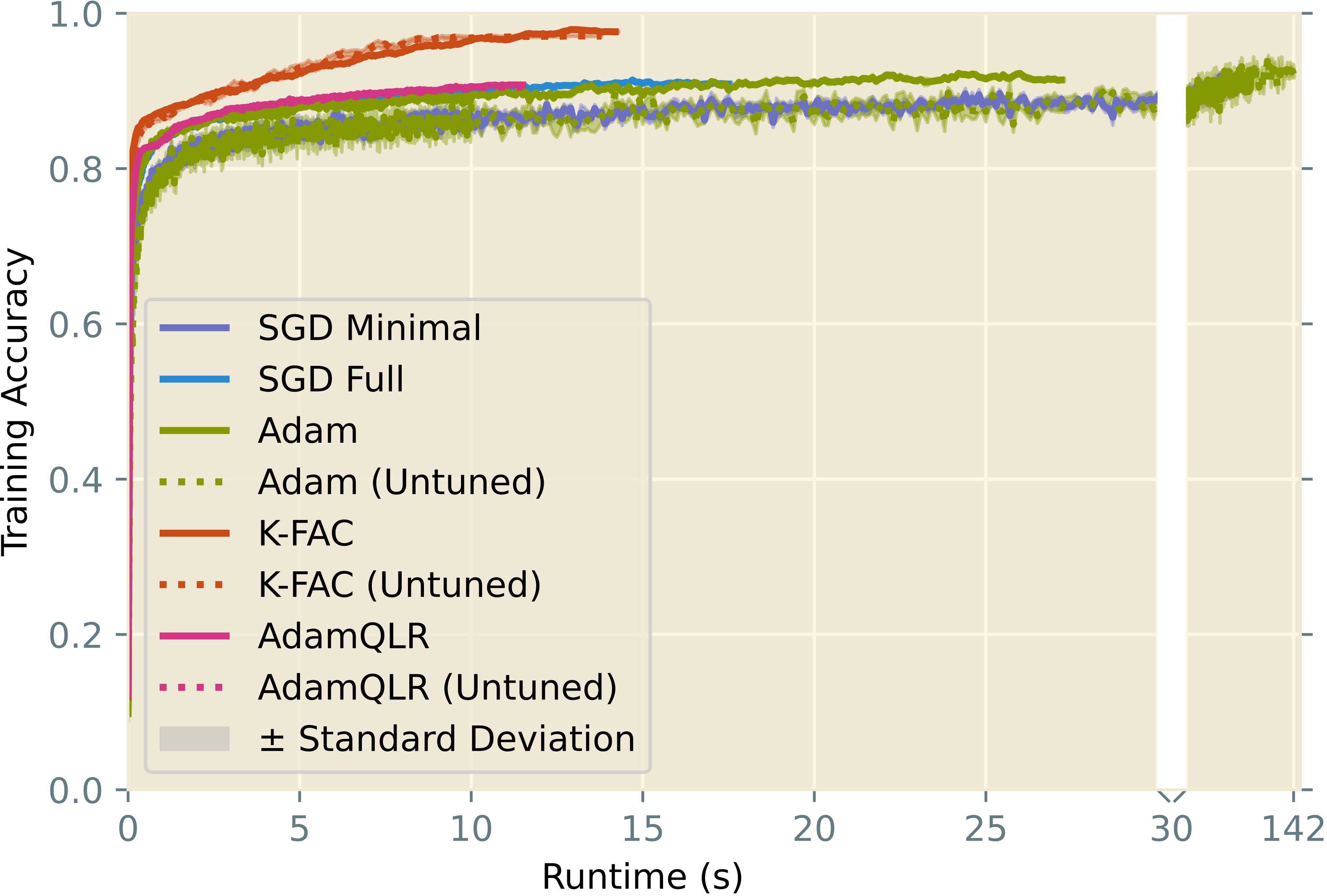}
        & \includegraphics[align=c,width=\linewidth]{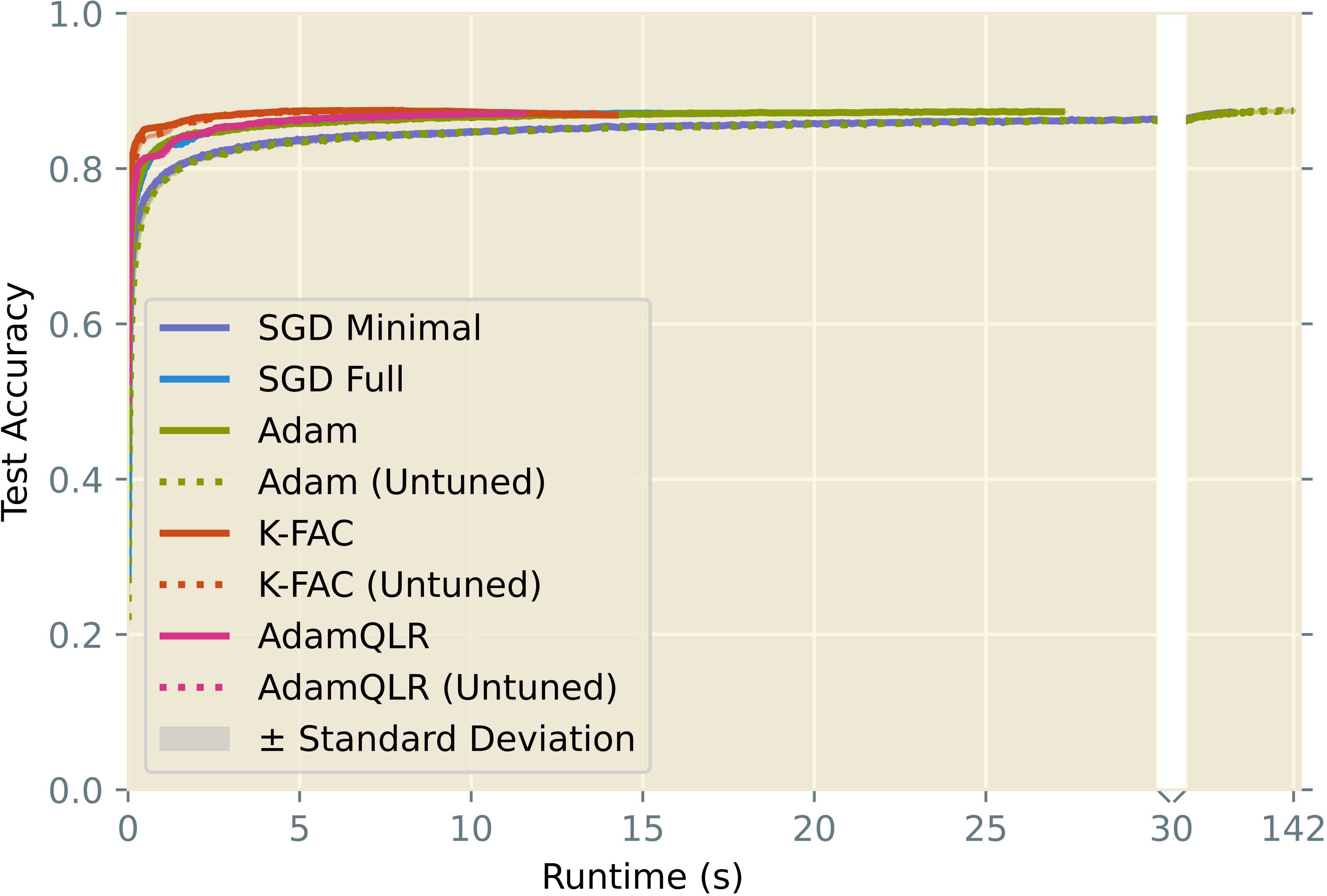} \vfill\\
        
        \rotatecaption{\subcaption{SVHN}\label{fig:SVHNAccuracies}}
        & \includegraphics[align=c,width=\linewidth]{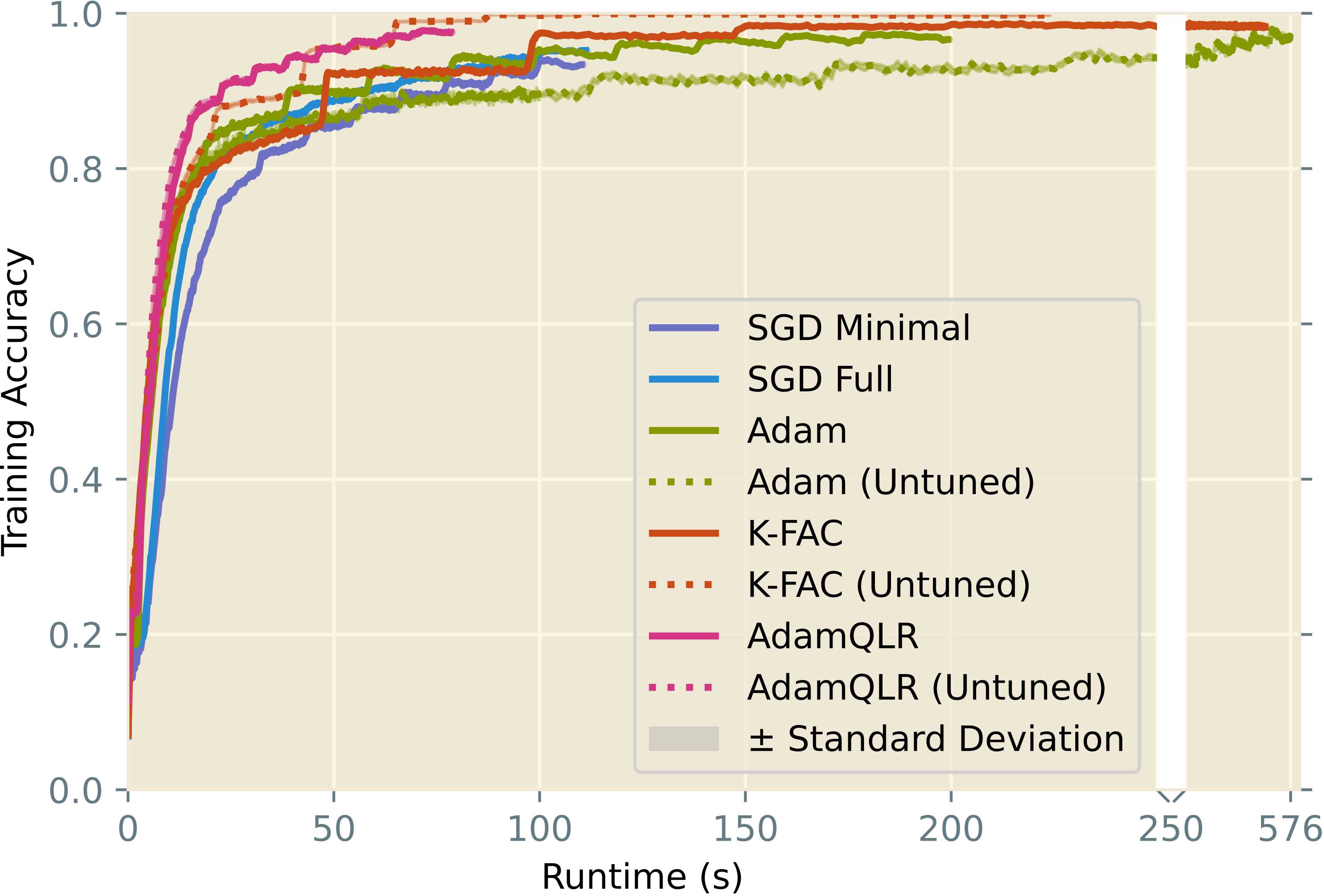}
        & \includegraphics[align=c,width=\linewidth]{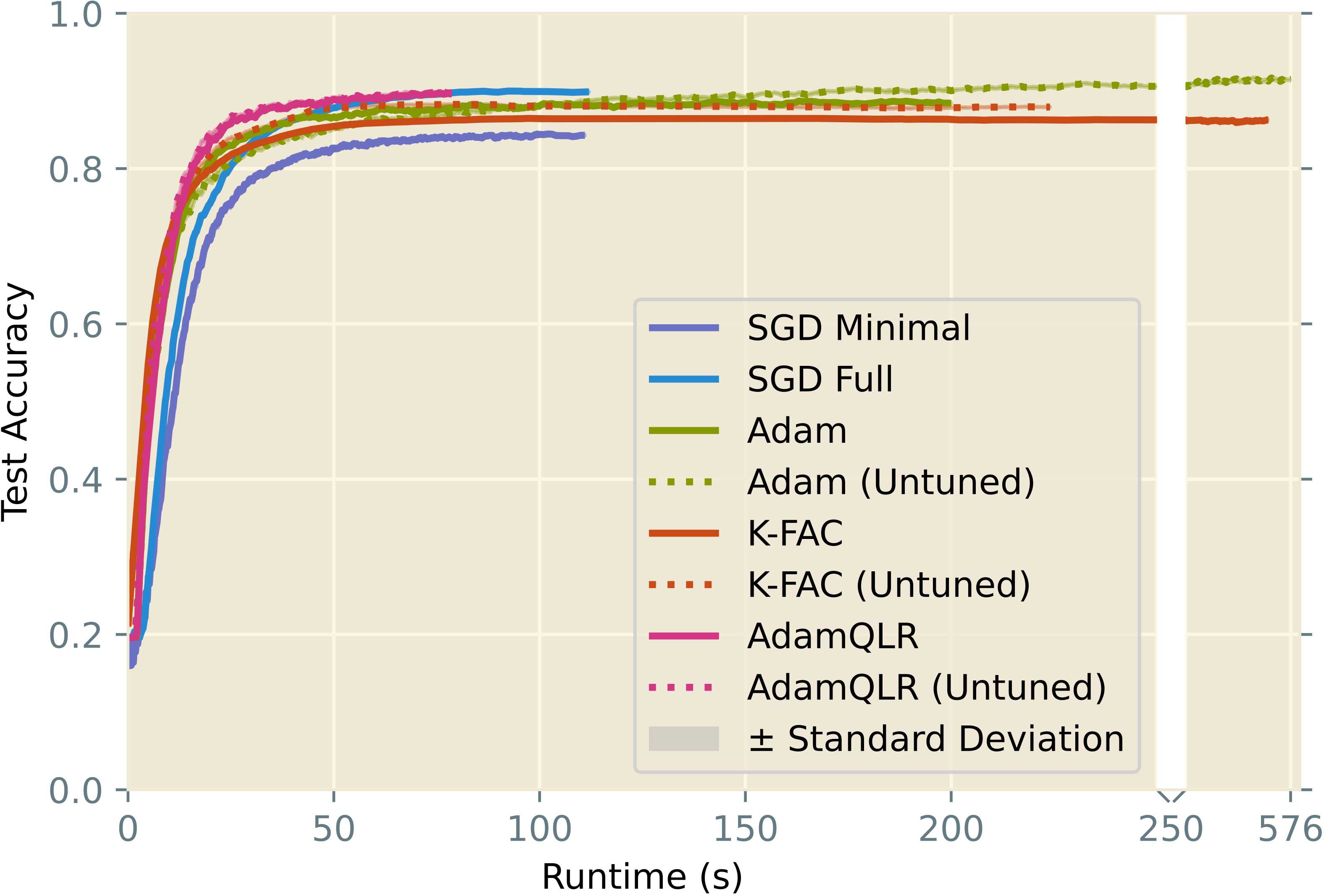} \vfill\\
        
        \rotatecaption{\subcaption{CIFAR-10}\label{fig:CIFAR10Accuracies}}
        & \includegraphics[align=c,width=\linewidth]{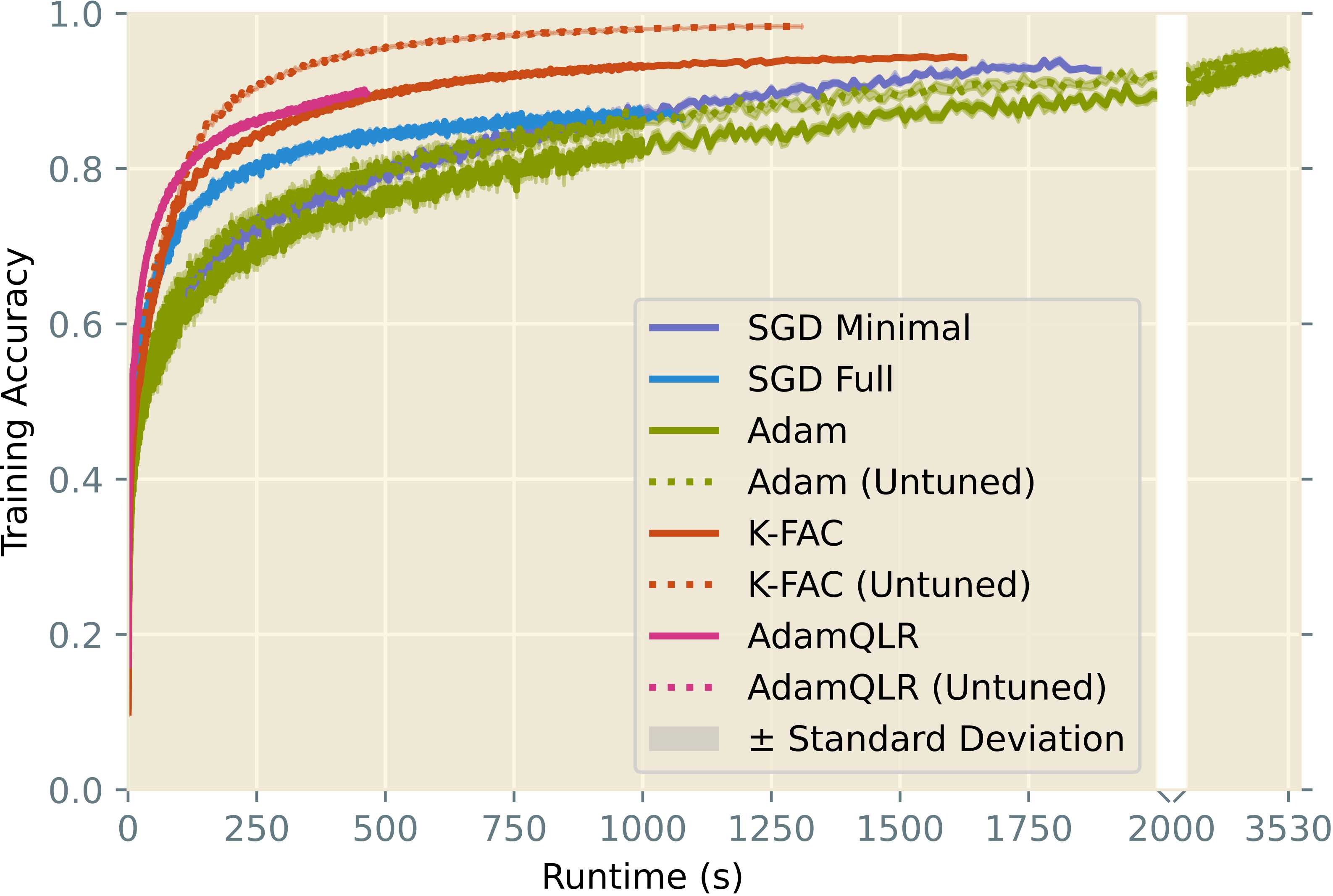}
        & \includegraphics[align=c,width=\linewidth]{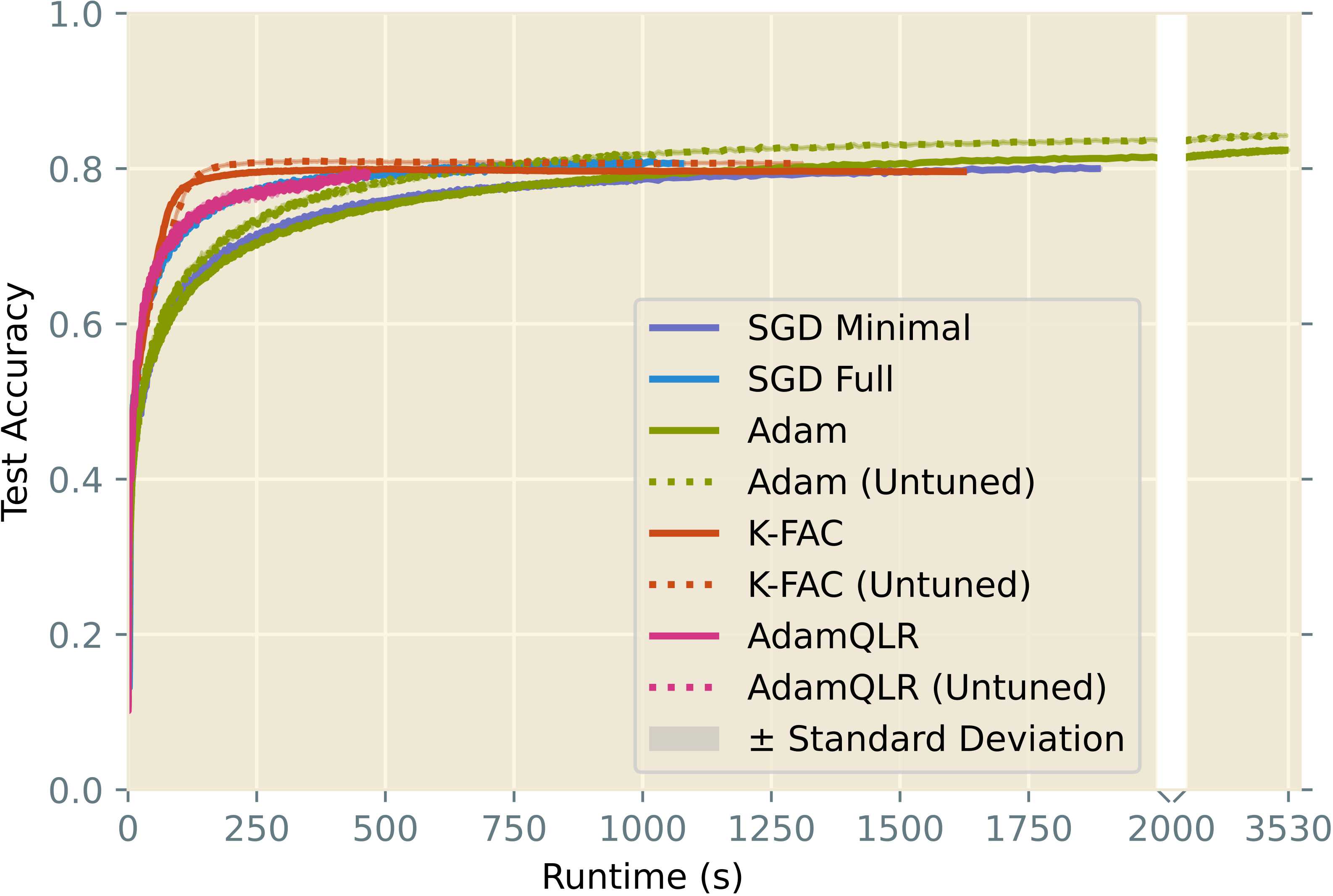} \vfill
    \end{tabularx}
    \caption{Median training (left) and test (right) performance trajectories, bootstrap-sampled over 50 repetitions per algorithm. Hyperparameters chosen by ASHA over 200 initialisations. Note changes of scale on the time axes. See also results on loss metrics and learning rate evolutions in Figures~\ref{fig:AlgorithmLosses} and~\ref{fig:AlgorithmLearningRates}, and numerical comparison in Table~\ref{tab:EpochConstrainedFinalResults}.}
    \label{fig:AlgorithmMixedResults}
\end{figure*}

UCI~Energy provides a low-dimensional regression task on a small dataset, which is amenable to hosting long experiments to explore convergence behaviour. We consider 4\,000 epochs of training and plot bootstrap-sampled median training and test loss trends in Figure~\ref{fig:UCIEnergyLosses}.

Our principal benchmarks fall much as we would expect: \emph{SGD Minimal} makes respectable, if sluggish, progress during optimisation, but is outclassed by the more rapid initial convergence of \emph{SGD Full} and \emph{Adam}. Both these latter methods achieve strong test performance, with \emph{SGD Full} achieving the best final test loss of all methods. Despite its rapid initial progress, \emph{K-FAC} quickly begins overfitting, reaching a final test loss similar to the \emph{AdamQLR} methods.

Generally, \emph{AdamQLR (Tuned)} performs comparably with the tuned \emph{Adam} baseline. The QLR computed learning rates accelerate initial progress, while the addition of damping provides some defence against overfitting, at the cost of a higher final training loss. However, on this trial, \emph{AdamQLR (Untuned)} diverged rapidly beyond the limits of Figure~\ref{fig:UCIEnergyLosses}. Since this phenomenon is not apparent with our other trials, and \emph{K-FAC} performs fine, we suppose the latter's curvature matrix smoothing must be important when using second-order information in this problem. Under full hyperparameter tuning, \emph{AdamQLR (Tuned)} does not convincingly improve over \emph{Adam} --- potentially the small-scale, full-batch nature of this setting responds poorly to this algorithm.

\subsection{UCI Protein}
UCI~Protein is another low-dimensional regression task, but with far more data points, allowing for a computationally-efficient study of a larger dataset. We show 200 epochs of training in Figure~\ref{fig:UCIProteinLosses}.

Here we see distinct generalisation trends for each algorithm. \emph{SGD Full} improves slightly over \emph{SGD Minimal}, but still lags behind the other methods. \emph{K-FAC} is now clearly the best-performing algorithm, as might perhaps be expected since it computes the most granular curvature approximation when choosing an update direction. However, we see meaningful gains from \emph{AdamQLR}, with the \emph{(Tuned)} variant comfortably outperforming \emph{Adam}. We observe \emph{AdamQLR}'s automatic learning rate selection is capable of outperforming methods which require a sensitive explicit choice of learning rate --- the \emph{Untuned} variant is clearly superior to tuned \emph{SGD} on this task and is only slightly worse than tuned \emph{Adam}. The clear ranking here indicates meaningful value is added to \emph{Adam} by including some heuristics from \emph{K-FAC}, but that \emph{K-FAC} offers further advantage even beyond this.

\subsection{Fashion-MNIST}
\label{sec:ExperimentsFashionMNIST}
Fashion-MNIST provides a first foray into higher-dimensional data, but at a scale still approachable by MLP models. Using a 10-epoch training window, we plot bootstrapped accuracy evolutions in Figure~\ref{fig:FashionMNISTAccuracies} and loss evolutions in Figure~\ref{fig:FashionMNISTLosses}.

While \emph{K-FAC} achieves the best final training loss, its loss evolution plots reveal a significant tendency to overfit. While this is a recognised issue with K-FAC \citep{martens_kronecker-factored_2018}, and the fundamental idea of minimising a test loss by optimising a training loss frustrates the application of naïvely-powerful optimisers, the impact is to question \emph{K-FAC}'s desirability in this application. \emph{SGD Full}, \emph{Adam} and \emph{AdamQLR} all perform very similarly, showing less overfitting but being essentially indistinguishable. \emph{Adam (Untuned)}'s underperformance of \emph{AdamQLR (Untuned)} thus demonstrates that the additional heuristics of the latter improve robustness more than outright performance. Surprisingly, adding further complexity in the form of \emph{K-FAC} seems to \emph{worsen} the overfitting problem.

\subsection{SVHN}
With SVHN, we progress to a full-colour image dataset and a substantially larger ResNet-18 model, which we tune for 10 epochs and present in Figures~\ref{fig:SVHNAccuracies} (accuracies) and~\ref{fig:SVHNLosses} (losses). The periodicity in these evolutions corresponds to individual epochs, and is simply a training artefact.

On this problem, both \emph{AdamQLR} variants achieve accuracy checkpoints slightly faster than \emph{Adam}. \emph{SGD Minimal} again forms a mediocre baseline, but \emph{SGD Full} performs admirably in this setting, only slightly trailing the other algorithms' initial loss convergence. Every method overfits losses in this setting. \emph{K-FAC} generalises less well than its impressive training performance might lead us to hope.

\emph{AdamQLR}'s tuned and untuned generalisation performances are rivalled only by \emph{SGD Full} and \emph{Adam (Untuned)}, the latter over a significantly longer timescale. Interestingly, this selection of heuristics from \emph{K-FAC} again seems to bring benefits which are nullified by the full \emph{K-FAC} setting, raising further questions about the latter's underlying dynamics.

\subsection{CIFAR-10}
Finally, in a simulation of larger-scale learning, we train a ResNet-18 on CIFAR-10 over 72 epochs. Here we include conventional data augmentation of 4-pixel padding, random cropping and random left-right flipping, displaying our accuracy results in Figure~\ref{fig:CIFAR10Accuracies} and loss results in Figure~\ref{fig:CIFAR10Losses}.

\emph{Adam} is now slower to converge in both training and test accuracy, suggesting this could be an ill-suited setting in which Adam can be expected to underperform \citep{balles_dissecting_2018}. However, it achieves the highest test accuracy of any method by the end of training, with \emph{Adam (Untuned)} curiously outperforming the tuned variant. This suggests our hyperparameter selection strategy may have been particularly noisy in this setting, in which case much of the difference between algorithms here is negligible. In this trial, \emph{AdamQLR} seems to close most of the performance gap between \emph{Adam} and \emph{K-FAC}, suggesting this problem benefits more from adaptive heuristics than from second-order curvature information.

%On this task, we see much the same pattern as before: \emph{K-FAC} overfits slightly and \emph{SGD Full} provides a strong yet stable baseline, but \emph{Adam} now performs worse than previously. Overall, the dynamics of CIFAR-10 training seem to substantially differ from our other datasets --- further work to better understand this phenomenon could be very beneficial, especially since Adam is known to sometimes underperform \citep{balles_dissecting_2018}. However, our ability to reach the same training losses as other methods in far less time lends support to our technique, and suggests the lackadaisical generalisation performance may reflect the training-test learning paradigm more than any particular optimiser.

\subsection{Sensitivity Studies}
\label{sec:SensitivityStudiesSummary}

\begin{figure}
    \centering
    \begin{subfigure}{0.82\linewidth}
        \includegraphics[width=\linewidth]{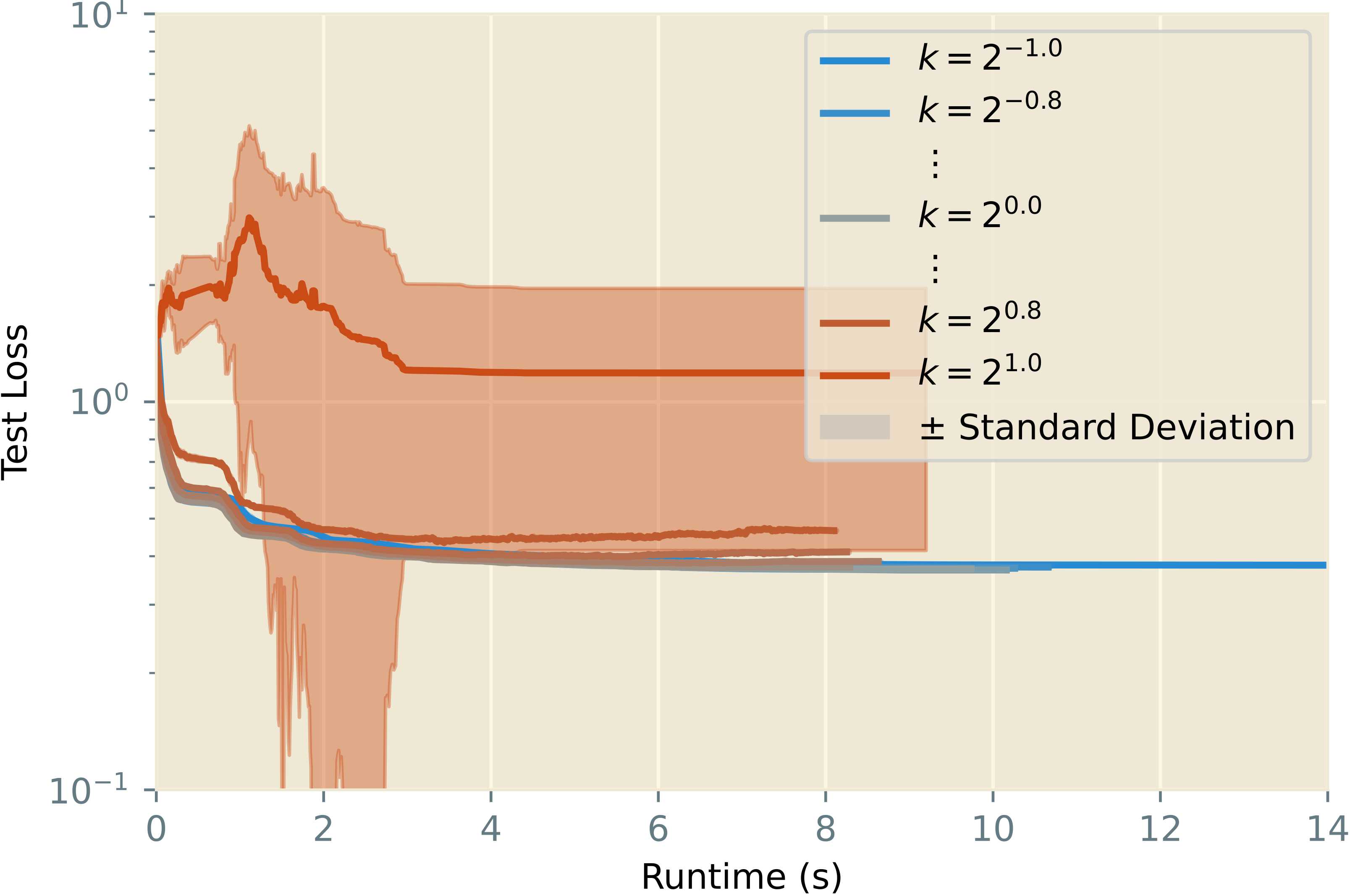}
        \caption{Learning Rate Rescaling ($\alpha \gets k\alpha$)}
        \label{fig:SensitivityLearningRate}
    \end{subfigure}

    \begin{subfigure}{0.82\linewidth}
        \includegraphics[width=\linewidth]{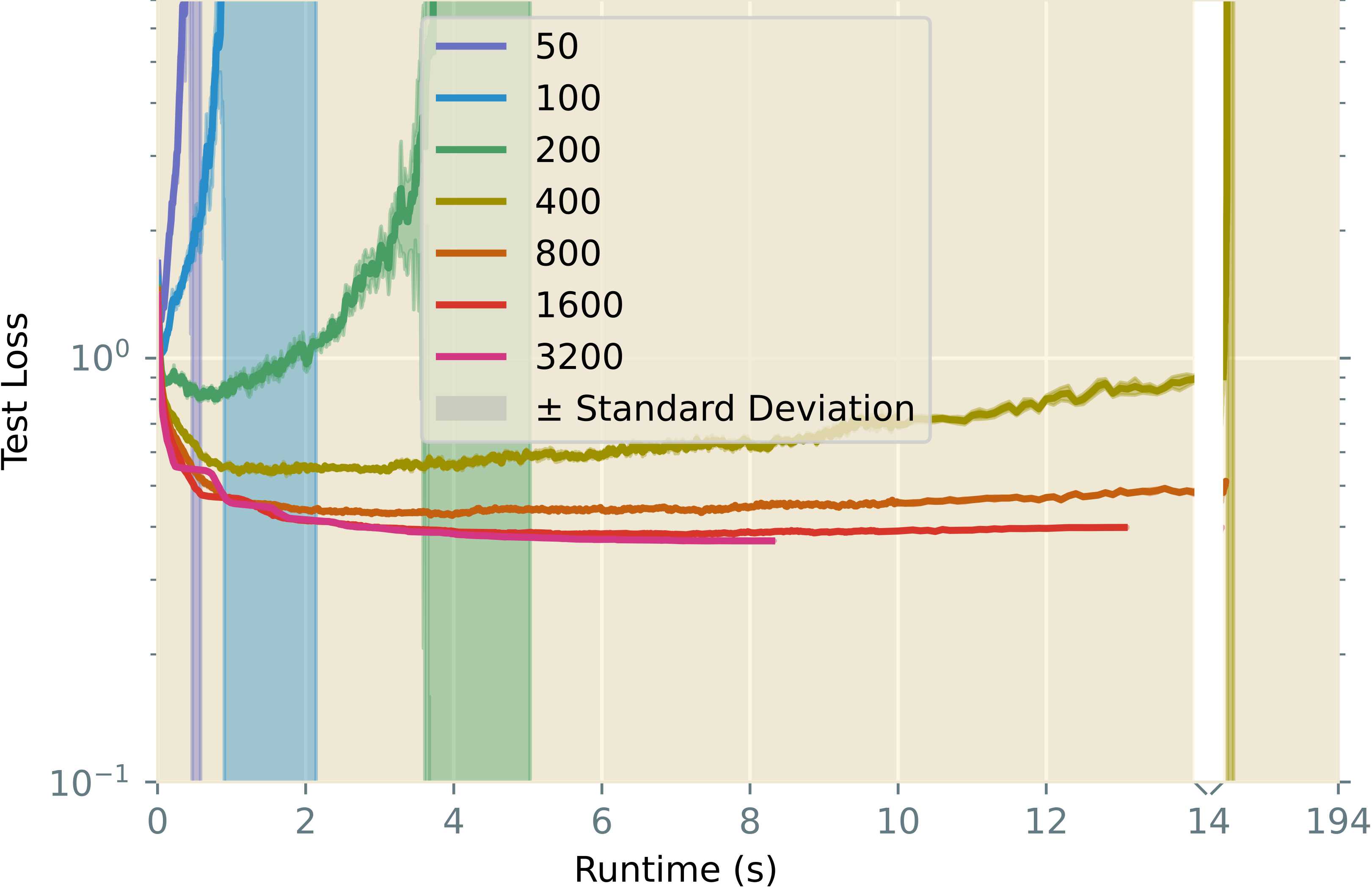}
        \caption{Batch Size}
        \label{fig:SensitivityBatchSize}
    \end{subfigure}

    \begin{subfigure}{0.82\linewidth}
        \includegraphics[width=\linewidth]{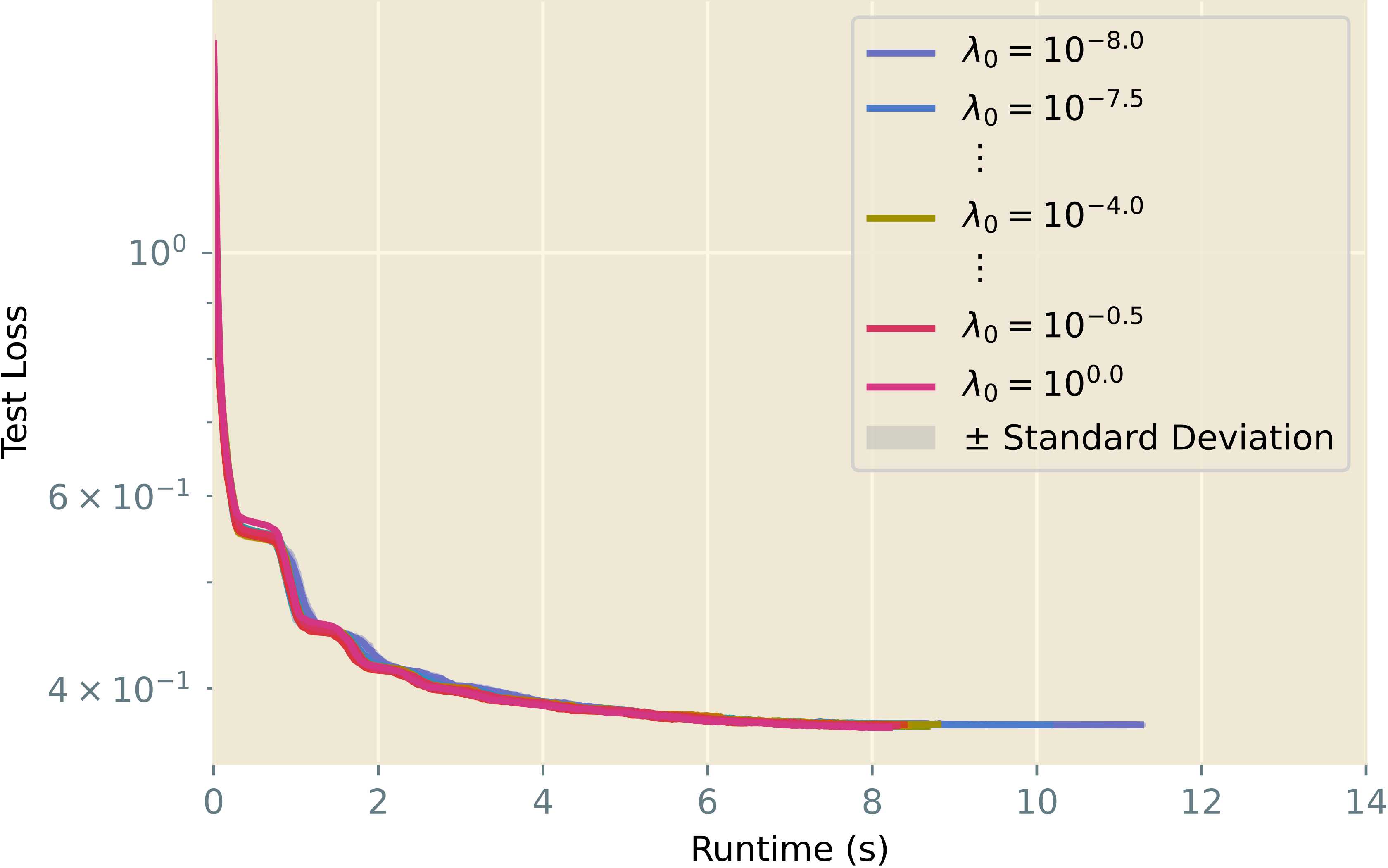}
        \caption{Initial Damping $\lambda_0$}
        \label{fig:SensitivityInitialDamping}
    \end{subfigure}
    
    \caption{Sensitivity studies for \emph{AdamQLR} on Fashion-MNIST over (a) learning rate rescaling, (b) batch size and  (c) initial damping, showing test losses.}
    \label{fig:SensitivitySummary}
\end{figure}

In Appendix~\ref{sec:SensitivityExperiments} we analyse the sensitivity of \emph{AdamQLR} on Fashion-MNIST by repeating the experiments of Section~\ref{sec:ExperimentsFashionMNIST} with a range of batch sizes, initial damping values and damping adjustment factors, and by replacing the approximately-optimal learning rate $\alpha$ from \eqref{eq:LearningRateUpdate} with the rescaled $k \alpha$, for various $k \in [0.5, 2.0]$. Figure~\ref{fig:SensitivitySummary} summarises our bootstrapped results for each intervention.

Our results inspire further confidence in AdamQLR as a correctly-configured adaptive method. Generalisation performance is optimised by choosing $k \approx 1$: constant rescaling of our proposed learning rates does not reduce test error, suggesting we adapt well to the local space and select performant update magnitudes for each direction $\vec{d}_t$ proposed by Adam. By contrast, AdamQLR is insensitive to the choice of initial damping $\lambda_0$ on this dataset, so while our ablation studies in Section~\ref{sec:DampingAblationExperiments} indicate damping is an important stabilising feature of our method, it appears the adaptive strategy of \eqref{eq:DampingUpdate} selects an appropriate damping magnitude regardless of its starting point. Finally, larger batch sizes increase generalisation performance. Since we depend implicitly on highly-parameterised curvature matrices, larger batch sizes would be expected to give a more performant average, but this also substantially decreases training time, owing to efficient GPU computation. All these results justify our \emph{AdamQLR (Untuned)} hyperparameter choices.
%Briefly, we observe best generalisation performance when $k=1$, indicating our algorithm is selecting the most effective step sizes for each update direction $\vec{d}_t$ proposed by Adam, and a general insensitivity to the initial damping $\lambda_0$, with perhaps a slight preference for larger values. We note larger batch sizes not only provide better generalisation, as we would expect from a method incorporating second-order curvature information, but also substantially reduce training time. These observations justify our use of batch size 3200 and initial damping unity in \emph{AdamQLR (Untuned)}.

\subsection{Learning Rate Evolution}
\begin{figure}
    \centering
    \newcommand{\rotatecaption}[1]{%
        \rotatebox[origin=c]{90}{\begin{minipage}{3cm}#1\end{minipage}} }
    \begin{tabularx}{0.82\linewidth}{p{2ex}X}
        \rotatecaption{\subcaption{UCI Energy}\label{fig:UCIEnergyLearningRates}}
        & \includegraphics[align=c,width=\linewidth]{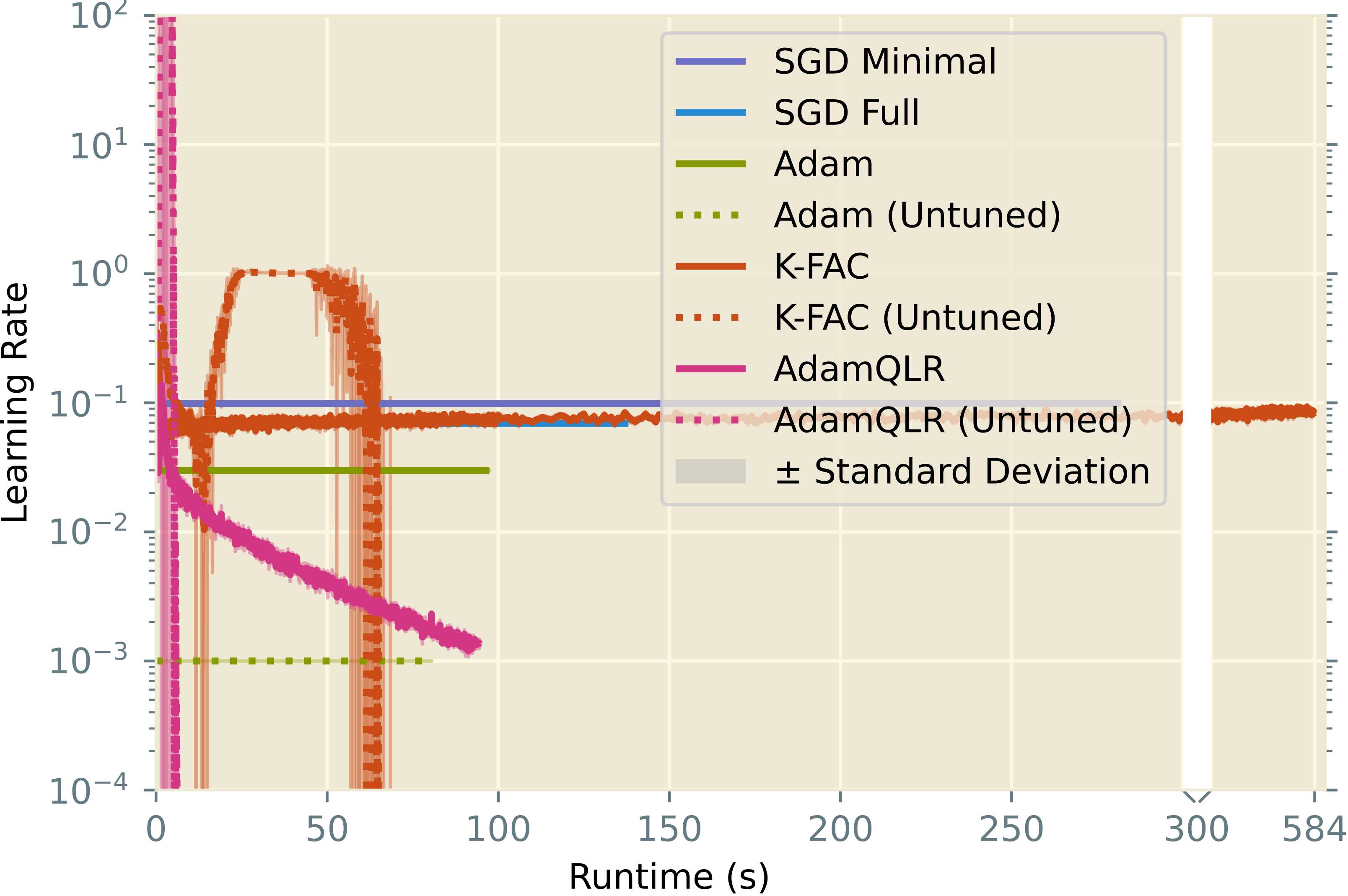} \vfill\\
        
        \rotatecaption{\subcaption{UCI Protein}\label{fig:UCIProteinLearningRates}}
        & \includegraphics[align=c,width=\linewidth]{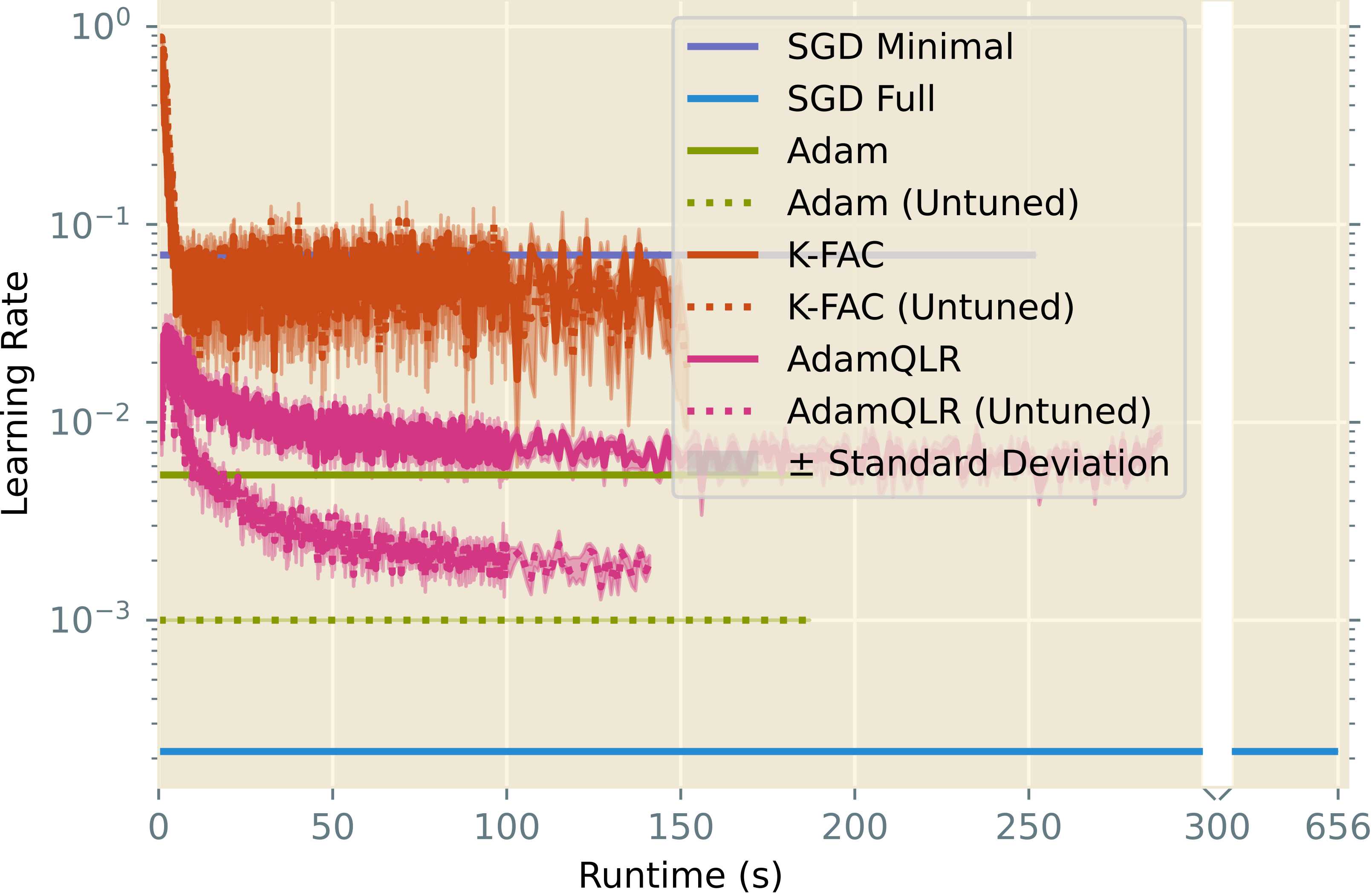} \vfill\\
        
        \rotatecaption{\subcaption{Fashion-MNIST}\label{fig:FashionMNISTLearningRates}}
        & \includegraphics[align=c,width=\linewidth]{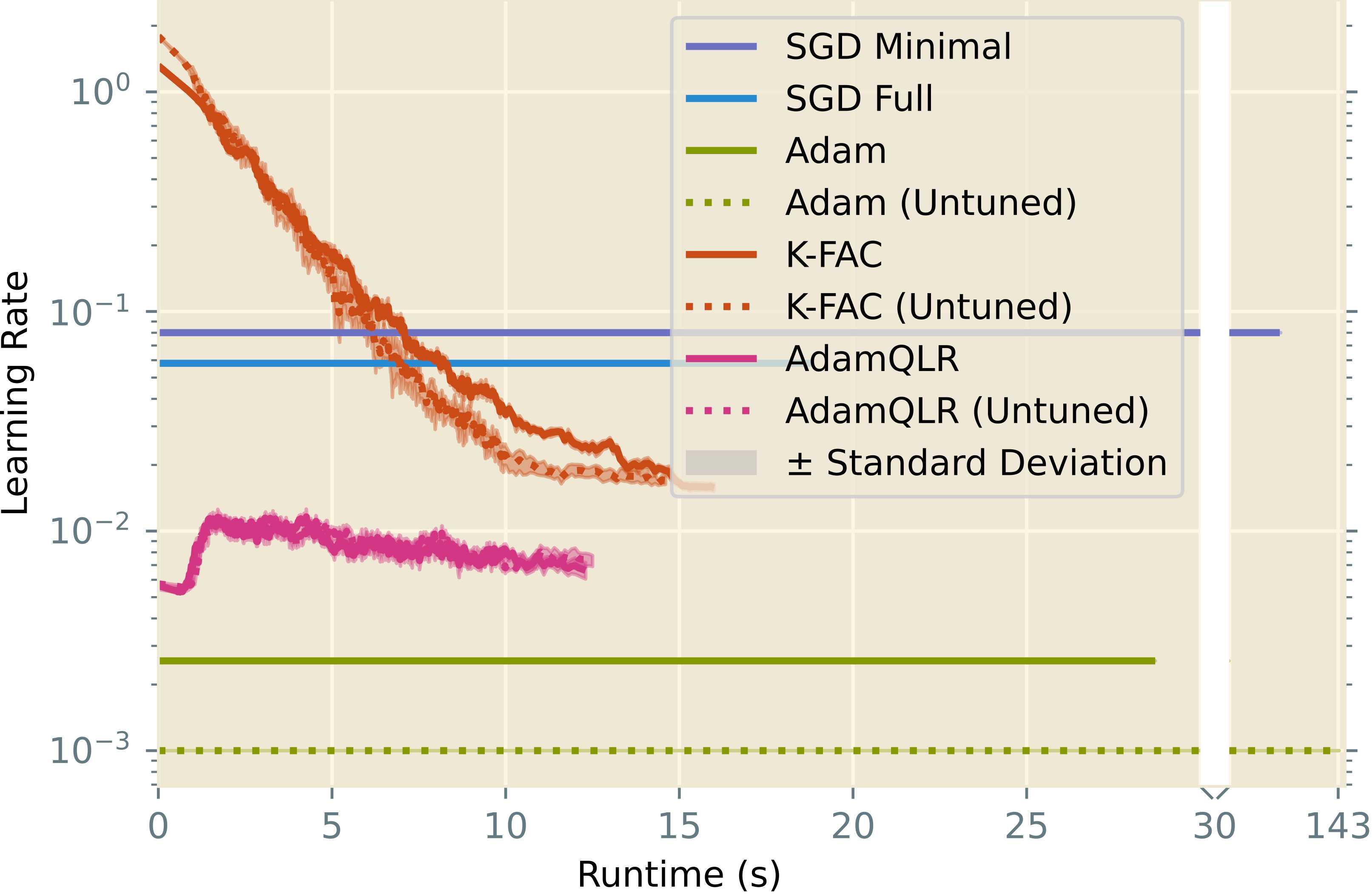} \vfill\\
        
        \rotatecaption{\subcaption{SVHN}\label{fig:SVHNLearningRates}}
        & \includegraphics[align=c,width=\linewidth]{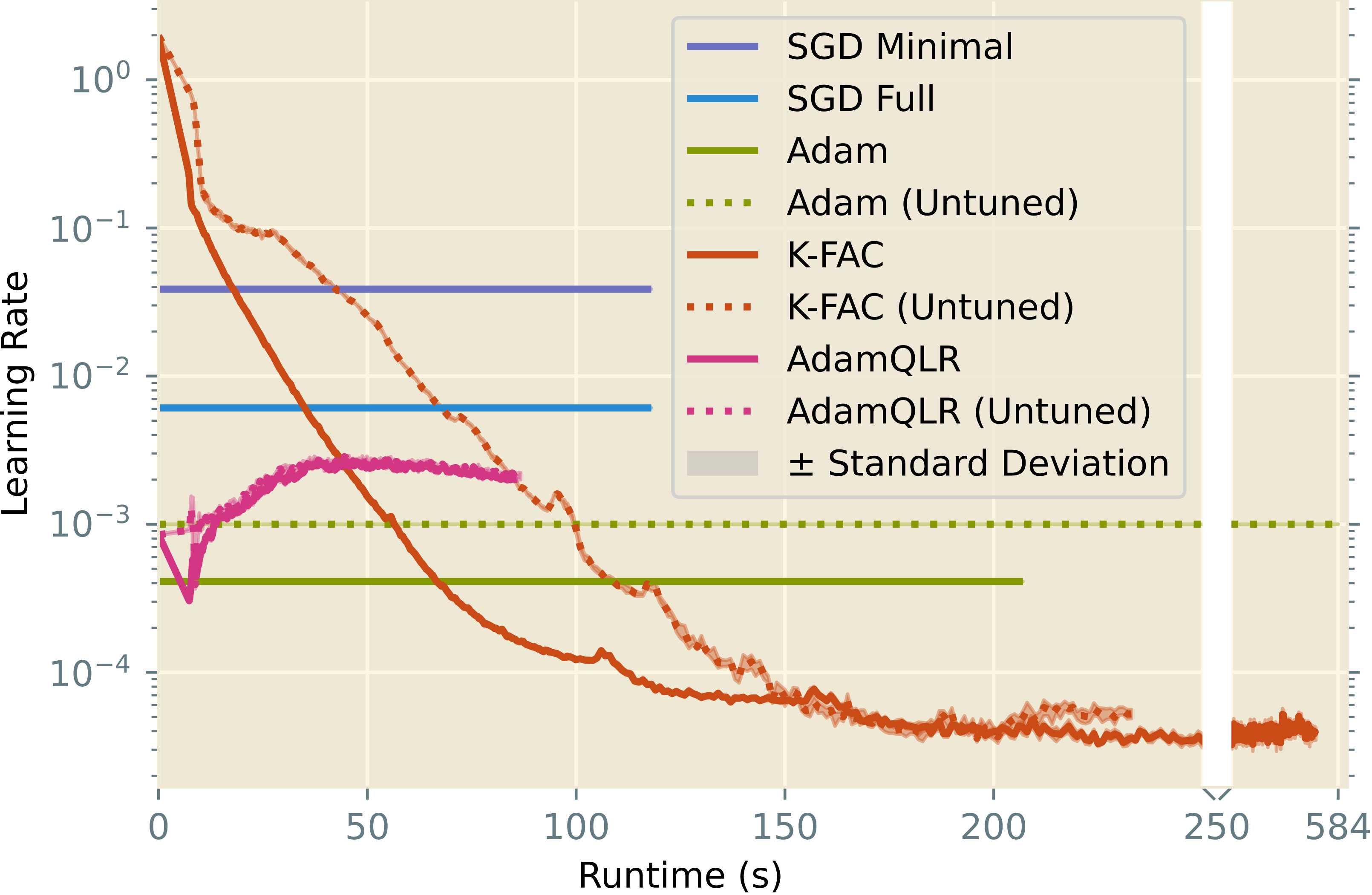} \vfill\\
        
        \rotatecaption{\subcaption{CIFAR-10}\label{fig:CIFAR10LearningRates}}
        & \includegraphics[align=c,width=\linewidth]{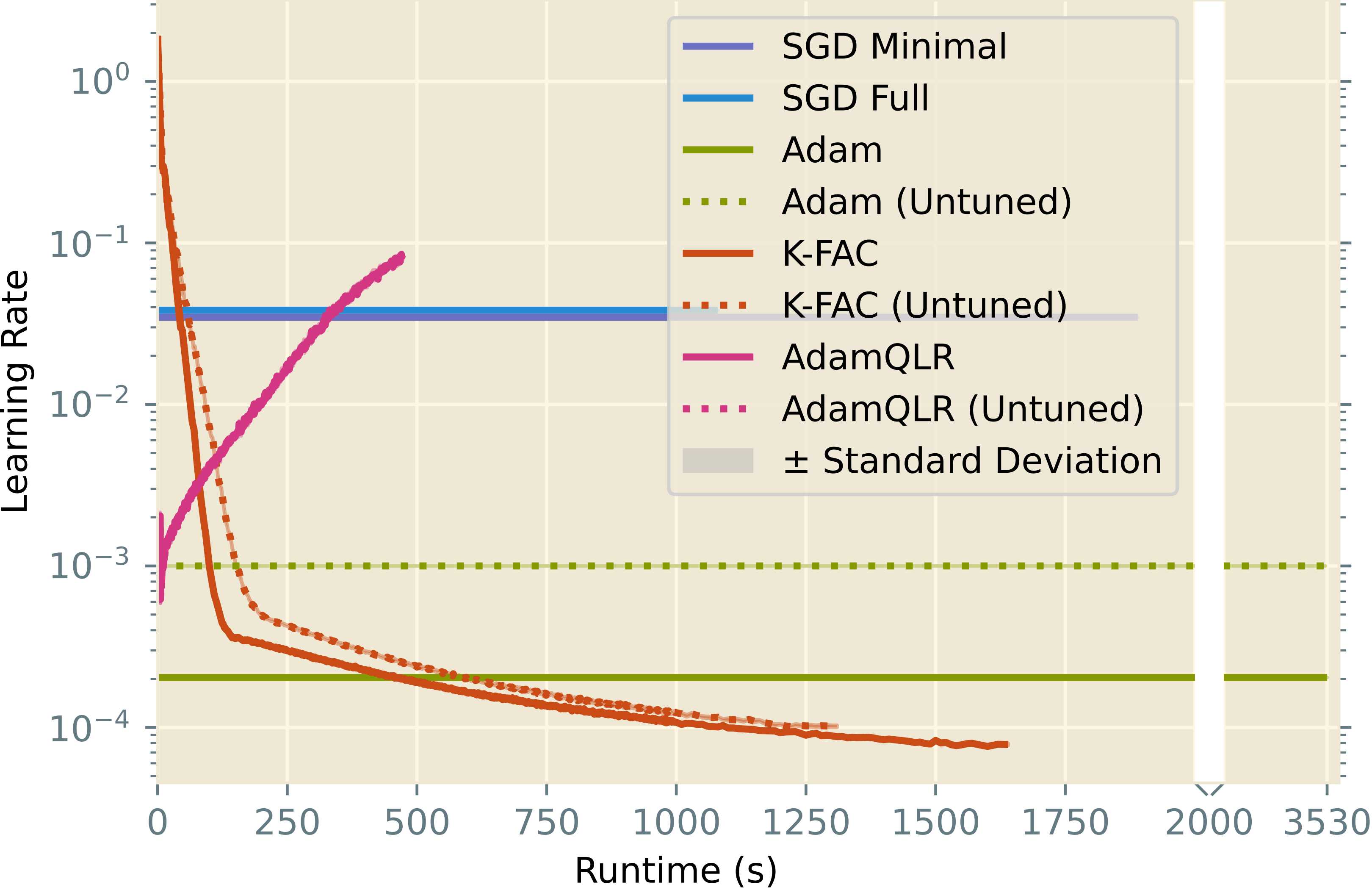} \vfill
    \end{tabularx}
    \caption{Median learning rate trajectories, bootstrap-sampled over 50 repetitions per algorithm. Hyperparameters chosen by ASHA over 200 initialisations. Note changes of scale on the time axes. See also our numerical presentation in Table~\ref{tab:EpochConstrainedFinalResults}.}
    \label{fig:AlgorithmLearningRates}
\end{figure}

In Figure~\ref{fig:AlgorithmLearningRates}, we plot the trajectories of average learning rates selected by AdamQLR and K-FAC against the fixed values used in SGD and Adam.

Learning rate schedules are widely known to be important in certain training problems, particularly at larger scales, so it is unsurprising that various algorithms' sense of the `optimal' learning rate varies over time. For the most part, the chosen schedules give an approximately exponential decay in learning rate, interestingly sometimes excluding the warm-up behaviour commonly specified in manually-designed schedules.

Broadly speaking, AdamQLR and K-FAC seem to adopt more aggressive learning rates than SGD or Adam. It is interesting to note that \emph{AdamQLR (Untuned)} chooses \emph{growing} learning rates on SVHN and CIFAR-10 which differ dramatically from those of other methods, yet achieves similar results in loss and accuracy space. In summation, these results suggest we might do well to explore other approaches to improving machine learning optimisers, beyond focussing on learning rates.

\section{Conclusion}
\label{sec:Conclusion}
In this paper we consider some of the heuristics of K-FAC by studying AdamQLR, an extension to Adam which borrows learning rate selection and adaptive damping strategies from second-order methods. Empirically, the effect of these heuristics seems to vary, being null or negative in some trials while outperforming both Adam and unablated K-FAC in others. This indicates the approach of K-FAC might be less generally applicable than that of common first-order optimisers, perhaps explaining its rare use in practice, and we think there could be great value in future work studying this phenomenon. Of tangential interest, we find an untuned version of AdamQLR, motivated by our sensitivity results, compares broadly favourably with tuned implementations of popular algorithms, while only rarely causing significant detriment to tuned-SGD or Adam baselines. Curiously, this occurs despite its use of large batch sizes conventionally understood to worsen generalisation performance.

We note challenging training-test dynamics from the CIFAR-10 results which merit further investigation, though we leave this to future work. Ultimately, we would like to better understand the workings of second-order methods like K-FAC, such that we can unify the benefits of first- and second-order optimisation to better serve the needs of the ML community, since these significantly differ from those of other optimisation practitioners. In future work, we hope to advance this line of research and better address this fundamental component of ML systems.

\clearpage
\section*{Acknowledgements}

We thank Baiyu Su for his valuable input to an earlier version of this work, which informed the rewrite presented in this paper.

% We acknowledge computation provided by the CSD3 operated by the
% University of Cambridge Research Computing Service (\url{www.csd3.cam.ac.uk}),
% provided by Dell~EMC and Intel using Tier-2 funding from the Engineering and
% Physical Sciences Research Countil (capital grant EP/P020259/1), and DiRAC
% funding from the Science and Technology Facilities Council
% (\url{www.dirac.ac.uk}).

Ross Clarke acknowledges funding from the Engineering and Physical Sciences
Research Council (project reference 2107369, grant EP/S515334/1), and support from Google DeepMind.

José Miguel Hernández-Lobato acknowledges support from a Turing AI
Fellowship under grant EP/V023756/1.

% \clearpage
\section*{Impact Statement}
Our work studies a general optimisation algorithm for neural networks, so is unlikely to influence a particular societal problem. However, increasing the effectiveness of optimisation methods makes it easier for both benevolent and malevolent actors to develop systems aligned with their goals, so this class of risk is unavoidable. Of additional concern is that the typical setting of seeking to optimise a test metric by minimising a training metric fundamentally misaligns our algorithms with our objectives, and that misalignment may cause unexpected downstream consequences if poorly understood by the model developer. Finally, it would be naïve to presume any one optimisation algorithm is a panacea for all settings, and any errant belief in this vein may cause promising research directions to be incorrectly dismissed if a supposedly `universal' optimiser happens to perform poorly on it.

\bibliographystyle{myicml2024}
\bibliography{ZoteroLibrary}

\clearpage
\appendix

\section{Notes}

\subsection{Reproducibility Statement}
We describe our algorithm fully in Section~\ref{sec:AdamQLR}, provide full source code to the reviewers and will publish this code to the community after deanonymisation. The descriptions in this paper describe all the modifications we make to Adam and provide a complete intuitive summary of our contribution, while the source code allows any fine detail of our implementation or experiments to be inspected.

\subsection{Limitations}
While we have evaluated our algorithm on a range of datasets and models, we have necessarily left many important settings untested. Thus, even though we hope our evaluation approach generalises well to other settings, we should recognise that it has not yet been tested in those settings. In particular, the learning rate selection strategy used by K-FAC and AdamQLR assumes the optimisation space is approximately convex and quadratic, which will not generally be true of machine learning problems --- this motivates our use of damping to defend against particularly ill-posed updates. With sufficient damping, we effectively define a `trust region' beyond which the surface can be non-quadratic without harming our method. Further, since Adam is known not to perform well in certain (poorly-understood) circumstances \citep{balles_dissecting_2018}, we might expect AdamQLR to have difficulty with the same class of problems.

\subsection{Reduction Ratio}
\label{sec:ReductionRatioCommentary}

Here, we give a more verbose commentary on the damping adjustment mechanism described in \eqref{eq:DampingUpdate}.

The definition of the reduction ratio $\rho$ is intuitive. When we update the model parameters from $\vec{\theta}_{t-1}$ to $\vec{\theta}_{t}$, we will observe some change in the loss metric $f(\vec{\theta}_{t}) - f(\vec{\theta}_{t-1})$. Similarly, our quadratic model $M$ will have proposed this parameter update predicting the loss will change by $M(\vec{\theta}_t) - M(\vec{\theta}_{t-1})$. Ideally, we would like our model to be a good fit for the true optimisation surface, in which case the observed and predicted changes will be similar, and we will find $\rho \approx 1$. Conversely, if the fit is poor, the observation and prediction will be very different, giving $\rho < 1$ if the observed change is much smaller than the model predicted, or $\rho > 1$ if the change is much larger than the model predicted.

If we find the fit of $M$ to be poor, we would like to adjust the damping to help rectify the situation, since a larger damping will generally bias the model towards expecting larger loss changes, thus proposing smaller parameter updates. Broadly speaking, $\rho > 1$ suggests the model is being too conservative, and we would benefit from decreasing damping to better reflect the underlying surface. Conversely, $\rho < 1$ suggests the model expects much more dramatic changes than we actually see, so we should increase damping to `reign in' the predictive behaviour.

As \citet{martens_optimizing_2015} note in the original presentation of K-FAC, the optimisation dynamics will change during training. In particular, as we approach a local minimum, the loss surface becomes more and more quadratic-like. Under these circumstances, damping slows down convergence by reducing parameter update sizes, without achieving any appreciable benefit. Even away from local minima, damping tends to trade convergence speed for stability. In both cases, there is a natural incentive to be biased towards decreasing damping if at all possible.

In this work, we retain \citet{martens_optimizing_2015}'s damping adjustment thresholds of $\rho > \nicefrac{3}{4}$ and $\rho < \nicefrac{1}{4}$, since these choices led to desirable performance from K-FAC. \Citeauthor{martens_optimizing_2015} articulate their preference for reducing $\lambda$ if possible, and we can understand their chosen thresholds in that light. It is for this reason that the thresholds are not centred about $\rho = 1$, as might have been our intuitive expectation.

\subsection{Convergence of AdamQLR}
\label{sec:ConvergenceOfAdamQLR}
While we have not studied the convergence of AdamQLR analytically, we believe it will converge under appropriate conditions by the following intuitive argument.

Let the update direction $\vec{d}_t$ proposed by Adam be arbitrary. By construction, AdamQLR selects the (possibly negative) learning rate $\alpha$ which minimises the value of some quadratic model $M(\vec{\theta})$ at the new parameters $\vec{\theta}_{t-1} - \alpha \vec{d}_t$, so we certainly have $M(\vec{\theta}_{t-1} - \alpha \vec{d}_t) \leq M(\vec{\theta}_{t-1})$. Whether this translates to non-increase of the objective function $f(\vec{\theta})$ depends on the quality of the approximation $M(\vec{\theta})$. For any given $\alpha$, some Lipschitz smoothness condition exists which guarantees $f(\vec{\theta}_{t-1} - \alpha \vec{d}_t) \leq f(\vec{\theta}_{t-1})$, so we should be able to guarantee convergence for a sufficiently smooth $f(\vec{\theta})$ if $\alpha$ is clipped to some maximum value. This intuition transfers to the unclipped case if we argue that, in practice, $\alpha$ will generally take values within some finite range.

\citet{shi_rmsprop_2020} prove the convergence of full-batch RMSprop for a diminishing learning rate schedule. While the more arbitrary adaptivity of AdamQLR's learning rate prevents a direct application of this proof, it does lend confidence that AdamQLR should behave as expected in common practical circumstances.

\subsection{Choice of Baselines}
\label{sec:ChoiceOfBaselines}

We now give some explanatory notes justifying our choice of baseline algorithms in Section~\ref{sec:Experiments}.

Weight decay was only included in \emph{SGD Full}, despite being theoretically applicable to any algorithm. To clarify the comparison, we studied vanilla versions of SGD, Adam and K-FAC as the most natural baselines for AdamQLR. Since SGD is very commonly used with momentum and weight decay, we include it in \emph{Minimal} and \emph{Full} forms, where the latter includes these additional components. \emph{SGD Full} is included primarily for background context, so its unique use of weight decay does not affect the comparability of the other algorithms. Moreover, since SGD typically underperformed other algorithms in our experiments, any advantage due to weight decay does not affect our results.

Similarly, the use of momentum in \emph{SGD Full} is primarily to provide background context from a common algorithm. \emph{Adam} contains momentum-like behaviour by construction, with the corresponding hyperparameters not usually tuned in practice (a convention we follow), and \emph{K-FAC} uses an adaptive momentum coefficient by default. \emph{AdamQLR} does not incorporate the adaptive momentum of \emph{K-FAC}, so momentum is only used by the inner Adam procedure to select the unscaled update direction, with the magnitude of each update computed independently at each iteration. Thus, any effect from the absence of momentum in AdamQLR is disadvantageous for this algorithm, so our conclusions are not invalidated.

\subsection{Hyperparameter Search Space}
We use similar hyperparameter search spaces (with unused hyperparameters removed) for each dataset and algorithm combination. These are detailed in Table~\ref{tab:HyperparameterTuning}.

\begin{table*}
    \caption{Hyperparameter search spaces for Section~\ref{sec:Experiments}}
    \label{tab:HyperparameterTuning}
    \centering
    \begin{tabular}{rl}
        \toprule
        Hyperparameter & Search Range\\
        \midrule
        Batch Size & Uniform in $\{50, 100, 200, 400, 800, 1\,600, 3\,200\}$ \\
        \multirow{2}{*}{Learning Rate $\alpha$} & \emph{SGD}: Logarithmic in $[10^{-6}, 10^{-1}]$ \\
        & \emph{Adam}: Logarithmic in $[10^{-6}, 1]$ \\
        Momentum & Logarithmic in $[10^{-4}, 0.3]$, subtracted from 1 \\
        Weight Decay & Logarithmic in $[10^{-10}, 1]$ \\
        Initial Damping $\lambda_0$ & Logarithmic in $[10^{-8}, 1]$ \\
        Damping Decrease Factor $\omega_\text{dec}$ & Logarithmic in $[0.5, 1.0]$ \\
        Damping Increase Factor $\omega_\text{inc}$ & Logarithmic in $[1.0, 4.0]$ \\
         \bottomrule
    \end{tabular}
\end{table*}

\subsection{Chosen Hyperparameters}
The best hyperparameters selected by ASHA for each setting considered in this work are indicated in Table~\ref{tab:Hyperparameters}.

\begin{table*}
    \centering
    \caption{Optimal hyperparameters used to produce the results of Section~\ref{sec:Experiments}, Appendix~\ref{sec:PennTreebank} and Appendix~\ref{sec:AblationExperiments}}
    \label{tab:Hyperparameters}
    \resizebox{0.85\linewidth}{!}{
    \begin{tabular}{cc
            S[table-format=4]
            S[table-format=1.3e-1]
            S[table-format=1.3]
            S[table-format=1.3e-1]
            S[table-format=1.3e-1]
            S[table-format=1.1]
            S[table-format=1.1]
            }
        \toprule
        Dataset & Algorithm & {\makecell{Batch\\ Size}} & {\makecell{Learning\\ Rate}} & {Momentum} & {\makecell{Weight\\ Decay}} & {\makecell{Initial\\ Damping}} & {\makecell{Damping\\ Decrease\\ Factor}} & {\makecell{Damping\\ Increase\\ Factor}}\\
        \midrule
        \multirow{8}{*}{Rosenbrock}
& GD Minimal 	& {---} 	& {---} 	& {---} 	& {---} 	& {---} 	&{---} 	& {---} \\
& GD Full 	& {---} 	& {---} 	& {---} 	& {---} 	& {---} 	&{---} 	& {---} \\
& Adam 	& {---} 	& 9.88e-02 	& {---} 	& {---} 	& {---} 	&{---} 	& {---} \\
& Adam (Untuned) 	& {---} 	& {---} 	& {---} 	& {---} 	& {---} 	&{---} 	& {---} \\
& AdamQLR (Tuned, Hessian) 	& {---} 	& {---} 	& {---} 	& {---} 	& 3.03e-06 	& 0.9 	& 2.1 \\
& AdamQLR (Hessian) 	& {---} 	& {---} 	& {---} 	& {---} 	& 4.14e-01 	& 0.7 	& 1.1 \\
& AdamQLR (Untuned, Hessian) 	& {---} 	& {---} 	& {---} 	& {---} 	& 1.00e+00 	& 0.9 	& 1.1 \\
& AdamQLR (Untuned) 	& {---} 	& {---} 	& {---} 	& {---} 	& 1.00e-03 	& 0.5 	& 2.0 \\
\midrule
\multirow{8}{*}{UCI Energy}
& SGD Minimal 	& 100 	& 9.88e-02 	& {---} 	& {---} 	& {---} 	&{---} 	& {---} \\
& SGD Full 	& 400 	& 6.92e-02 	& 0.996 	& 1.29e-04 	& {---} 	&{---} 	& {---} \\
& Adam 	& 800 	& 2.99e-02 	& {---} 	& {---} 	& {---} 	&{---} 	& {---} \\
& Adam (Untuned) 	& 588 	& {---} 	& {---} 	& {---} 	& {---} 	&{---} 	& {---} \\
& K-FAC 	& 50 	& {---} 	& {---} 	& {---} 	& 1.00e-02 	& {---} 	& {---} \\
& K-FAC (Untuned) 	& 3200 	& {---} 	& {---} 	& {---} 	& {---} 	&{---} 	& {---} \\
& AdamQLR 	& 800 	& {---} 	& {---} 	& {---} 	& 5.39e-03 	& 0.6 	& 1.3 \\
& AdamQLR (Untuned) 	& 3200 	& {---} 	& {---} 	& {---} 	& {---} 	&0.5 	& 2.0 \\
\midrule
\multirow{8}{*}{UCI Protein}
& SGD Minimal 	& 400 	& 7.00e-02 	& {---} 	& {---} 	& {---} 	&{---} 	& {---} \\
& SGD Full 	& 100 	& 2.17e-04 	& 0.997 	& 1.54e-08 	& {---} 	&{---} 	& {---} \\
& Adam 	& 800 	& 5.42e-03 	& {---} 	& {---} 	& {---} 	&{---} 	& {---} \\
& Adam (Untuned) 	& 1000 	& {---} 	& {---} 	& {---} 	& {---} 	&{---} 	& {---} \\
& K-FAC 	& 3200 	& {---} 	& {---} 	& {---} 	& 2.11e-01 	& {---} 	& {---} \\
& K-FAC (Untuned) 	& 3200 	& {---} 	& {---} 	& {---} 	& {---} 	&{---} 	& {---} \\
& AdamQLR 	& 400 	& {---} 	& {---} 	& {---} 	& 1.49e-04 	& 0.6 	& 1.7 \\
& AdamQLR (Untuned) 	& 3200 	& {---} 	& {---} 	& {---} 	& {---} 	&0.5 	& 2.0 \\
\midrule
\multirow{11}{*}{Fashion-MNIST}
& SGD Minimal 	& 100 	& 8.01e-02 	& {---} 	& {---} 	& {---} 	&{---} 	& {---} \\
& SGD Full 	& 800 	& 5.81e-02 	& 0.929 	& 1.65e-08 	& {---} 	&{---} 	& {---} \\
& Adam 	& 400 	& 2.56e-03 	& {---} 	& {---} 	& {---} 	&{---} 	& {---} \\
& Adam (Untuned) 	& 50 	& {---} 	& {---} 	& {---} 	& {---} 	&{---} 	& {---} \\
& K-FAC 	& 3200 	& {---} 	& {---} 	& {---} 	& 1.92e-01 	& {---} 	& {---} \\
& K-FAC (Untuned) 	& 3200 	& {---} 	& {---} 	& {---} 	& {---} 	&{---} 	& {---} \\
& AdamQLR (Hessian) 	& 3200 	& {---} 	& {---} 	& {---} 	& 3.04e-02 	& 0.8 	& 1.9 \\
& AdamQLR (Undamped) 	& 3200 	& {---} 	& {---} 	& {---} 	& {---} 	&{---} 	& {---} \\
& AdamQLR 	& 3200 	& {---} 	& {---} 	& {---} 	& 1.17e-05 	& 0.7 	& 1.8 \\
& AdamQLR (Untuned) 	& 3200 	& {---} 	& {---} 	& {---} 	& {---} 	&0.5 	& 2.0 \\
& AdamQLR (Fisher) 	& 3200 	& {---} 	& {---} 	& {---} 	& 1.17e-05 	& 0.7 	& 1.8 \\
\midrule
\multirow{8}{*}{SVHN}
& SGD Minimal 	& 1600 	& 3.86e-02 	& {---} 	& {---} 	& {---} 	&{---} 	& {---} \\
& SGD Full 	& 1600 	& 6.10e-03 	& 0.986 	& 8.61e-07 	& {---} 	&{---} 	& {---} \\
& Adam 	& 800 	& 4.10e-04 	& {---} 	& {---} 	& {---} 	&{---} 	& {---} \\
& Adam (Untuned) 	& 256 	& {---} 	& {---} 	& {---} 	& {---} 	&{---} 	& {---} \\
& K-FAC 	& 800 	& {---} 	& {---} 	& {---} 	& 6.40e-01 	& {---} 	& {---} \\
& K-FAC (Untuned) 	& 3200 	& {---} 	& {---} 	& {---} 	& {---} 	&{---} 	& {---} \\
& AdamQLR 	& 3200 	& {---} 	& {---} 	& {---} 	& 4.73e-01 	& 0.6 	& 1.1 \\
& AdamQLR (Untuned) 	& 3200 	& {---} 	& {---} 	& {---} 	& {---} 	&0.5 	& 2.0 \\
\midrule
\multirow{11}{*}{CIFAR-10}
& SGD Minimal 	& 200 	& 3.47e-02 	& {---} 	& {---} 	& {---} 	&{---} 	& {---} \\
& SGD Full 	& 400 	& 3.83e-02 	& 0.920 	& 8.74e-04 	& {---} 	&{---} 	& {---} \\
& Adam 	& 100 	& 2.04e-04 	& {---} 	& {---} 	& {---} 	&{---} 	& {---} \\
& Adam (Untuned) 	& 128 	& {---} 	& {---} 	& {---} 	& {---} 	&{---} 	& {---} \\
& K-FAC 	& 1600 	& {---} 	& {---} 	& {---} 	& 9.03e-01 	& {---} 	& {---} \\
& K-FAC (Untuned) 	& 3200 	& {---} 	& {---} 	& {---} 	& {---} 	&{---} 	& {---} \\
& AdamQLR (Hessian) 	& 3200 	& {---} 	& {---} 	& {---} 	& 6.00e-01 	& 0.9 	& 3.7 \\
& AdamQLR (Undamped) 	& 1600 	& {---} 	& {---} 	& {---} 	& {---} 	&{---} 	& {---} \\
& AdamQLR 	& 3200 	& {---} 	& {---} 	& {---} 	& 2.99e-05 	& 0.7 	& 1.3 \\
& AdamQLR (Untuned) 	& 3200 	& {---} 	& {---} 	& {---} 	& {---} 	&0.5 	& 2.0 \\
& AdamQLR (Fisher) 	& 3200 	& {---} 	& {---} 	& {---} 	& 2.99e-05 	& 0.7 	& 1.3 \\
\bottomrule
    \end{tabular}
    }
\end{table*}

\subsection{Compute Used}
\begin{table*}
  \centering
  \caption{System configurations used to run our experiments.}
  \label{tab:ExperimentalHardware}
  \begin{tabular}{ccccccc}
    \toprule
    Type & CPU & GPU (NVIDIA) & Python & JAX & CUDA & cuDNN \\
    \midrule
    Consumer Desktop & Intel Core i7-3930K & RTX 2080GTX  & 3.10.11 & 0.3.25 & 11.4 & 8.05\\
    Local Cluster & Intel Core i9-10900X & RTX 2080GTX & 3.10.11 & 0.3.25 & 11.8 & 8.05\\
    \bottomrule
  \end{tabular}
\end{table*}

Our experiments were performed on one of the two sets of hardware shown in Table~\ref{tab:ExperimentalHardware}. All runtime comparisons were performed on like-for-like hardware. We make use of GPU acceleration throughout, with the JAX \citep{bradbury_jax_2018}, Haiku \citep{hennigan_haiku_2020} and KFAC-JAX \citep{botev_kfac-jax_2022} libraries, along with various related components of the DeepMind JAX Ecosystem \citep{babuschkin_deepmind_2020}.

%Producing experimental data for every plot in this paper required approximately 228.3 GPU-hours on the Local Cluster and 9.5 GPU-hours on the Consumer Desktop. This accounts for performing multiple trials in parallel on the same GPU where capacity exists and for hyperparameter search, but excludes development, debugging and unit testing time, which would substantially increase these figures.

\subsection{Datasets}
The datasets we use are all standard in the ML literature; we outline their usage conditions in Table~\ref{tab:DatasetLicences}.

\begin{table*}
    \centering
    \caption{Licences under which we use datasets in this work}
    \label{tab:DatasetLicences}
    \resizebox{\linewidth}{!}{
    \begin{tabular}{cccccS[table-format=5]}
        \toprule
        Dataset & Licence & Source & Input & Output & {Total Size} \\
        \midrule
        UCI Energy & \makecell{Creative Commons Attribution 4.0\\International (CC BY 4.0)}   & \makecell{\citet{tsanas_accurate_2012};\\ \citet{gal_dropout_2016}} & $8$-Vector & Scalar & 692 \\
        UCI Protein & None specified & \makecell{\citet{rana_uci_2013};\\ \citet{gal_dropout_2016}} & $9$-Vector & Scalar & 45730\\
        Fashion-MNIST & MIT & \citet{xiao_fashion-mnist_2017} & $28 \times 28$ Image & Class (from 10) & 60000\\
        CIFAR-10 & None specified & \citet{krizhevsky_learning_2009} & $32 \times 32$ Image & Class (from 10) & 60000\\
        SVHN & None specified & \citet{netzer_reading_2011} & $32 \times 32$ Image & Class (from 10) & 99289\\
        \bottomrule
    \end{tabular}
    }
\end{table*}

\section{Additional Experiments}

\subsection{Algorithm Comparisons}
In this Section, we provide some additional viewpoints into our main results of Section~\ref{sec:Experiments}.

\subsubsection{Fashion-MNIST, SVHN and CIFAR-10 Losses}
\label{sec:LossResults}
\begin{figure*}
    \centering
    \newcommand{\rotatecaption}[1]{%
        \rotatebox[origin=c]{90}{\begin{minipage}{3cm}#1\end{minipage}} }
    \begin{tabularx}{0.82\linewidth}{p{2ex}XX}
        \rotatecaption{\subcaption{Fashion-MNIST}\label{fig:FashionMNISTLosses}}
        & \includegraphics[align=c,width=\linewidth]{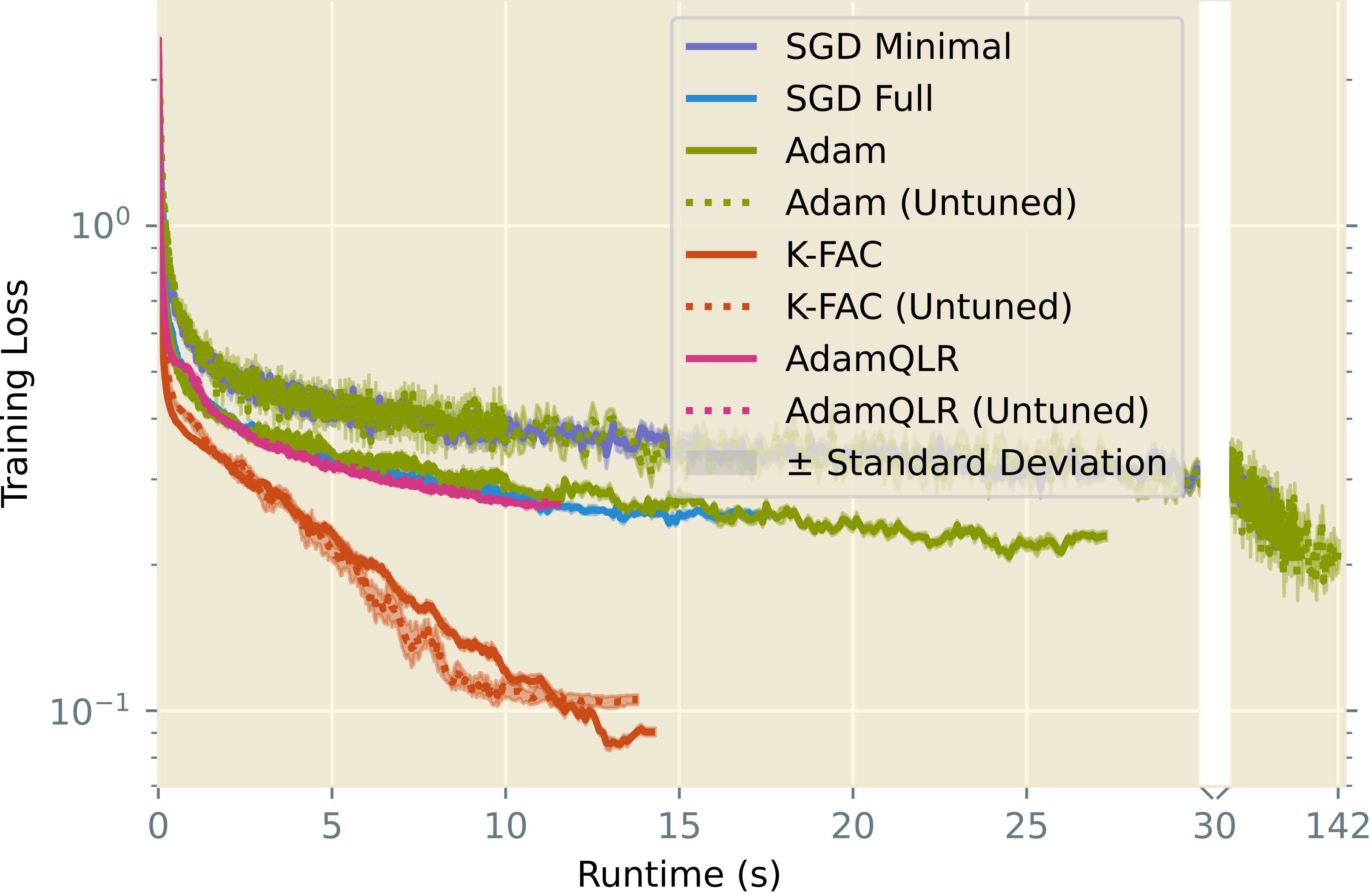}
        & \includegraphics[align=c,width=\linewidth]{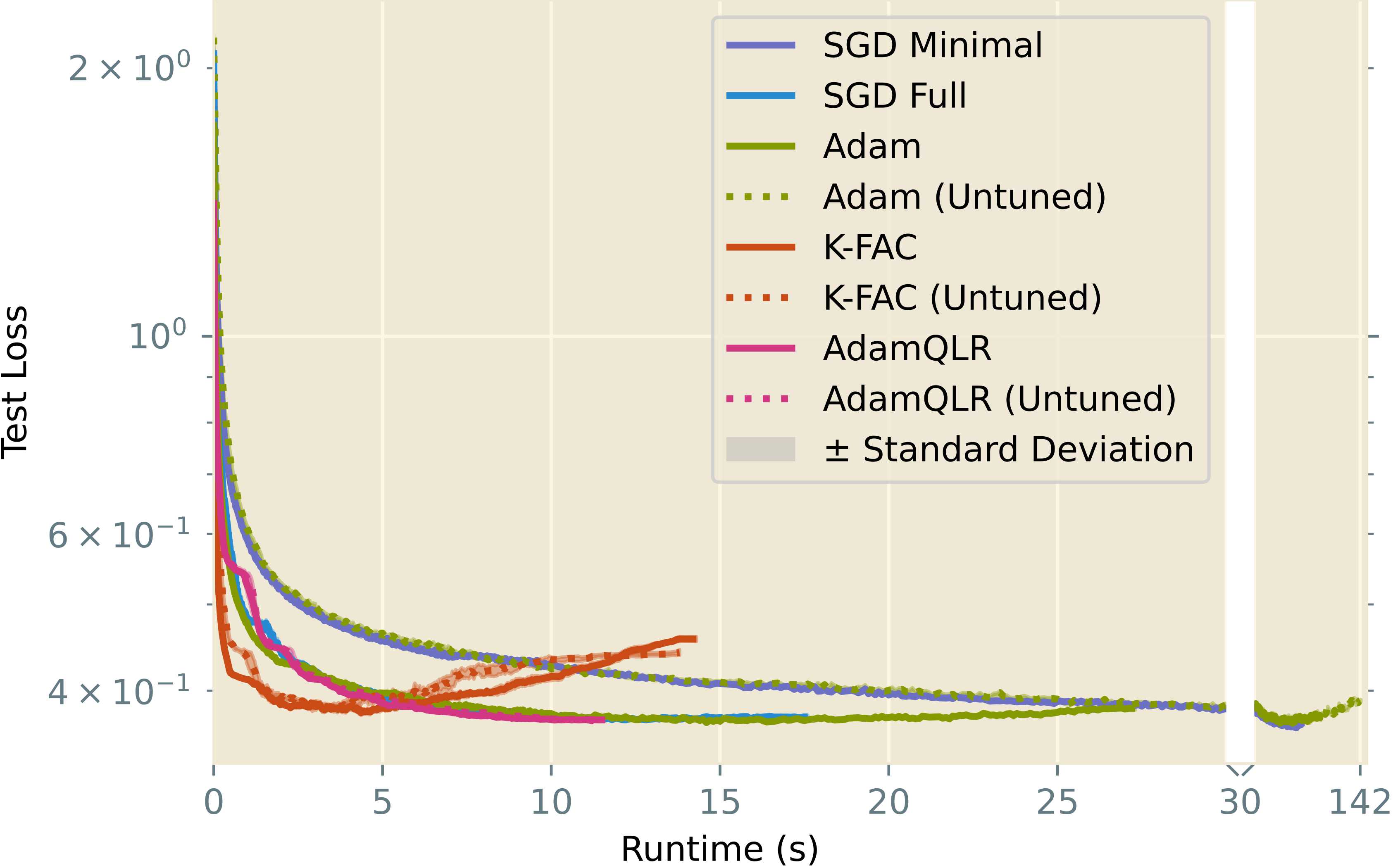} \vfill\\
        
        \rotatecaption{\subcaption{SVHN}\label{fig:SVHNLosses}}
        & \includegraphics[align=c,width=\linewidth]{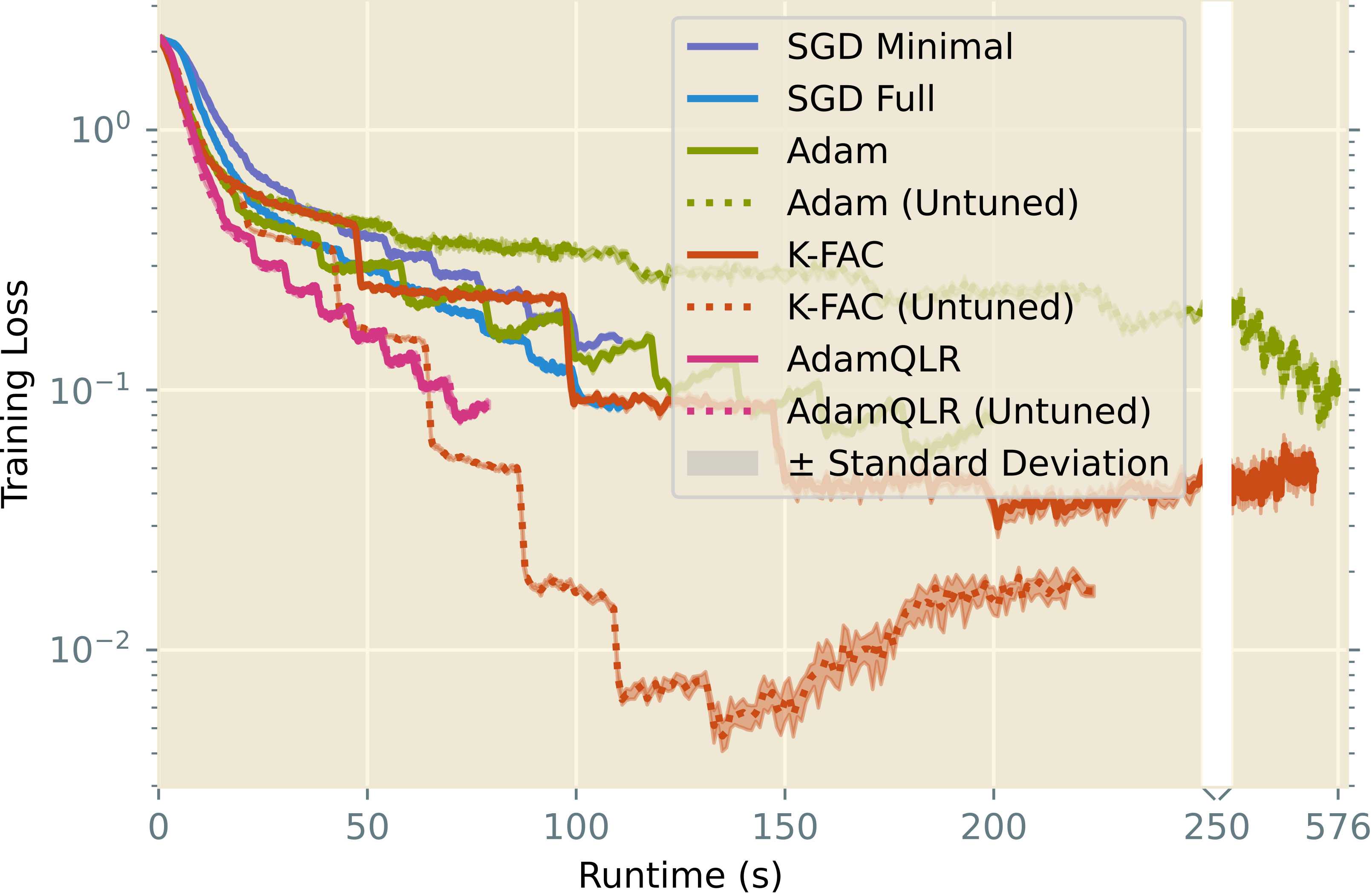}
        & \includegraphics[align=c,width=\linewidth]{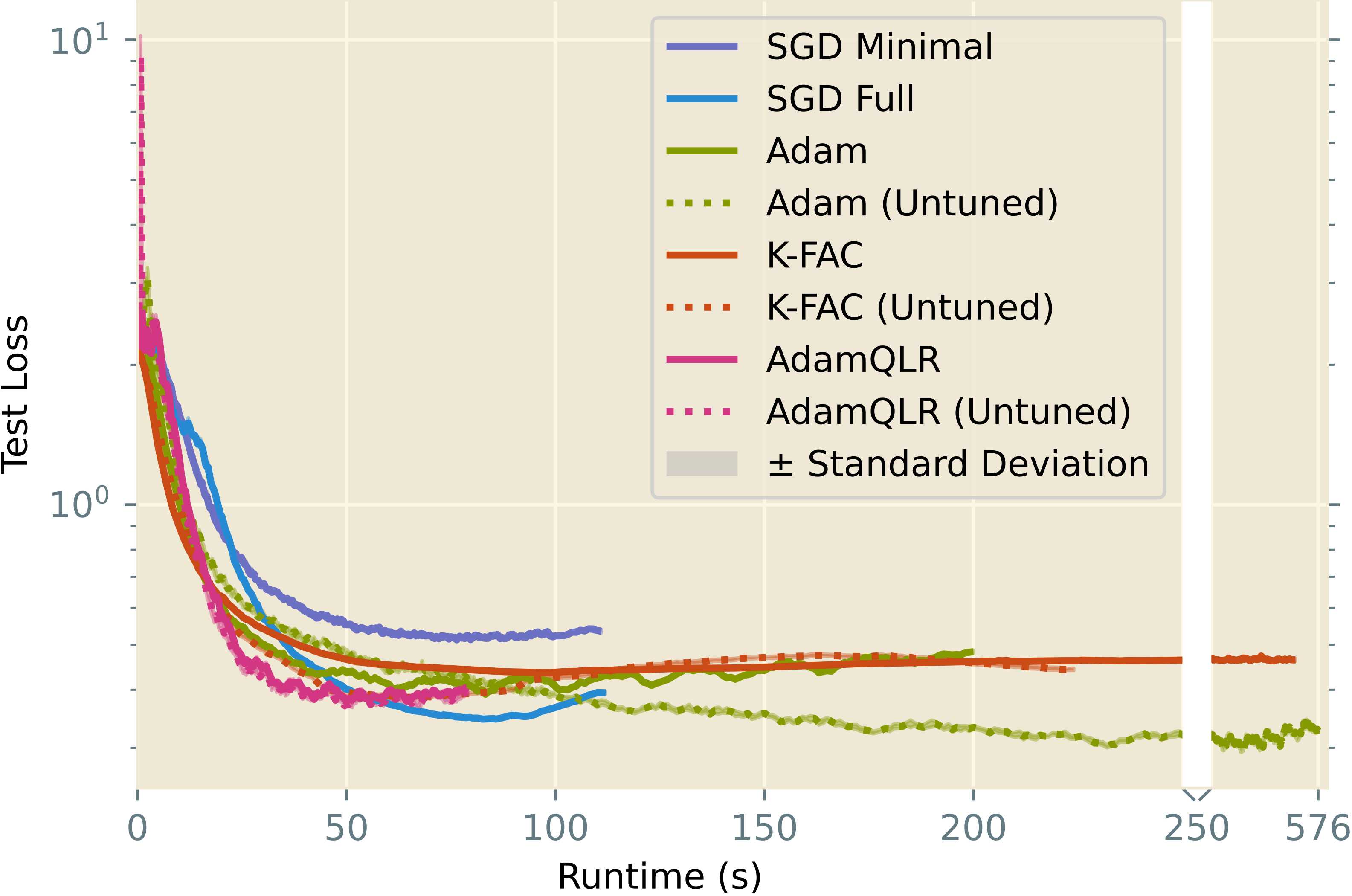} \vfill\\
        
        \rotatecaption{\subcaption{CIFAR-10}\label{fig:CIFAR10Losses}}
        & \includegraphics[align=c,width=\linewidth]{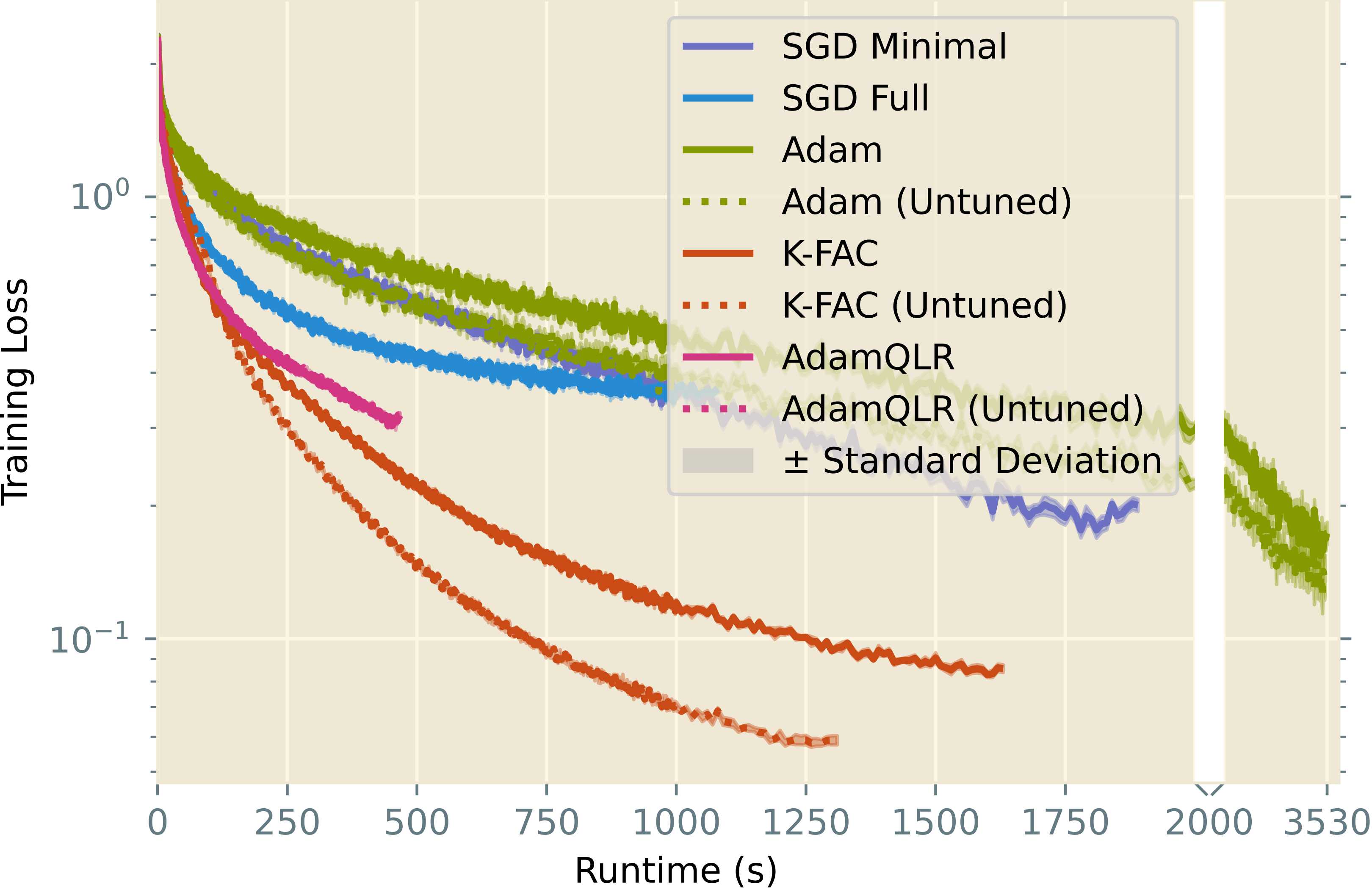}
        & \includegraphics[align=c,width=\linewidth]{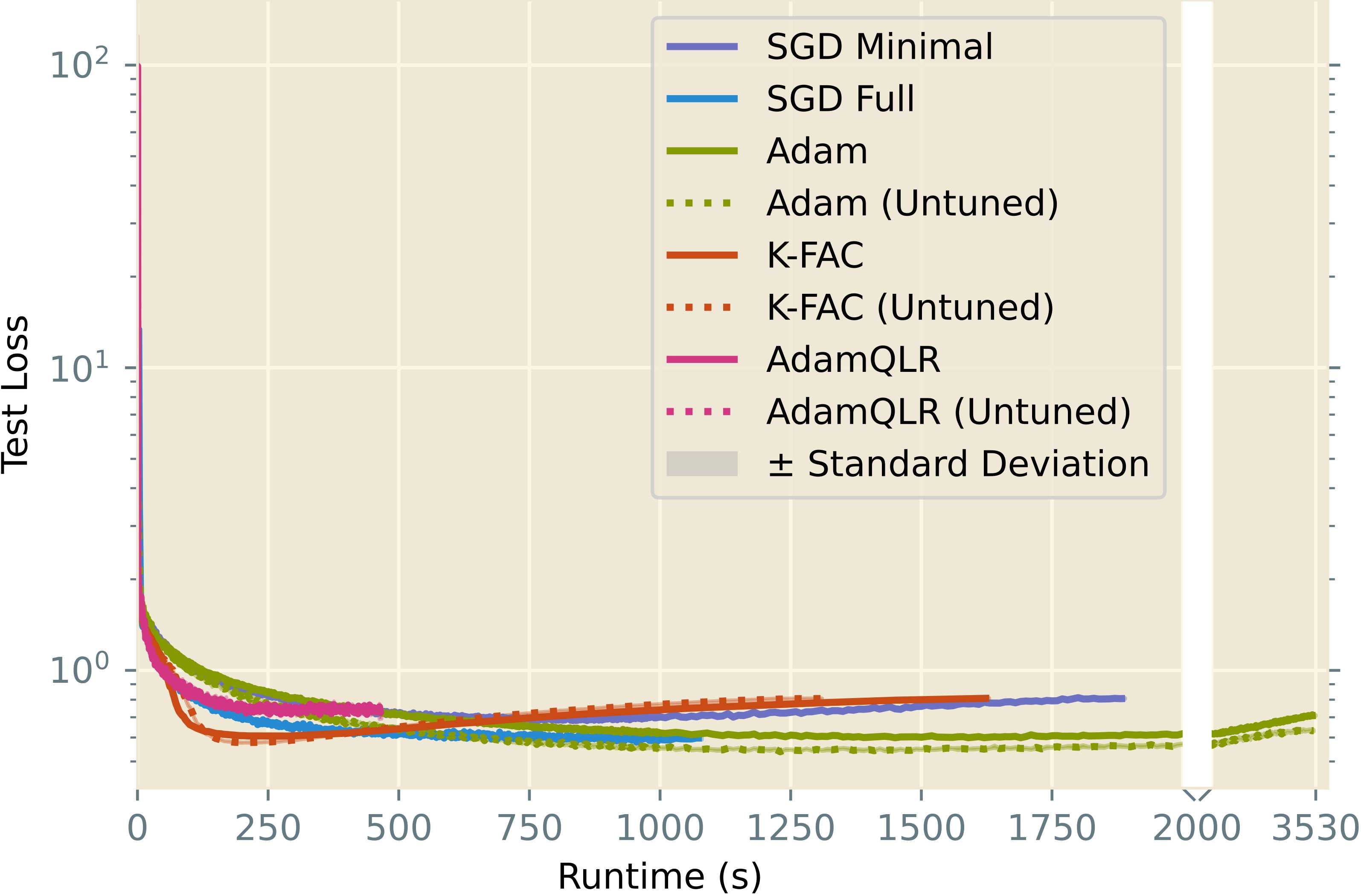} \vfill
    \end{tabularx}
    \caption{Median training (left) and test (right) loss trajectories, bootstrap-sampled over 50 repetitions per algorithm. Hyperparameters chosen by ASHA over 200 initialisations. Note changes of scale on the time axes. See also our numerical comparison in Table~\ref{tab:EpochConstrainedFinalResults}.}
    \label{fig:AlgorithmLosses}
\end{figure*}

In Figure~\ref{fig:AlgorithmMixedResults}, we plotted experimental results in terms of the loss metric used during training. For Fashion-MNIST, SVHN and CIFAR-10, we also present classification accuracy in metrics in Figure~\ref{fig:AlgorithmLosses} and Table~\ref{tab:EpochConstrainedFinalResults}. These illustrate broadly the same patterns as we discussed in the main body of the paper.

\subsubsection{Penn Treebank}
\label{sec:PennTreebank}
\begin{figure*}
    \centering
    \includegraphics[width=0.45\linewidth]{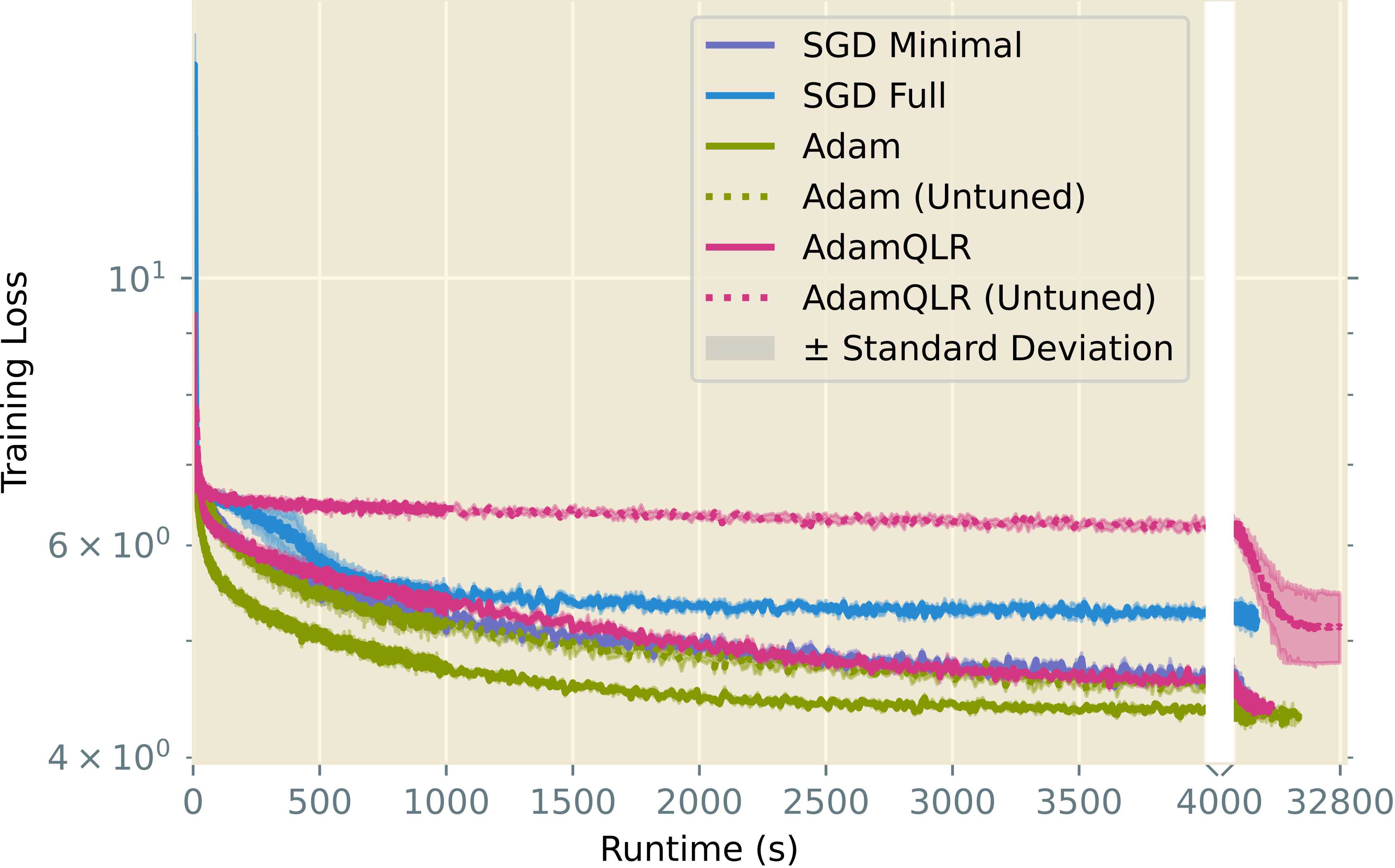}
    % \hfill
    \hspace*{0.01\linewidth}
    \includegraphics[width=0.45\linewidth]{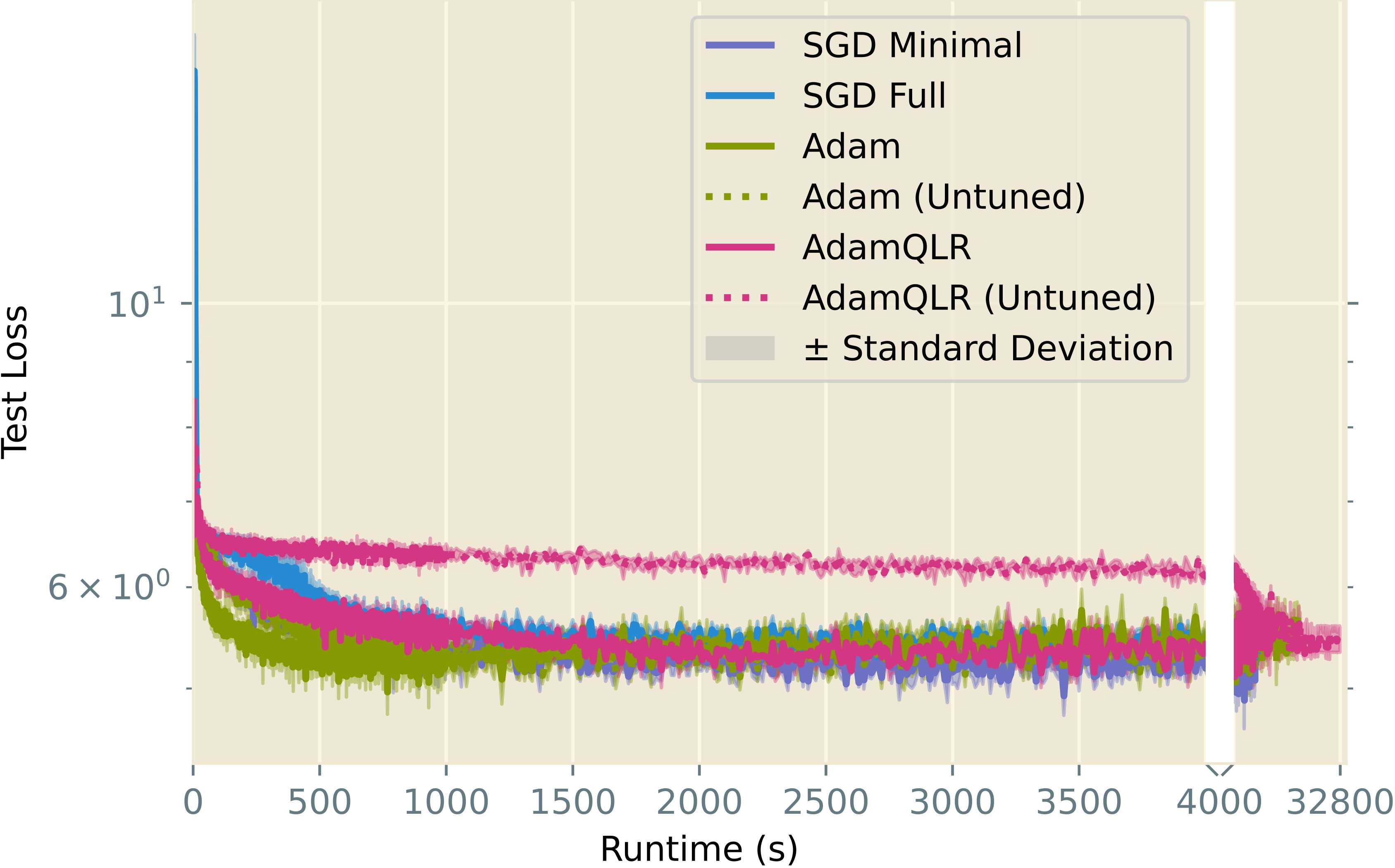}
    \caption{Median training (left) and test (right) loss trajectories for Penn Treebank on GPT-2, bootstrap-sampled over 10 repetitions per algorithm. Hyperparameters chosen by ASHA over 200 initialisations. Note changes of scale on the time axes. See also our numerical presentation in Table~\ref{tab:EpochConstrainedFinalResults}.}
    \label{fig:PennTreebank}
\end{figure*}

As an additional baseline, we consider training the standard Penn Treebank subset \citep{marcus_mitchell_p_treebank-3_1999,marcus_treebank-3_1999} on the GPT-2 model \citep{radford_language_2019}, as implemented by Hugging Face. We interpret the batch size as the number of token subsequences considered in parallel, chosen over $\{5, 10, 20, 35, 50, 100, 200\}$, and also choose the length of subsequences considered from $\{10, 20, 30, 40, 50, 60, 70, 80, 90, 100\}$, and we set the HPO runtime limit to 1~hour. Otherwise, our hyperparameter optimisation is identical to that of Section~\ref{sec:Experiments}. For time efficiency, we perform 10 repetitions of training with the best hyperparameters found, rather than 50 as in Section~\ref{sec:Experiments}, with each repetition comprising 100 epochs of training, and show our results in Figure~\ref{fig:PennTreebank} and Table~\ref{tab:EpochConstrainedFinalResults}. Our \emph{AdamQLR (Untuned)} setting uses a batch size of 30 and subsequence length of 70, chosen based on the largest values which fit on our GPUs.

Interestingly, our results reflect the observation that transformer training dynamics are quite different from other NN model classes. The addition of momentum and weight decay to \emph{SGD Full} seems to hinder it in comparison to \emph{SGD Minimal}, with the latter exhibiting superior training and generalisation performance. Both are ultimately beaten by \emph{Adam} on training performance, but the latter shows a greater tendency to overfit, with a gradually increasing test loss after around 1000\,s which neither \emph{SGD} algorithm exhibits. 

\emph{AdamQLR (Tuned)} performs very similarly to \emph{SGD Minimal} on this setting, albeit now with a slight tendency to overfit towards the end of training. Further, it achieves similar final training losses to \emph{Adam}, though the latter reaches these losses much faster. On the other hand, \emph{AdamQLR (Untuned)} shows a much greater distinction from \emph{AdamQLR (Tuned)} here than in our other experiments, suggesting the default hyperparameters we propose are not as immediately applicable to transformer models. However, it is reassuring that this algorithm achieves monotonically decreasing training and test losses --- combined with \emph{AdamQLR (Untuned)}'s robustness on other experiments, this leads us to suspect that a transformer-specific choice of default hyperparameters would provide similar robustness in this setting. We leave an investigation of these alternative defaults to future work.

\subsubsection{Numerical Results}

In Table~\ref{tab:EpochConstrainedFinalResults}, we give a numerical presentation of the results in Figures~\ref{fig:AlgorithmMixedResults}, their corresponding loss plots from Figure~\ref{fig:AlgorithmLosses} (Appendix~\ref{sec:LossResults}) and our additional Penn Treebank study from Figure~\ref{fig:PennTreebank} (Appendix~\ref{sec:PennTreebank}). We use a similar bootstrapping technique to Section~\ref{sec:Experiments} to give estimates for typical runtimes and numbers of steps completed.

\begin{table*}
    \centering
    \caption{Numerical study of the results shown in Figures~\ref{fig:AlgorithmMixedResults}, \ref{fig:AlgorithmLosses} and~\ref{fig:PennTreebank}: final statistics after epoch-constrained training on our benchmark tasks.}
    \label{tab:EpochConstrainedFinalResults}
    \resizebox{\linewidth}{!}{
    \begin{tabular}{cc
        S[table-format=1.7]
        U
        S[table-format=1.5]
        U
        S[table-format=1.6]
        U
        S[table-format=1.5]
        U
        S[table-format=1.6]
        U
        S[table-format=5.1]
        U
        S[table-format=5.3]
        U
    }
        \toprule
        Dataset & Algorithm & \multicolumn{2}{c}{Training Loss} & \multicolumn{2}{c}{Training Accuracy} & \multicolumn{2}{c}{Test Loss} & \multicolumn{2}{c}{Test Accuracy} & \multicolumn{2}{c}{Generalisation Gap} & \multicolumn{2}{c}{Total Steps} & \multicolumn{2}{c}{Total Time (s)} \\
        \midrule
        % \midrule
\multirow{8}{*}{UCI Energy}
& SGD Minimal 	& 0.000674 & $\pm$ \num{0.000031} 	& \multicolumn{2}{c}{---}	& 0.001245 & $\pm$ \num{0.000049} 	& \multicolumn{2}{c}{---}	& 0.000571 & $\pm$ \num{0.000081} 	& 24000 & $\pm$ \num{0} 	& 276.79 & $\pm$ \num{0.30} \\
& SGD Full 	& 0.000431 & $\pm$ \num{0.000019} 	& \multicolumn{2}{c}{---}	& 0.000657 & $\pm$ \num{0.000016} 	& \multicolumn{2}{c}{---}	& 0.000226 & $\pm$ \num{0.000035} 	& 8000 & $\pm$ \num{0} 	& 134.37 & $\pm$ \num{0.34} \\
& Adam 	& 0.000282 & $\pm$ \num{0.000026} 	& \multicolumn{2}{c}{---}	& 0.000768 & $\pm$ \num{0.000029} 	& \multicolumn{2}{c}{---}	& 0.000486 & $\pm$ \num{0.000054} 	& 4000 & $\pm$ \num{0} 	& 94.05 & $\pm$ \num{0.11} \\
& Adam (Untuned) 	& 0.000596 & $\pm$ \num{0.000021} 	& \multicolumn{2}{c}{---}	& 0.001147 & $\pm$ \num{0.000058} 	& \multicolumn{2}{c}{---}	& 0.000550 & $\pm$ \num{0.000079} 	& 4000 & $\pm$ \num{0} 	& 77.92 & $\pm$ \num{0.49} \\
& K-FAC 	& 0.0001349 & $\pm$ \num{0.0000056} 	& \multicolumn{2}{c}{---}	& 0.000843 & $\pm$ \num{0.000027} 	& \multicolumn{2}{c}{---}	& 0.000708 & $\pm$ \num{0.000032} 	& 48000 & $\pm$ \num{0} 	& 580.38 & $\pm$ \num{0.46} \\
& K-FAC (Untuned) 	& 0.0002953 & $\pm$ \num{0.0000094} 	& \multicolumn{2}{c}{---}	& 0.001069 & $\pm$ \num{0.000039} 	& \multicolumn{2}{c}{---}	& 0.000774 & $\pm$ \num{0.000048} 	& 1027 & $\pm$ \num{20} 	& 64.7 & $\pm$ \num{1.1} \\
& AdamQLR 	& 0.0003330 & $\pm$ \num{0.0000087} 	& \multicolumn{2}{c}{---}	& 0.000873 & $\pm$ \num{0.000021} 	& \multicolumn{2}{c}{---}	& 0.000540 & $\pm$ \num{0.000030} 	& 4000 & $\pm$ \num{0} 	& 92.49 & $\pm$ \num{0.10} \\
& AdamQLR (Untuned) 	& \num{1.6e29} & $\pm$ \num{5.3e28} 	& \multicolumn{2}{c}{---}	& \num{1.3e29} & $\pm$ \num{4.2e28} 	& \multicolumn{2}{c}{---}	& \num{-3.6e28} & $\pm$ \num{9.5e28} 	& 175.0 & $\pm$ \num{2.6} 	& 10.82 & $\pm$ \num{0.32} \\
\midrule
\multirow{8}{*}{UCI Protein}
& SGD Minimal 	& 0.2583 & $\pm$ \num{0.0041} 	& \multicolumn{2}{c}{---}	& 0.2714 & $\pm$ \num{0.0019} 	& \multicolumn{2}{c}{---}	& 0.0132 & $\pm$ \num{0.0060} 	& 17600 & $\pm$ \num{0} 	& 249.16 & $\pm$ \num{0.75} \\
& SGD Full 	& 0.2401 & $\pm$ \num{0.0088} 	& \multicolumn{2}{c}{---}	& 0.2497 & $\pm$ \num{0.0016} 	& \multicolumn{2}{c}{---}	& 0.010 & $\pm$ \num{0.010} 	& 70000 & $\pm$ \num{0} 	& 648.69 & $\pm$ \num{0.52} \\
& Adam 	& 0.2370 & $\pm$ \num{0.0029} 	& \multicolumn{2}{c}{---}	& 0.2475 & $\pm$ \num{0.0011} 	& \multicolumn{2}{c}{---}	& 0.0105 & $\pm$ \num{0.0040} 	& 8800 & $\pm$ \num{0} 	& 185.62 & $\pm$ \num{0.56} \\
& Adam (Untuned) 	& 0.2511 & $\pm$ \num{0.0014} 	& \multicolumn{2}{c}{---}	& 0.26249 & $\pm$ \num{0.00052} 	& \multicolumn{2}{c}{---}	& 0.0114 & $\pm$ \num{0.0019} 	& 7000 & $\pm$ \num{0} 	& 182.31 & $\pm$ \num{0.75} \\
& K-FAC 	& 0.2015 & $\pm$ \num{0.0010} 	& \multicolumn{2}{c}{---}	& 0.23018 & $\pm$ \num{0.00077} 	& \multicolumn{2}{c}{---}	& 0.0287 & $\pm$ \num{0.0018} 	& 2200 & $\pm$ \num{0} 	& 146.89 & $\pm$ \num{0.60} \\
& K-FAC (Untuned) 	& 0.2010 & $\pm$ \num{0.0010} 	& \multicolumn{2}{c}{---}	& 0.22989 & $\pm$ \num{0.00090} 	& \multicolumn{2}{c}{---}	& 0.0289 & $\pm$ \num{0.0020} 	& 2200 & $\pm$ \num{0} 	& 148.47 & $\pm$ \num{0.66} \\
& AdamQLR 	& 0.2241 & $\pm$ \num{0.0048} 	& \multicolumn{2}{c}{---}	& 0.23733 & $\pm$ \num{0.00076} 	& \multicolumn{2}{c}{---}	& 0.0132 & $\pm$ \num{0.0056} 	& 17600 & $\pm$ \num{0} 	& 284.14 & $\pm$ \num{0.30} \\
& AdamQLR (Untuned) 	& 0.2451 & $\pm$ \num{0.0017} 	& \multicolumn{2}{c}{---}	& 0.25732 & $\pm$ \num{0.00056} 	& \multicolumn{2}{c}{---}	& 0.0122 & $\pm$ \num{0.0022} 	& 2200 & $\pm$ \num{0} 	& 138.13 & $\pm$ \num{0.45} \\
\midrule
\multirow{8}{*}{Fashion-MNIST}
& SGD Minimal 	& 0.244 & $\pm$ \num{0.011} 	& 0.9070 & $\pm$ \num{0.0066} 	& 0.3678 & $\pm$ \num{0.0013} 	& 0.87222 & $\pm$ \num{0.00042} 	& 0.124 & $\pm$ \num{0.013} 	& 5000 & $\pm$ \num{0} 	& 79.671 & $\pm$ \num{0.064} \\
& SGD Full 	& 0.2530 & $\pm$ \num{0.0042} 	& 0.9088 & $\pm$ \num{0.0023} 	& 0.3738 & $\pm$ \num{0.0016} 	& 0.87115 & $\pm$ \num{0.00056} 	& 0.1209 & $\pm$ \num{0.0058} 	& 630 & $\pm$ \num{0} 	& 15.16 & $\pm$ \num{0.37} \\
& Adam 	& 0.2292 & $\pm$ \num{0.0052} 	& 0.9142 & $\pm$ \num{0.0020} 	& 0.3822 & $\pm$ \num{0.0015} 	& 0.87320 & $\pm$ \num{0.00072} 	& 0.1530 & $\pm$ \num{0.0067} 	& 1250 & $\pm$ \num{0} 	& 26.286 & $\pm$ \num{0.057} \\
& Adam (Untuned) 	& 0.214 & $\pm$ \num{0.011} 	& 0.9214 & $\pm$ \num{0.0045} 	& 0.3899 & $\pm$ \num{0.0011} 	& 0.87437 & $\pm$ \num{0.00074} 	& 0.176 & $\pm$ \num{0.012} 	& 10000 & $\pm$ \num{0} 	& 140.47 & $\pm$ \num{0.26} \\
& K-FAC 	& 0.0904 & $\pm$ \num{0.0016} 	& 0.97612 & $\pm$ \num{0.00089} 	& 0.4572 & $\pm$ \num{0.0028} 	& 0.86892 & $\pm$ \num{0.00064} 	& 0.3668 & $\pm$ \num{0.0044} 	& 160 & $\pm$ \num{0} 	& 13.378 & $\pm$ \num{0.082} \\
& K-FAC (Untuned) 	& 0.1054 & $\pm$ \num{0.0023} 	& 0.97058 & $\pm$ \num{0.00072} 	& 0.4416 & $\pm$ \num{0.0033} 	& 0.87070 & $\pm$ \num{0.00047} 	& 0.3361 & $\pm$ \num{0.0056} 	& 160 & $\pm$ \num{0} 	& 9.66 & $\pm$ \num{0.35} \\
& AdamQLR 	& 0.2664 & $\pm$ \num{0.0023} 	& 0.9076 & $\pm$ \num{0.0012} 	& 0.37100 & $\pm$ \num{0.00078} 	& 0.87040 & $\pm$ \num{0.00021} 	& 0.1046 & $\pm$ \num{0.0030} 	& 160 & $\pm$ \num{0} 	& 10.53 & $\pm$ \num{0.20} \\
& AdamQLR (Untuned) 	& 0.2692 & $\pm$ \num{0.0029} 	& 0.9050 & $\pm$ \num{0.0010} 	& 0.37028 & $\pm$ \num{0.00060} 	& 0.87100 & $\pm$ \num{0.00028} 	& 0.1011 & $\pm$ \num{0.0035} 	& 160 & $\pm$ \num{0} 	& 10.656 & $\pm$ \num{0.075} \\
\midrule
\multirow{8}{*}{SVHN}
& SGD Minimal 	& 0.1544 & $\pm$ \num{0.0019} 	& 0.9337 & $\pm$ \num{0.0013} 	& 0.5344 & $\pm$ \num{0.0042} 	& 0.84287 & $\pm$ \num{0.00081} 	& 0.3800 & $\pm$ \num{0.0061} 	& 390 & $\pm$ \num{0} 	& 110.28 & $\pm$ \num{0.22} \\
& SGD Full 	& 0.0871 & $\pm$ \num{0.0021} 	& 0.95225 & $\pm$ \num{0.00058} 	& 0.3939 & $\pm$ \num{0.0024} 	& 0.89883 & $\pm$ \num{0.00056} 	& 0.3068 & $\pm$ \num{0.0044} 	& 390 & $\pm$ \num{0} 	& 110.52 & $\pm$ \num{0.15} \\
& Adam 	& 0.0776 & $\pm$ \num{0.0033} 	& 0.96636 & $\pm$ \num{0.00086} 	& 0.4827 & $\pm$ \num{0.0038} 	& 0.8842 & $\pm$ \num{0.0010} 	& 0.4051 & $\pm$ \num{0.0071} 	& 770 & $\pm$ \num{0} 	& 198.50 & $\pm$ \num{0.13} \\
& Adam (Untuned) 	& 0.1109 & $\pm$ \num{0.0097} 	& 0.9696 & $\pm$ \num{0.0031} 	& 0.3275 & $\pm$ \num{0.0037} 	& 0.91472 & $\pm$ \num{0.00060} 	& 0.217 & $\pm$ \num{0.013} 	& 2390 & $\pm$ \num{0} 	& 570.6 & $\pm$ \num{3.8} \\
& K-FAC 	& 0.0480 & $\pm$ \num{0.0038} 	& 0.9828 & $\pm$ \num{0.0011} 	& 0.4638 & $\pm$ \num{0.0034} 	& 0.86270 & $\pm$ \num{0.00067} 	& 0.4158 & $\pm$ \num{0.0072} 	& 770 & $\pm$ \num{0} 	& 500.8 & $\pm$ \num{1.2} \\
& K-FAC (Untuned) 	& 0.01684 & $\pm$ \num{0.00078} 	& 0.99762 & $\pm$ \num{0.00024} 	& 0.4425 & $\pm$ \num{0.0023} 	& 0.87926 & $\pm$ \num{0.00046} 	& 0.4256 & $\pm$ \num{0.0031} 	& 200 & $\pm$ \num{0} 	& 221.72 & $\pm$ \num{0.17} \\
& AdamQLR 	& 0.0892 & $\pm$ \num{0.0027} 	& 0.97446 & $\pm$ \num{0.00080} 	& 0.4034 & $\pm$ \num{0.0054} 	& 0.89595 & $\pm$ \num{0.00080} 	& 0.3142 & $\pm$ \num{0.0081} 	& 200 & $\pm$ \num{0} 	& 77.03 & $\pm$ \num{0.71} \\
& AdamQLR (Untuned) 	& 0.0876 & $\pm$ \num{0.0029} 	& 0.97505 & $\pm$ \num{0.00069} 	& 0.3952 & $\pm$ \num{0.0060} 	& 0.89760 & $\pm$ \num{0.00100} 	& 0.3076 & $\pm$ \num{0.0089} 	& 200 & $\pm$ \num{0} 	& 78.04 & $\pm$ \num{0.39} \\
\midrule
\multirow{8}{*}{CIFAR-10}
& SGD Minimal 	& 0.2005 & $\pm$ \num{0.0064} 	& 0.9261 & $\pm$ \num{0.0035} 	& 0.8074 & $\pm$ \num{0.0035} 	& 0.80003 & $\pm$ \num{0.00083} 	& 0.6069 & $\pm$ \num{0.0099} 	& 16200 & $\pm$ \num{0} 	& 1846 & $\pm$ \num{11} \\
& SGD Full 	& 0.3683 & $\pm$ \num{0.0087} 	& 0.8633 & $\pm$ \num{0.0027} 	& 0.5975 & $\pm$ \num{0.0082} 	& 0.8065 & $\pm$ \num{0.0024} 	& 0.229 & $\pm$ \num{0.017} 	& 8136 & $\pm$ \num{0} 	& 1058.6 & $\pm$ \num{2.7} \\
& Adam 	& 0.173 & $\pm$ \num{0.010} 	& 0.9344 & $\pm$ \num{0.0065} 	& 0.7095 & $\pm$ \num{0.0031} 	& 0.82364 & $\pm$ \num{0.00055} 	& 0.536 & $\pm$ \num{0.013} 	& 32400 & $\pm$ \num{0} 	& 3480.1 & $\pm$ \num{4.0} \\
& Adam (Untuned) 	& 0.140 & $\pm$ \num{0.015} 	& 0.9475 & $\pm$ \num{0.0044} 	& 0.6322 & $\pm$ \num{0.0039} 	& 0.84242 & $\pm$ \num{0.00070} 	& 0.493 & $\pm$ \num{0.019} 	& 25344 & $\pm$ \num{0} 	& 3310 & $\pm$ \num{180} \\
& K-FAC 	& 0.0856 & $\pm$ \num{0.0017} 	& 0.94301 & $\pm$ \num{0.00091} 	& 0.8087 & $\pm$ \num{0.0024} 	& 0.79593 & $\pm$ \num{0.00052} 	& 0.7231 & $\pm$ \num{0.0041} 	& 2088 & $\pm$ \num{0} 	& 1601 & $\pm$ \num{16} \\
& K-FAC (Untuned) 	& 0.0589 & $\pm$ \num{0.0014} 	& 0.98277 & $\pm$ \num{0.00061} 	& 0.8068 & $\pm$ \num{0.0020} 	& 0.80661 & $\pm$ \num{0.00071} 	& 0.7479 & $\pm$ \num{0.0034} 	& 1080 & $\pm$ \num{0} 	& 1227.0 & $\pm$ \num{1.9} \\
& AdamQLR 	& 0.3202 & $\pm$ \num{0.0040} 	& 0.8934 & $\pm$ \num{0.0012} 	& 0.733 & $\pm$ \num{0.011} 	& 0.7931 & $\pm$ \num{0.0027} 	& 0.412 & $\pm$ \num{0.015} 	& 1080 & $\pm$ \num{0} 	& 463.73 & $\pm$ \num{0.66} \\
& AdamQLR (Untuned) 	& 0.3213 & $\pm$ \num{0.0035} 	& 0.89507 & $\pm$ \num{0.00098} 	& 0.706 & $\pm$ \num{0.015} 	& 0.7986 & $\pm$ \num{0.0032} 	& 0.385 & $\pm$ \num{0.018} 	& 1080 & $\pm$ \num{0} 	& 461.80 & $\pm$ \num{0.42} \\
\midrule
\multirow{6}{*}{Penn Treebank}
& SGD Minimal 	& 4.413 & $\pm$ \num{0.028} 	& \multicolumn{2}{c}{---}	& 5.394 & $\pm$ \num{0.098} 	& \multicolumn{2}{c}{---}	& 0.98 & $\pm$ \num{0.13} 	& 51600 & $\pm$ \num{0} 	& 10090 & $\pm$ \num{190} \\
& SGD Full 	& 5.177 & $\pm$ \num{0.045} 	& \multicolumn{2}{c}{---}	& 5.330 & $\pm$ \num{0.063} 	& \multicolumn{2}{c}{---}	& 0.15 & $\pm$ \num{0.11} 	& 53100 & $\pm$ \num{0} 	& 10030 & $\pm$ \num{140} \\
& Adam 	& 4.310 & $\pm$ \num{0.026} 	& \multicolumn{2}{c}{---}	& 5.52 & $\pm$ \num{0.11} 	& \multicolumn{2}{c}{---}	& 1.21 & $\pm$ \num{0.13} 	& 37100 & $\pm$ \num{0} 	& 9001 & $\pm$ \num{14} \\
& Adam (Untuned) 	& 4.322 & $\pm$ \num{0.029} 	& \multicolumn{2}{c}{---}	& 5.64 & $\pm$ \num{0.17} 	& \multicolumn{2}{c}{---}	& 1.31 & $\pm$ \num{0.20} 	& 44200 & $\pm$ \num{0} 	& 21130 & $\pm$ \num{110} \\
& AdamQLR 	& 4.418 & $\pm$ \num{0.029} 	& \multicolumn{2}{c}{---}	& 5.56 & $\pm$ \num{0.13} 	& \multicolumn{2}{c}{---}	& 1.15 & $\pm$ \num{0.16} 	& 29500 & $\pm$ \num{0} 	& 13930 & $\pm$ \num{190} \\
& AdamQLR (Untuned) 	& 5.13 & $\pm$ \num{0.33} 	& \multicolumn{2}{c}{---}	& 5.46 & $\pm$ \num{0.12} 	& \multicolumn{2}{c}{---}	& 0.33 & $\pm$ \num{0.46} 	& 44200 & $\pm$ \num{0} 	& 19200 & $\pm$ \num{3000} \\
\bottomrule
    \end{tabular}
    }
\end{table*}

\subsubsection{Fixed-Runtime Comparisons}
\label{sec:ASHATimeExperiments}

\begin{figure*}
    \centering
    \newcommand{\rotatecaption}[1]{%
        \rotatebox[origin=c]{90}{\begin{minipage}{3cm}#1\end{minipage}} }
    \begin{tabularx}{0.82\linewidth}{p{2ex}XX}
        \rotatecaption{\subcaption{UCI Energy}}
        & \includegraphics[align=c,width=\linewidth]{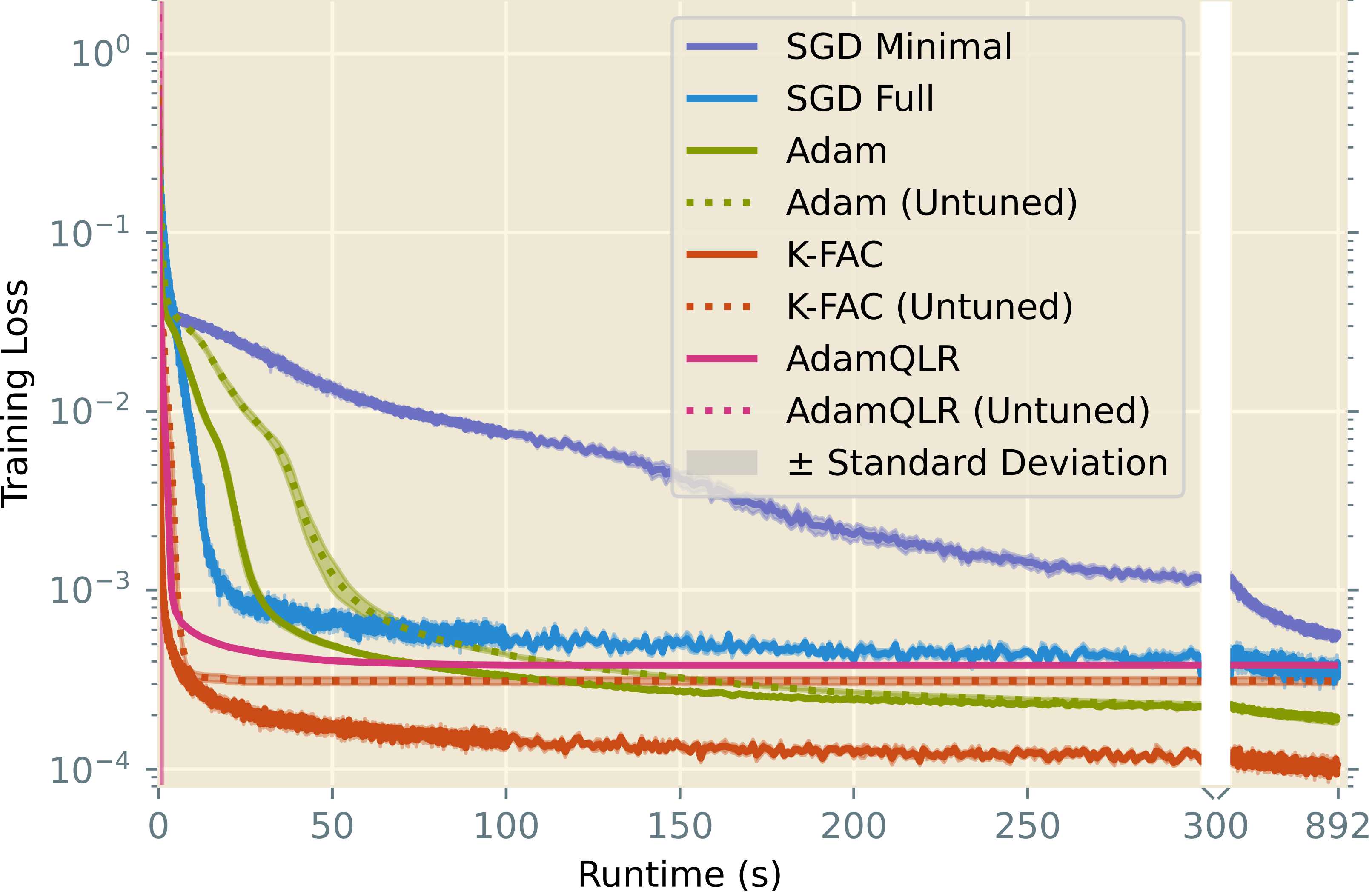}
        & \includegraphics[align=c,width=\linewidth]{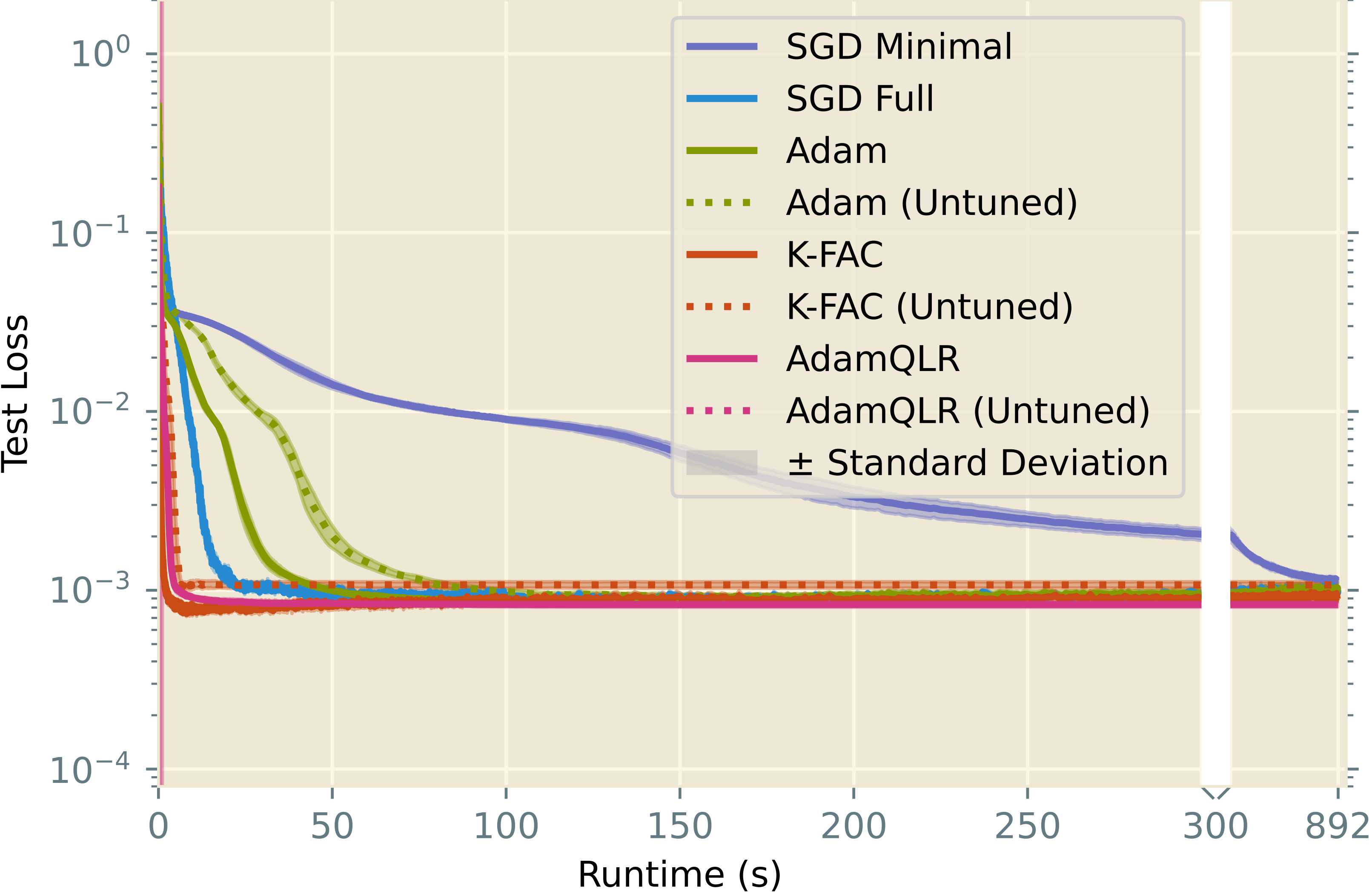} \vfill\\
        
        \rotatecaption{\subcaption{UCI Protein}}
        & \includegraphics[align=c,width=\linewidth]{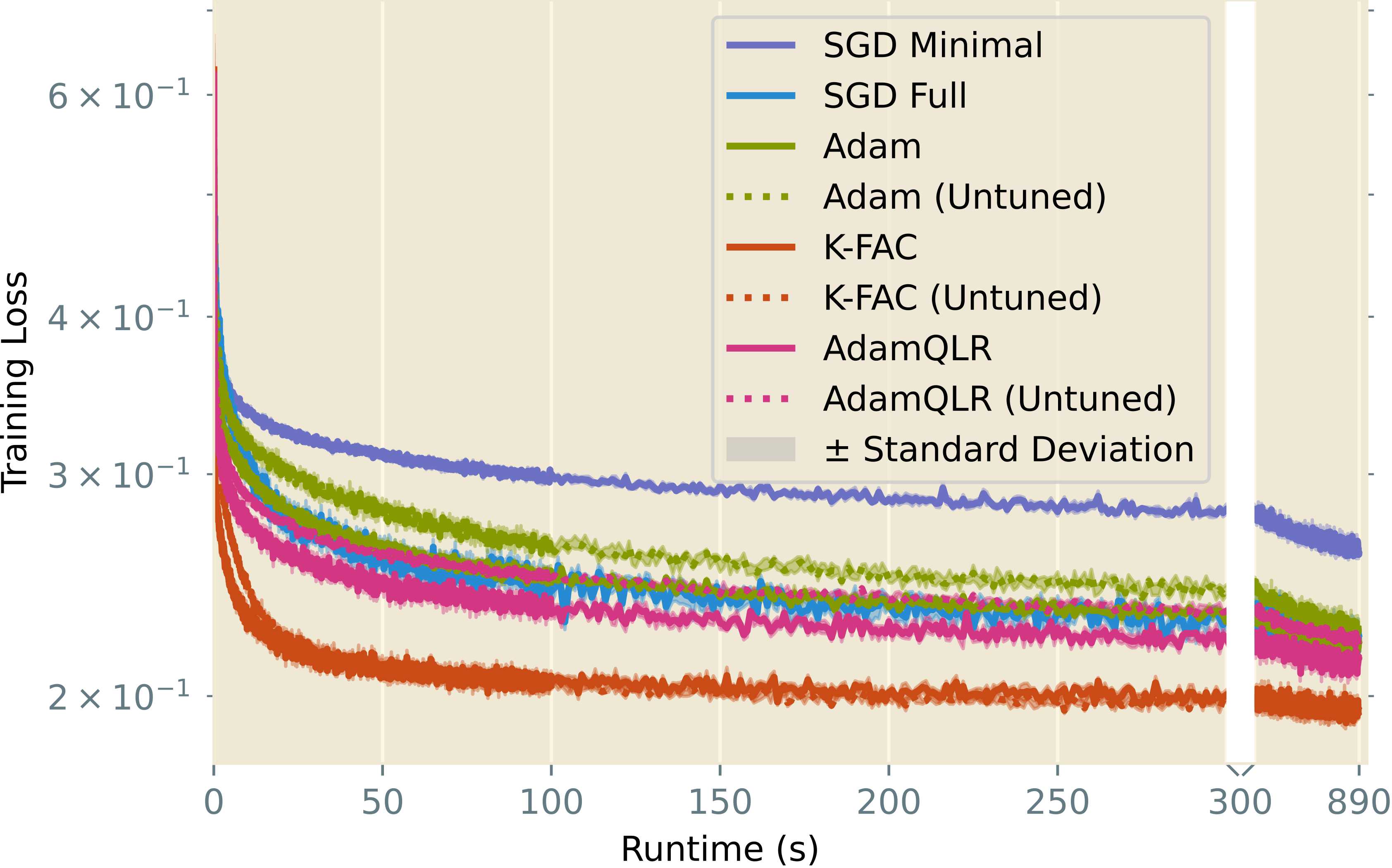}
        & \includegraphics[align=c,width=\linewidth]{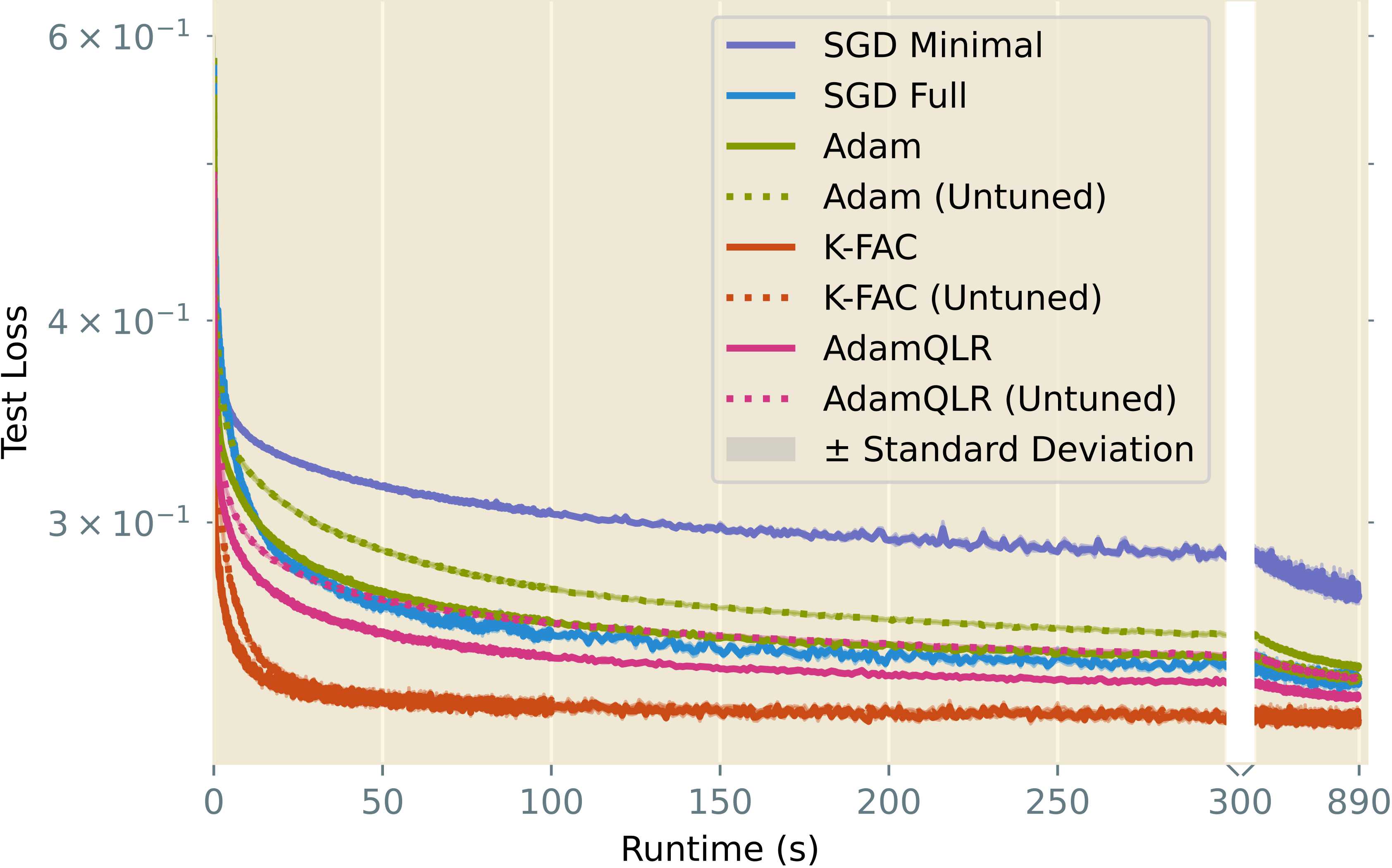} \vfill\\
        
        \rotatecaption{\subcaption{Fashion-MNIST}}
        & \includegraphics[align=c,width=\linewidth]{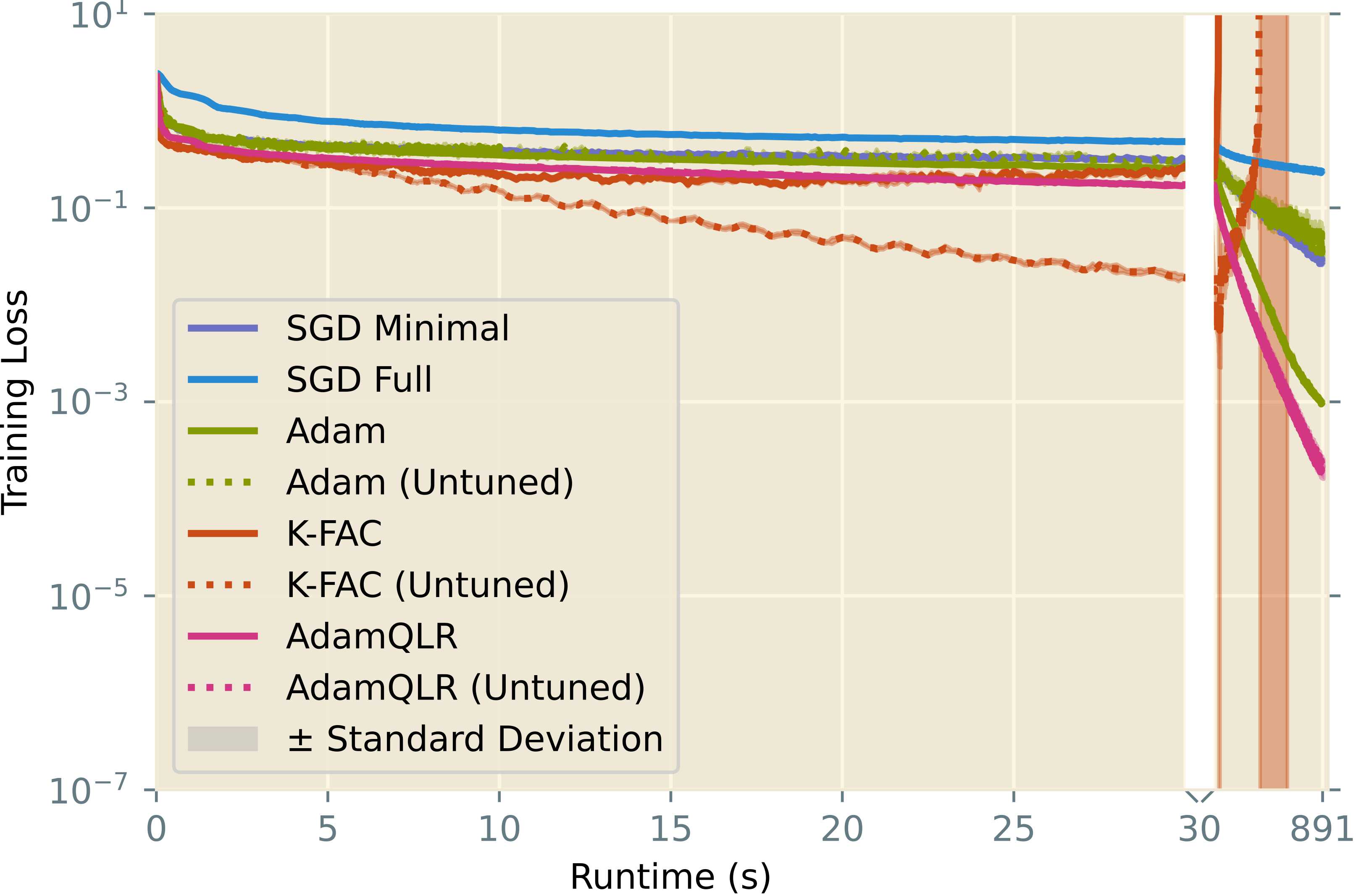}
        & \includegraphics[align=c,width=\linewidth]{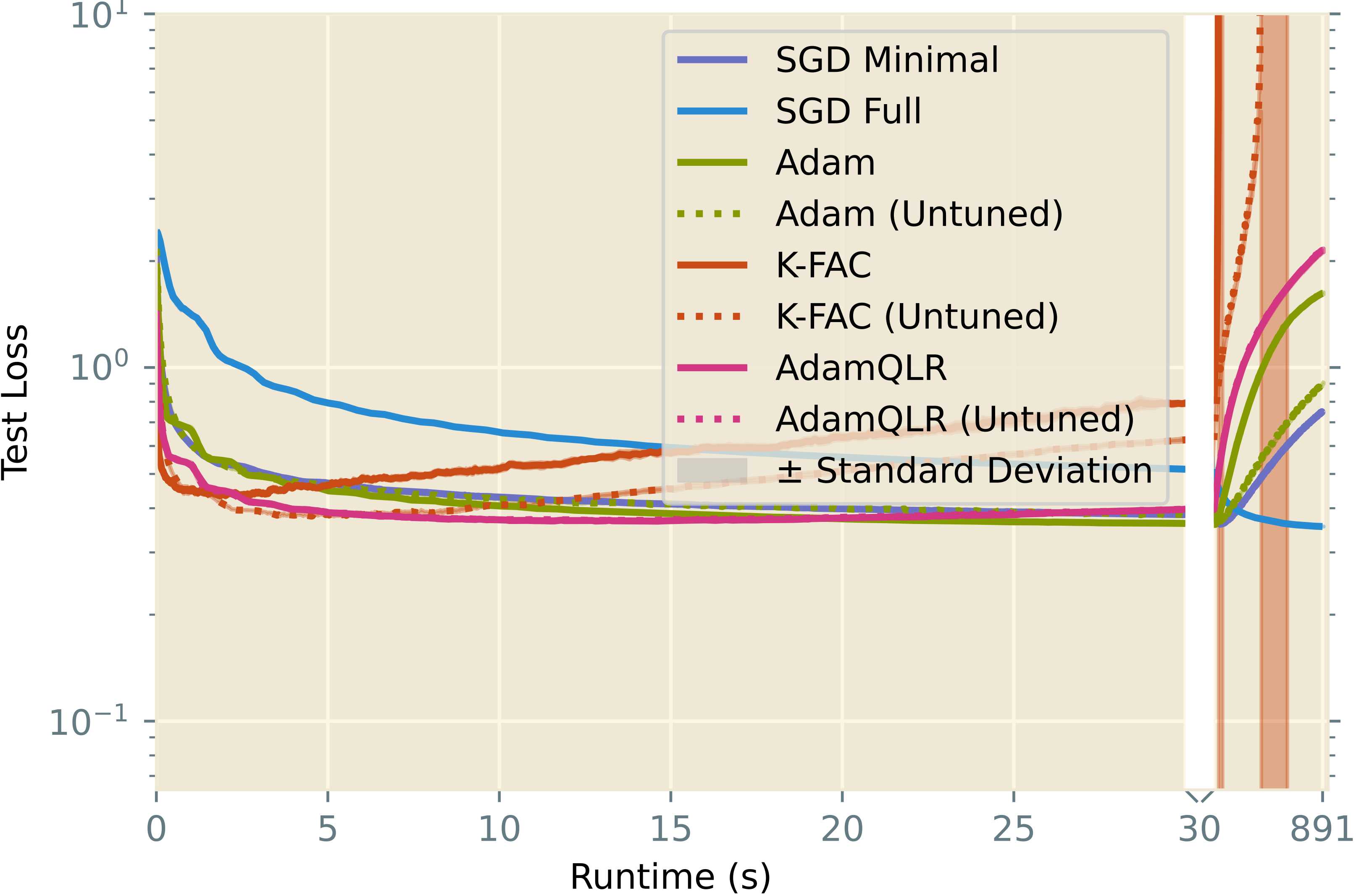} \vfill\\
        
        \rotatecaption{\subcaption{SVHN}}
        & \includegraphics[align=c,width=\linewidth]{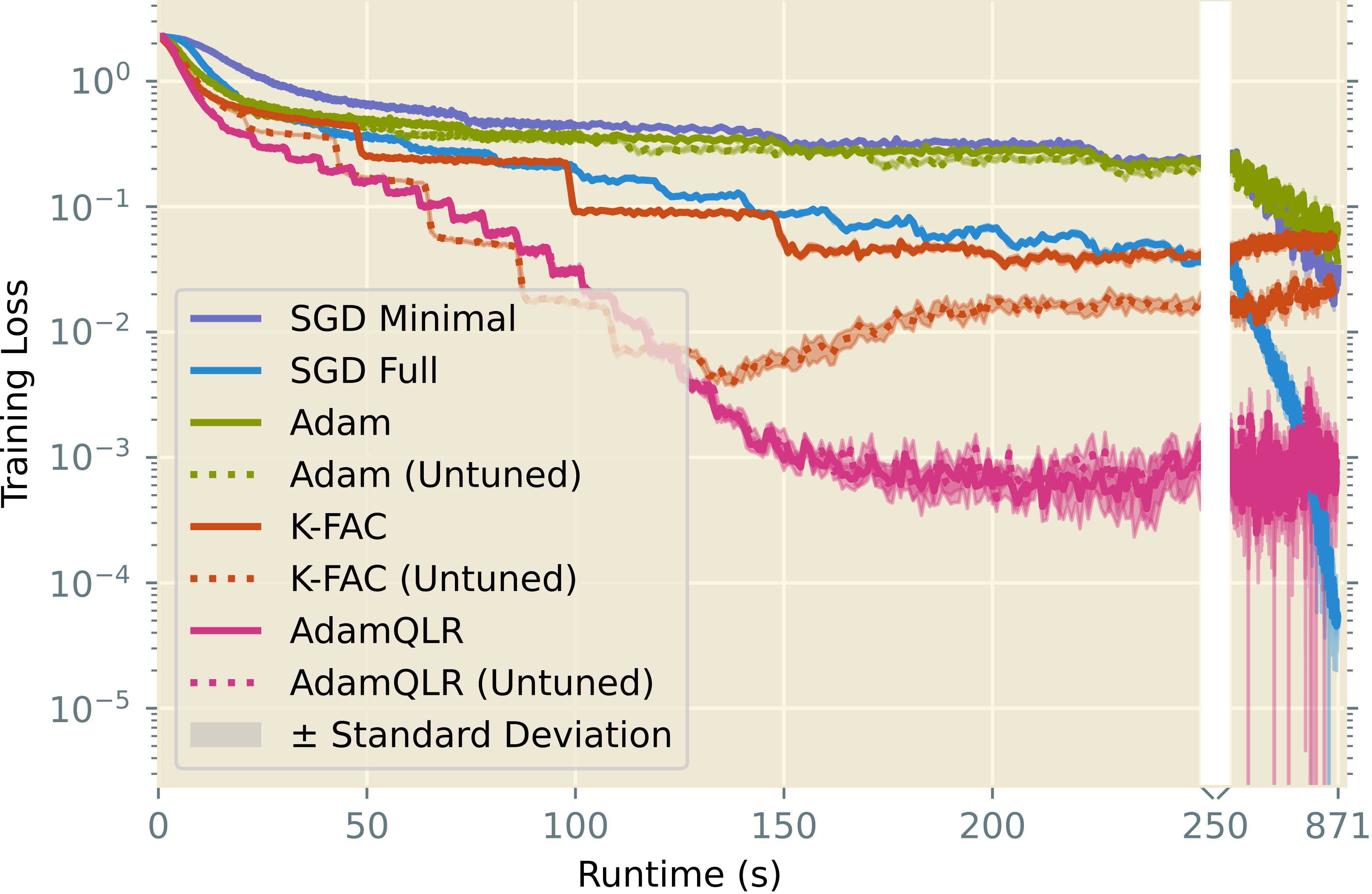}
        & \includegraphics[align=c,width=\linewidth]{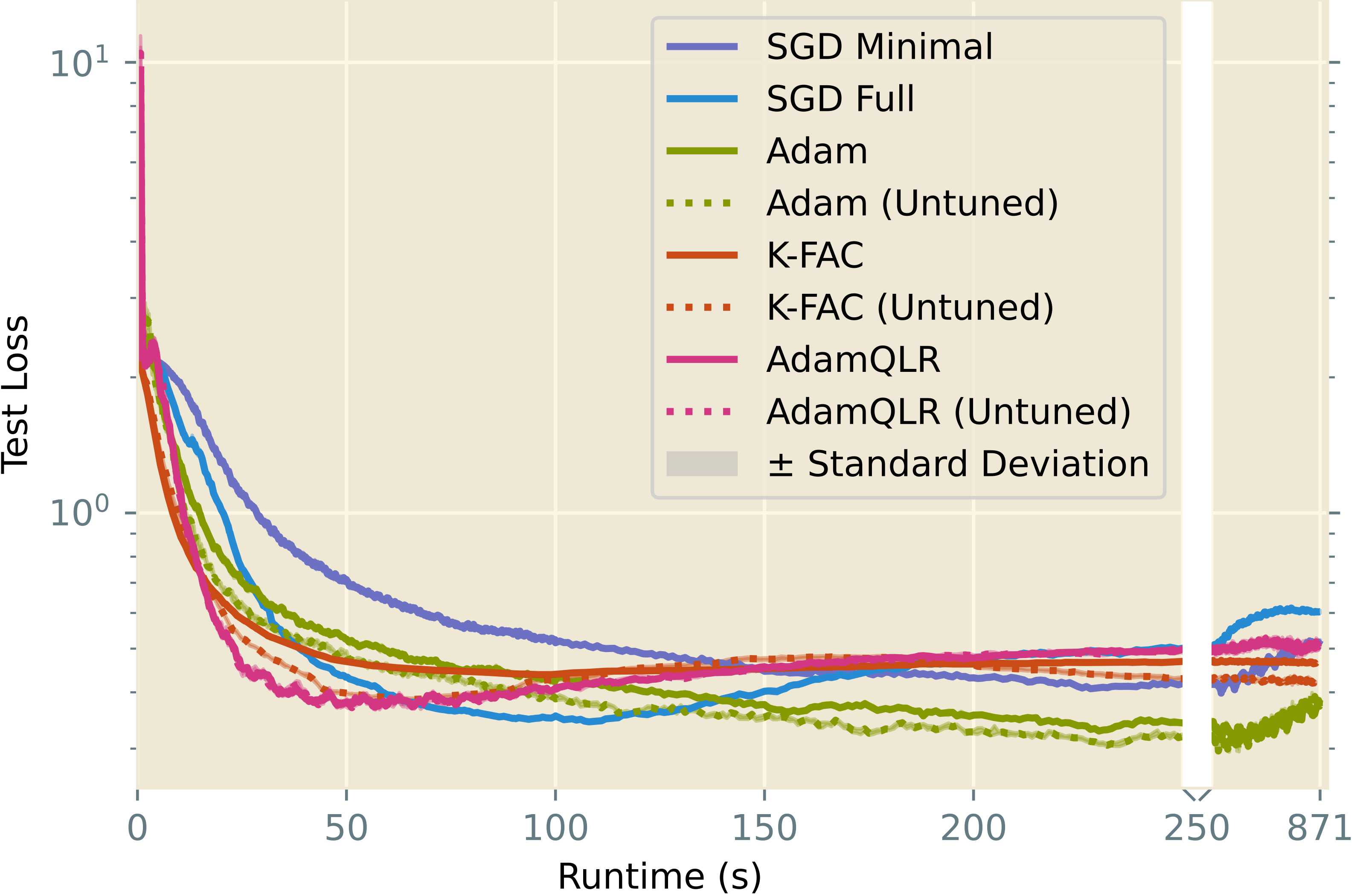} \vfill\\
        
        \rotatecaption{\subcaption{CIFAR-10}}
        & \includegraphics[align=c,width=\linewidth]{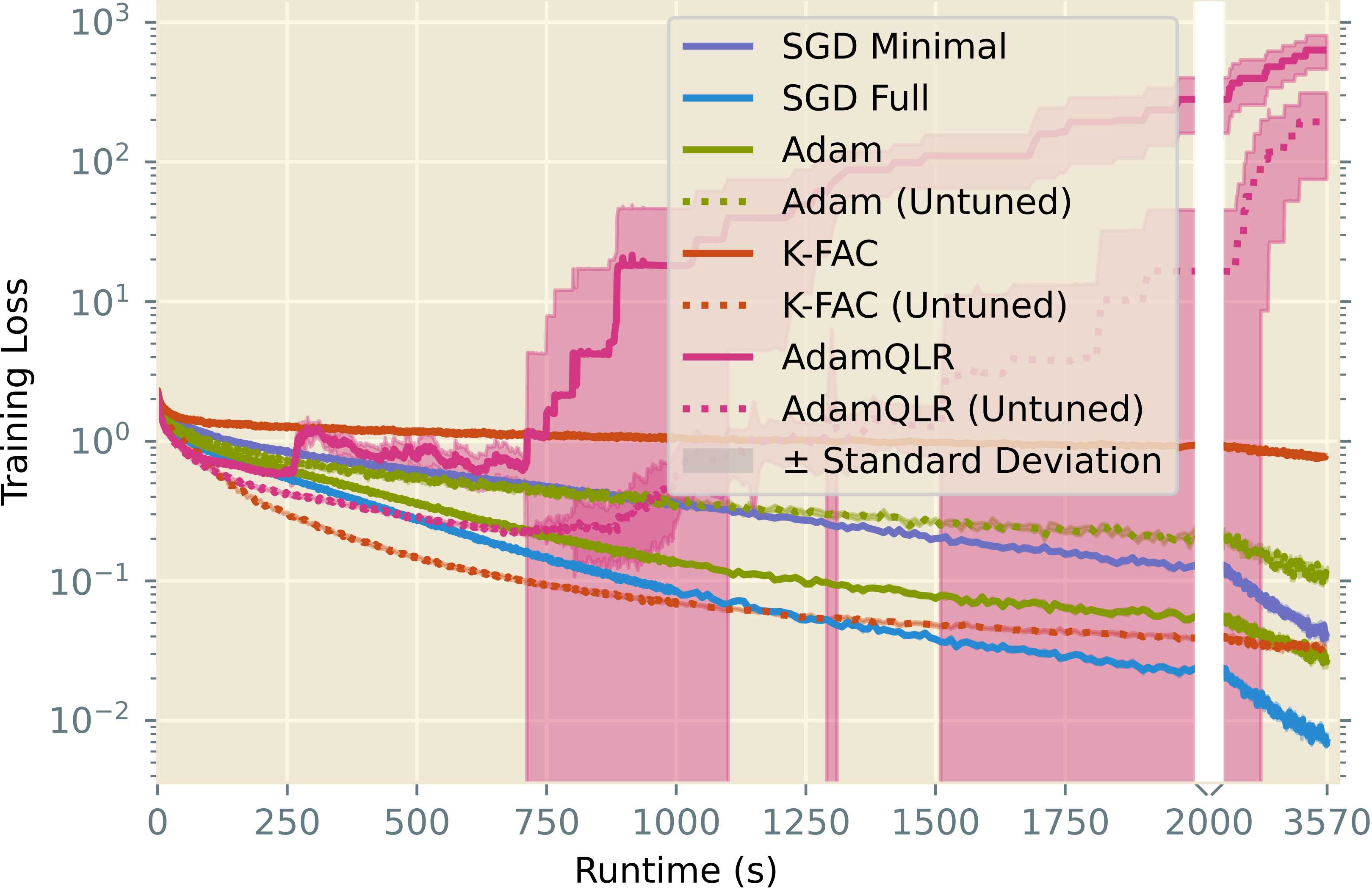}
        & \includegraphics[align=c,width=\linewidth]{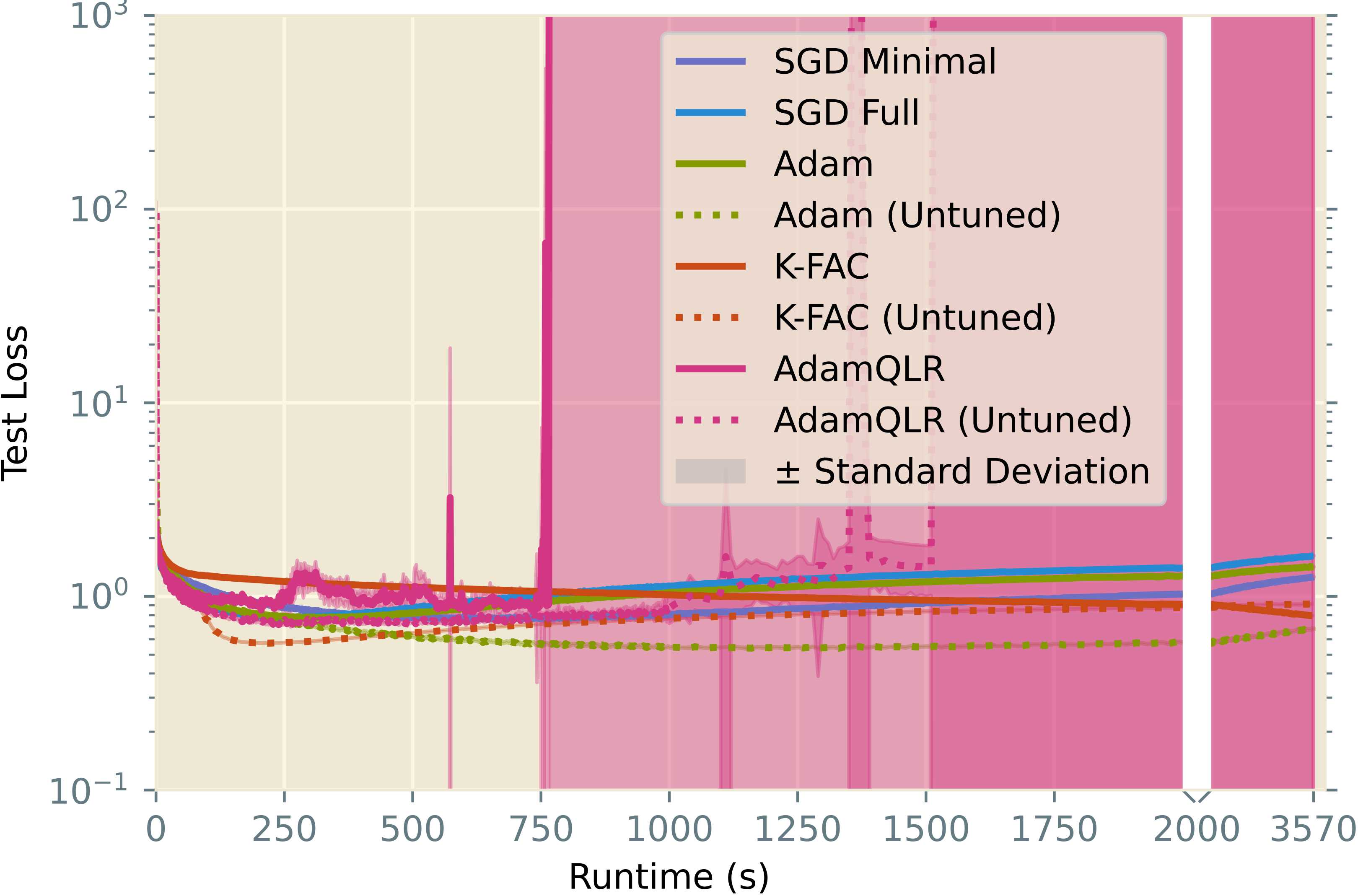} \vfill\\
    \end{tabularx}
    \caption{Median training (left) and test (right) loss trajectories, bootstrap-sampled over 50 repetitions per algorithm. Hyperparameters chosen by ASHA over 200 initialisations to minimise validation loss after a fixed runtime of 15~minutes. Note changes of scale on the time axes. See also our numerical comparison in Table~\ref{tab:ASHATimeValidationFinalResults}.}
    \label{fig:ASHATimeValidation}
\end{figure*}

\begin{figure*}
    \centering
    \newcommand{\rotatecaption}[1]{%
        \rotatebox[origin=c]{90}{\begin{minipage}{3cm}#1\end{minipage}} }
    \begin{tabularx}{0.82\linewidth}{p{2ex}XX}
        \rotatecaption{\subcaption{UCI Energy}}
        & \includegraphics[align=c,width=\linewidth]{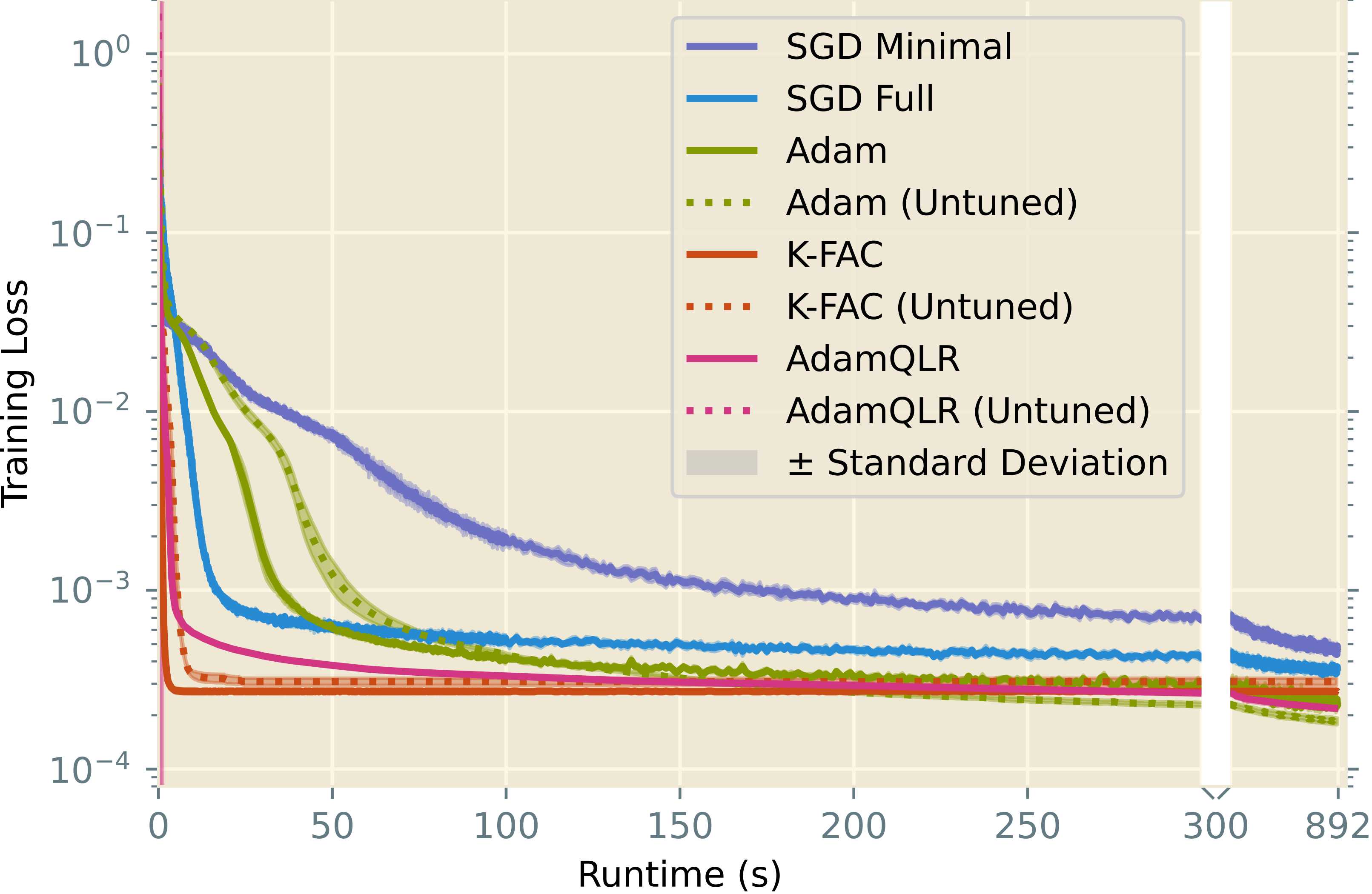}
        & \includegraphics[align=c,width=\linewidth]{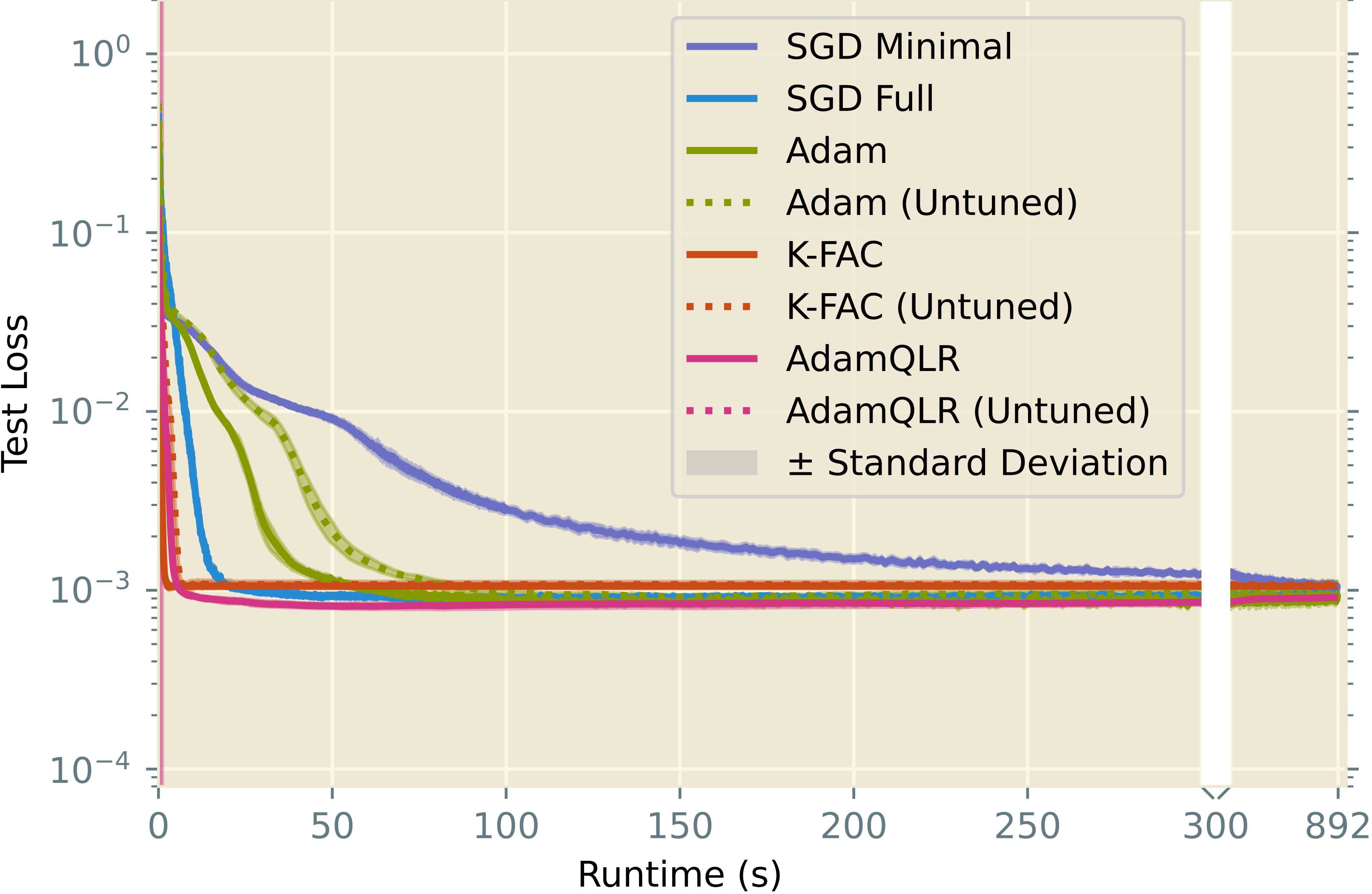} \vfill\\
        
        \rotatecaption{\subcaption{UCI Protein}}
        & \includegraphics[align=c,width=\linewidth]{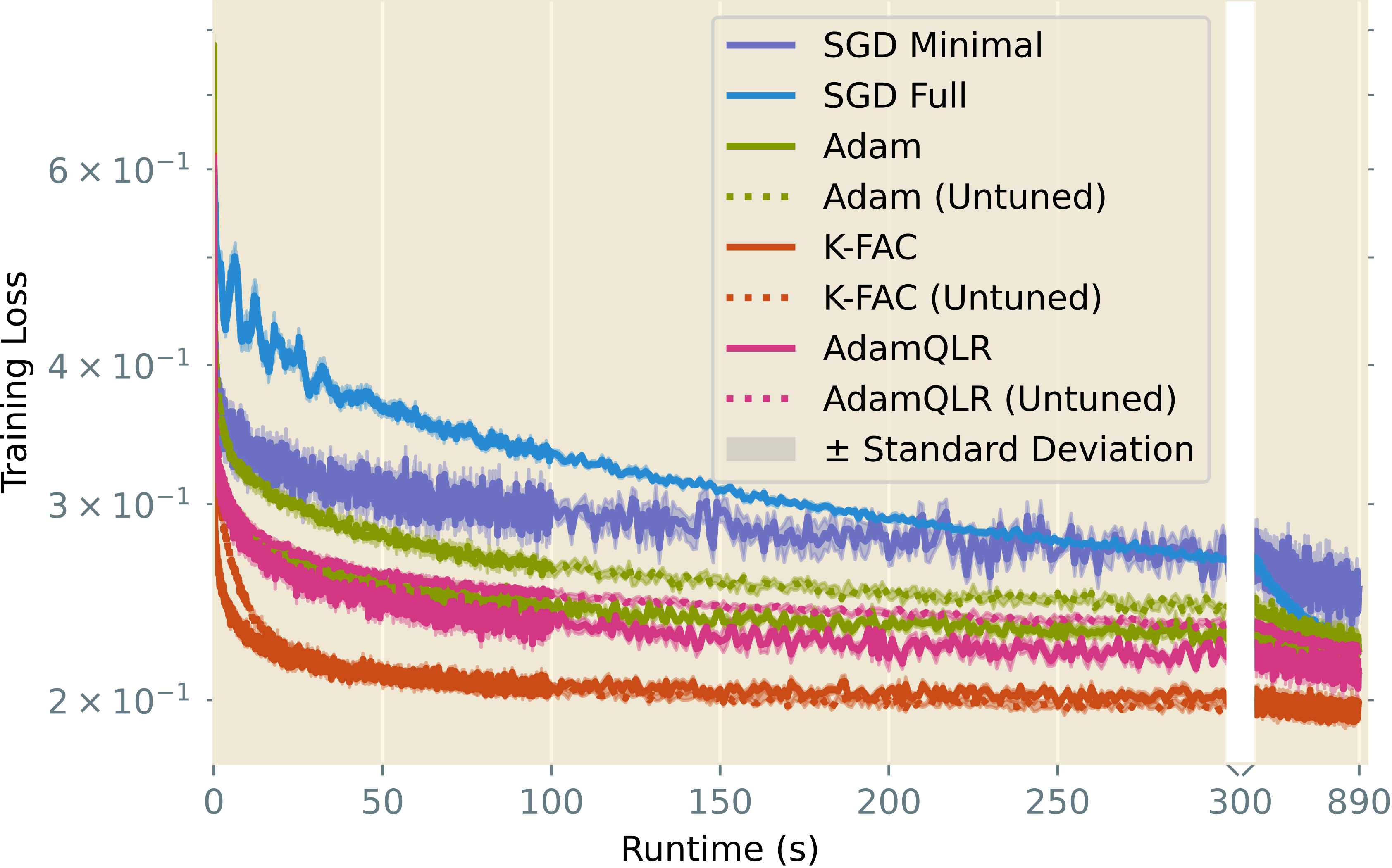}
        & \includegraphics[align=c,width=\linewidth]{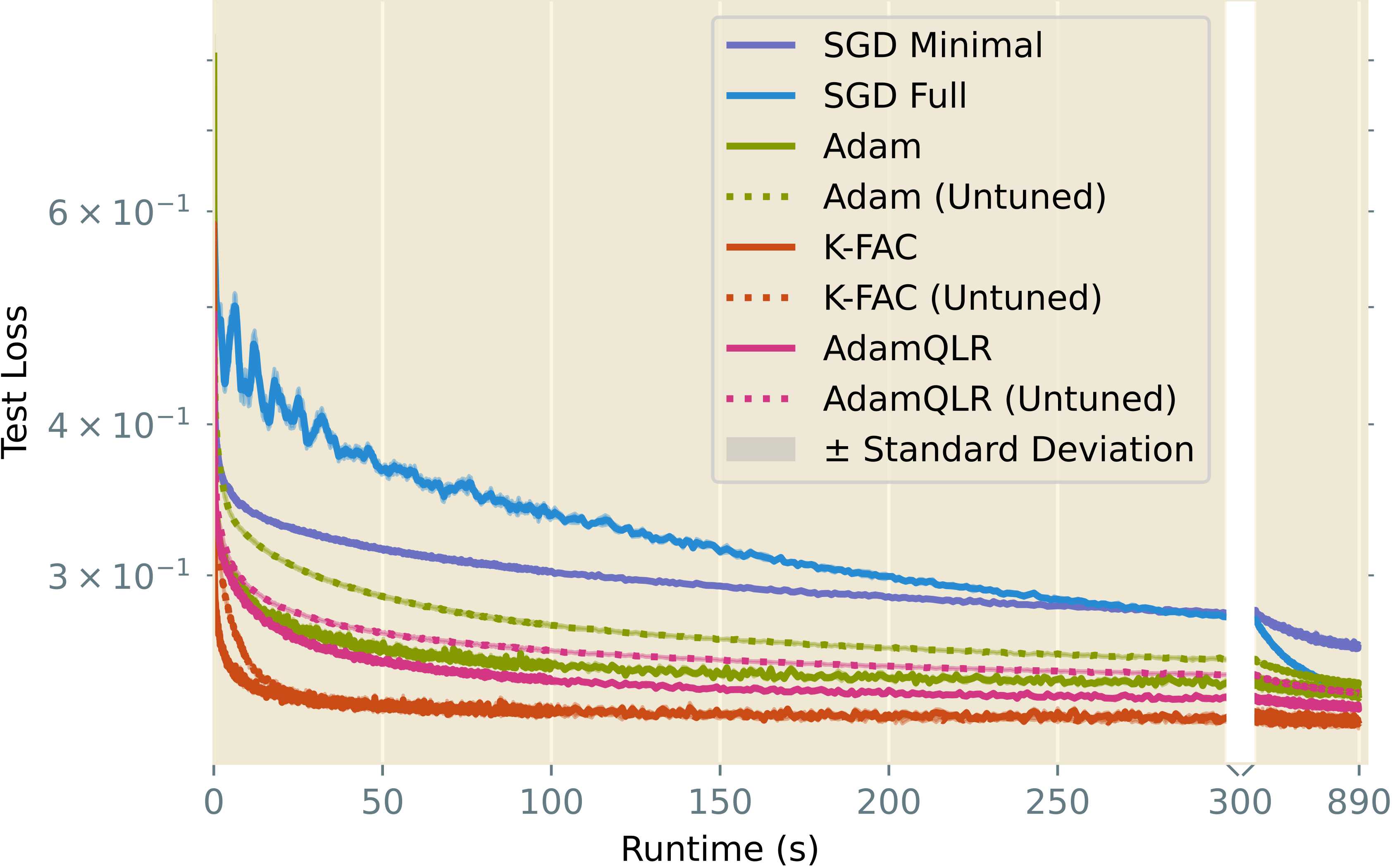} \vfill\\
        
        \rotatecaption{\subcaption{Fashion-MNIST}}
        & \includegraphics[align=c,width=\linewidth]{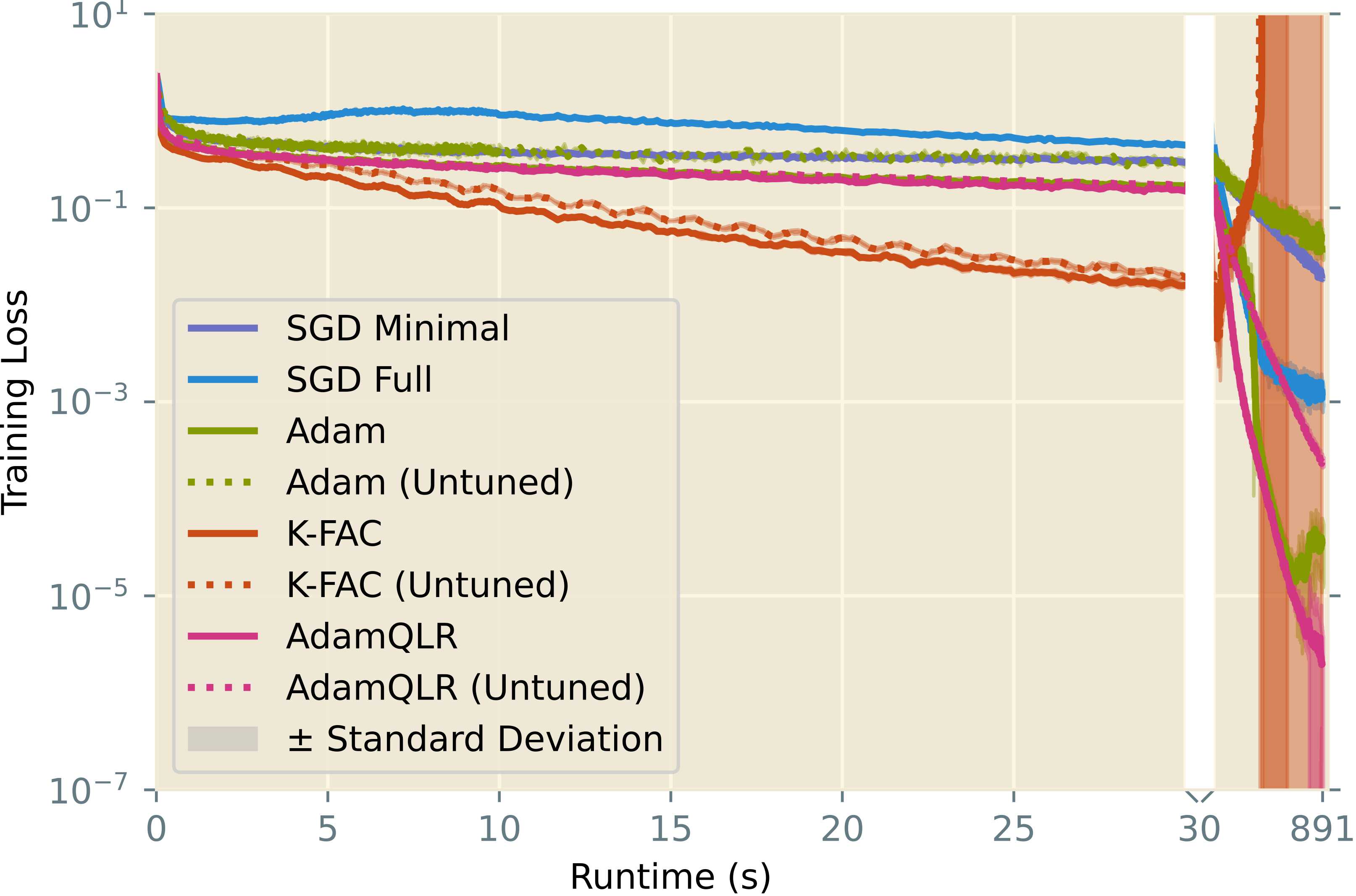}
        & \includegraphics[align=c,width=\linewidth]{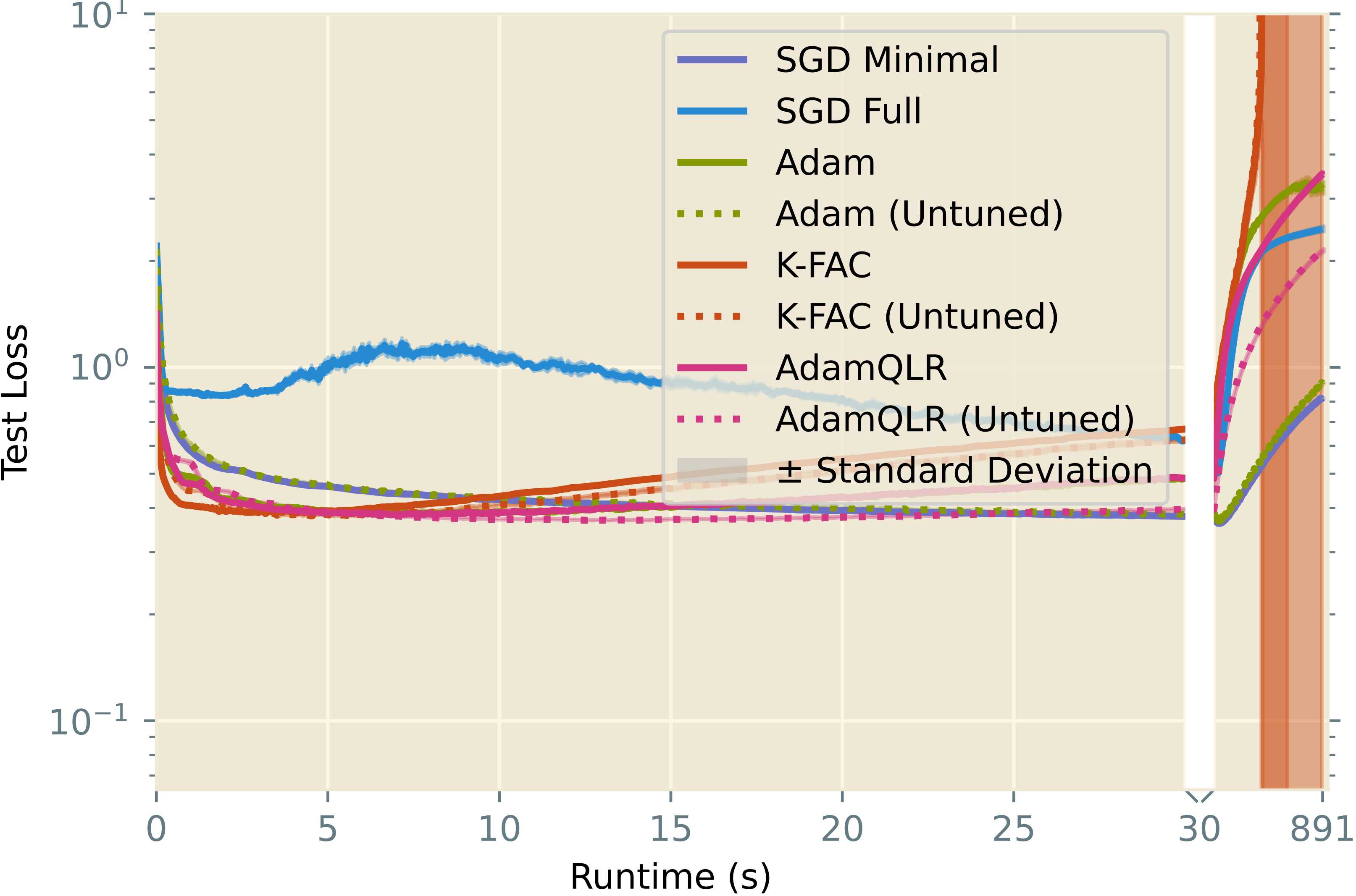} \vfill\\
        
        \rotatecaption{\subcaption{SVHN}}
        & \includegraphics[align=c,width=\linewidth]{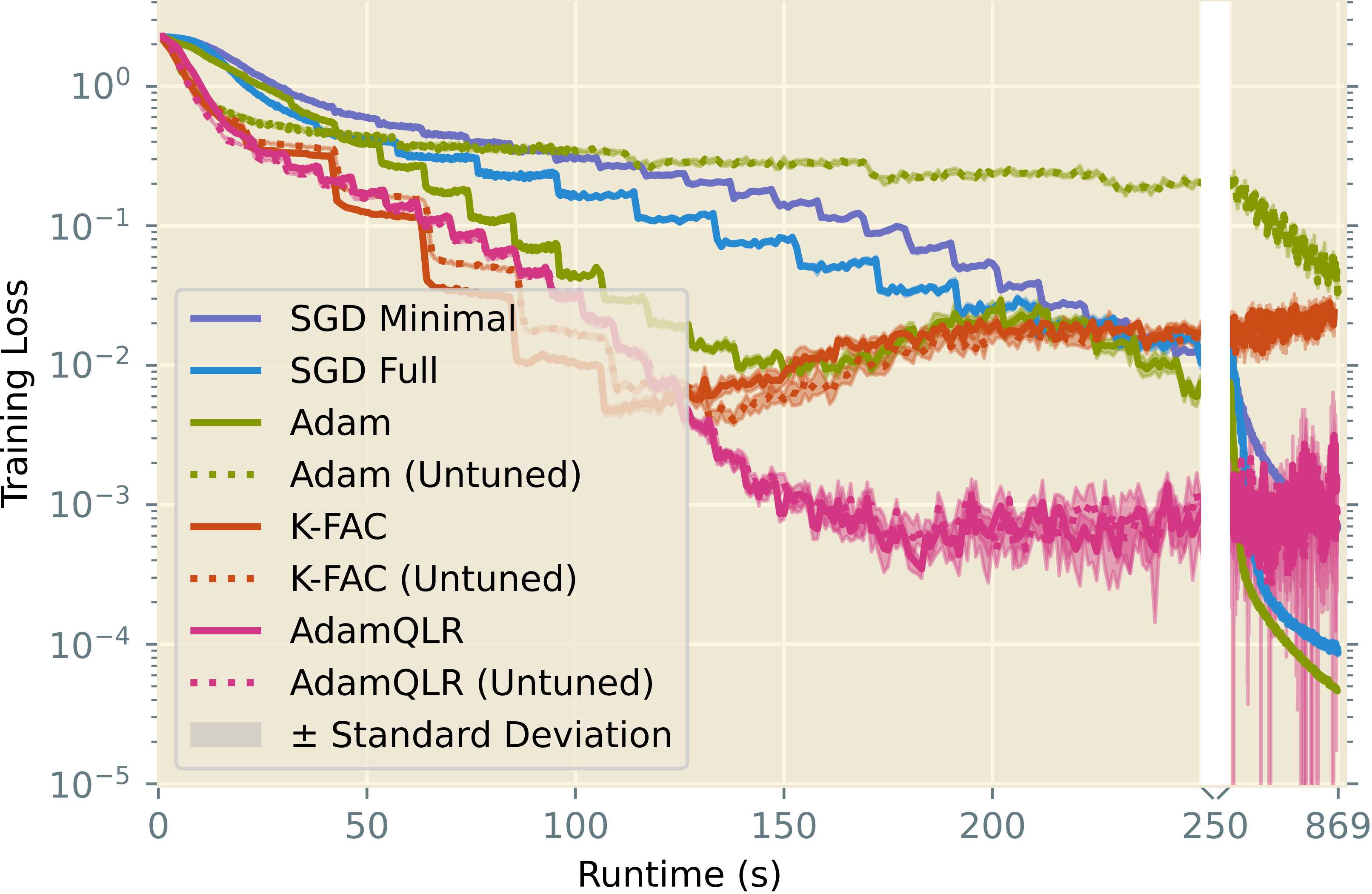}
        & \includegraphics[align=c,width=\linewidth]{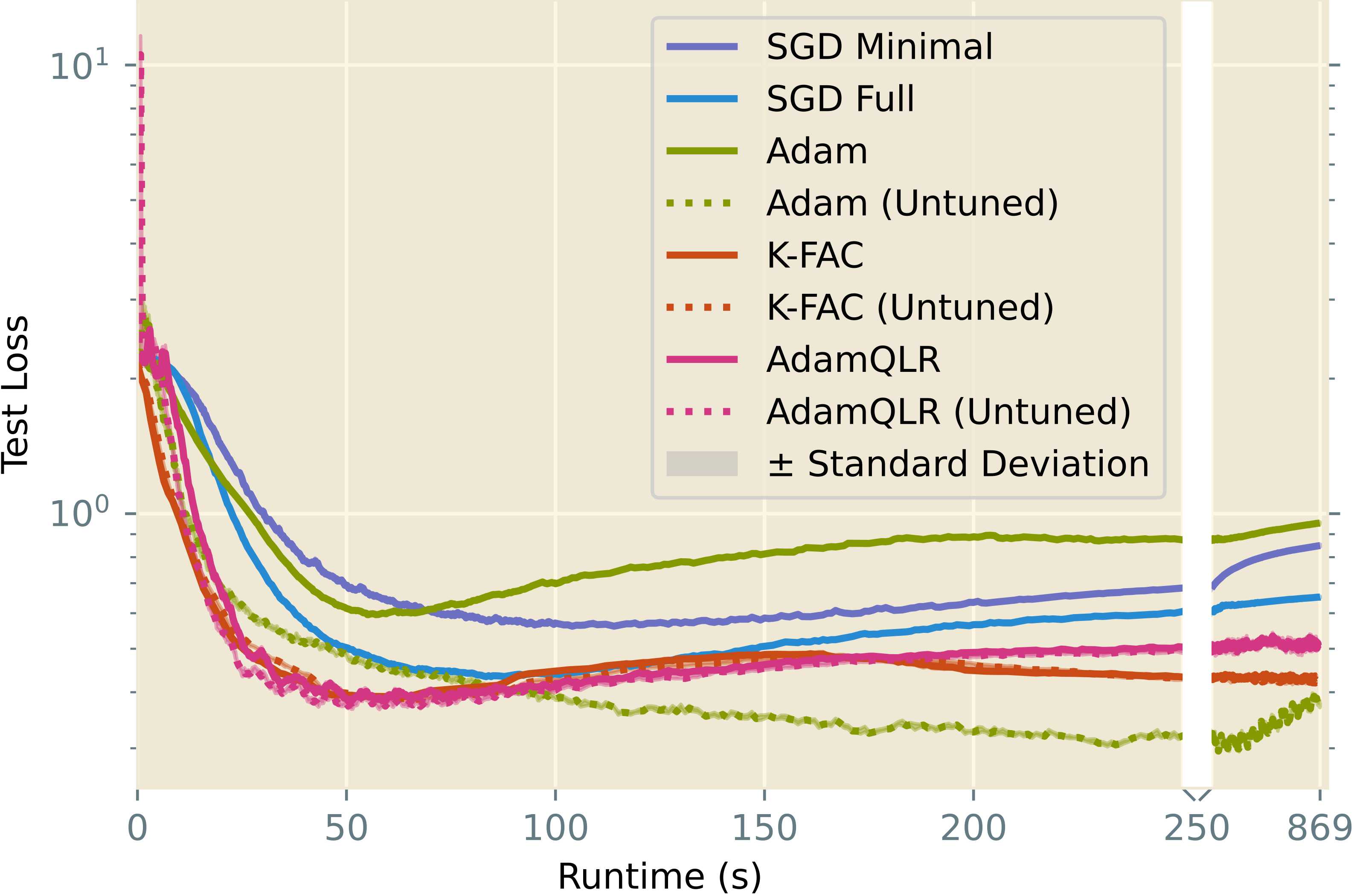} \vfill\\
        
        \rotatecaption{\subcaption{CIFAR-10}}
        & \includegraphics[align=c,width=\linewidth]{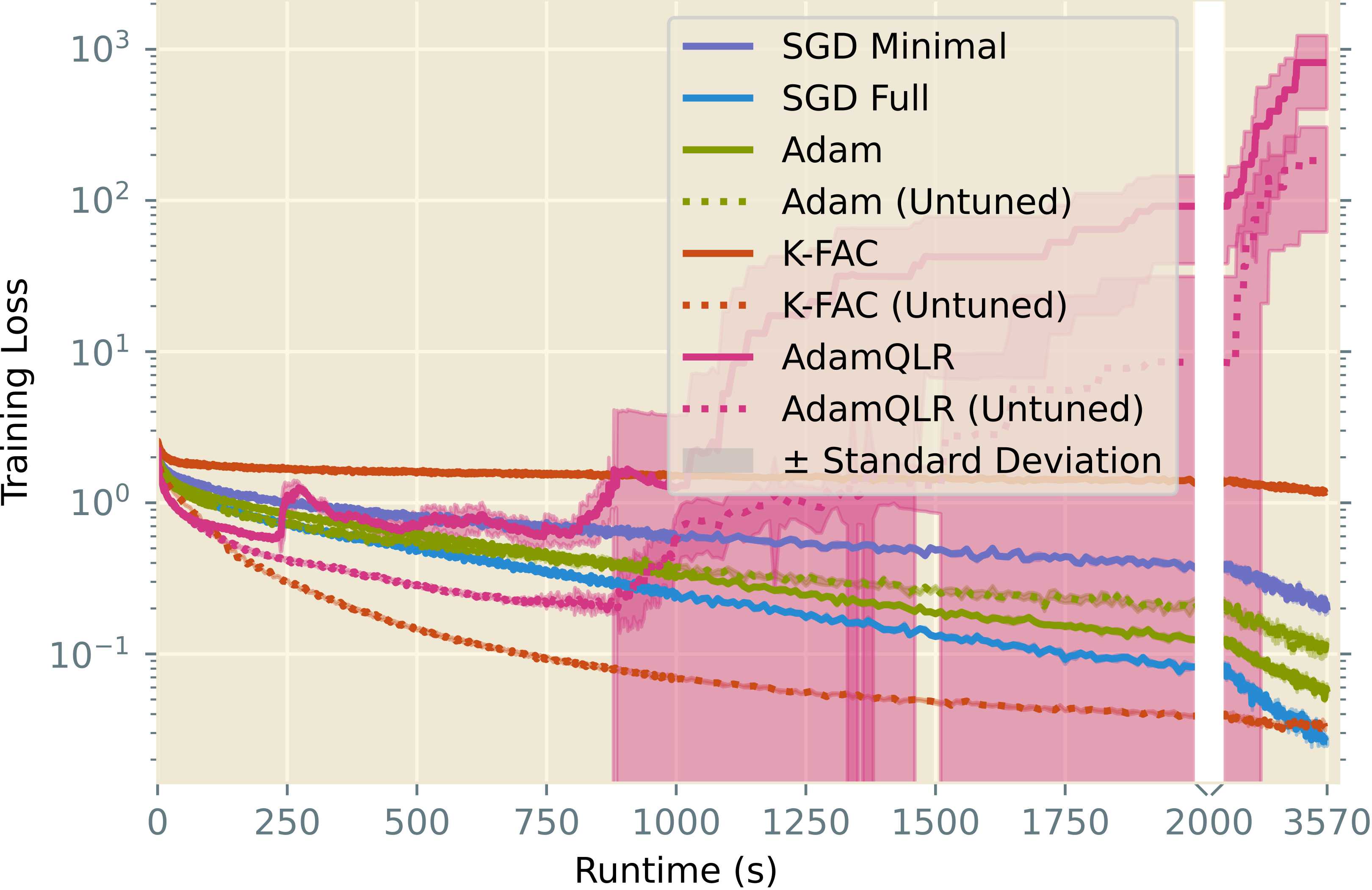}
        & \includegraphics[align=c,width=\linewidth]{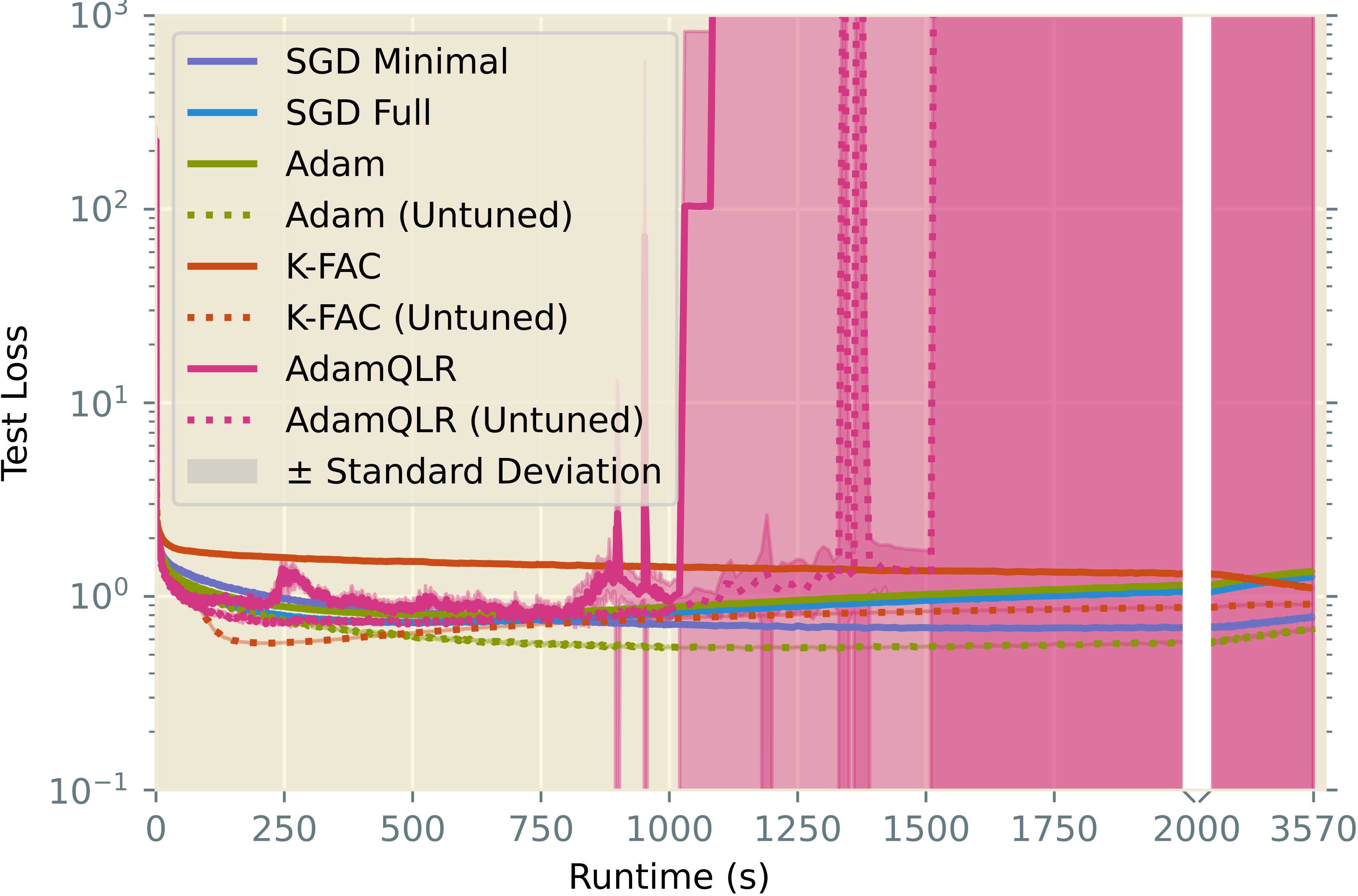} \vfill\\
    \end{tabularx}
    \caption{Median training (left) and test (right) loss trajectories, bootstrap-sampled over 50 repetitions per algorithm. Hyperparameters chosen by ASHA over 200 initialisations to minimise \emph{training} loss after a fixed runtime of 15~minutes, characterising the naïve power of each algorithm. Note changes of scale on the time axes. See also our numerical comparison in Table~\ref{tab:ASHATimeTrainingFinalResults}}
    \label{fig:ASHATimeTraining}
\end{figure*}

Our main results in Section~\ref{sec:Experiments} impose a primary constraint of a fixed number of epochs, with a secondary constraint of a runtime limit. Further, since our hyperparameter optimisation sought a minimal validation loss, the training loss evolutions display an early-stopping-like behaviour in which they do not fully converge. To develop additional context on AdamQLR's performance, we repeat these experiments without the primary number-of-epochs constraint, such that our hyperparameter tuning directly optimises for the best loss attained after the 15~minute runtime limit, and the algorithms are evaluated on the same metric. Figure~\ref{fig:ASHATimeValidation} and Table~\ref{tab:ASHATimeValidationFinalResults} show results where we optimised for final validation loss, while Figure~\ref{fig:ASHATimeTraining} and Table~\ref{tab:ASHATimeTrainingFinalResults} show results where the hyperparameters were optimised to minimise final \emph{training} loss. This latter setting allows us to compare the naïve power of each optimiser to optimise the given objective in isolation.

These results display an interesting tendency for \emph{K-FAC} to wildly diverge in the later phases of training on Fashion-MNIST, an effect which \emph{AdamQLR} is largely able to avoid. Broadly speaking, \emph{AdamQLR} gives competitive generalisation performance on UCI~Energy and UCI~Protein in Figure~\ref{fig:ASHATimeValidation}, with a more pronounced overfitting behaviour on larger datasets. However, on CIFAR-10 \emph{AdamQLR (Tuned)} diverges severely. We additionally see an effective demonstration of \emph{AdamQLR}'s optimisation power in Figure~\ref{fig:ASHATimeTraining} --- although training performance on Fashion-MNIST again lags behind \emph{Adam} in this setting, larger datasets achieve particularly strong training loss evolutions, though CIFAR-10 remains unstable.

\begin{table*}
    \centering
    \caption{Numerical study of the results shown in Figure~\ref{fig:ASHATimeValidation}: final statistics after runtime-constrained training on our benchmark tasks, with hyperparameters optimised to minimise \emph{validation} loss.}
    \label{tab:ASHATimeValidationFinalResults}
    \resizebox{\linewidth}{!}{
    \begin{tabular}{cc
        S[table-format=3.7]
        U
        S[table-format=1.5]
        U
        S[table-format=1.6]
        U
        S[table-format=1.5]
        U
        S[table-format=1.6]
        U
        S[table-format=5.1]
        U
        S[table-format=4.3]
        U
    }
        \toprule
        Dataset & Algorithm & \multicolumn{2}{c}{Training Loss} & \multicolumn{2}{c}{Training Accuracy} & \multicolumn{2}{c}{Test Loss} & \multicolumn{2}{c}{Test Accuracy} & \multicolumn{2}{c}{Generalisation Gap} & \multicolumn{2}{c}{Total Steps} & \multicolumn{2}{c}{Total Time (s)} \\
        \midrule
        % \midrule
\multirow{8}{*}{UCI Energy}
& SGD Minimal 	& 0.000559 & $\pm$ \num{0.000017} 	& \multicolumn{2}{c}{---}	& 0.001133 & $\pm$ \num{0.000033} 	& \multicolumn{2}{c}{---}	& 0.000574 & $\pm$ \num{0.000049} 	& 67771 & $\pm$ \num{33} 	& 891.225 & $\pm$ \num{0.023} \\
& SGD Full 	& 0.000384 & $\pm$ \num{0.000028} 	& \multicolumn{2}{c}{---}	& 0.001000 & $\pm$ \num{0.000051} 	& \multicolumn{2}{c}{---}	& 0.000616 & $\pm$ \num{0.000079} 	& 79380 & $\pm$ \num{67} 	& 889.581 & $\pm$ \num{0.054} \\
& Adam 	& 0.0001903 & $\pm$ \num{0.0000052} 	& \multicolumn{2}{c}{---}	& 0.001004 & $\pm$ \num{0.000043} 	& \multicolumn{2}{c}{---}	& 0.000814 & $\pm$ \num{0.000049} 	& 38460 & $\pm$ \num{110} 	& 889.118 & $\pm$ \num{0.070} \\
& Adam (Untuned) 	& 0.0001849 & $\pm$ \num{0.0000060} 	& \multicolumn{2}{c}{---}	& 0.001053 & $\pm$ \num{0.000051} 	& \multicolumn{2}{c}{---}	& 0.000868 & $\pm$ \num{0.000057} 	& 45570 & $\pm$ \num{220} 	& 889.297 & $\pm$ \num{0.053} \\
& K-FAC 	& 0.0001034 & $\pm$ \num{0.0000069} 	& \multicolumn{2}{c}{---}	& 0.000895 & $\pm$ \num{0.000040} 	& \multicolumn{2}{c}{---}	& 0.000791 & $\pm$ \num{0.000047} 	& 67983 & $\pm$ \num{70} 	& 887.048 & $\pm$ \num{0.057} \\
& K-FAC (Untuned) 	& 0.000311 & $\pm$ \num{0.000014} 	& \multicolumn{2}{c}{---}	& 0.001071 & $\pm$ \num{0.000044} 	& \multicolumn{2}{c}{---}	& 0.000760 & $\pm$ \num{0.000059} 	& 1011 & $\pm$ \num{28} 	& 63.7 & $\pm$ \num{1.7} \\
& AdamQLR 	& 0.0003808 & $\pm$ \num{0.0000090} 	& \multicolumn{2}{c}{---}	& 0.000832 & $\pm$ \num{0.000028} 	& \multicolumn{2}{c}{---}	& 0.000451 & $\pm$ \num{0.000037} 	& 36463 & $\pm$ \num{31} 	& 887.179 & $\pm$ \num{0.061} \\
& AdamQLR (Untuned) 	& \num{1.9e29} & $\pm$ \num{2.9e28} 	& \multicolumn{2}{c}{---}	& \num{1.7e29} & $\pm$ \num{3.3e28} 	& \multicolumn{2}{c}{---}	& \num{-2.2e28} & $\pm$ \num{6.2e28} 	& 175.0 & $\pm$ \num{2.8} 	& 10.63 & $\pm$ \num{0.19} \\
\midrule
\multirow{8}{*}{UCI Protein}
& SGD Minimal 	& 0.2595 & $\pm$ \num{0.0017} 	& \multicolumn{2}{c}{---}	& 0.2696 & $\pm$ \num{0.0024} 	& \multicolumn{2}{c}{---}	& 0.0101 & $\pm$ \num{0.0041} 	& 13849 & $\pm$ \num{61} 	& 887.052 & $\pm$ \num{0.031} \\
& SGD Full 	& 0.2227 & $\pm$ \num{0.0042} 	& \multicolumn{2}{c}{---}	& 0.23898 & $\pm$ \num{0.00080} 	& \multicolumn{2}{c}{---}	& 0.0163 & $\pm$ \num{0.0050} 	& 54190 & $\pm$ \num{120} 	& 886.931 & $\pm$ \num{0.032} \\
& Adam 	& 0.2211 & $\pm$ \num{0.0014} 	& \multicolumn{2}{c}{---}	& 0.24047 & $\pm$ \num{0.00072} 	& \multicolumn{2}{c}{---}	& 0.0194 & $\pm$ \num{0.0021} 	& 24520 & $\pm$ \num{130} 	& 886.767 & $\pm$ \num{0.052} \\
& Adam (Untuned) 	& 0.2264 & $\pm$ \num{0.0024} 	& \multicolumn{2}{c}{---}	& 0.24404 & $\pm$ \num{0.00060} 	& \multicolumn{2}{c}{---}	& 0.0176 & $\pm$ \num{0.0030} 	& 34000 & $\pm$ \num{220} 	& 886.737 & $\pm$ \num{0.054} \\
& K-FAC 	& 0.1951 & $\pm$ \num{0.0020} 	& \multicolumn{2}{c}{---}	& 0.22597 & $\pm$ \num{0.00069} 	& \multicolumn{2}{c}{---}	& 0.0309 & $\pm$ \num{0.0027} 	& 22466 & $\pm$ \num{90} 	& 884.757 & $\pm$ \num{0.089} \\
& K-FAC (Untuned) 	& 0.1932 & $\pm$ \num{0.0013} 	& \multicolumn{2}{c}{---}	& 0.2251 & $\pm$ \num{0.0012} 	& \multicolumn{2}{c}{---}	& 0.0320 & $\pm$ \num{0.0025} 	& 12890 & $\pm$ \num{48} 	& 884.557 & $\pm$ \num{0.074} \\
& AdamQLR 	& 0.2147 & $\pm$ \num{0.0026} 	& \multicolumn{2}{c}{---}	& 0.23418 & $\pm$ \num{0.00015} 	& \multicolumn{2}{c}{---}	& 0.0195 & $\pm$ \num{0.0028} 	& 38623 & $\pm$ \num{57} 	& 884.319 & $\pm$ \num{0.096} \\
& AdamQLR (Untuned) 	& 0.2218 & $\pm$ \num{0.0010} 	& \multicolumn{2}{c}{---}	& 0.24002 & $\pm$ \num{0.00062} 	& \multicolumn{2}{c}{---}	& 0.0182 & $\pm$ \num{0.0016} 	& 14089 & $\pm$ \num{20} 	& 886.631 & $\pm$ \num{0.068} \\
\midrule
\multirow{8}{*}{Fashion-MNIST}
& SGD Minimal 	& 0.0282 & $\pm$ \num{0.0013} 	& 0.9942 & $\pm$ \num{0.0011} 	& 0.7486 & $\pm$ \num{0.0029} 	& 0.86706 & $\pm$ \num{0.00038} 	& 0.7204 & $\pm$ \num{0.0043} 	& 44110 & $\pm$ \num{200} 	& 886.24 & $\pm$ \num{0.21} \\
& SGD Full 	& 0.2351 & $\pm$ \num{0.0028} 	& 0.9195 & $\pm$ \num{0.0011} 	& 0.35519 & $\pm$ \num{0.00057} 	& 0.87504 & $\pm$ \num{0.00019} 	& 0.1201 & $\pm$ \num{0.0034} 	& 20940 & $\pm$ \num{29} 	& 886.380 & $\pm$ \num{0.052} \\
& Adam 	& 0.000996 & $\pm$ \num{0.000023} 	& 1.0 & $\pm$ \num{0} 	& 1.618 & $\pm$ \num{0.013} 	& 0.86059 & $\pm$ \num{0.00042} 	& 1.617 & $\pm$ \num{0.013} 	& 12073 & $\pm$ \num{24} 	& 886.212 & $\pm$ \num{0.081} \\
& Adam (Untuned) 	& 0.0345 & $\pm$ \num{0.0038} 	& 0.9936 & $\pm$ \num{0.0092} 	& 0.9026 & $\pm$ \num{0.0038} 	& 0.86793 & $\pm$ \num{0.00047} 	& 0.8681 & $\pm$ \num{0.0076} 	& 62801 & $\pm$ \num{88} 	& 884.692 & $\pm$ \num{0.087} \\
& K-FAC 	& \num{3.9e9} & $\pm$ \num{3.3e9} 	& 0.045 & $\pm$ \num{0.017} 	& \num{3.2e9} & $\pm$ \num{2.7e9} 	& 0.10 & $\pm$ \num{0} 	& \num{-7.0e8} & $\pm$ \num{6.0e9} 	& 1888 & $\pm$ \num{48} 	& 69.8 & $\pm$ \num{1.7} \\
& K-FAC (Untuned) 	& \num{4.0e13} & $\pm$ \num{1.9e14} 	& 0.084 & $\pm$ \num{0.047} 	& \num{5.0e13} & $\pm$ \num{2.2e14} 	& 0.10 & $\pm$ \num{0} 	& \num{1.0e13} & $\pm$ \num{4.1e14} 	& 5120 & $\pm$ \num{240} 	& 425 & $\pm$ \num{22} \\
& AdamQLR 	& 0.000188 & $\pm$ \num{0.000025} 	& 1.0 & $\pm$ \num{0} 	& 2.137 & $\pm$ \num{0.026} 	& 0.85740 & $\pm$ \num{0.00043} 	& 2.137 & $\pm$ \num{0.026} 	& 11907 & $\pm$ \num{47} 	& 883.45 & $\pm$ \num{0.14} \\
& AdamQLR (Untuned) 	& 0.000231 & $\pm$ \num{0.000020} 	& 1.0 & $\pm$ \num{0} 	& 2.143 & $\pm$ \num{0.022} 	& 0.85780 & $\pm$ \num{0.00061} 	& 2.142 & $\pm$ \num{0.022} 	& 11934 & $\pm$ \num{34} 	& 886.09 & $\pm$ \num{0.19} \\
\midrule
\multirow{8}{*}{SVHN}
& SGD Minimal 	& 0.0344 & $\pm$ \num{0.0035} 	& 0.98995 & $\pm$ \num{0.00080} 	& 0.5214 & $\pm$ \num{0.0015} 	& 0.88478 & $\pm$ \num{0.00043} 	& 0.4870 & $\pm$ \num{0.0050} 	& 3551 & $\pm$ \num{16} 	& 869.955 & $\pm$ \num{0.059} \\
& SGD Full 	& 0.000055 & $\pm$ \num{0.000015} 	& 1.0 & $\pm$ \num{0} 	& 0.6038 & $\pm$ \num{0.0047} 	& 0.91360 & $\pm$ \num{0.00042} 	& 0.6037 & $\pm$ \num{0.0047} 	& 3282 & $\pm$ \num{12} 	& 867.930 & $\pm$ \num{0.079} \\
& Adam 	& 0.0681 & $\pm$ \num{0.0066} 	& 0.9788 & $\pm$ \num{0.0025} 	& 0.3845 & $\pm$ \num{0.0028} 	& 0.91066 & $\pm$ \num{0.00067} 	& 0.3164 & $\pm$ \num{0.0094} 	& 3516 & $\pm$ \num{13} 	& 868.24 & $\pm$ \num{0.16} \\
& Adam (Untuned) 	& 0.0345 & $\pm$ \num{0.0024} 	& 0.9887 & $\pm$ \num{0.0011} 	& 0.3737 & $\pm$ \num{0.0023} 	& 0.91863 & $\pm$ \num{0.00042} 	& 0.3392 & $\pm$ \num{0.0047} 	& 3635 & $\pm$ \num{25} 	& 867.911 & $\pm$ \num{0.078} \\
& K-FAC 	& 0.0534 & $\pm$ \num{0.0032} 	& 0.99207 & $\pm$ \num{0.00083} 	& 0.4658 & $\pm$ \num{0.0011} 	& 0.86104 & $\pm$ \num{0.00040} 	& 0.4125 & $\pm$ \num{0.0043} 	& 1296.0 & $\pm$ \num{4.9} 	& 848.48 & $\pm$ \num{0.14} \\
& K-FAC (Untuned) 	& 0.0212 & $\pm$ \num{0.0020} 	& 0.99717 & $\pm$ \num{0.00041} 	& 0.4200 & $\pm$ \num{0.0025} 	& 0.88035 & $\pm$ \num{0.00058} 	& 0.3988 & $\pm$ \num{0.0045} 	& 770.0 & $\pm$ \num{5.6} 	& 843.65 & $\pm$ \num{0.33} \\
& AdamQLR 	& 0.00073 & $\pm$ \num{0.00041} 	& 0.99982 & $\pm$ \num{0.00015} 	& 0.5068 & $\pm$ \num{0.0091} 	& 0.91898 & $\pm$ \num{0.00084} 	& 0.5061 & $\pm$ \num{0.0095} 	& 2184.0 & $\pm$ \num{7.8} 	& 860.16 & $\pm$ \num{0.19} \\
& AdamQLR (Untuned) 	& 0.00108 & $\pm$ \num{0.00053} 	& 0.99972 & $\pm$ \num{0.00023} 	& 0.5068 & $\pm$ \num{0.0082} 	& 0.91854 & $\pm$ \num{0.00097} 	& 0.5058 & $\pm$ \num{0.0088} 	& 2168 & $\pm$ \num{11} 	& 862.534 & $\pm$ \num{0.076} \\
\midrule
\multirow{8}{*}{CIFAR-10}
& SGD Minimal 	& 0.0400 & $\pm$ \num{0.0028} 	& 0.9868 & $\pm$ \num{0.0010} 	& 1.2623 & $\pm$ \num{0.0053} 	& 0.77728 & $\pm$ \num{0.00036} 	& 1.2222 & $\pm$ \num{0.0082} 	& 26157 & $\pm$ \num{70} 	& 3567.73 & $\pm$ \num{0.20} \\
& SGD Full 	& 0.00743 & $\pm$ \num{0.00057} 	& 0.99752 & $\pm$ \num{0.00019} 	& 1.6225 & $\pm$ \num{0.0021} 	& 0.78408 & $\pm$ \num{0.00071} 	& 1.6151 & $\pm$ \num{0.0027} 	& 14950 & $\pm$ \num{29} 	& 3564.70 & $\pm$ \num{0.21} \\
& Adam 	& 0.0274 & $\pm$ \num{0.0017} 	& 0.98987 & $\pm$ \num{0.00036} 	& 1.4217 & $\pm$ \num{0.0048} 	& 0.7760 & $\pm$ \num{0.0010} 	& 1.3943 & $\pm$ \num{0.0064} 	& 20855 & $\pm$ \num{28} 	& 3565.16 & $\pm$ \num{0.13} \\
& Adam (Untuned) 	& 0.1175 & $\pm$ \num{0.0077} 	& 0.9555 & $\pm$ \num{0.0042} 	& 0.6830 & $\pm$ \num{0.0031} 	& 0.8434 & $\pm$ \num{0.0010} 	& 0.566 & $\pm$ \num{0.011} 	& 31420 & $\pm$ \num{180} 	& 3566.670 & $\pm$ \num{0.059} \\
& K-FAC 	& 0.792 & $\pm$ \num{0.015} 	& 0.7256 & $\pm$ \num{0.0042} 	& 0.7967 & $\pm$ \num{0.0032} 	& 0.72517 & $\pm$ \num{0.00084} 	& 0.005 & $\pm$ \num{0.018} 	& 8609 & $\pm$ \num{34} 	& 3548.41 & $\pm$ \num{0.18} \\
& K-FAC (Untuned) 	& 0.0325 & $\pm$ \num{0.0012} 	& 0.99196 & $\pm$ \num{0.00037} 	& 0.9141 & $\pm$ \num{0.0044} 	& 0.80656 & $\pm$ \num{0.00052} 	& 0.8816 & $\pm$ \num{0.0055} 	& 3097 & $\pm$ \num{16} 	& 3541.10 & $\pm$ \num{0.14} \\
& AdamQLR 	& 630 & $\pm$ \num{170} 	& 0.1283 & $\pm$ \num{0.0018} 	& \num{1.4e12} & $\pm$ \num{2.7e12} 	& 0.10 & $\pm$ \num{0} 	& \num{1.4e12} & $\pm$ \num{2.7e12} 	& 4100 & $\pm$ \num{760} 	& 1120 & $\pm$ \num{210} \\
& AdamQLR (Untuned) 	& 190 & $\pm$ \num{120} 	& 0.171 & $\pm$ \num{0.043} 	& \num{9.0e10} & $\pm$ \num{4.5e11} 	& 0.13 & $\pm$ \num{0.12} 	& \num{9.0e10} & $\pm$ \num{4.5e11} 	& 5700 & $\pm$ \num{1000} 	& 2450 & $\pm$ \num{450} \\
\bottomrule
    \end{tabular}
    }
\end{table*}

\begin{table*}
    \centering
    \caption{Numerical study of the results shown in Figure~\ref{fig:ASHATimeTraining}: final statistics after runtime-constrained training on our benchmark tasks, with hyperparameters optimised to minimise \emph{training} loss.}
    \label{tab:ASHATimeTrainingFinalResults}
    \resizebox{\linewidth}{!}{
    \begin{tabular}{cc
        S[table-format=3.7]
        U
        S[table-format=1.5]
        U
        S[table-format=1.6]
        U
        S[table-format=1.5]
        U
        S[table-format=1.6]
        U
        S[table-format=5.2]
        U
        S[table-format=4.3]
        U
    }
        \toprule
        Dataset & Algorithm & \multicolumn{2}{c}{Training Loss} & \multicolumn{2}{c}{Training Accuracy} & \multicolumn{2}{c}{Test Loss} & \multicolumn{2}{c}{Test Accuracy} & \multicolumn{2}{c}{Generalisation Gap} & \multicolumn{2}{c}{Total Steps} & \multicolumn{2}{c}{Total Time (s)} \\
        \midrule
        % \midrule
\multirow{8}{*}{UCI Energy}
& SGD Minimal 	& 0.000458 & $\pm$ \num{0.000015} 	& \multicolumn{2}{c}{---}	& 0.001029 & $\pm$ \num{0.000028} 	& \multicolumn{2}{c}{---}	& 0.000571 & $\pm$ \num{0.000044} 	& 68023 & $\pm$ \num{29} 	& 891.103 & $\pm$ \num{0.017} \\
& SGD Full 	& 0.000355 & $\pm$ \num{0.000013} 	& \multicolumn{2}{c}{---}	& 0.000947 & $\pm$ \num{0.000052} 	& \multicolumn{2}{c}{---}	& 0.000592 & $\pm$ \num{0.000065} 	& 53935 & $\pm$ \num{25} 	& 889.305 & $\pm$ \num{0.049} \\
& Adam 	& 0.000232 & $\pm$ \num{0.000011} 	& \multicolumn{2}{c}{---}	& 0.000952 & $\pm$ \num{0.000035} 	& \multicolumn{2}{c}{---}	& 0.000720 & $\pm$ \num{0.000046} 	& 13486.0 & $\pm$ \num{6.3} 	& 888.054 & $\pm$ \num{0.063} \\
& Adam (Untuned) 	& 0.0001853 & $\pm$ \num{0.0000074} 	& \multicolumn{2}{c}{---}	& 0.001064 & $\pm$ \num{0.000051} 	& \multicolumn{2}{c}{---}	& 0.000879 & $\pm$ \num{0.000059} 	& 45640 & $\pm$ \num{190} 	& 889.305 & $\pm$ \num{0.064} \\
& K-FAC 	& 0.0002710 & $\pm$ \num{0.0000077} 	& \multicolumn{2}{c}{---}	& 0.001055 & $\pm$ \num{0.000032} 	& \multicolumn{2}{c}{---}	& 0.000784 & $\pm$ \num{0.000040} 	& 971 & $\pm$ \num{25} 	& 23.47 & $\pm$ \num{0.46} \\
& K-FAC (Untuned) 	& 0.000308 & $\pm$ \num{0.000014} 	& \multicolumn{2}{c}{---}	& 0.001075 & $\pm$ \num{0.000043} 	& \multicolumn{2}{c}{---}	& 0.000767 & $\pm$ \num{0.000057} 	& 1018 & $\pm$ \num{28} 	& 64.0 & $\pm$ \num{2.0} \\
& AdamQLR 	& 0.0002183 & $\pm$ \num{0.0000048} 	& \multicolumn{2}{c}{---}	& 0.000912 & $\pm$ \num{0.000025} 	& \multicolumn{2}{c}{---}	& 0.000693 & $\pm$ \num{0.000030} 	& 38402 & $\pm$ \num{63} 	& 889.021 & $\pm$ \num{0.051} \\
& AdamQLR (Untuned) 	& \num{1.9e29} & $\pm$ \num{3.2e28} 	& \multicolumn{2}{c}{---}	& \num{1.7e29} & $\pm$ \num{2.8e28} 	& \multicolumn{2}{c}{---}	& \num{-2.3e28} & $\pm$ \num{6.0e28} 	& 175.0 & $\pm$ \num{3.1} 	& 10.62 & $\pm$ \num{0.19} \\
\midrule
\multirow{8}{*}{UCI Protein}
& SGD Minimal 	& 0.2535 & $\pm$ \num{0.0091} 	& \multicolumn{2}{c}{---}	& 0.26285 & $\pm$ \num{0.00088} 	& \multicolumn{2}{c}{---}	& 0.0094 & $\pm$ \num{0.0099} 	& 75471 & $\pm$ \num{52} 	& 886.875 & $\pm$ \num{0.094} \\
& SGD Full 	& 0.2214 & $\pm$ \num{0.0014} 	& \multicolumn{2}{c}{---}	& 0.24121 & $\pm$ \num{0.00066} 	& \multicolumn{2}{c}{---}	& 0.0198 & $\pm$ \num{0.0020} 	& 13813.0 & $\pm$ \num{9.3} 	& 886.756 & $\pm$ \num{0.057} \\
& Adam 	& 0.2209 & $\pm$ \num{0.0013} 	& \multicolumn{2}{c}{---}	& 0.23747 & $\pm$ \num{0.00067} 	& \multicolumn{2}{c}{---}	& 0.0166 & $\pm$ \num{0.0020} 	& 39106 & $\pm$ \num{20} 	& 886.605 & $\pm$ \num{0.056} \\
& Adam (Untuned) 	& 0.2266 & $\pm$ \num{0.0030} 	& \multicolumn{2}{c}{---}	& 0.24419 & $\pm$ \num{0.00058} 	& \multicolumn{2}{c}{---}	& 0.0176 & $\pm$ \num{0.0036} 	& 34000 & $\pm$ \num{200} 	& 886.724 & $\pm$ \num{0.047} \\
& K-FAC 	& 0.1994 & $\pm$ \num{0.0018} 	& \multicolumn{2}{c}{---}	& 0.2262 & $\pm$ \num{0.0012} 	& \multicolumn{2}{c}{---}	& 0.0268 & $\pm$ \num{0.0030} 	& 22381 & $\pm$ \num{18} 	& 884.61 & $\pm$ \num{0.11} \\
& K-FAC (Untuned) 	& 0.1927 & $\pm$ \num{0.0013} 	& \multicolumn{2}{c}{---}	& 0.2254 & $\pm$ \num{0.0014} 	& \multicolumn{2}{c}{---}	& 0.0327 & $\pm$ \num{0.0027} 	& 12907 & $\pm$ \num{47} 	& 884.588 & $\pm$ \num{0.083} \\
& AdamQLR 	& 0.2109 & $\pm$ \num{0.0026} 	& \multicolumn{2}{c}{---}	& 0.23311 & $\pm$ \num{0.00055} 	& \multicolumn{2}{c}{---}	& 0.0223 & $\pm$ \num{0.0032} 	& 54825 & $\pm$ \num{37} 	& 886.587 & $\pm$ \num{0.060} \\
& AdamQLR (Untuned) 	& 0.2215 & $\pm$ \num{0.0010} 	& \multicolumn{2}{c}{---}	& 0.23981 & $\pm$ \num{0.00057} 	& \multicolumn{2}{c}{---}	& 0.0183 & $\pm$ \num{0.0016} 	& 14088 & $\pm$ \num{26} 	& 886.631 & $\pm$ \num{0.061} \\
\midrule
\multirow{8}{*}{Fashion-MNIST}
& SGD Minimal 	& 0.01921 & $\pm$ \num{0.00083} 	& 0.99763 & $\pm$ \num{0.00052} 	& 0.8154 & $\pm$ \num{0.0046} 	& 0.86604 & $\pm$ \num{0.00059} 	& 0.7962 & $\pm$ \num{0.0054} 	& 43837 & $\pm$ \num{51} 	& 881.95 & $\pm$ \num{0.69} \\
& SGD Full 	& 0.00119 & $\pm$ \num{0.00027} 	& 0.99978 & $\pm$ \num{0.00014} 	& 2.492 & $\pm$ \num{0.051} 	& 0.85416 & $\pm$ \num{0.00054} 	& 2.490 & $\pm$ \num{0.051} 	& 12106.0 & $\pm$ \num{5.3} 	& 886.384 & $\pm$ \num{0.070} \\
& Adam 	& 0.000034 & $\pm$ \num{0.000021} 	& 1.0 & $\pm$ \num{0} 	& 3.26 & $\pm$ \num{0.14} 	& 0.86524 & $\pm$ \num{0.00048} 	& 3.26 & $\pm$ \num{0.14} 	& 20870 & $\pm$ \num{46} 	& 885.83 & $\pm$ \num{0.10} \\
& Adam (Untuned) 	& 0.0342 & $\pm$ \num{0.0031} 	& 0.9948 & $\pm$ \num{0.0086} 	& 0.9023 & $\pm$ \num{0.0038} 	& 0.86801 & $\pm$ \num{0.00036} 	& 0.8681 & $\pm$ \num{0.0070} 	& 62787 & $\pm$ \num{88} 	& 884.72 & $\pm$ \num{0.12} \\
& K-FAC 	& \num{9.0e13} & $\pm$ \num{2.8e14} 	& 0.038 & $\pm$ \num{0.010} 	& 1.1e14 & $\pm$ \num{3.9e14} 	& 0.10 & $\pm$ \num{0} 	& \num{2.0e13} & $\pm$ \num{6.7e14} 	& 5130 & $\pm$ \num{110} 	& 437 & $\pm$ \num{13} \\
& K-FAC (Untuned) 	& \num{2.7e13} & $\pm$ \num{8.2e13} 	& 0.096 & $\pm$ \num{0.061} 	& \num{1.3e14} & $\pm$ \num{4.4e14} 	& 0.10 & $\pm$ \num{0} 	& \num{1.1e14} & $\pm$ \num{5.2e14} 	& 5170 & $\pm$ \num{240} 	& 433 & $\pm$ \num{24} \\
& AdamQLR 	& 0.0000019 & $\pm$ \num{0.0000027} 	& 1.0 & $\pm$ \num{0} 	& 3.500 & $\pm$ \num{0.055} 	& 0.86091 & $\pm$ \num{0.00087} 	& 3.500 & $\pm$ \num{0.055} 	& 20872 & $\pm$ \num{47} 	& 885.711 & $\pm$ \num{0.072} \\
& AdamQLR (Untuned) 	& 0.000239 & $\pm$ \num{0.000025} 	& 1.0 & $\pm$ \num{0} 	& 2.148 & $\pm$ \num{0.024} 	& 0.85797 & $\pm$ \num{0.00061} 	& 2.148 & $\pm$ \num{0.024} 	& 11937 & $\pm$ \num{35} 	& 886.09 & $\pm$ \num{0.22} \\
\midrule
\multirow{8}{*}{SVHN}
& SGD Minimal 	& 0.000694 & $\pm$ \num{0.000019} 	& 1.0 & $\pm$ \num{0} 	& 0.8492 & $\pm$ \num{0.0058} 	& 0.83171 & $\pm$ \num{0.00044} 	& 0.8485 & $\pm$ \num{0.0058} 	& 3190.00 & $\pm$ \num{0.28} 	& 867.716 & $\pm$ \num{0.069} \\
& SGD Full 	& 0.0000895 & $\pm$ \num{0.0000040} 	& 1.0 & $\pm$ \num{0} 	& 0.6520 & $\pm$ \num{0.0025} 	& 0.88326 & $\pm$ \num{0.00038} 	& 0.6519 & $\pm$ \num{0.0025} 	& 3479.00 & $\pm$ \num{0.47} 	& 867.876 & $\pm$ \num{0.089} \\
& Adam 	& 0.00004722 & $\pm$ \num{0.00000095} 	& 1.0 & $\pm$ \num{0} 	& 0.9528 & $\pm$ \num{0.0053} 	& 0.84417 & $\pm$ \num{0.00090} 	& 0.9528 & $\pm$ \num{0.0053} 	& 3154.00 & $\pm$ \num{0.51} 	& 865.862 & $\pm$ \num{0.078} \\
& Adam (Untuned) 	& 0.0349 & $\pm$ \num{0.0023} 	& 0.9888 & $\pm$ \num{0.0014} 	& 0.3743 & $\pm$ \num{0.0026} 	& 0.91850 & $\pm$ \num{0.00045} 	& 0.3394 & $\pm$ \num{0.0050} 	& 3641 & $\pm$ \num{25} 	& 867.911 & $\pm$ \num{0.068} \\
& K-FAC 	& 0.0254 & $\pm$ \num{0.0021} 	& 0.99702 & $\pm$ \num{0.00036} 	& 0.4344 & $\pm$ \num{0.0026} 	& 0.8775 & $\pm$ \num{0.0011} 	& 0.4090 & $\pm$ \num{0.0047} 	& 786.00 & $\pm$ \num{0.34} 	& 844.76 & $\pm$ \num{0.18} \\
& K-FAC (Untuned) 	& 0.0208 & $\pm$ \num{0.0015} 	& 0.99721 & $\pm$ \num{0.00035} 	& 0.4202 & $\pm$ \num{0.0032} 	& 0.88047 & $\pm$ \num{0.00062} 	& 0.3994 & $\pm$ \num{0.0047} 	& 770.0 & $\pm$ \num{5.4} 	& 843.69 & $\pm$ \num{0.37} \\
& AdamQLR 	& 0.0016 & $\pm$ \num{0.0011} 	& 0.99961 & $\pm$ \num{0.00018} 	& 0.5103 & $\pm$ \num{0.0082} 	& 0.91582 & $\pm$ \num{0.00069} 	& 0.5088 & $\pm$ \num{0.0093} 	& 2180.0 & $\pm$ \num{6.2} 	& 860.22 & $\pm$ \num{0.15} \\
& AdamQLR (Untuned) 	& 0.00099 & $\pm$ \num{0.00059} 	& 0.99974 & $\pm$ \num{0.00021} 	& 0.5074 & $\pm$ \num{0.0060} 	& 0.9185 & $\pm$ \num{0.0010} 	& 0.5064 & $\pm$ \num{0.0066} 	& 2167 & $\pm$ \num{12} 	& 862.534 & $\pm$ \num{0.092} \\
\midrule
\multirow{8}{*}{CIFAR-10}
& SGD Minimal 	& 0.1939 & $\pm$ \num{0.0081} 	& 0.9262 & $\pm$ \num{0.0057} 	& 0.7827 & $\pm$ \num{0.0034} 	& 0.7961 & $\pm$ \num{0.0010} 	& 0.589 & $\pm$ \num{0.012} 	& 33701 & $\pm$ \num{21} 	& 3568.63 & $\pm$ \num{0.10} \\
& SGD Full 	& 0.0264 & $\pm$ \num{0.0014} 	& 0.99245 & $\pm$ \num{0.00035} 	& 1.2644 & $\pm$ \num{0.0064} 	& 0.79274 & $\pm$ \num{0.00050} 	& 1.2380 & $\pm$ \num{0.0078} 	& 26676 & $\pm$ \num{12} 	& 3566.73 & $\pm$ \num{0.13} \\
& Adam 	& 0.0559 & $\pm$ \num{0.0039} 	& 0.9817 & $\pm$ \num{0.0019} 	& 1.3471 & $\pm$ \num{0.0034} 	& 0.75620 & $\pm$ \num{0.00057} 	& 1.2912 & $\pm$ \num{0.0073} 	& 26696 & $\pm$ \num{11} 	& 3565.90 & $\pm$ \num{0.11} \\
& Adam (Untuned) 	& 0.1192 & $\pm$ \num{0.0058} 	& 0.9561 & $\pm$ \num{0.0036} 	& 0.6822 & $\pm$ \num{0.0036} 	& 0.8433 & $\pm$ \num{0.0011} 	& 0.5630 & $\pm$ \num{0.0094} 	& 31390 & $\pm$ \num{180} 	& 3566.676 & $\pm$ \num{0.069} \\
& K-FAC 	& 1.180 & $\pm$ \num{0.019} 	& 0.5787 & $\pm$ \num{0.0082} 	& 1.1067 & $\pm$ \num{0.0080} 	& 0.6116 & $\pm$ \num{0.0020} 	& -0.074 & $\pm$ \num{0.027} 	& 9660.0 & $\pm$ \num{1.7} 	& 3548.50 & $\pm$ \num{0.13} \\
& K-FAC (Untuned) 	& 0.0327 & $\pm$ \num{0.0016} 	& 0.99203 & $\pm$ \num{0.00034} 	& 0.9137 & $\pm$ \num{0.0043} 	& 0.80652 & $\pm$ \num{0.00049} 	& 0.8810 & $\pm$ \num{0.0059} 	& 3100 & $\pm$ \num{14} 	& 3541.06 & $\pm$ \num{0.17} \\
& AdamQLR 	& 820 & $\pm$ \num{420} 	& 0.1268 & $\pm$ \num{0.0028} 	& \num{7.0e12} & $\pm$ \num{1.7e13} 	& 0.10 & $\pm$ \num{0} 	& \num{7.0e12} & $\pm$ \num{1.7e13} 	& 5130 & $\pm$ \num{960} 	& 1390 & $\pm$ \num{260} \\
& AdamQLR (Untuned) 	& 180 & $\pm$ \num{120} 	& 0.174 & $\pm$ \num{0.048} 	& \num{8.0e10} & $\pm$ \num{5.5e11} 	& 0.123 & $\pm$ \num{0.088} 	& \num{8.0e10} & $\pm$ \num{5.5e11} 	& 5770 & $\pm$ \num{970} 	& 2490 & $\pm$ \num{420} \\
\bottomrule
    \end{tabular}
    }
\end{table*}

\subsection{Sensitivity Studies}
\label{sec:SensitivityExperiments}
To justify our configurations and further demonstrate the utility of our algorithm, we conduct a range of sensitivity experiments for \emph{AdamQLR (Tuned)} trained on Fashion-MNIST under the same conditions as in Section~\ref{sec:ExperimentsFashionMNIST}. All hyperparameters except for the one under investigation are fixed at the best values found for ASHA in those experiments. Again, our plots show the averages of median trends of bootstrap-sampled sets of 50 repetitions for each configuration considered.

\subsubsection{Learning Rate Rescaling}
\begin{figure*}
    \centering
    \includegraphics[width=0.45\linewidth]{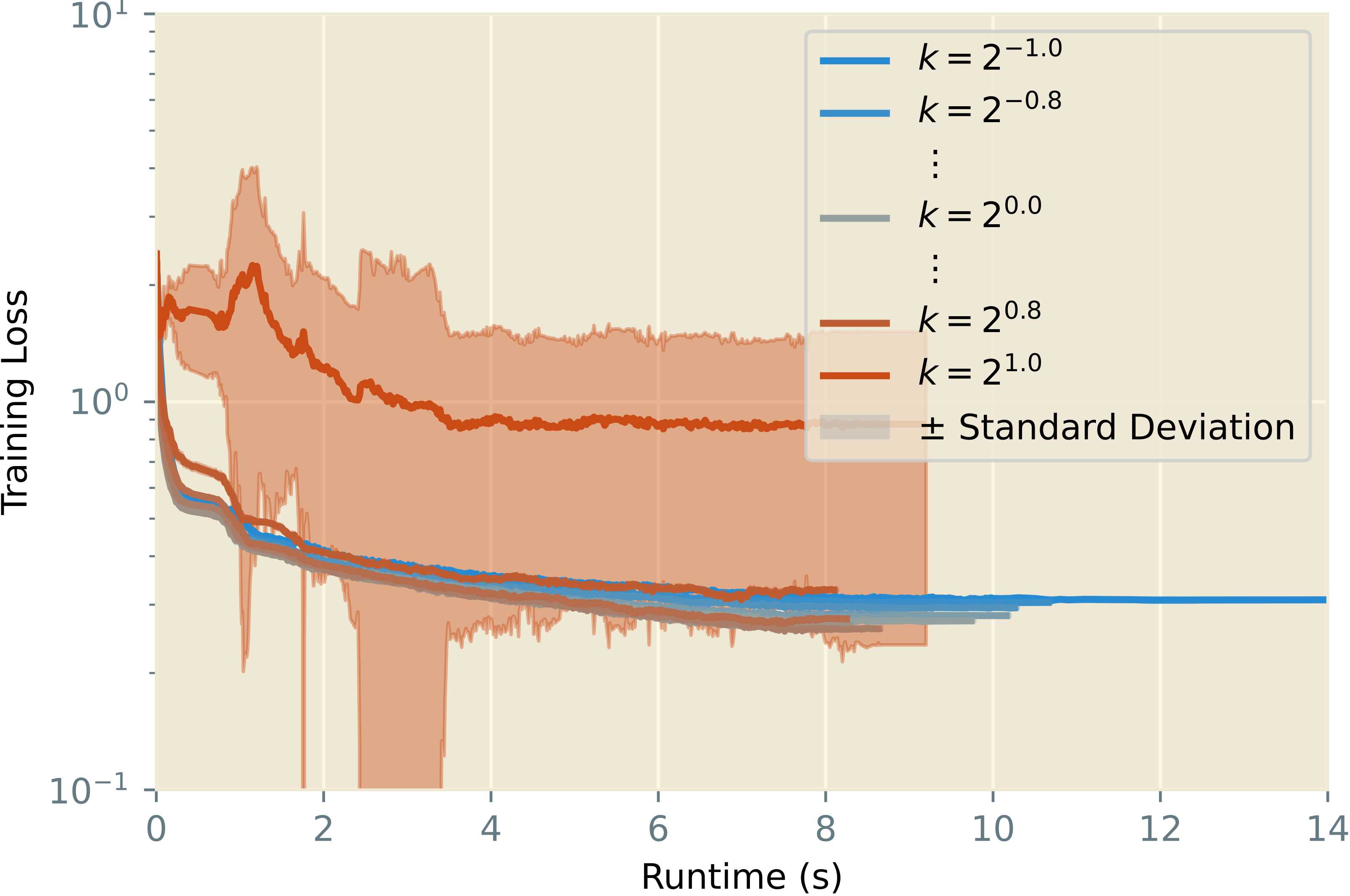}
    % \hfill
    \hspace*{0.01\linewidth}
    \includegraphics[width=0.45\linewidth]{Figures/Sensitivity_Amplification_TestLoss.pdf.jpg}
    \caption{Sensitivity studies over learning rate, which is scaled by a variety of constant factors $k$ for our Fashion-MNIST trial from Section~\ref{sec:ExperimentsFashionMNIST}.}
    \label{fig:AblationLearningRate}
\end{figure*}

Firstly, we analyse the accuracy of our learning rate selection strategy by executing our algorithm as normal, but setting $\alpha \gets k \alpha$ for each $k$ in $\{2^{-1.0}, 2^{-0.8}, 2^{-0.6}, \cdots, 2^{1.0}\}$. In effect, we investigate the potential for systemic bias in our learning rate selection by asking if our results would improve with a constant scaling factor on those learning rates. 

Our results in Figure~\ref{fig:AblationLearningRate} show the $k=2^{1.0}$ case exhibiting large variance due to unstable runs, while the best training losses are obtained for $k$ slightly larger than unity. This makes sense given our use of damping: if stability can be achieved without damping for any given update, then the damping will serve only to downsize our proposed update step, so we should expect the best results to be obtained by slightly increasing it again. However, test loss appears generally less sensitive to $k$, with the lowest value obtained for $k=1$: this would also be expected under damping, since we would hope the damping would increase generalisation performance. In aggregate, these results confirm our approach accurately selects the correct learning rate to use for any given optimisation step.

\subsubsection{Initial Damping}
\begin{figure*}
    \centering
    \includegraphics[width=0.45\linewidth]{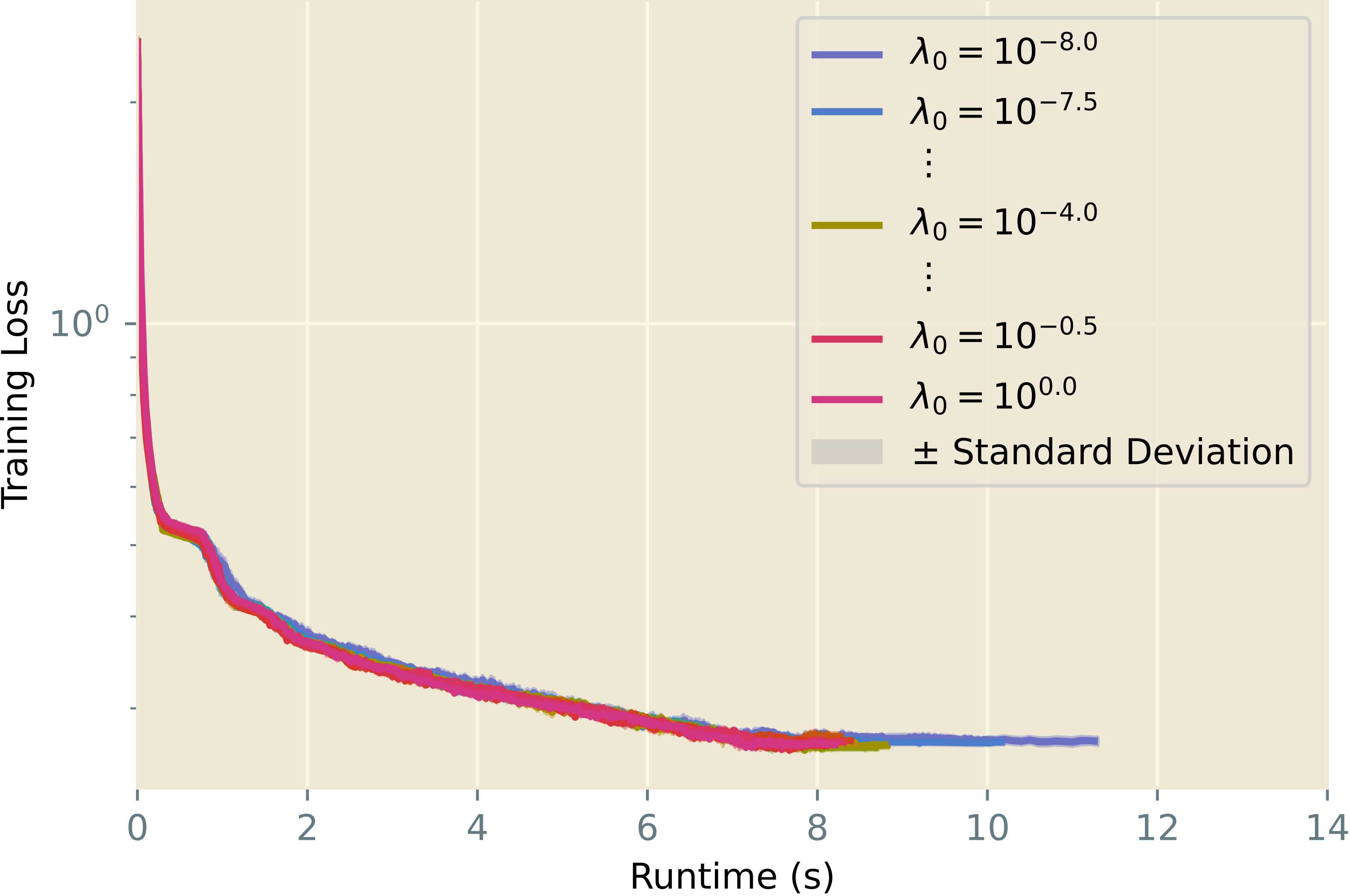}
    % \hfill
    \hspace*{0.01\linewidth}
    \includegraphics[width=0.45\linewidth]{Figures/Sensitivity_InitialDamping_TestLoss.pdf.jpg}
    \caption{Sensitivity studies over initial damping value $\lambda_0$ for our Fashion-MNIST trial from Section~\ref{sec:ExperimentsFashionMNIST}.}
    \label{fig:AblationInitialDamping}
\end{figure*}

Next, we consider the initial value $\lambda_0$ assigned to our Levenberg-Marquardt damping term $\lambda$, testing values in $\{10^{-8.0}, 10^{-7.5}, 10^{-7.0}, \cdots, 10^{0.0}\}$. Here, we seek to quantify the trade-off between damping's stabilising effect and its tendency to worsen training loss. Figure~\ref{fig:AblationInitialDamping} presents our results.

With the exception of the very smallest values, we see our performance is largely insensitive to $\lambda_0$. This matches our empirical observation that damping becomes most important for larger-scale problems than our Fashion-MNIST setting, and thus has minimal effect here. However, given its substantial importance in these more complex experiments, it is reassuring that the inclusion of damping does not dramatically worsen performance when its influence is not required.

\subsubsection{Batch Size}
\begin{figure*}
    \centering
    \includegraphics[width=0.45\linewidth]{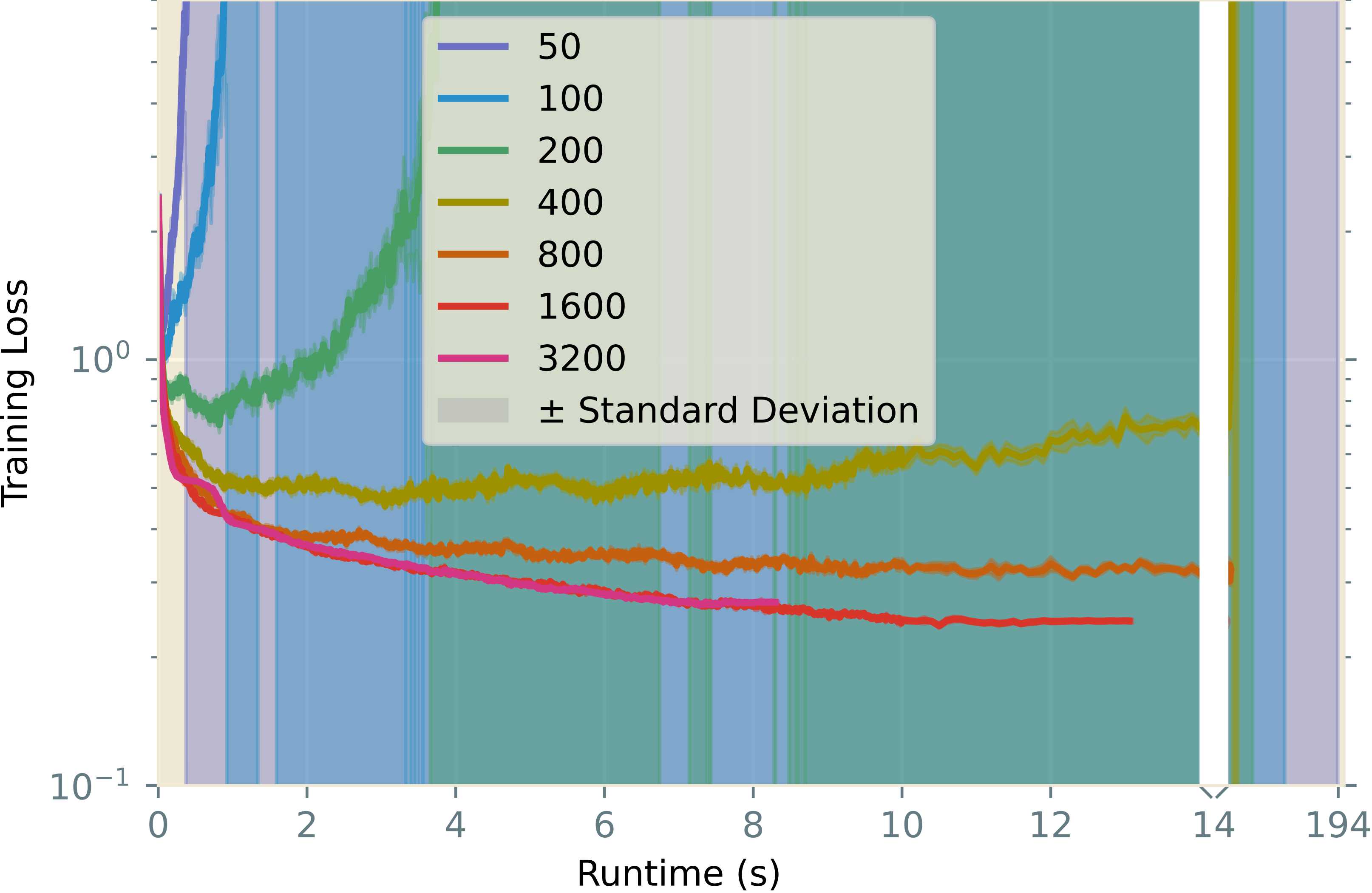}
    % \hfill
    \hspace*{0.01\linewidth}
    \includegraphics[width=0.45\linewidth]{Figures/Sensitivity_BatchSize_TestLoss.pdf.jpg}
    \caption{Sensitivity studies over batch size for our Fashion-MNIST trial from Section~\ref{sec:ExperimentsFashionMNIST}.}
    \label{fig:AblationBatchSize}
\end{figure*}

In Figure~\ref{fig:AblationBatchSize}, we consider each batch size available to ASHA in Section~\ref{sec:ExperimentsFashionMNIST} ($\{50, 100, 200, 400, 800, 1\,600, 3\,200\}$) to investigate the effect of this hyperparameter on our algorithm.

Since the optimal batch size selected by ASHA for \emph{AdamQLR} was generally large (3\,200 in this case), it is perhaps unsurprising that we see divergence from smaller batches. This also matches our intuition: unlike classical first-order methods, \emph{AdamQLR} uses each batch to (implicitly) construct a full curvature matrix for the optimisation surface, which magnifies the importance of having a low-bias sample of the training data. Empirically, we found the computational benefits of fewer batches outweighed the increased cost of computing each batch, so this preference for larger batch sizes aligns with our desire to minimise runtime. Thus, our results show a clear trend that larger batch sizes give greater training and generalisation performance.

\subsubsection{Damping Stepping Factor}
\begin{figure*}
    \centering
    \includegraphics[width=0.45\linewidth]{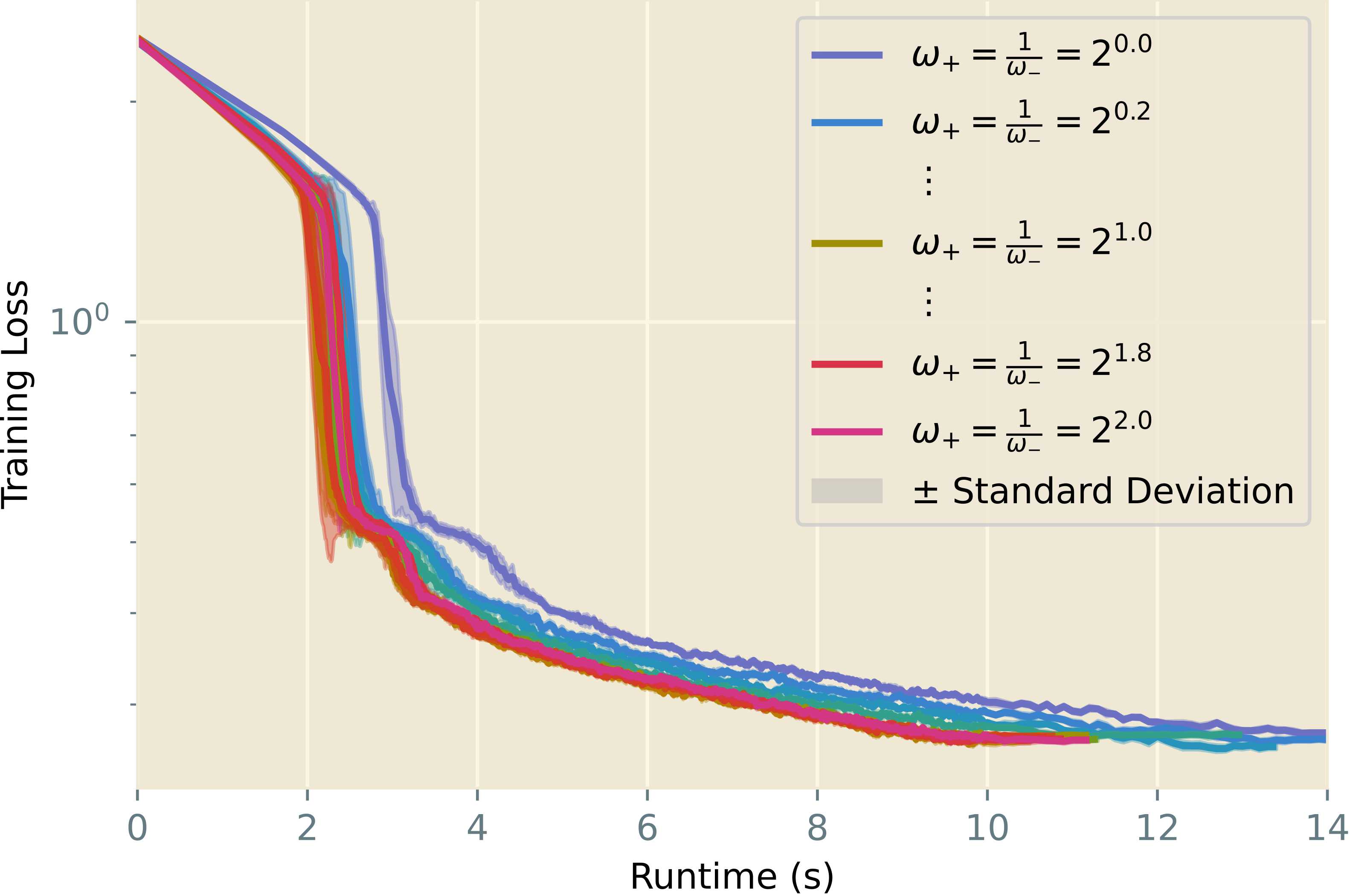}
    % \hfill
    \hspace*{0.01\linewidth}
    \includegraphics[width=0.45\linewidth]{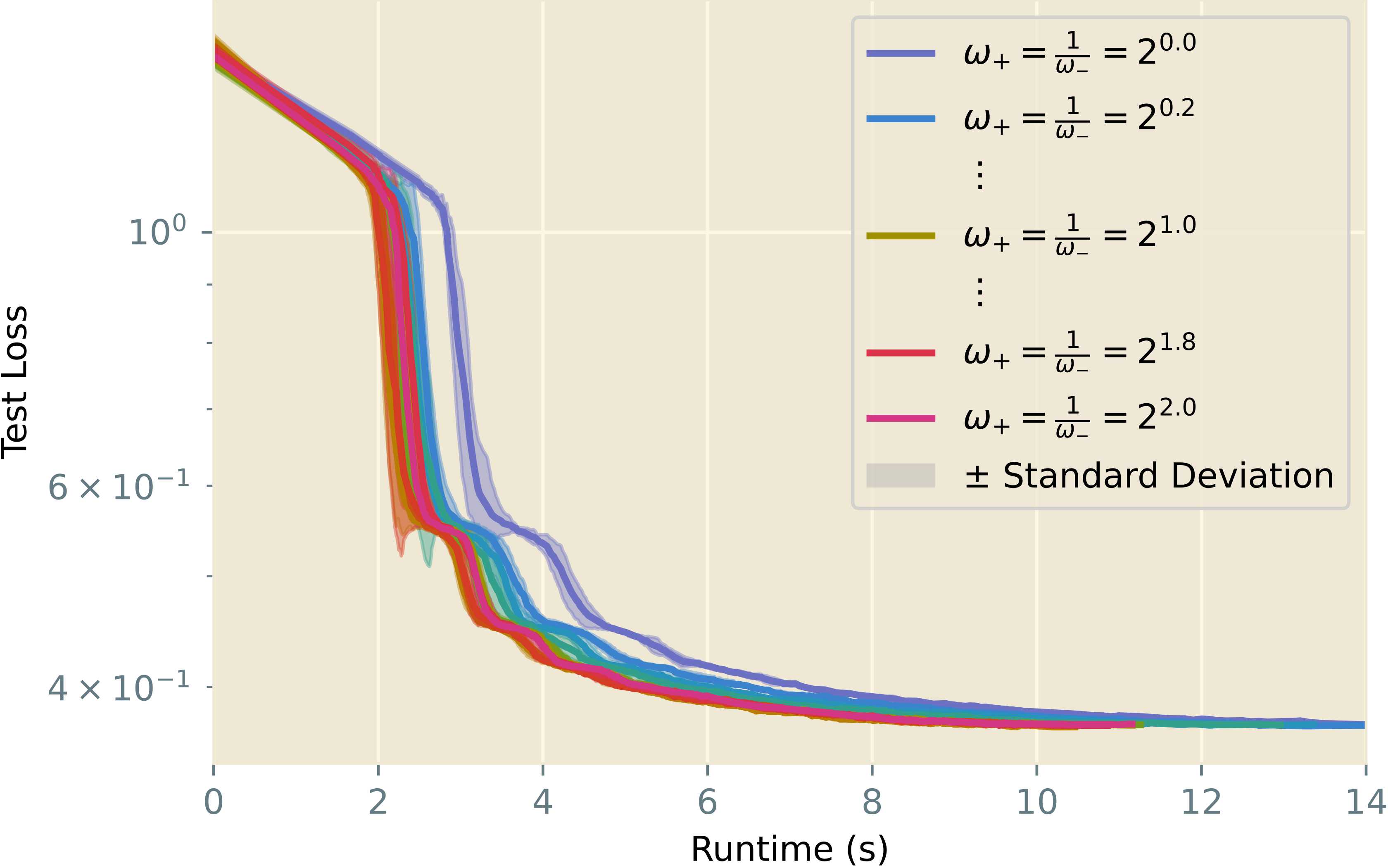}
    \caption{Sensitivity studies over damping stepping factor for our Fashion-MNIST trial from Section~\ref{sec:ExperimentsFashionMNIST}.}
    \label{fig:AblationSteppingFactor}
\end{figure*}

Finally, we explore the effect of different stepping factors by setting $\omega_\text{inc}$ to values in $\{2^{0.0}, 2^{0.2}, 2^{0.4}, \cdots, 2^{2.0}\}$, then choosing a symmetric $\omega_\text{dec} = \frac{1}{\omega_\text{inc}}$. Our results are plotted in Figure~\ref{fig:AblationSteppingFactor}.

The impact of different damping stepping factors becomes most apparent when damping plays a key role in stabilising the optimiser, which does not happen in this Fashion-MNIST test case. However, the plots match our subjective observation that the behaviour at the very start of training is critical to defining the optimisation trajectory, with a high variance at around 2\,s of runtime indicating an increased sensitivity here. Moreover, the results reinforce our intuition that the exact factor by which the damping $\lambda$ is modified is not crucially important, so long as AdamQLR is capable of making rapid adjustments over successive optimisation iterations when this becomes necessary.

\subsubsection{Adam (Batch Size and Learning Rate) and K-FAC (Batch Size and Initial Damping}
\begin{figure*}
    \centering
    \includegraphics[width=0.45\linewidth]{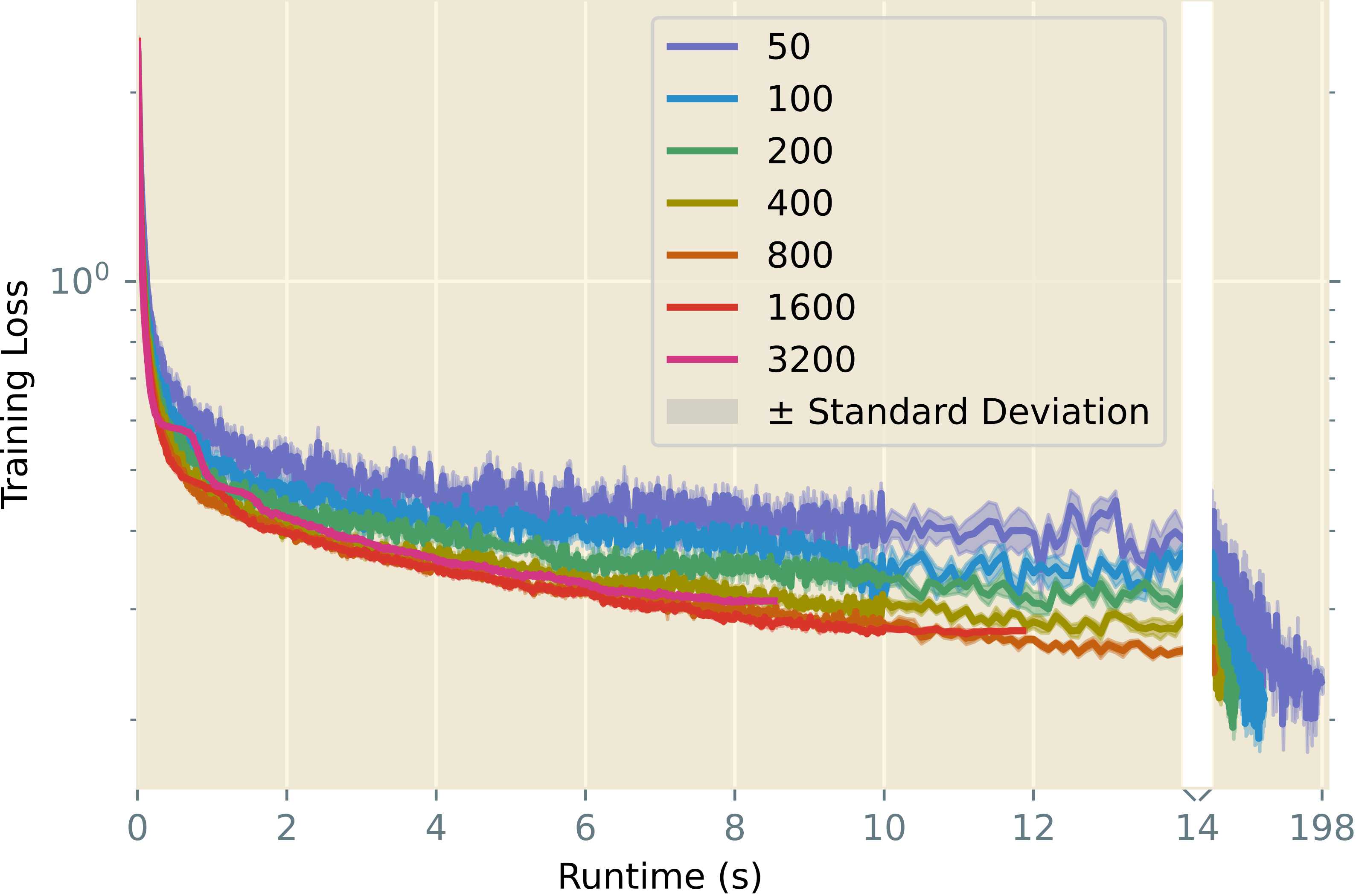}
    % \hfill
    \hspace*{0.01\linewidth}
    \includegraphics[width=0.45\linewidth]{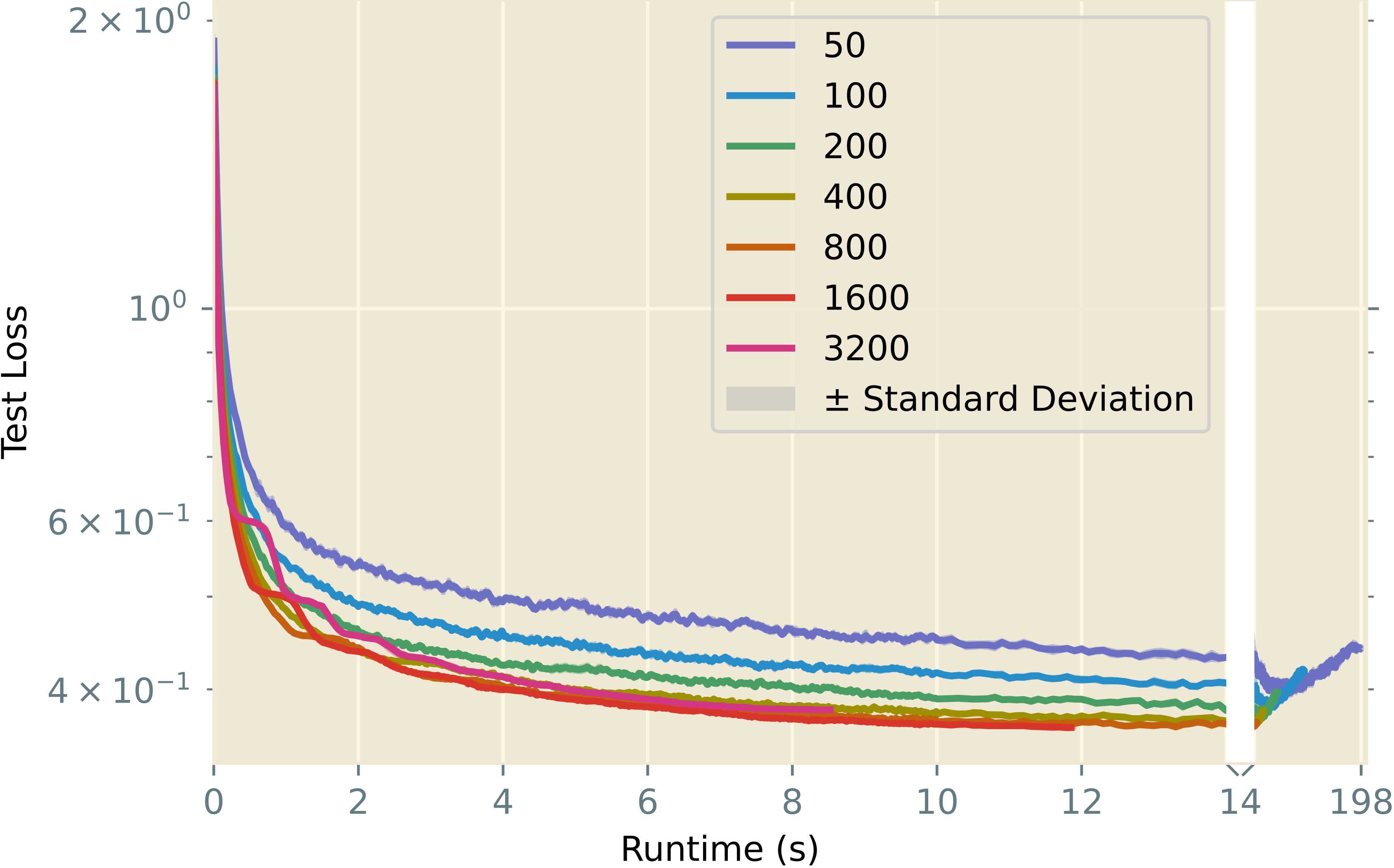}
    \caption{Sensitivity studies over batch size for our Fashion-MNIST trial from Section~\ref{sec:ExperimentsFashionMNIST}, considering now the \emph{Adam} setting.}
    \label{fig:SensitivityAdamBatchSize}
\end{figure*}

\begin{figure*}
    \centering
    \includegraphics[width=0.45\linewidth]{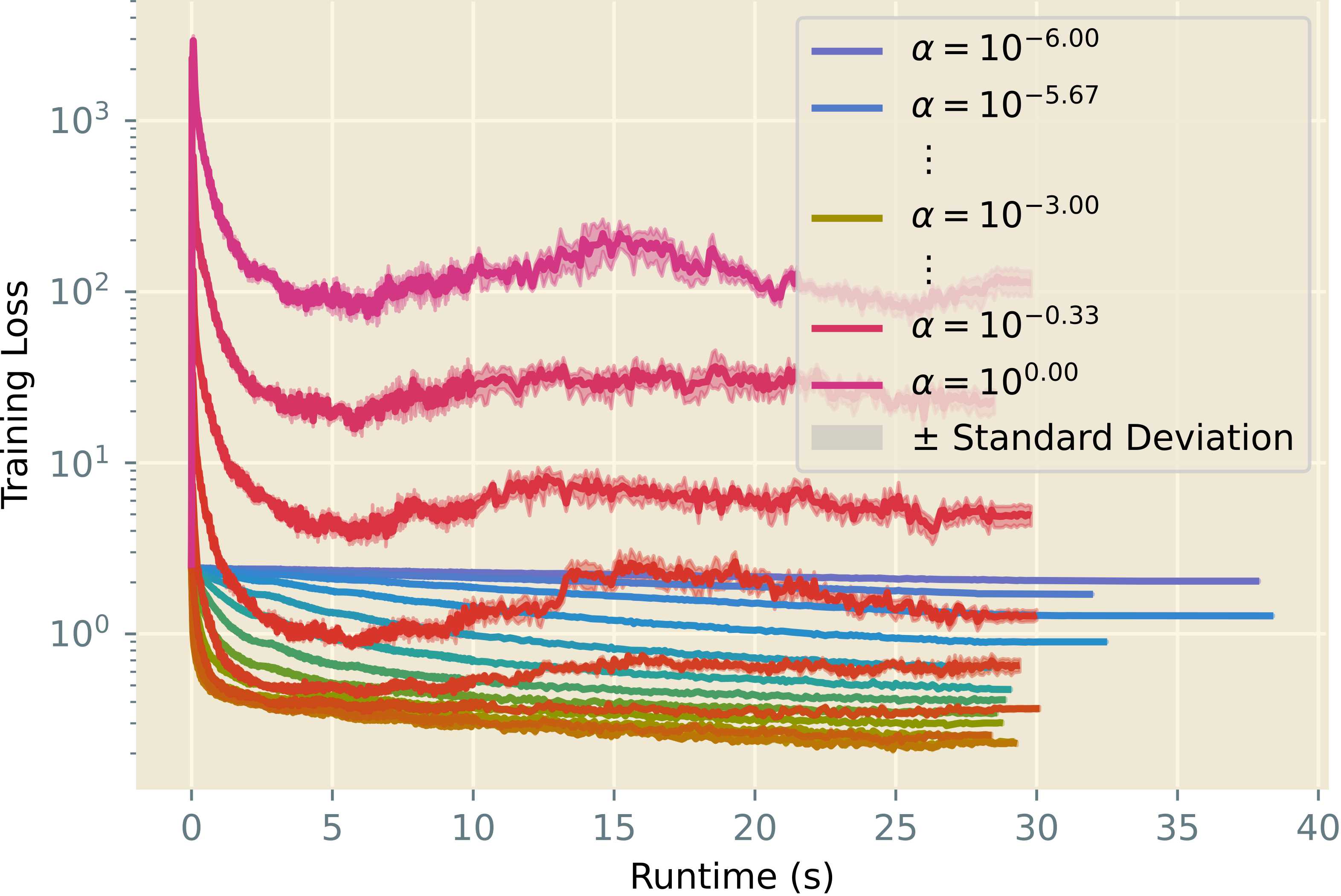}
    % \hfill
    \hspace*{0.01\linewidth}
    \includegraphics[width=0.45\linewidth]{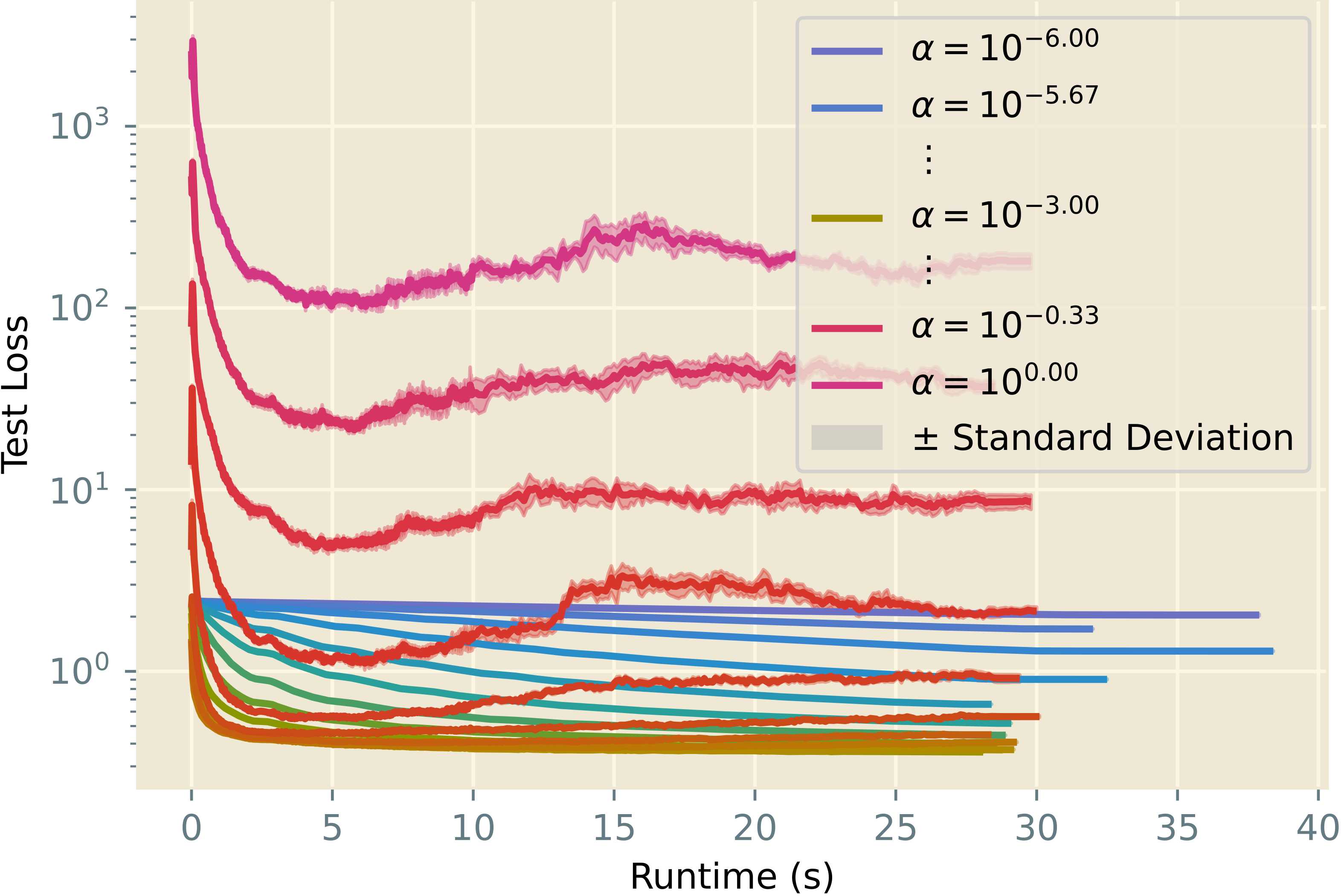}
    \caption{Sensitivity studies over learning rates for our Fashion-MNIST trial from Section~\ref{sec:ExperimentsFashionMNIST}, considering now the \emph{Adam} setting.}
    \label{fig:SensitivityAdamLearningRate}
\end{figure*}

\begin{figure*}
    \centering
    \includegraphics[width=0.45\linewidth]{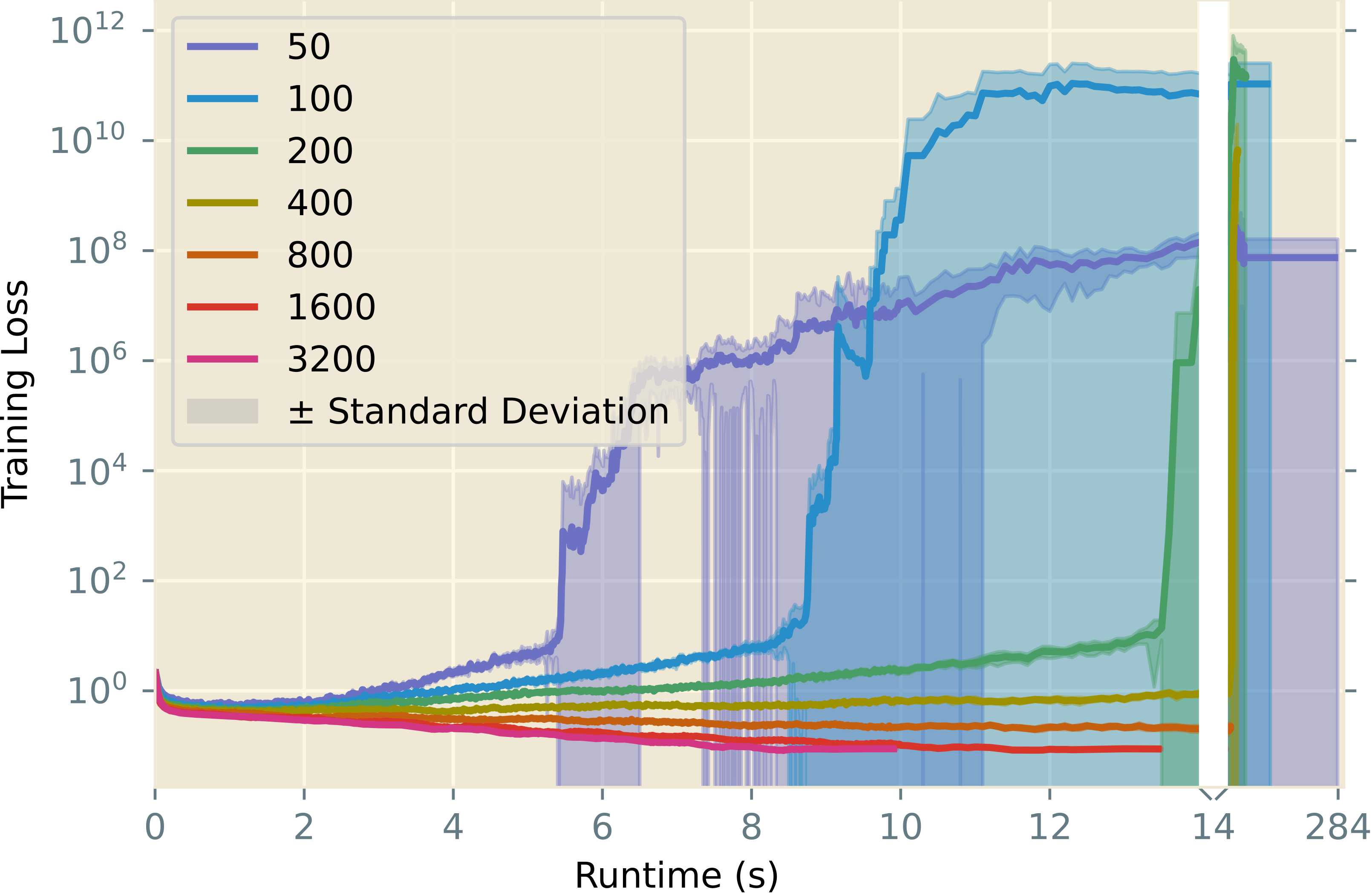}
    % \hfill
    \hspace*{0.01\linewidth}
    \includegraphics[width=0.45\linewidth]{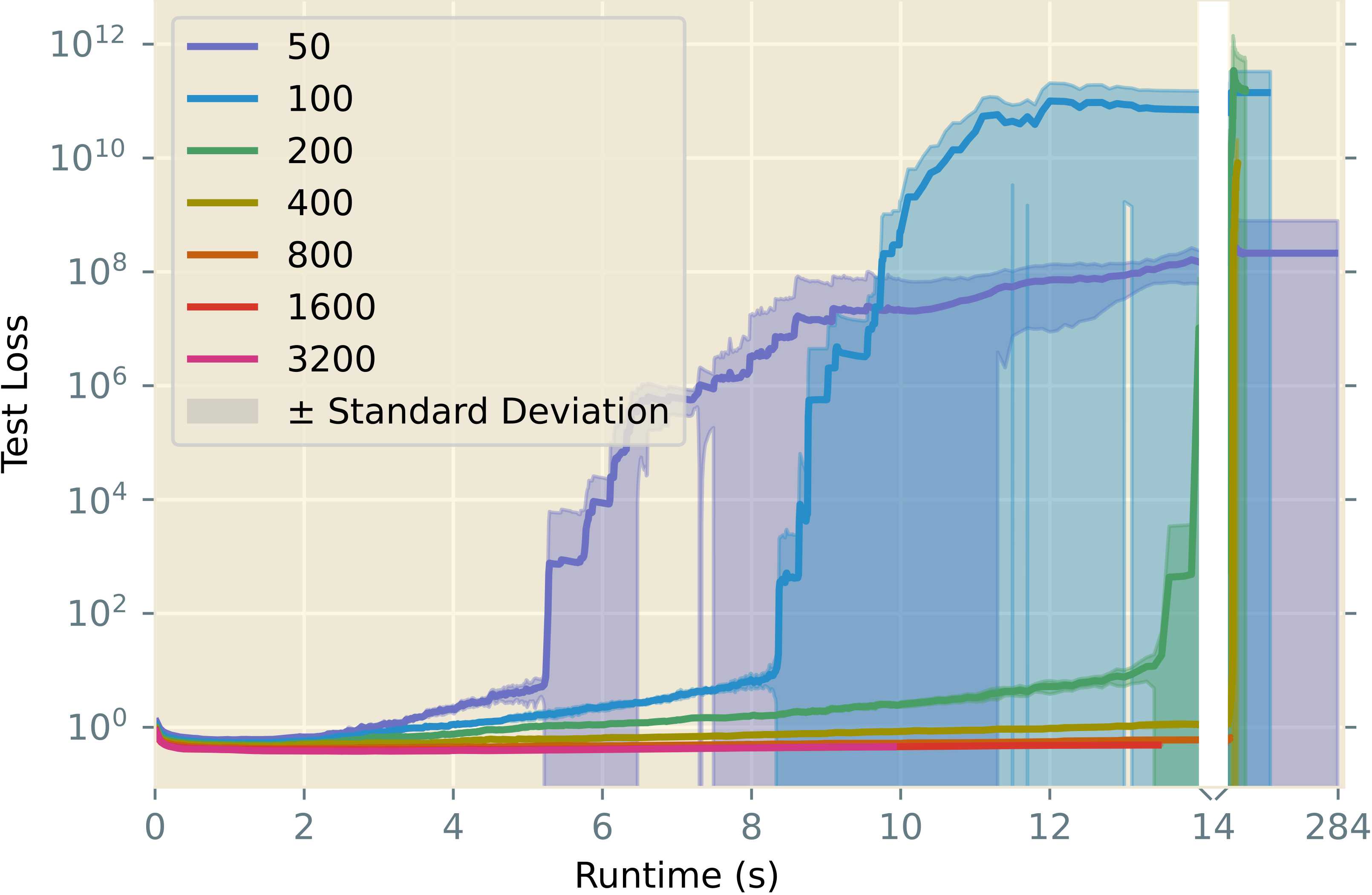}
    \caption{Sensitivity studies over batch size for our Fashion-MNIST trial from Section~\ref{sec:ExperimentsFashionMNIST}, considering now the \emph{K-FAC} setting.}
    \label{fig:SensitivityKFACBatchSize}
\end{figure*}

\begin{figure*}
    \centering
    \includegraphics[width=0.45\linewidth]{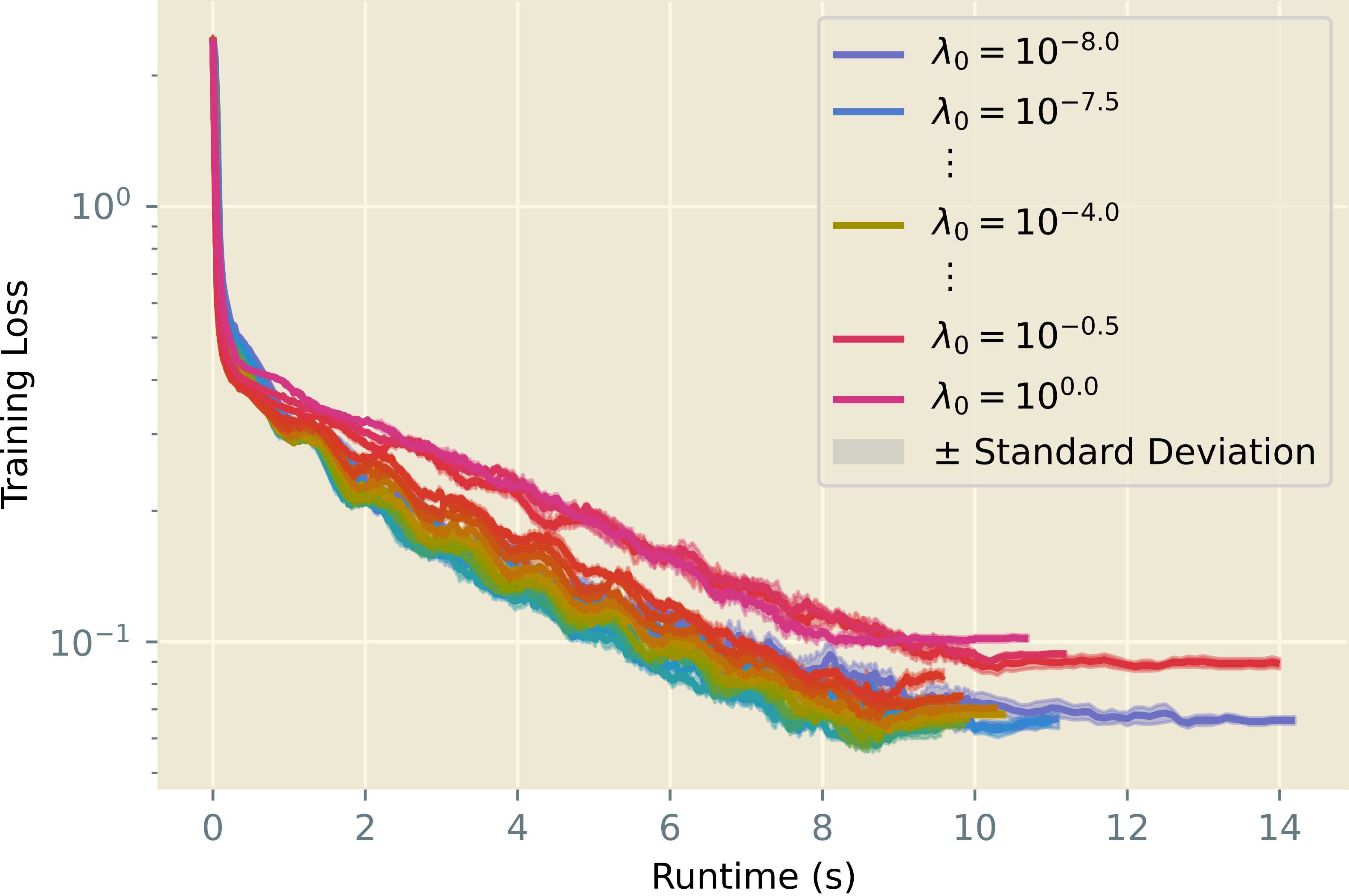}
    % \hfill
    \hspace*{0.01\linewidth}
    \includegraphics[width=0.45\linewidth]{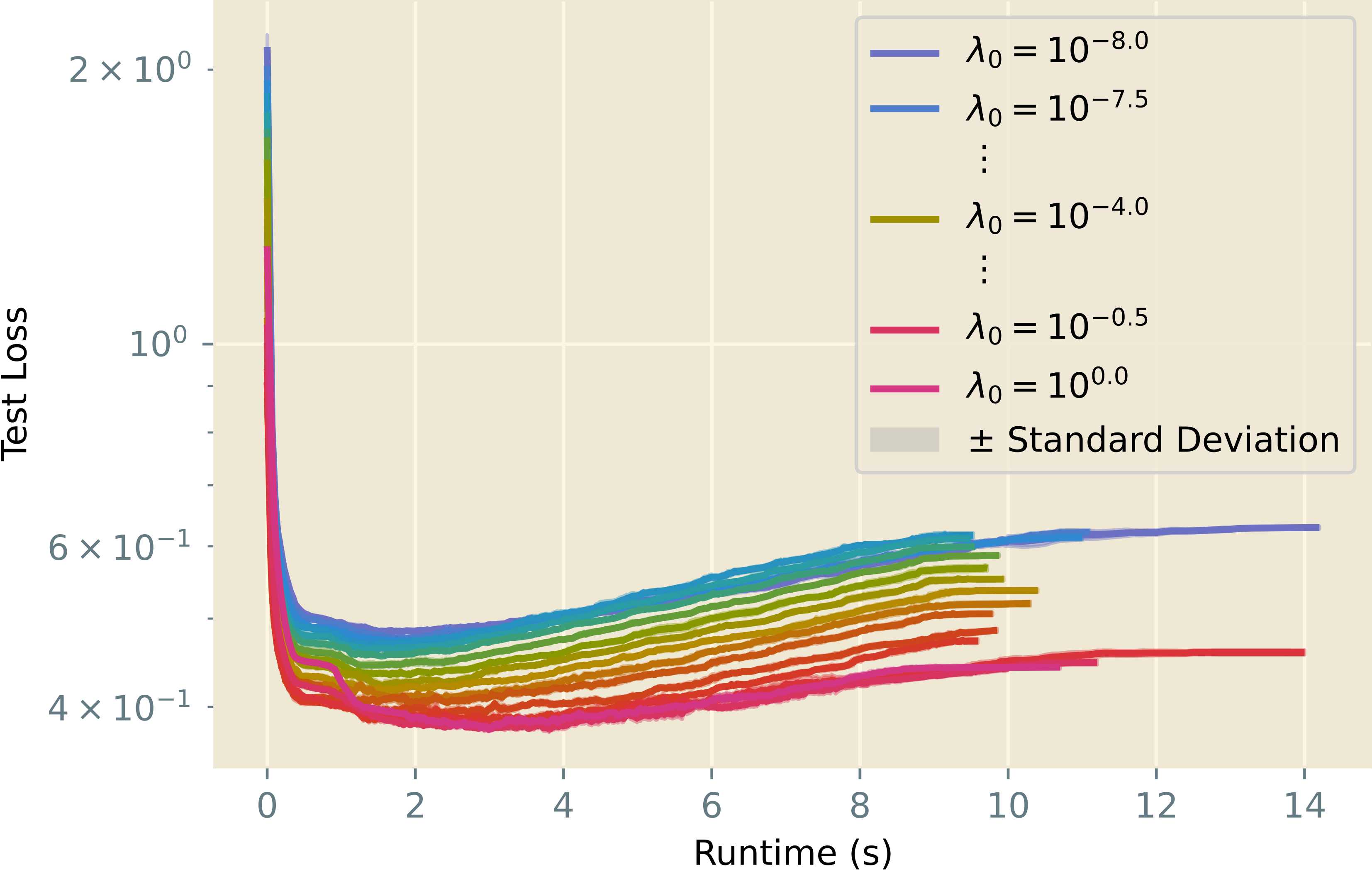}
    \caption{Sensitivity studies over initial damping for our Fashion-MNIST trial from Section~\ref{sec:ExperimentsFashionMNIST}, considering now the \emph{K-FAC} setting.}
    \label{fig:SensitivityKFACInitialDamping}
\end{figure*}

To reinforce our conclusions about the robustness of AdamQLR, we repeat our sensitivity study on batch size applied to \emph{Adam}, and perform an additional study over learning rates, setting $\alpha$ to values in $\{10^{-6.00}, 10^{-5.67}, 10^{-5.33}, \cdots, 10^{0.00}  \}$. Our results are plotted in Figures~\ref{fig:SensitivityAdamBatchSize} and \ref{fig:SensitivityAdamLearningRate}, respectively.

Similarly, we examine the robustness of \emph{K-FAC} by means of repeated sensitivity studies over batch size and initial damping. These results are plotted in Figures~\ref{fig:SensitivityKFACBatchSize} and \ref{fig:SensitivityKFACInitialDamping}, respectively.

We see Adam is substantially more sensitive to its learning rate than AdamQLR and K-FAC are to any of their hyperparameters, while AdamQLR strikes a middle ground between Adam and K-FAC in its sensitivity to batch size. Despite AdamQLR's near-invariance to initial damping, this hyperparameter has a sizeable impact on K-FAC. These results support the conclusion that AdamQLR is at least as robust as --- and for some particularly important hyperparameters is more robust than --- Adam and K-FAC.

\subsection{Ablation Studies}
\label{sec:AblationExperiments}

In addition to the algorithms plotted in Section~\ref{sec:Experiments}, we conduct additional experiments to study the impact of different components of \emph{AdamQLR} on its overall performance. Specifically, we examine the effects of Levenberg-Marquardt damping and the choice of curvature matrix used to construct our quadratic model. We use the same experimental configuration as in Section~\ref{sec:Experiments}, including hyperparameter tuning with ASHA, and plot bootstrapped average trends over 50 repetitions of the best hyperparameters found.

The new results presented in this section were computed on a mix of the Consumer Desktop and Local Cluster, using one GPU at a time. Since the GPUs are of the same model, we expect any performance difference due to this discrepancy to be minor.

\subsubsection{Levenberg-Marquardt Damping}
\label{sec:DampingAblationExperiments}
\begin{figure*}
    \centering
    \includegraphics[width=0.45\linewidth]{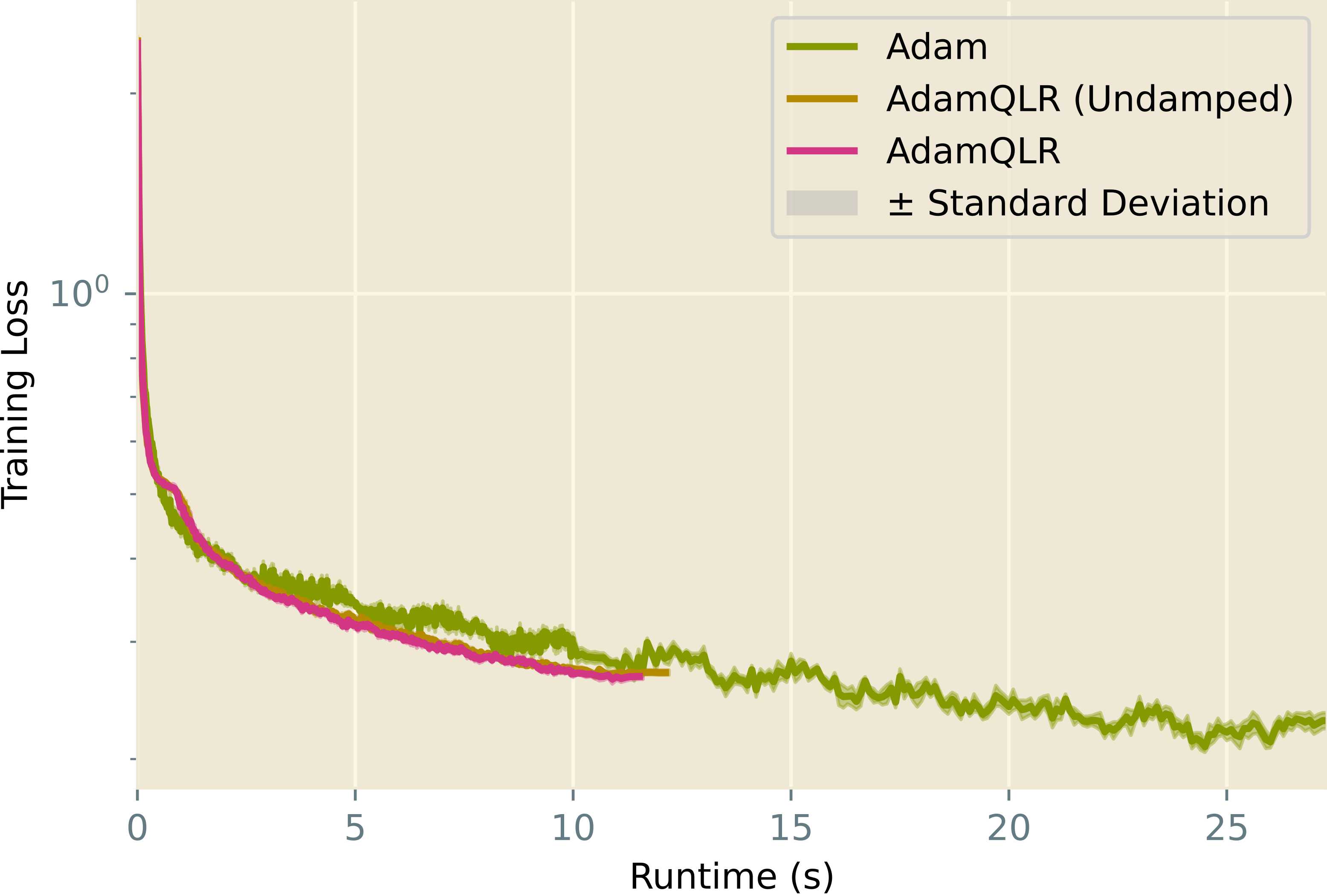}
    % \hfill
    \hspace*{0.01\linewidth}
    \includegraphics[width=0.45\linewidth]{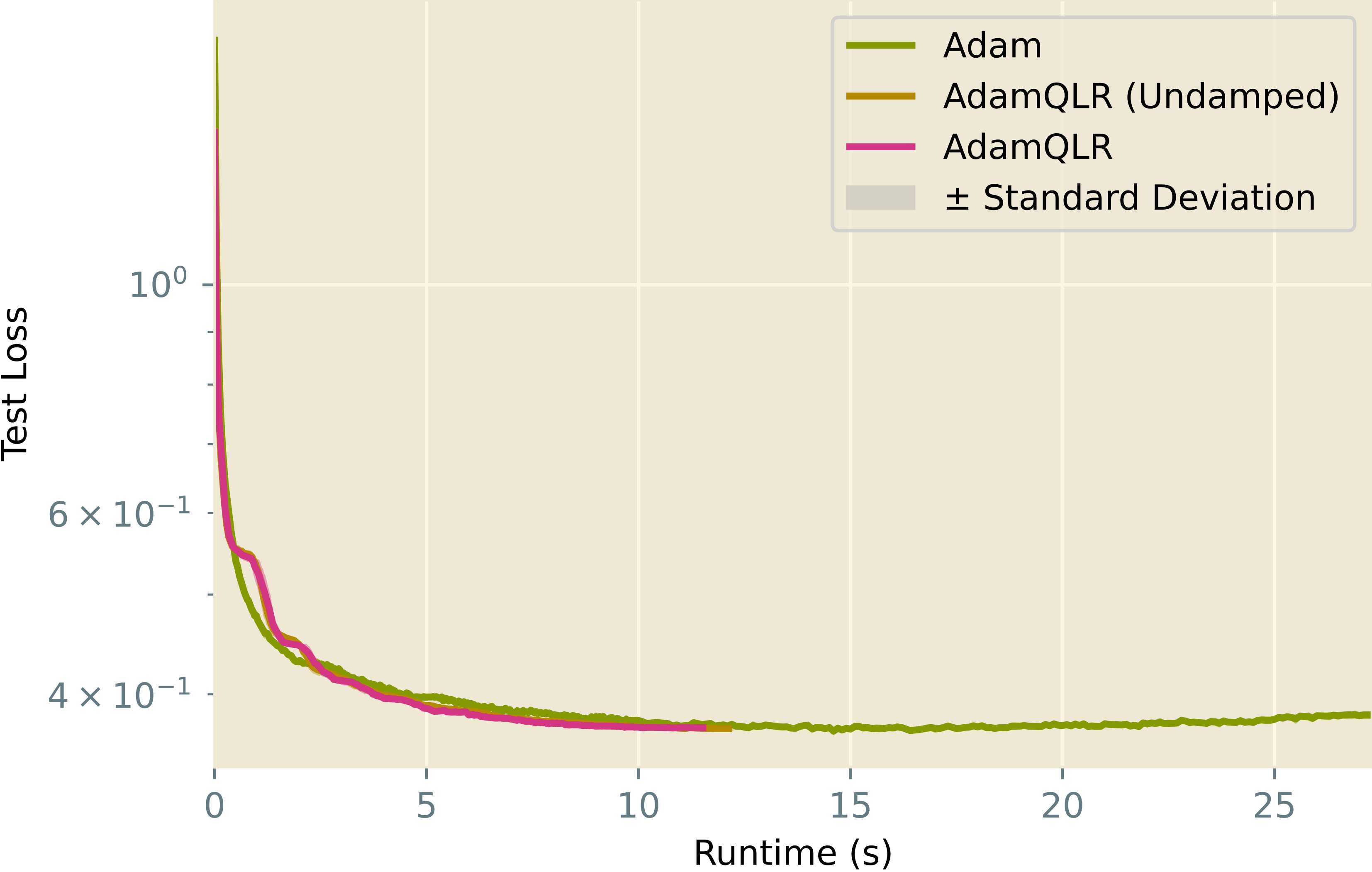}
    
    \includegraphics[width=0.45\linewidth]{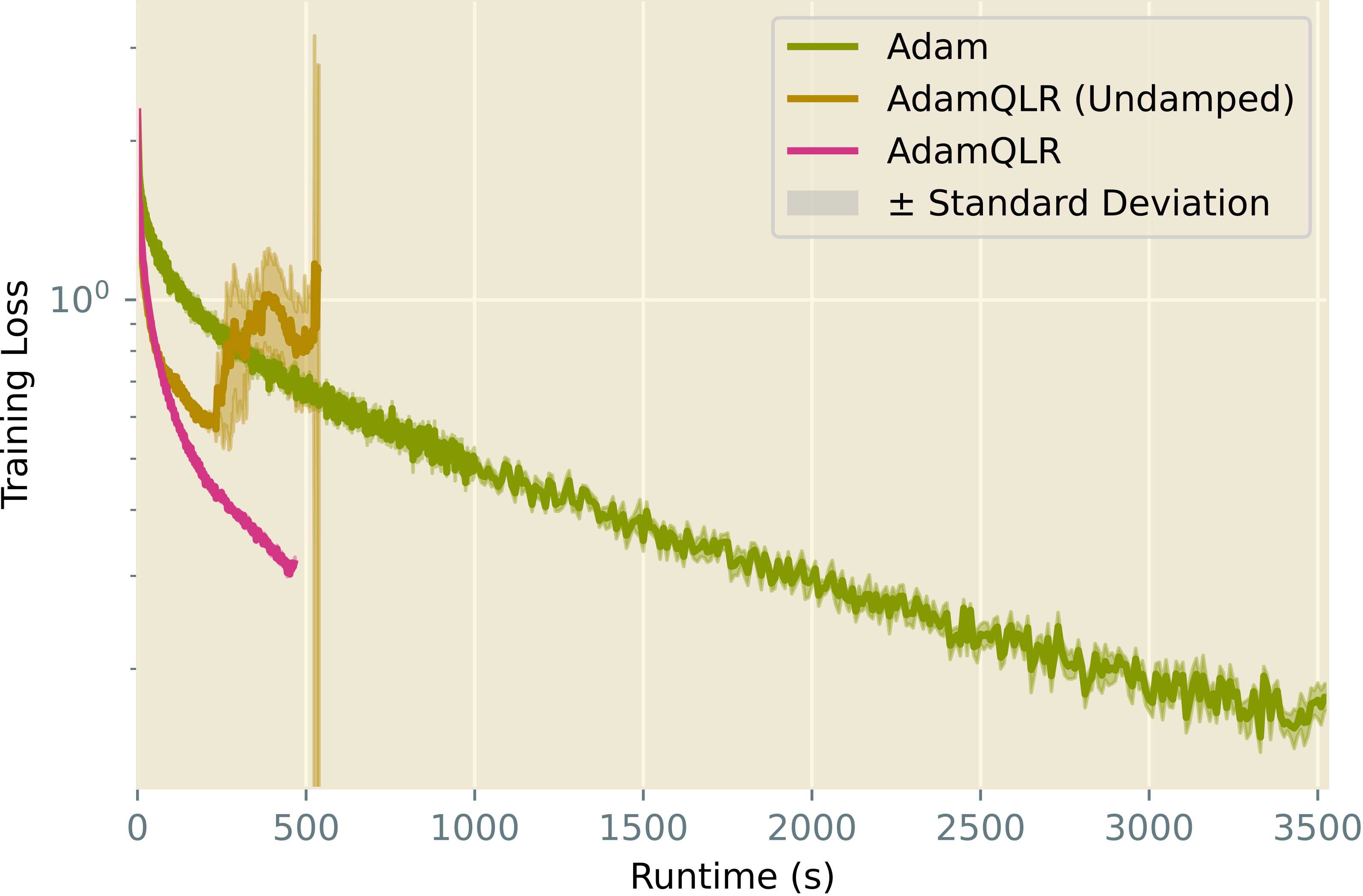}
    % \hfill
    \hspace*{0.01\linewidth}
    \includegraphics[width=0.45\linewidth]{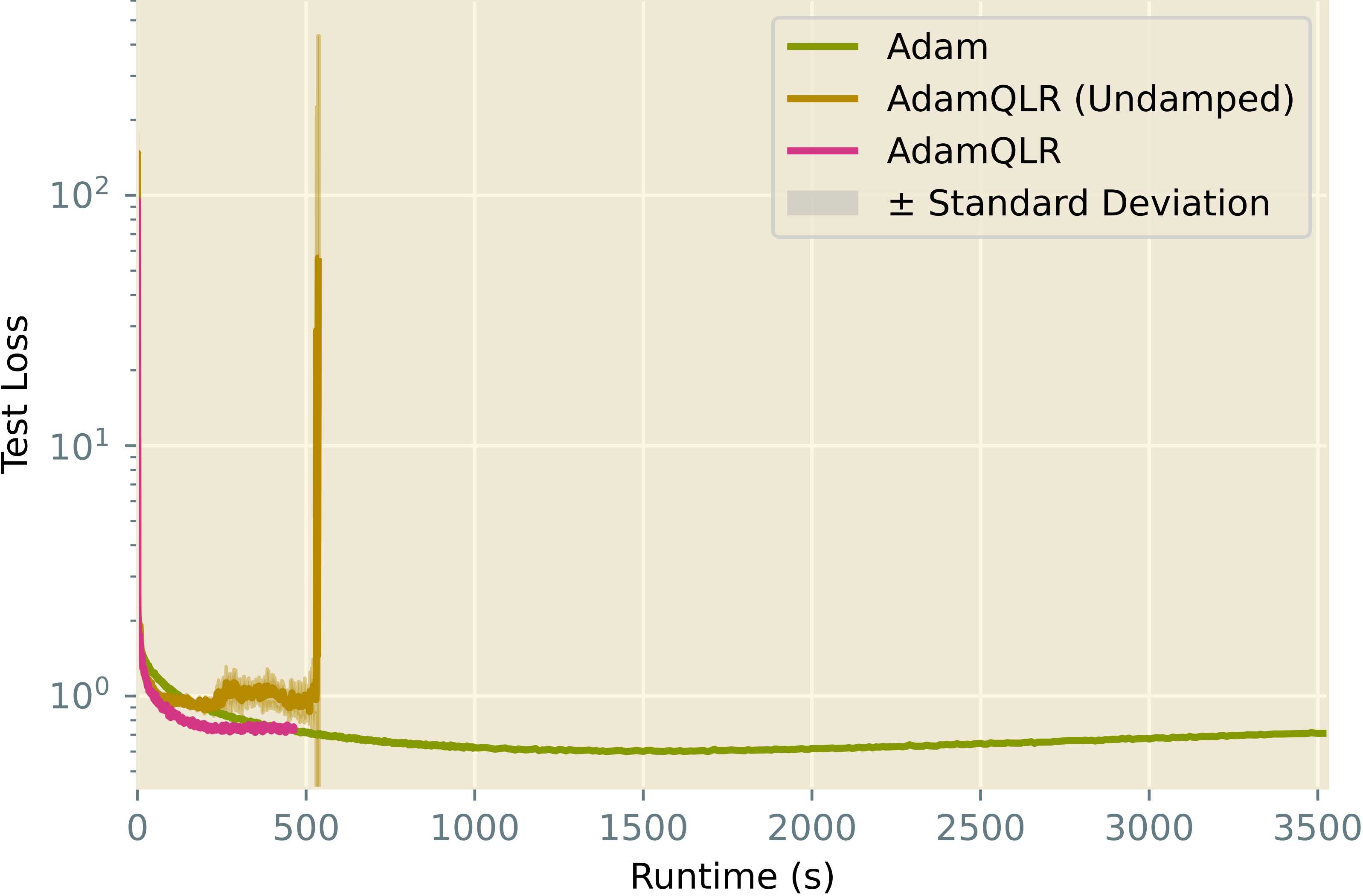}
    \caption{Evolution of Levelberg-Marquardt damping, as measured by Training (left) and Test (right) loss on Fashion-MNIST (top) and CIFAR-10 (bottom)}
    \label{fig:AblationDamping}
\end{figure*}

Appropriate damping is viewed as a necessity in many second-order algorithms in order to defend against degenerate parameter updates, and Figure~\ref{fig:AblationDamping} examines its inclusion in \emph{AdamQLR}. We consider vanilla \emph{Adam} alongside two versions of \emph{AdamQLR}: one which includes damping, and another which excludes it, and perform hyperparameter optimisation as before on each algorithm.

On Fashion-MNIST, we see minimal effect from the inclusion of damping, as the problem does not suffer greatly from degenerate parameter updates. Thus, especially when the internal model of objective space performs well and damping is pushed to very low values, the damping makes a proportionally very small difference to the updates we take. As such, while we do benefit slightly from damping here, the advantage is very slight.

On CIFAR-10, however, we see more dramatic differences from the inclusion of damping, though we note the difference in horizontal scale is likely due to different optimal batch sizes chosen by ASHA. Adjusting for this factor, we see the undamped version of AdamQLR is substantially less stable than the standard damped setting. This result is intuitive --- since the model is larger and is substantially more overparameterised than in the Fashion-MNIST case, there are likely to be more parameters to which the output of our network is insensitive, corresponding to low-curvature directions of optimisation space. These low-curvature directions correspond to small eigenvalues of the curvature matrix, so a naïve curvature-based approach would take very large steps in these directions. Because the problem is inherently non-convex and non-quadratic, such large steps would not be well-motivated, and we would suffer a resulting penalty in our rapidly-excursing loss. Indeed, during our development of AdamQLR, we observed damping to play an important role in avoiding the destabilisation of training. Further, damping clearly stabilises the algorithm here to allow for more aggressive optimisation over time. This evidence justifies our use of damping in the default AdamQLR approach.

\subsubsection{Curvature Matrix}
\begin{figure*}
    \centering
    \includegraphics[width=0.45\linewidth]{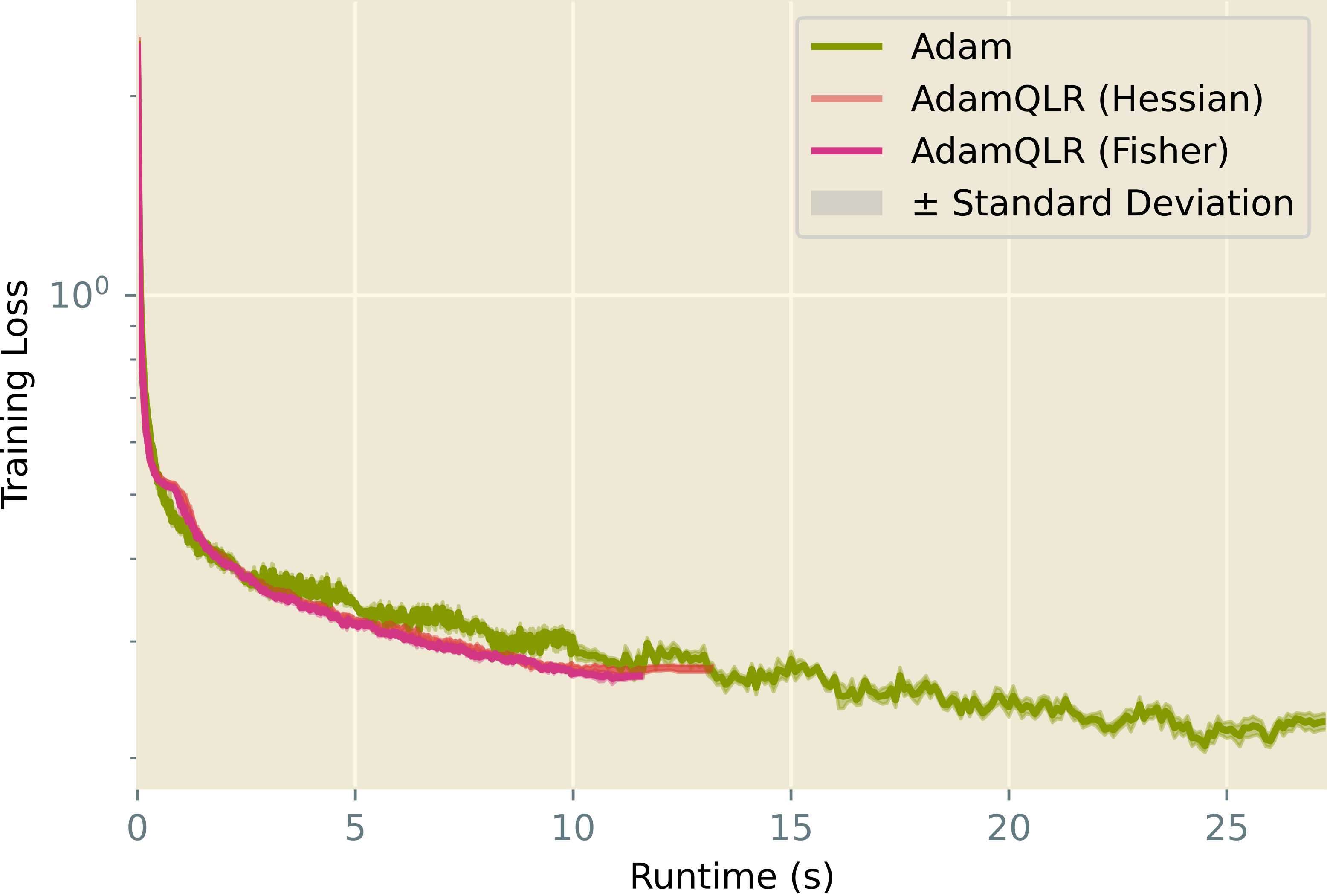}
    % \hfill
    \hspace*{0.01\linewidth}
    \includegraphics[width=0.45\linewidth]{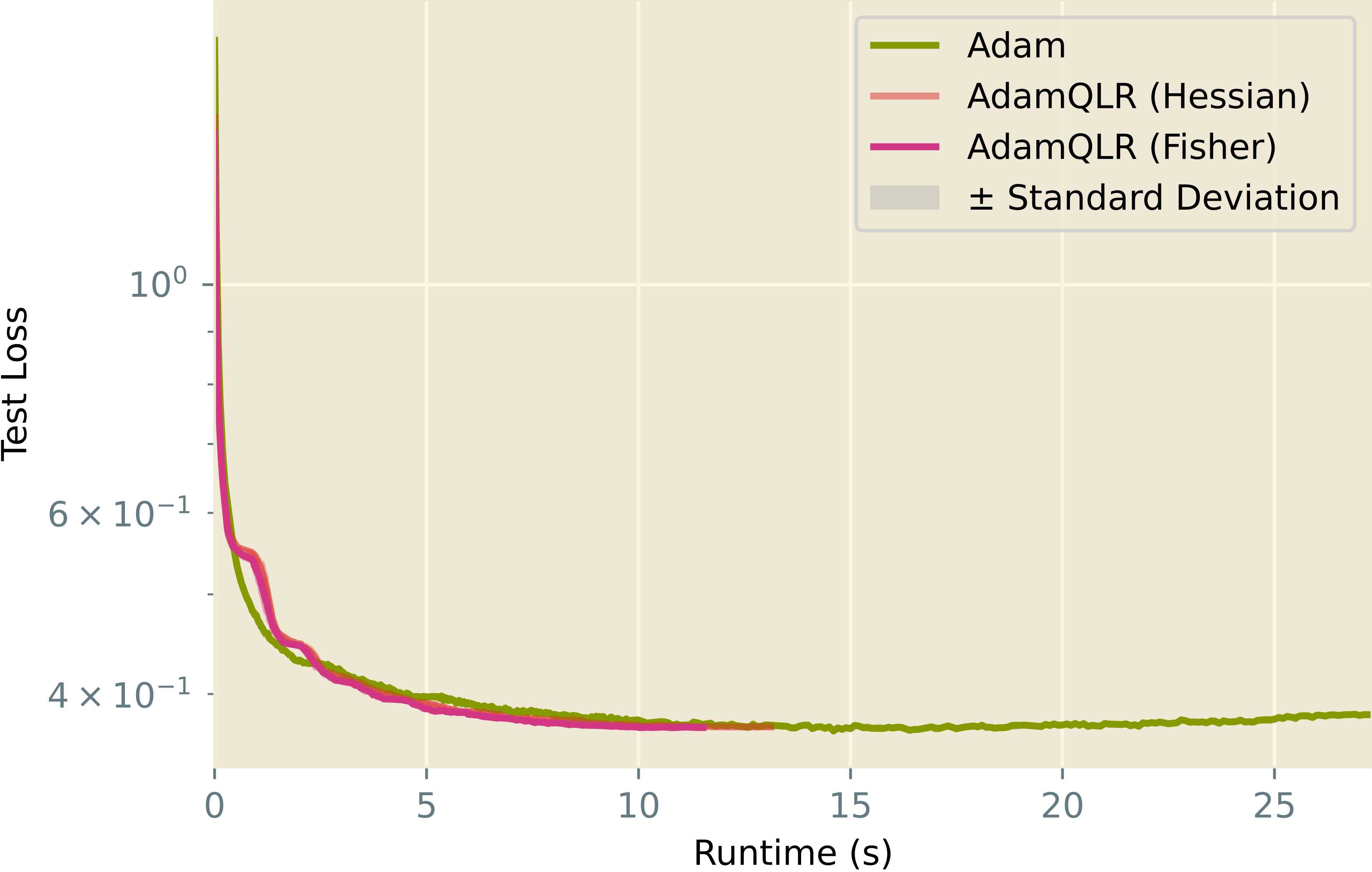}
    
    \includegraphics[width=0.45\linewidth]{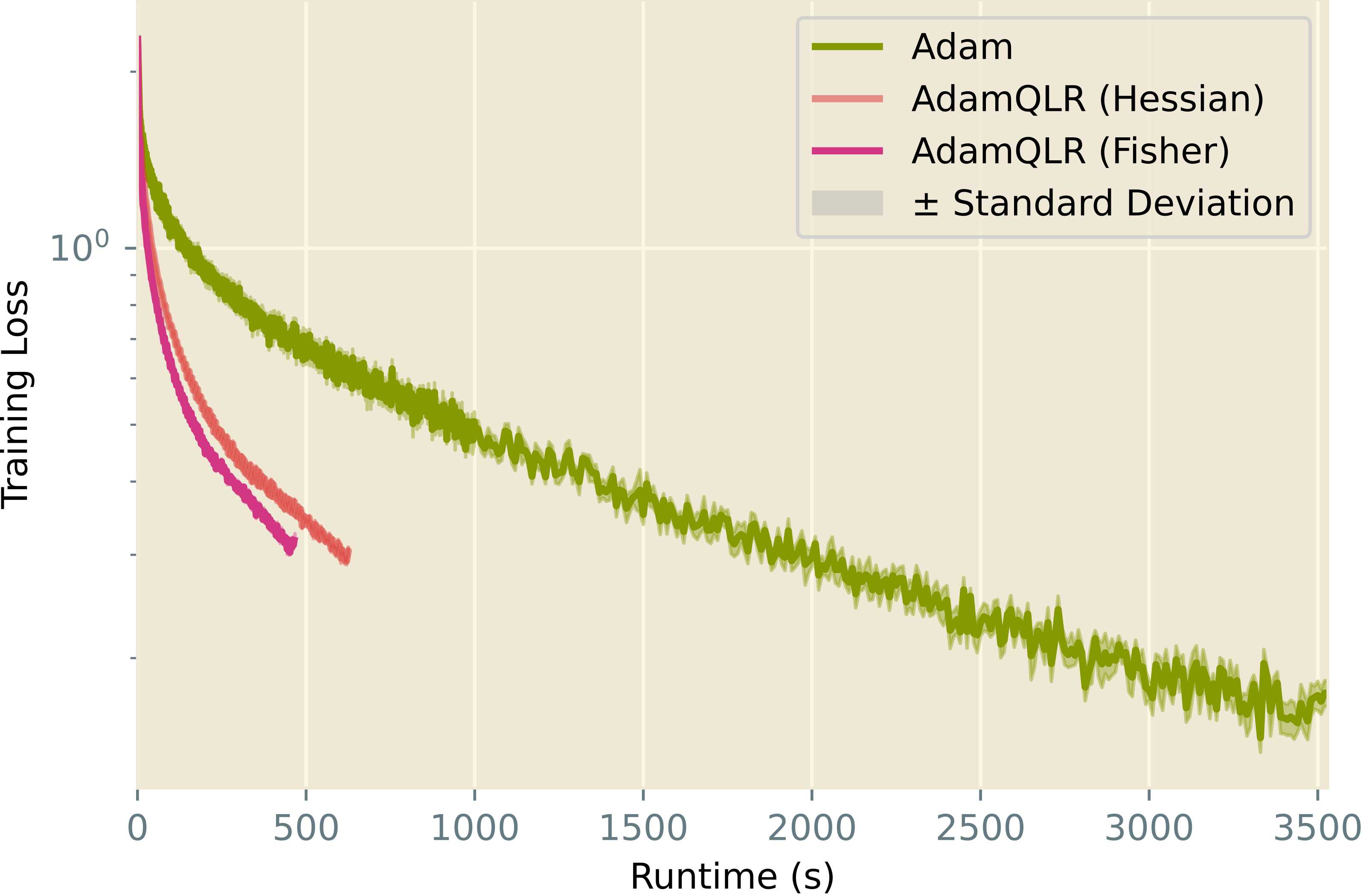}
    % \hfill
    \hspace*{0.01\linewidth}
    \includegraphics[width=0.45\linewidth]{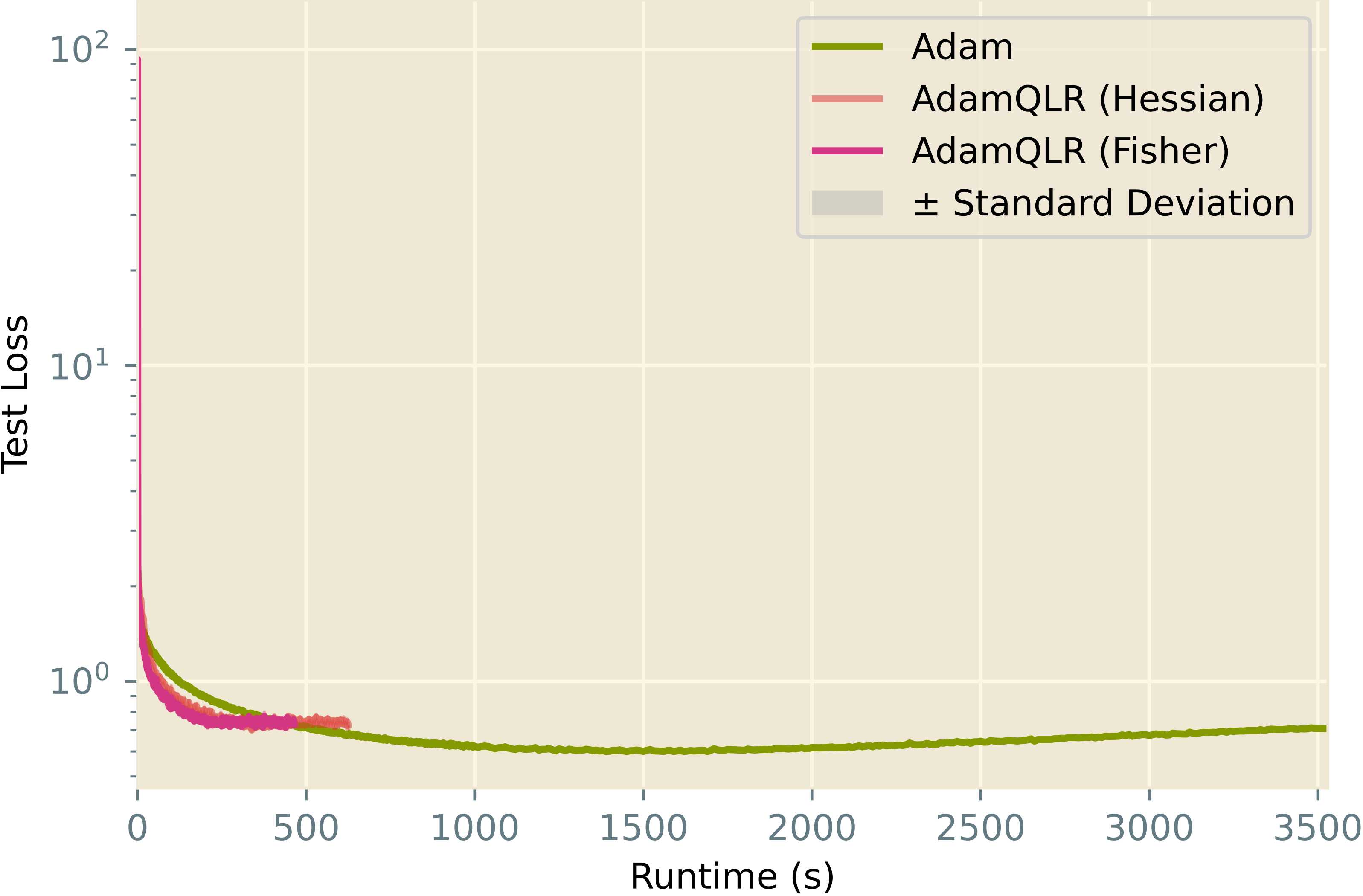}
    \caption{Evaluation of the choice of curvature matrix for the learning rate and damping calculations in \emph{AdamQLR}}
    \label{fig:AblationLearningRateCurvature}
\end{figure*}
As discussed in Appendix~\ref{sec:CurvatureMatrices}, there is good reason to motivate both the Hessian and the Fisher matrices as curvatures to use to select the learning rate $\alpha$ at each update step. To explore their relative merits, we consider two versions of \emph{AdamQLR}: one which uses Hessian curvature to compute a learning rate and update damping, and another which uses Fisher curvature for the same purposes. The performance of hyperparameter-optimised versions of each setting is compared alongside vanilla \emph{Adam} in Figure~\ref{fig:AblationLearningRateCurvature}.

On Fashion-MNIST, we see a slight advantage for Fisher curvature compared to the Hessian curvature, both of which generalise very slightly better than vanilla \emph{Adam}.
While the generalisation benefit does not transfer to CIFAR-10, we still see a small speed advantage for Fisher-based curvature.
Again, we note that different optimal batch sizes are likely responsible for most of the horizontal scaling difference.
The similarity of these results, combined with the subjectively greater stability of the Fisher version of \emph{AdamQLR} in our development process, justify our use of the Fisher curvature as the default in our algorithm.
While Fisher-vector products are more intricate than Hessian-vector products, requiring a rederived component for each loss function, a relatively small number of different loss functions see regular use in practice, so we accept this additional burden.

\subsubsection{Alternative \emph{Adam (Tuned $\epsilon$)} Setting}
\begin{figure*}[p]
    \centering
    \newcommand{\rotatecaption}[1]{%
        \rotatebox[origin=c]{90}{\begin{minipage}{3cm}#1\end{minipage}} }
    \begin{tabularx}{0.82\linewidth}{p{2ex}XX}
        \rotatecaption{\subcaption{UCI Energy}}
        & \includegraphics[align=c,width=\linewidth]{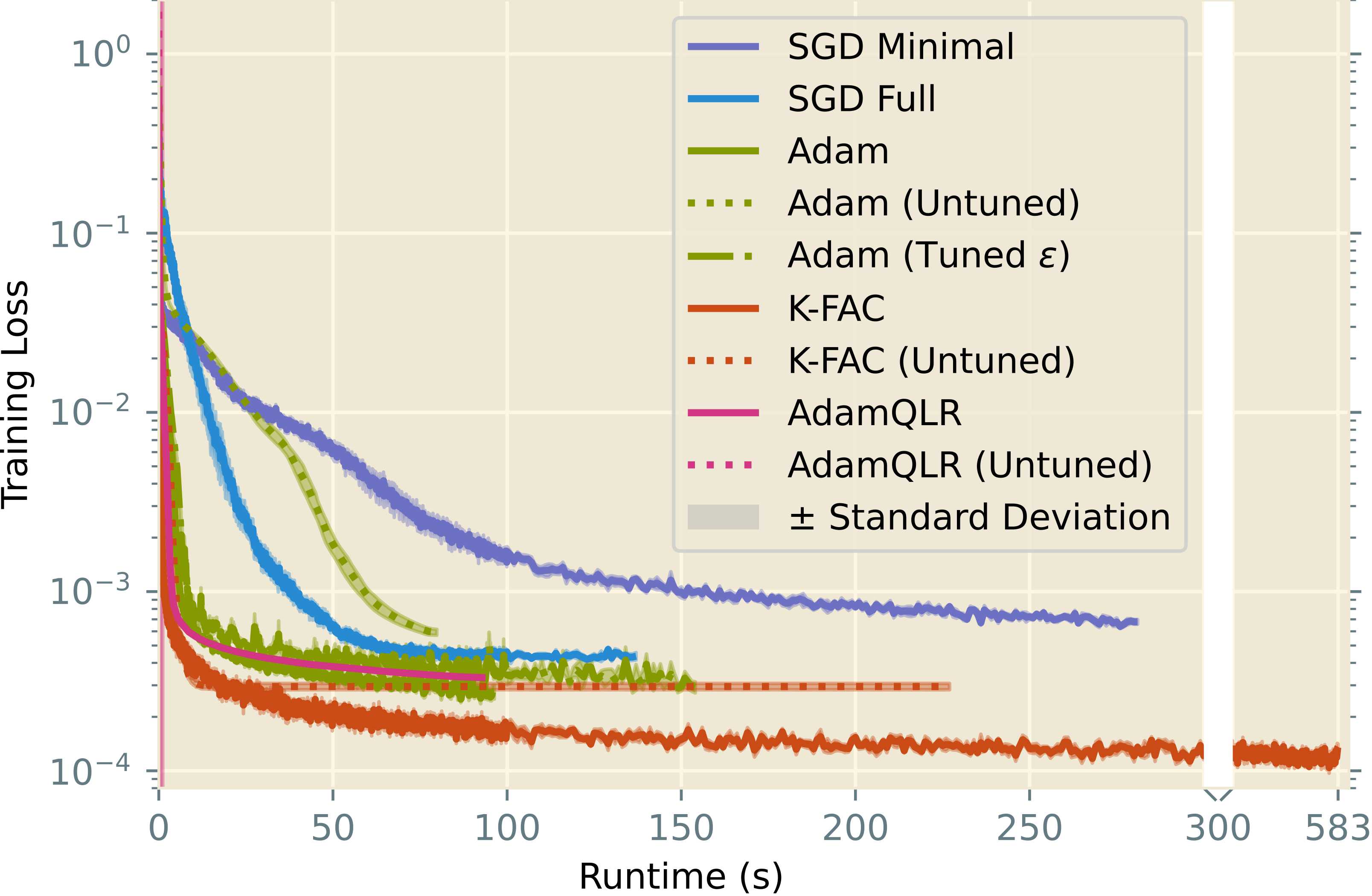}
        & \includegraphics[align=c,width=\linewidth]{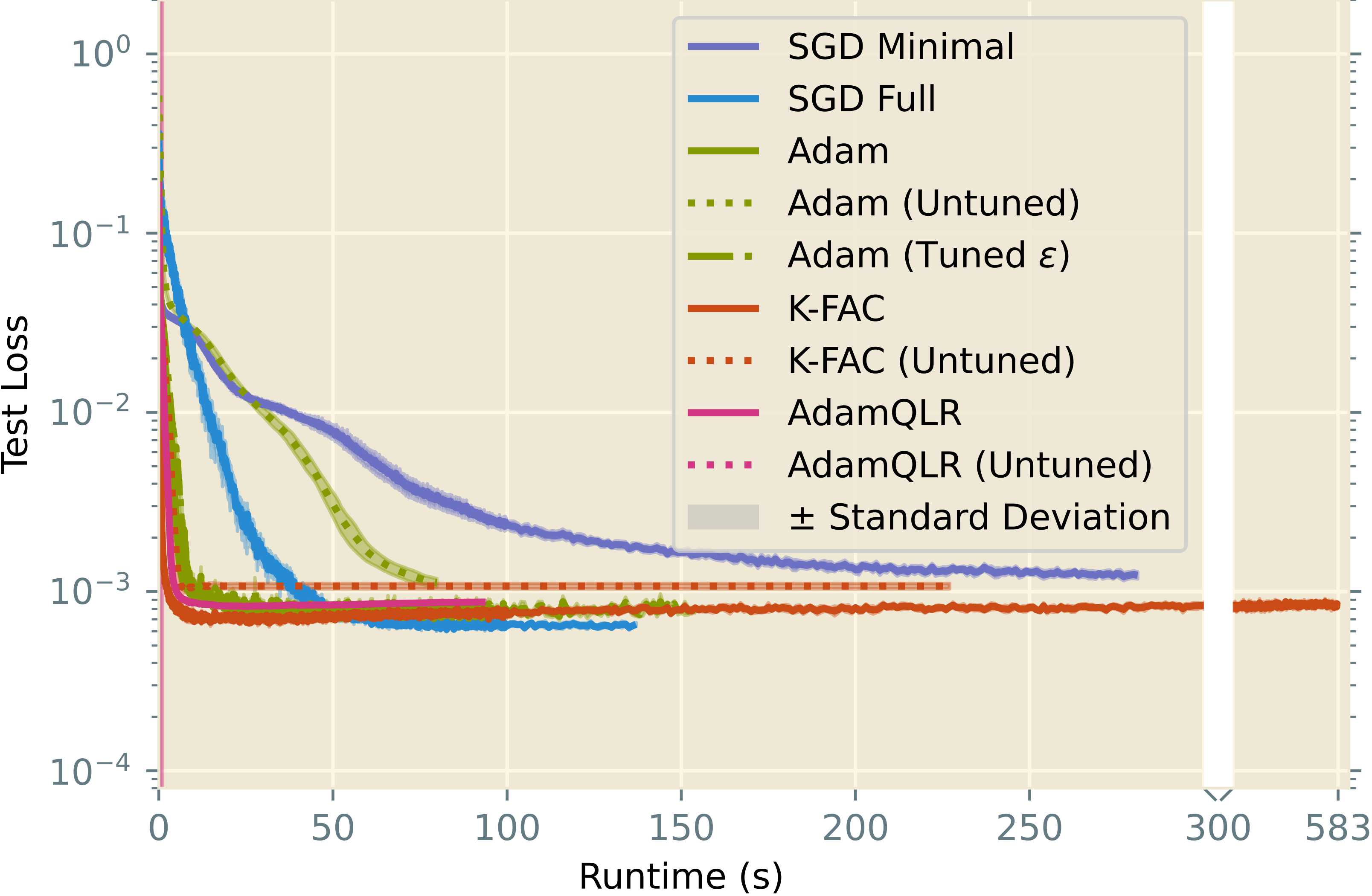} \vfill\\
        
        \rotatecaption{\subcaption{UCI Protein}}
        & \includegraphics[align=c,width=\linewidth]{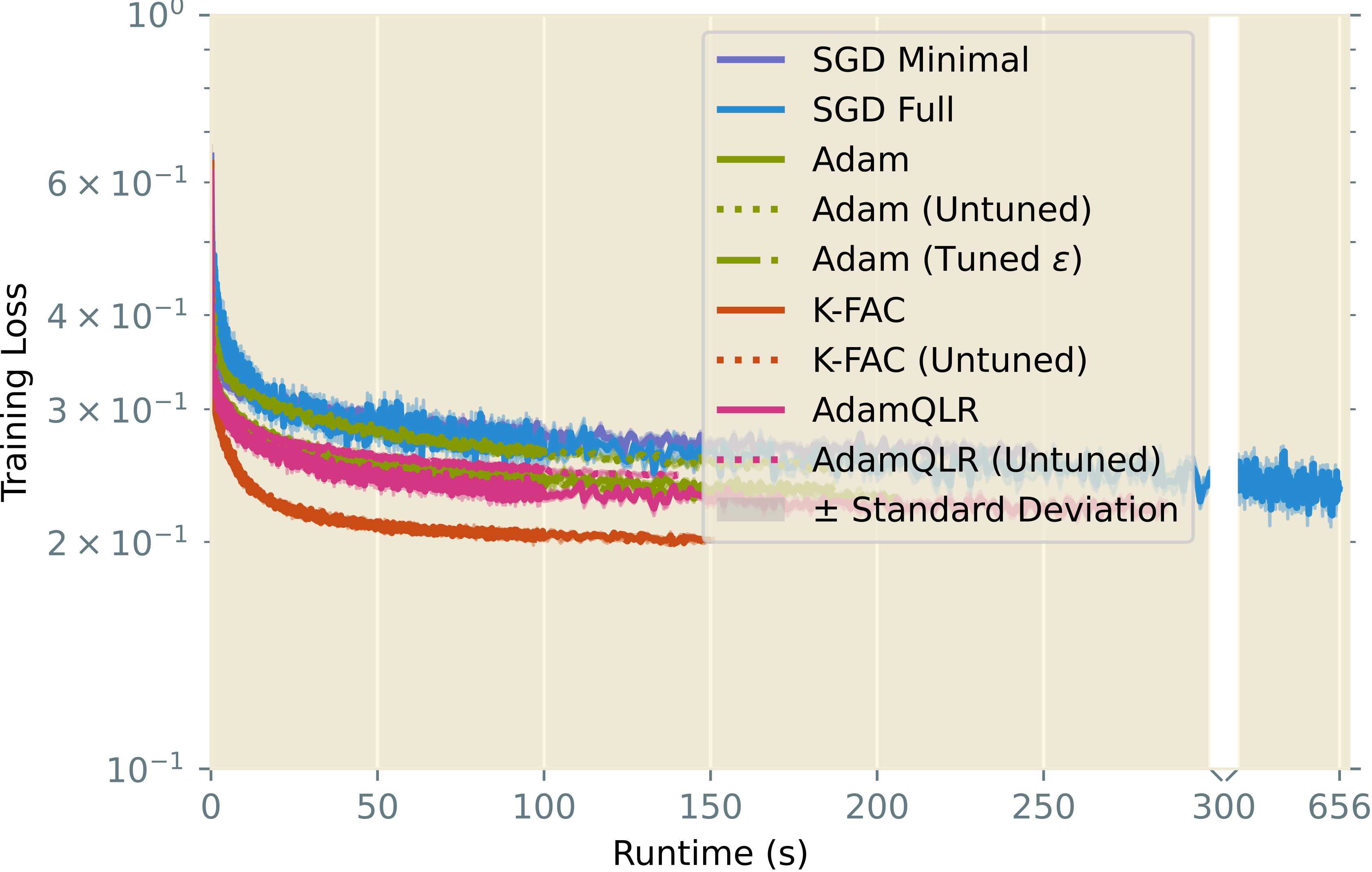}
        & \includegraphics[align=c,width=\linewidth]{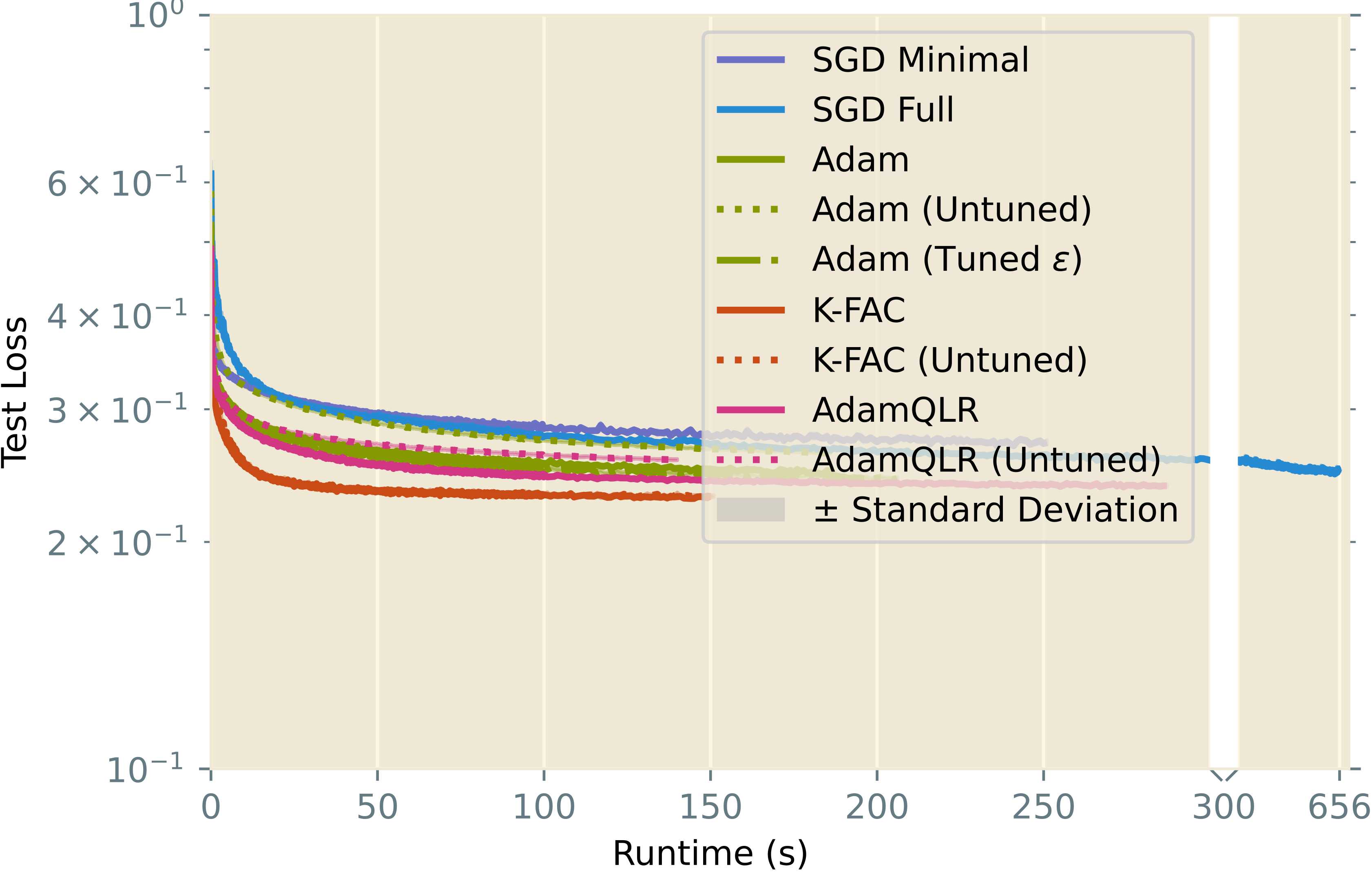} \vfill\\
        
        \rotatecaption{\subcaption{Fashion-MNIST}}
        & \includegraphics[align=c,width=\linewidth]{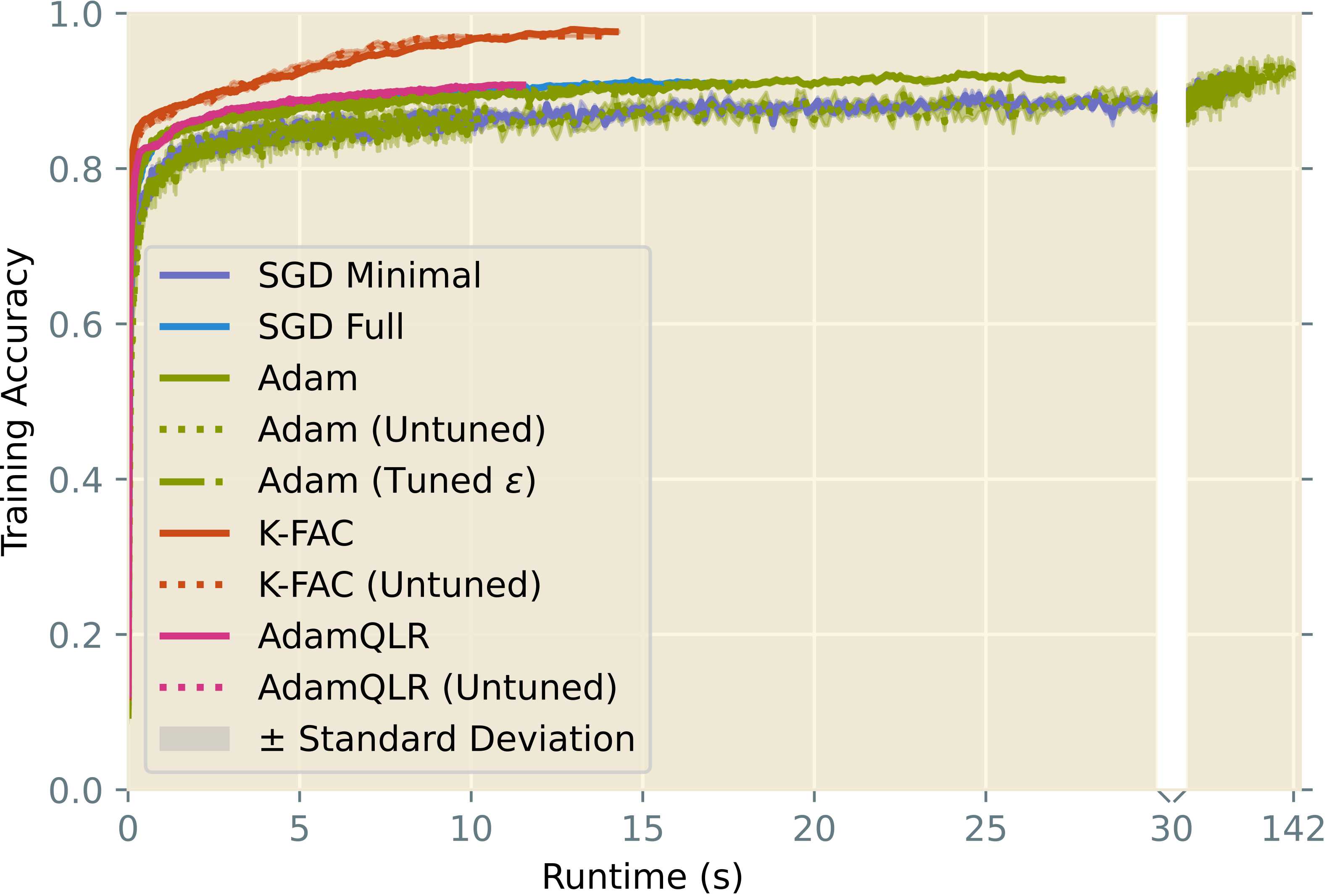}
        & \includegraphics[align=c,width=\linewidth]{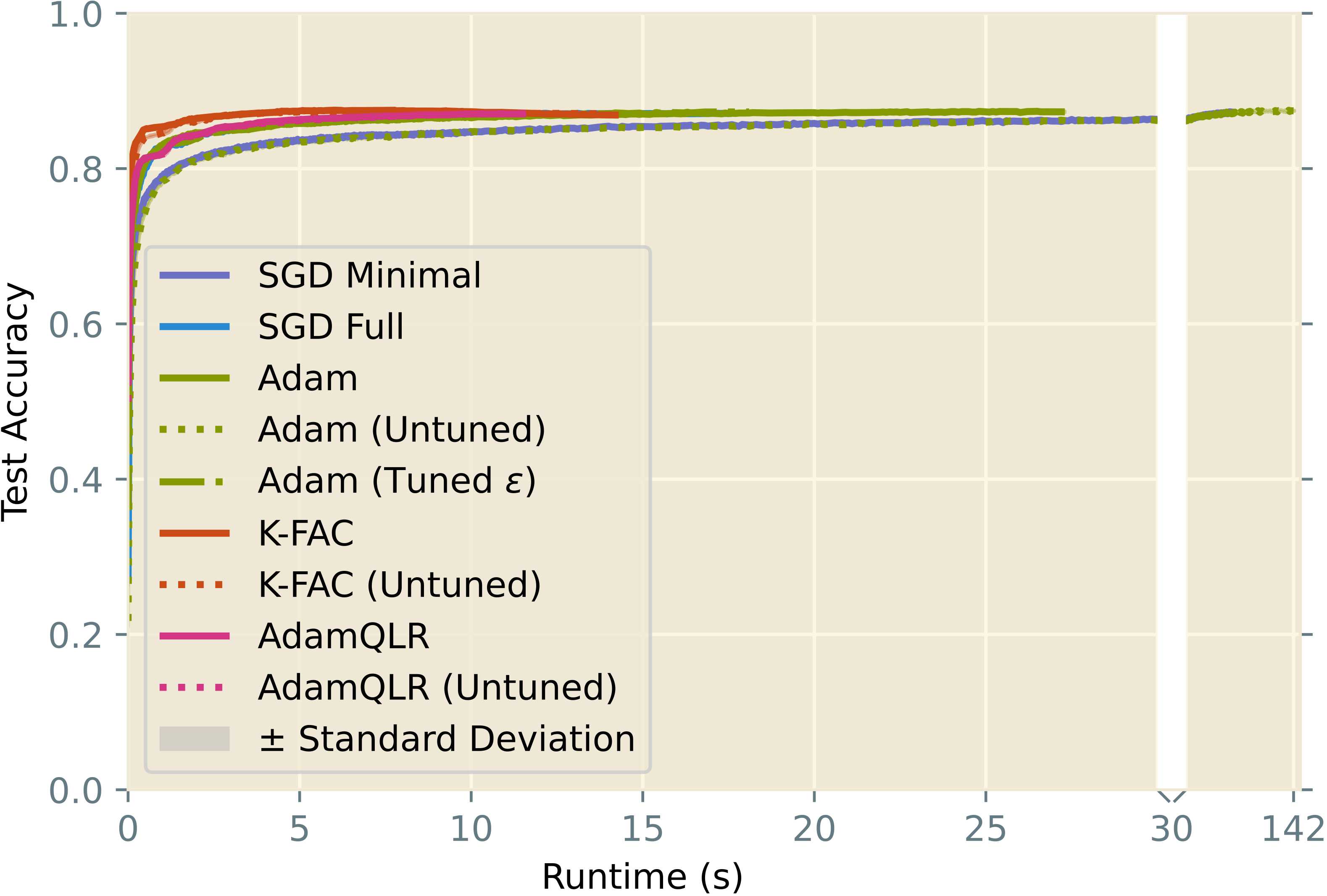} \vfill\\
        
        \rotatecaption{\subcaption{SVHN}}
        & \includegraphics[align=c,width=\linewidth]{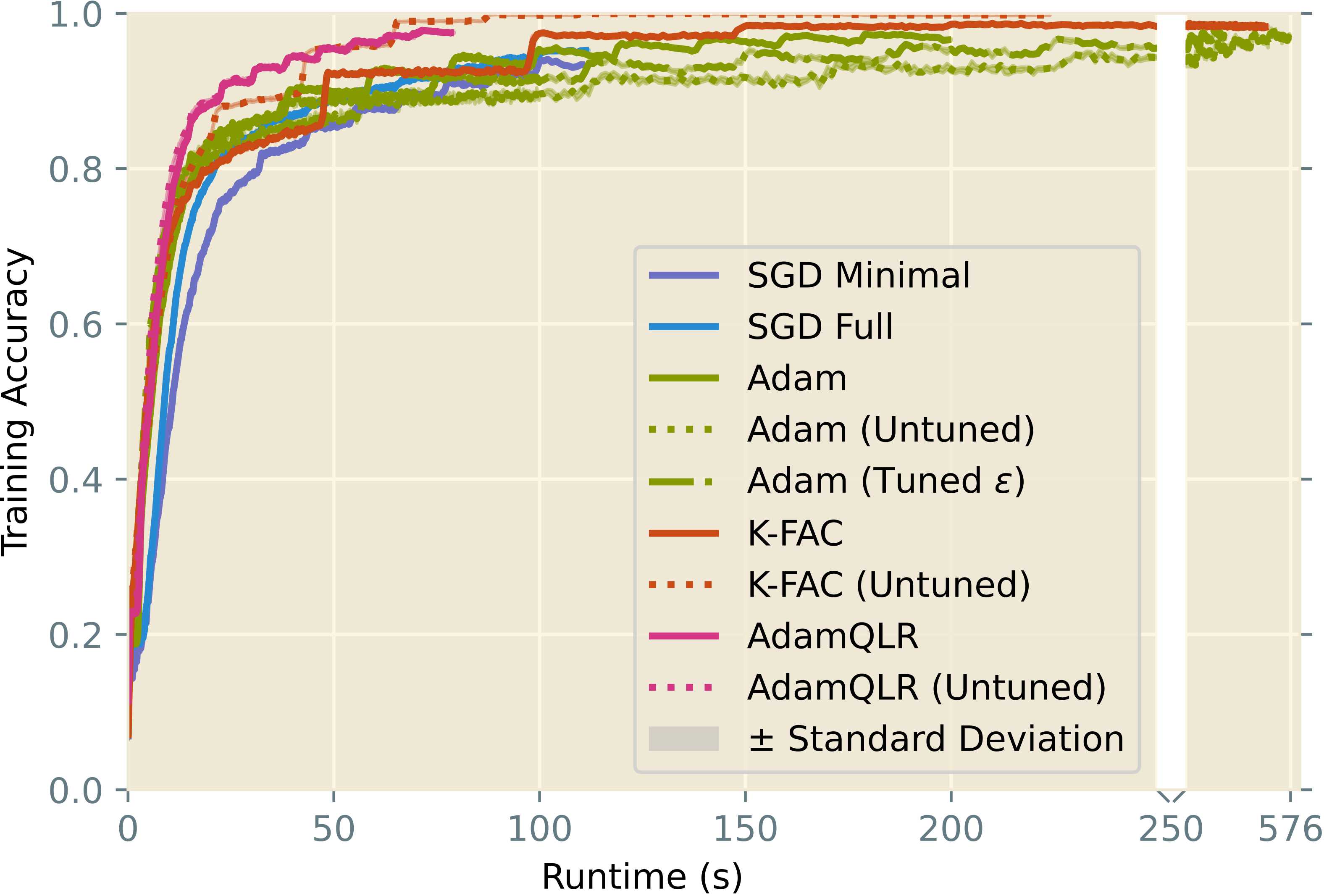}
        & \includegraphics[align=c,width=\linewidth]{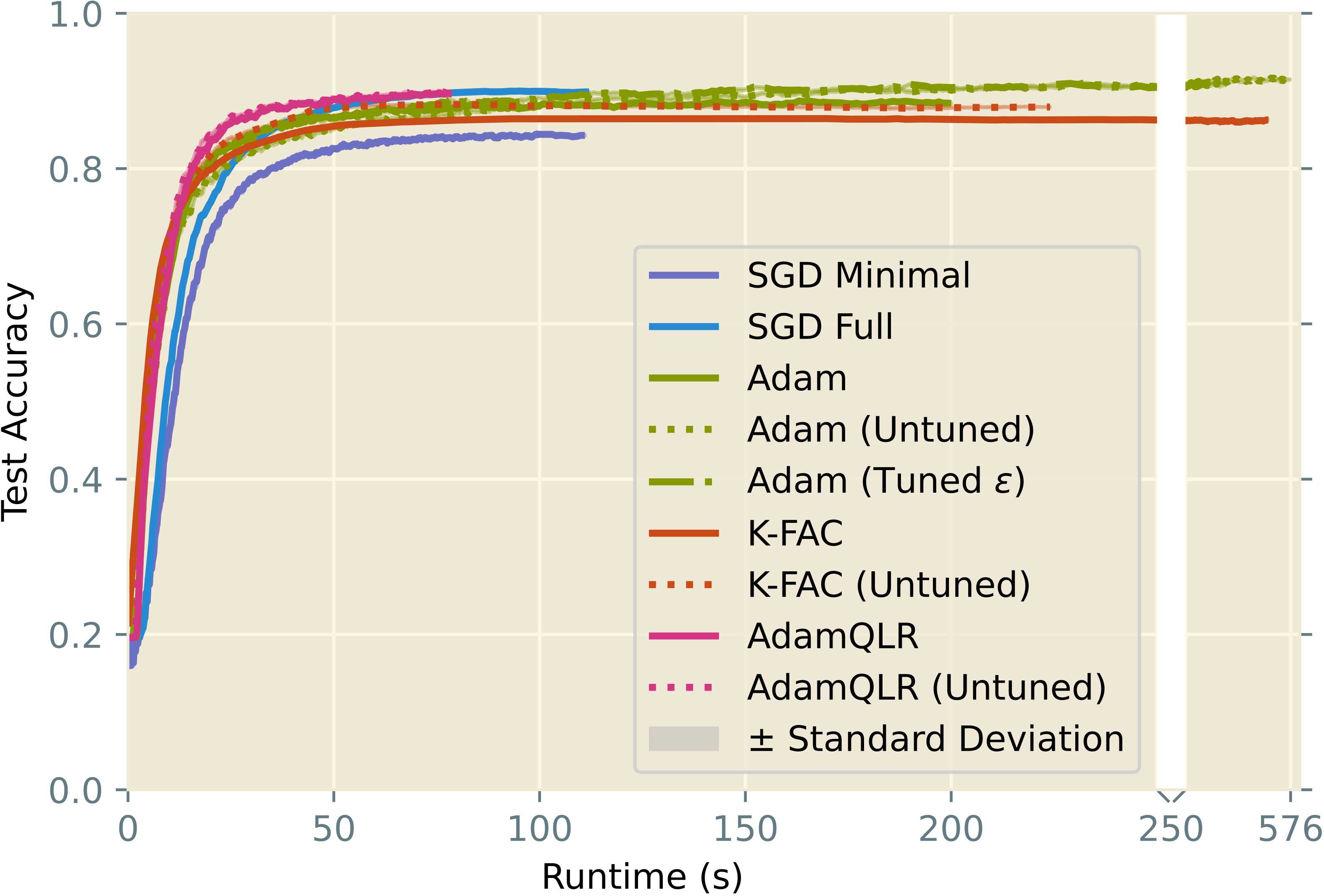} \vfill
    \end{tabularx}
    \caption{Reprise of Figure~\ref{fig:AlgorithmMixedResults}, with additional results on an \emph{Adam (Tuned $\epsilon$)} setting extending \emph{Adam} by additionally tuning the $\epsilon$ hyperparameter.}
    \label{fig:AdamTunedEpsilonResults}
\end{figure*}

While the Adam optimiser is often invoked without tuning the $\epsilon$ hyperparameter, its interpretation as a form of `damping' for Adam's `curvature estimate` invites the alternative \emph{Adam (Tuned $\epsilon$)} setting, in which our hyperparameter optimisation approach also selects $\epsilon$ in the logarithmic range $[10^{-8}, 10^{0}]$. Our results in Figure~\ref{fig:AdamTunedEpsilonResults} demonstrate that this implementation is generally interchangeable with the \emph{Adam} baseline of Section~\ref{sec:Experiments} in terms of performance.

\subsubsection{K-FAC Adaptive Heuristics}
\label{sec:KFACAblations}
\begin{figure*}[p]
    \centering
    \newcommand{\rotatecaption}[1]{%
        \rotatebox[origin=c]{90}{\begin{minipage}{3cm}#1\end{minipage}} }
    \begin{tabularx}{0.82\linewidth}{p{2ex}XX}
        \rotatecaption{\subcaption{UCI Energy}}
        & \includegraphics[align=c,width=\linewidth]{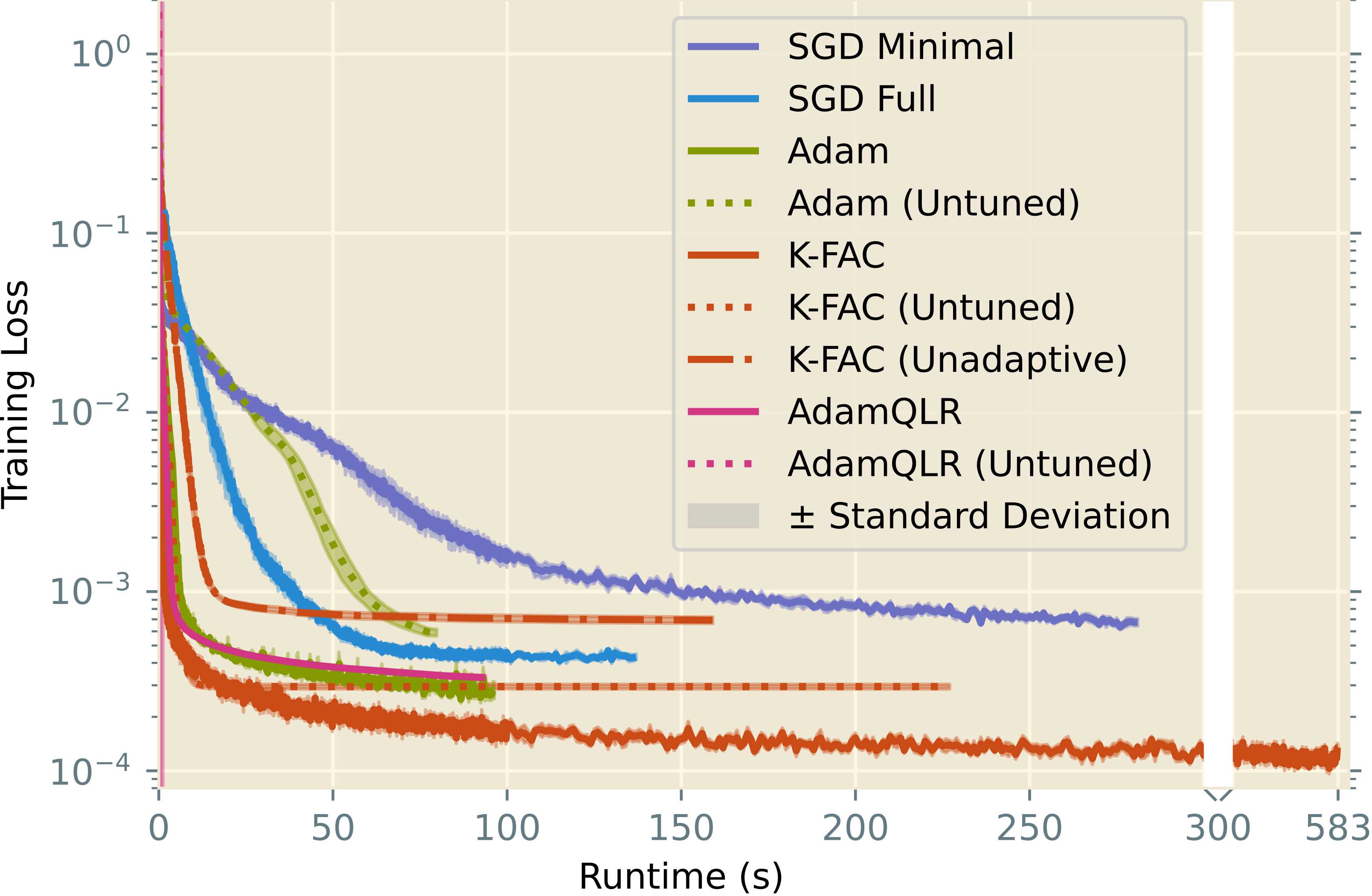}
        & \includegraphics[align=c,width=\linewidth]{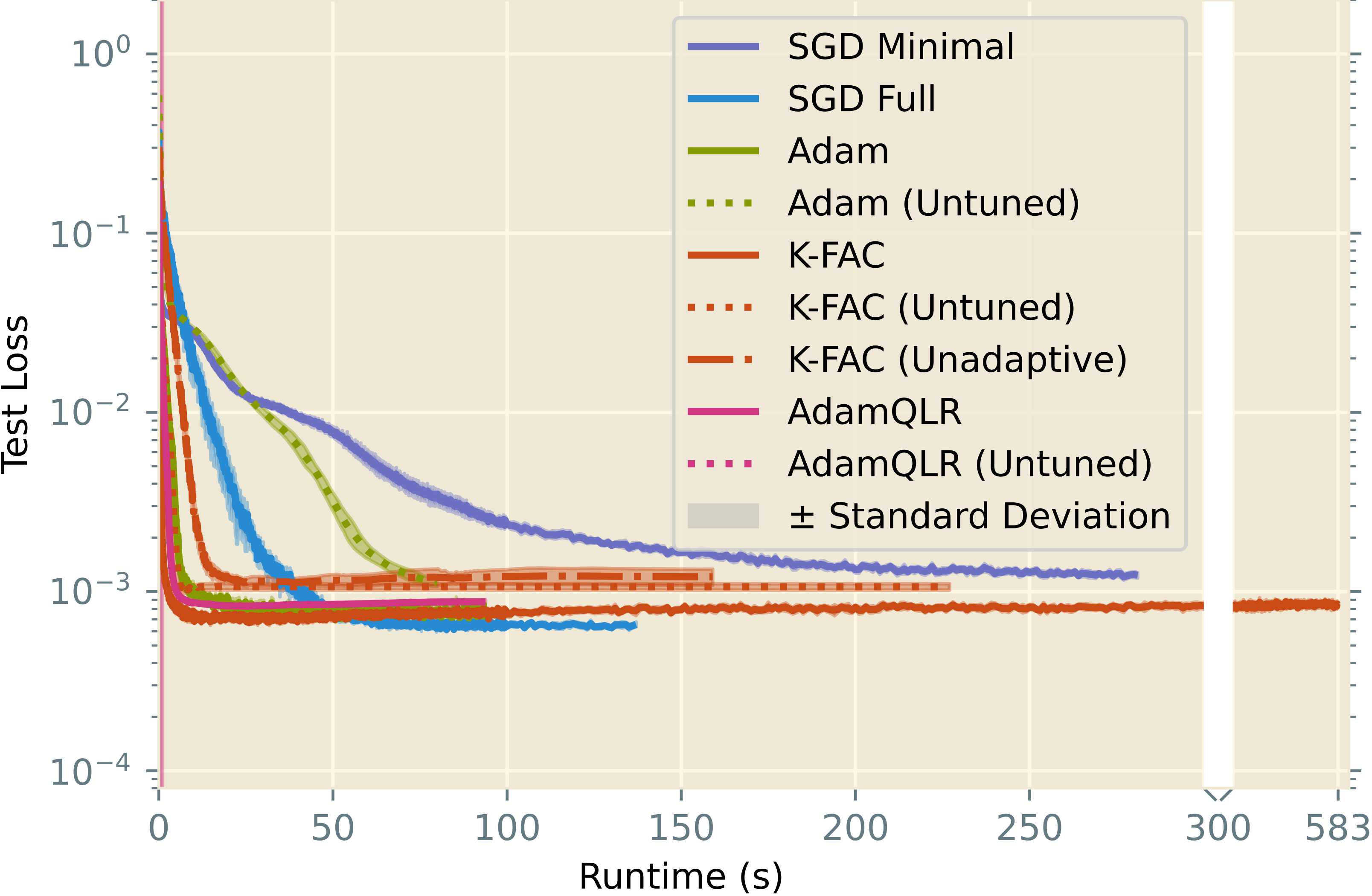} \vfill\\
        
        \rotatecaption{\subcaption{UCI Protein}}
        & \includegraphics[align=c,width=\linewidth]{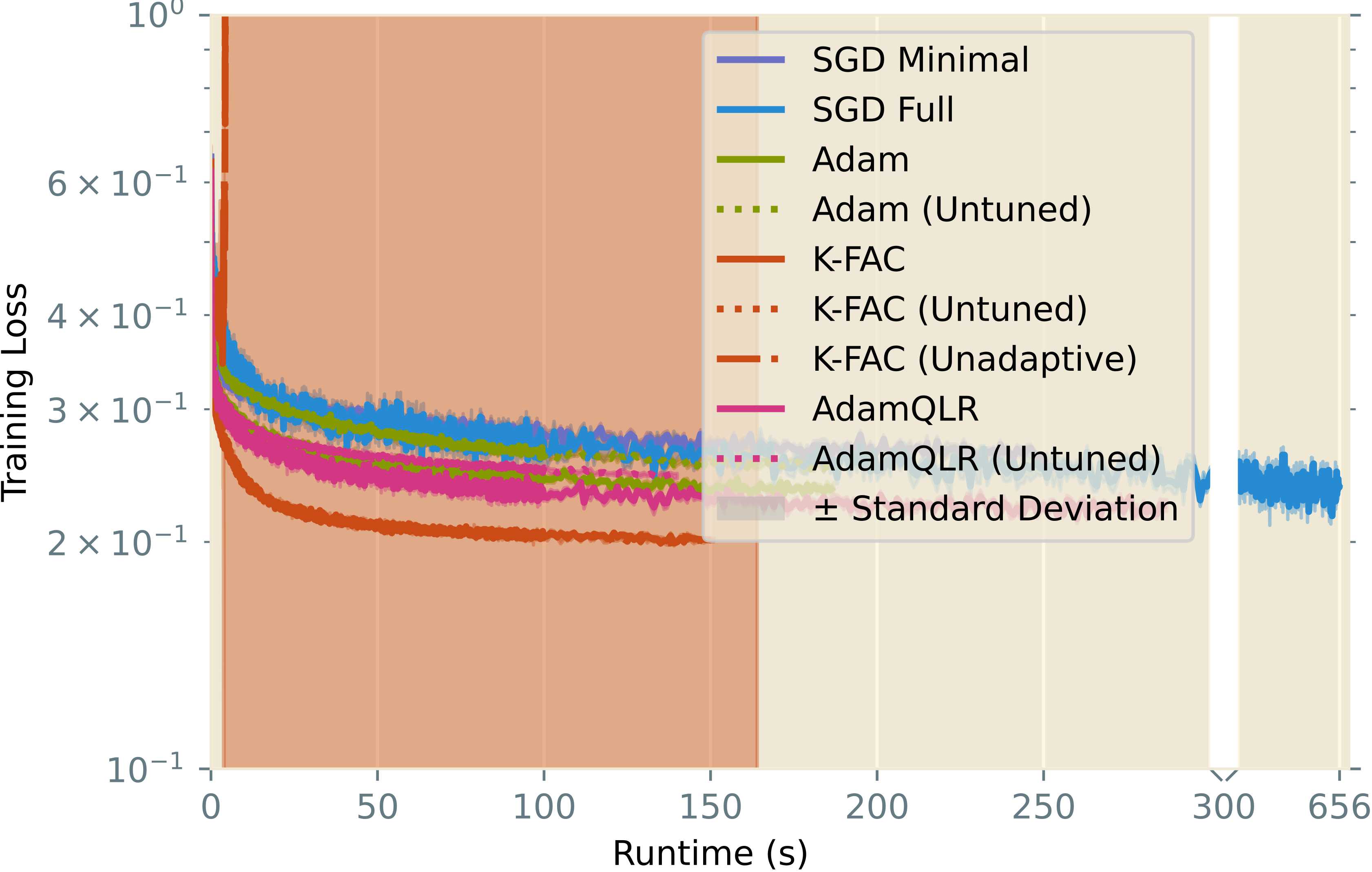}
        & \includegraphics[align=c,width=\linewidth]{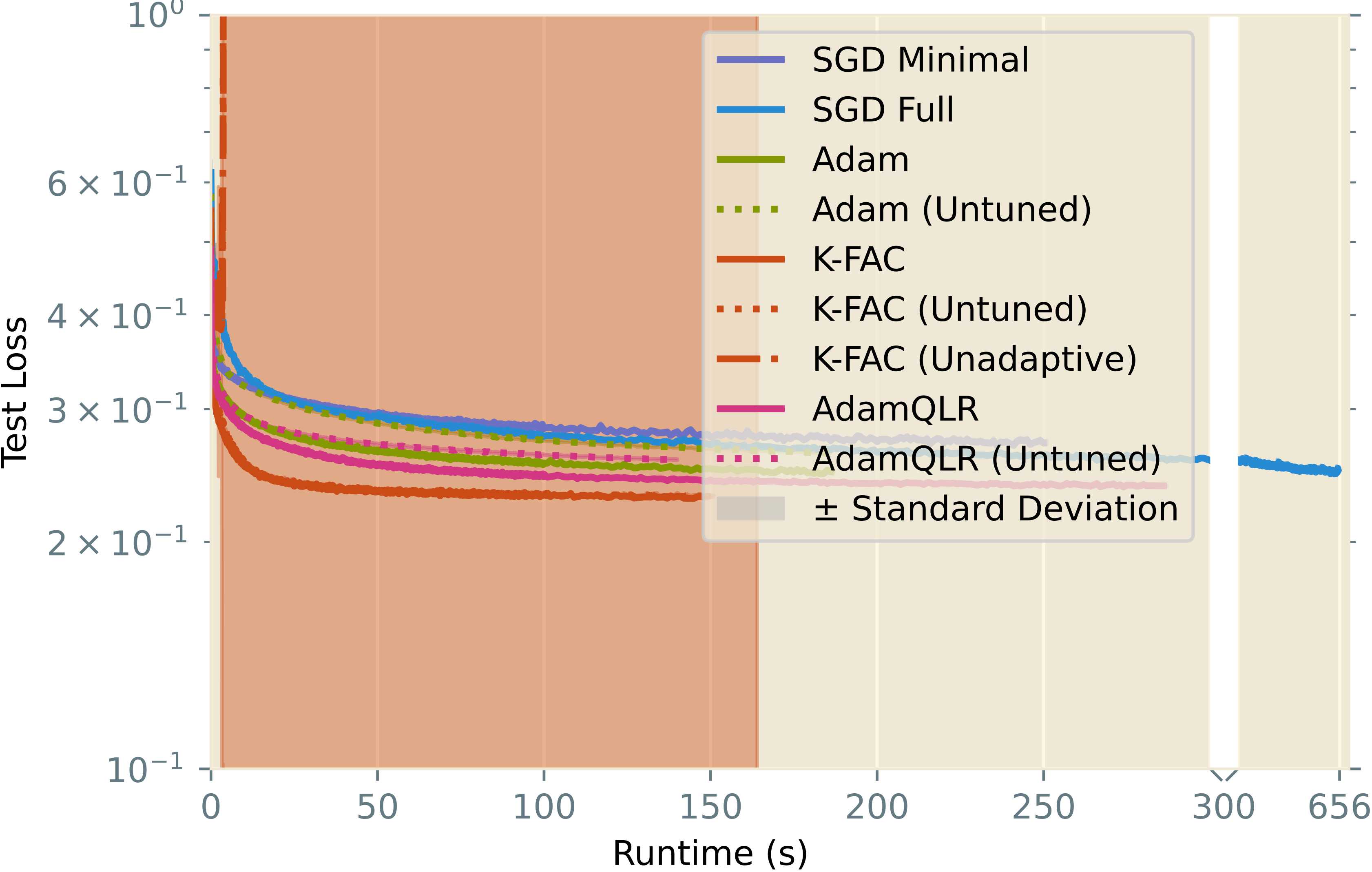} \vfill\\
        
        \rotatecaption{\subcaption{Fashion-MNIST}}
        & \includegraphics[align=c,width=\linewidth]{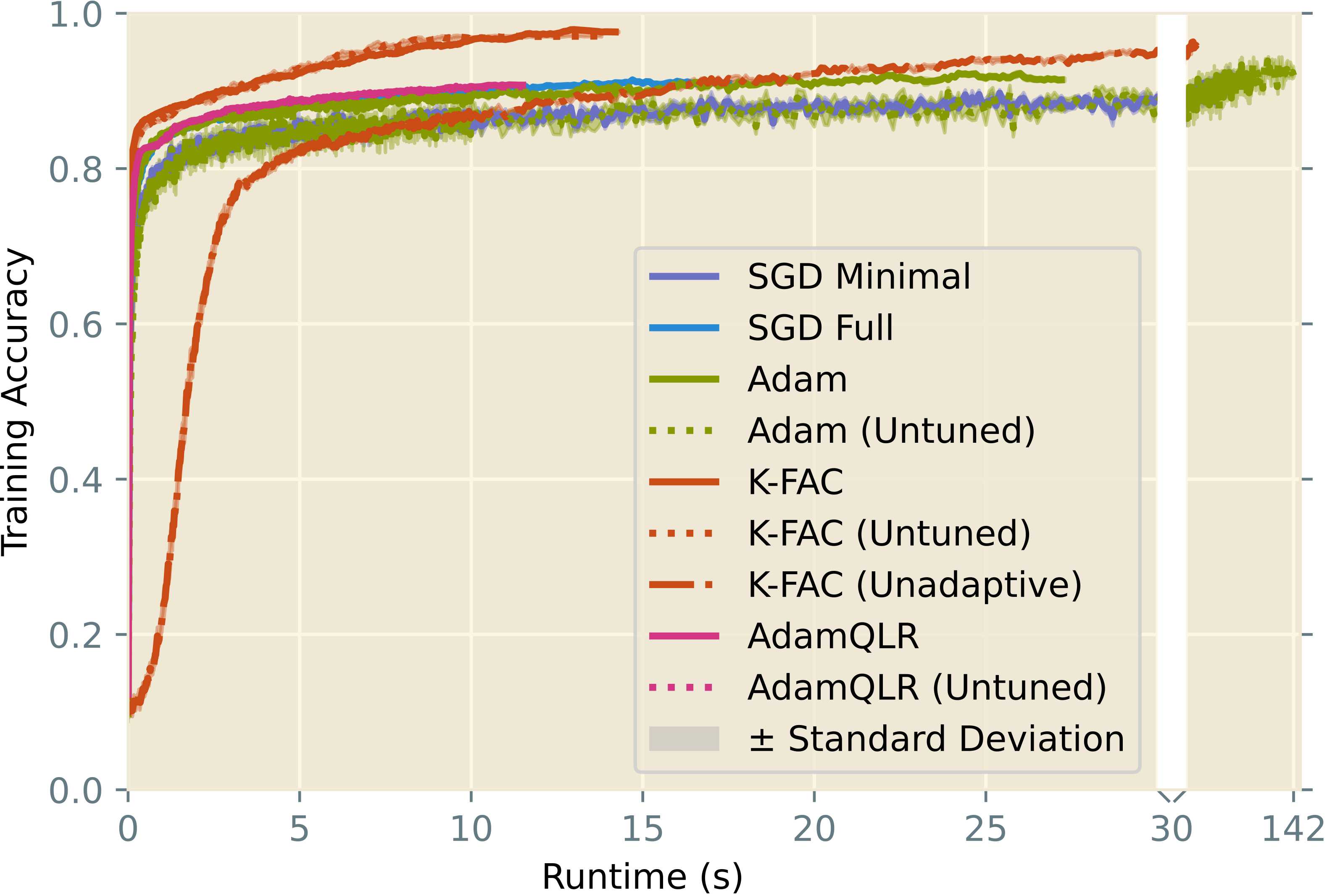}
        & \includegraphics[align=c,width=\linewidth]{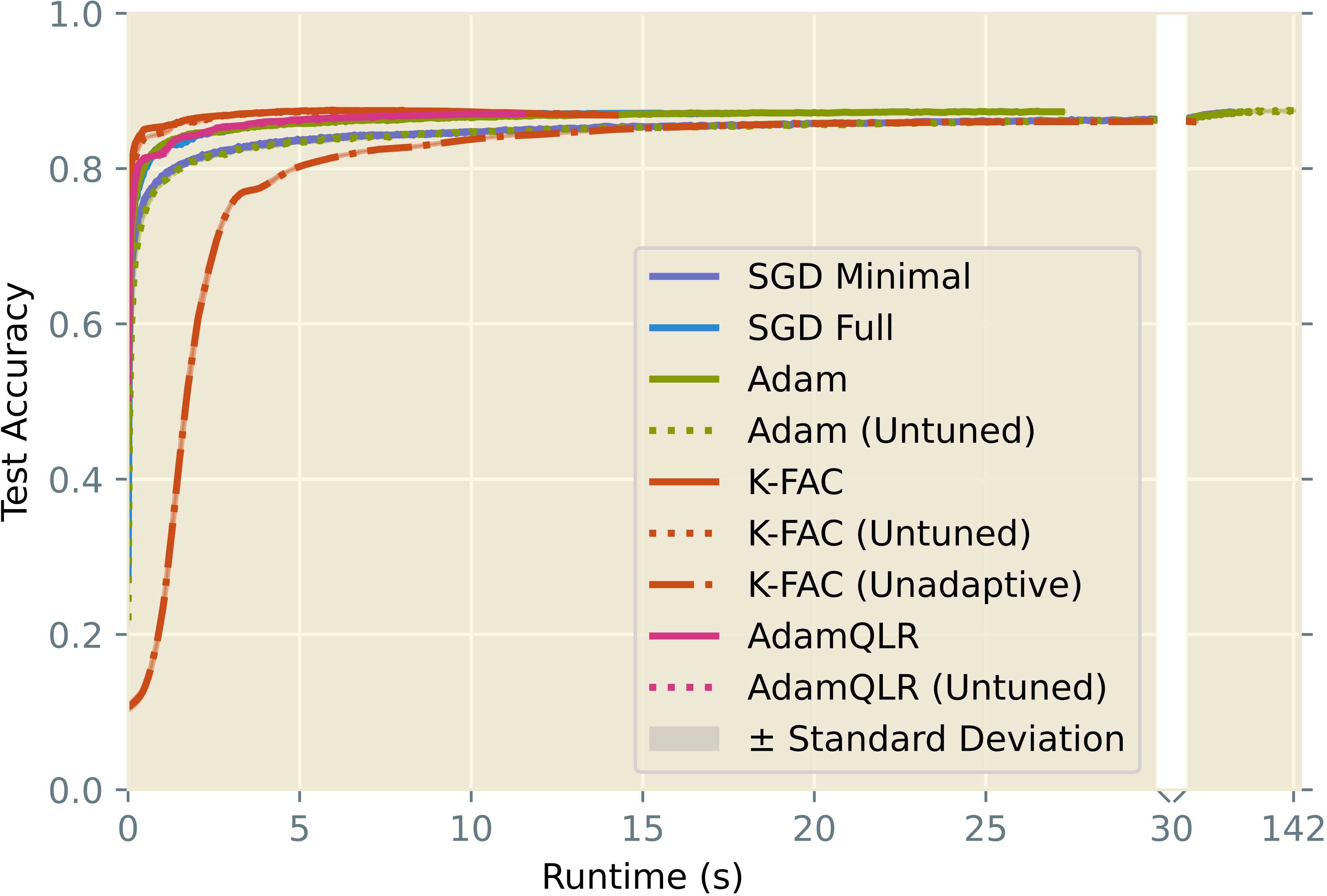} \vfill\\
        
        \rotatecaption{\subcaption{SVHN}}
        & \includegraphics[align=c,width=\linewidth]{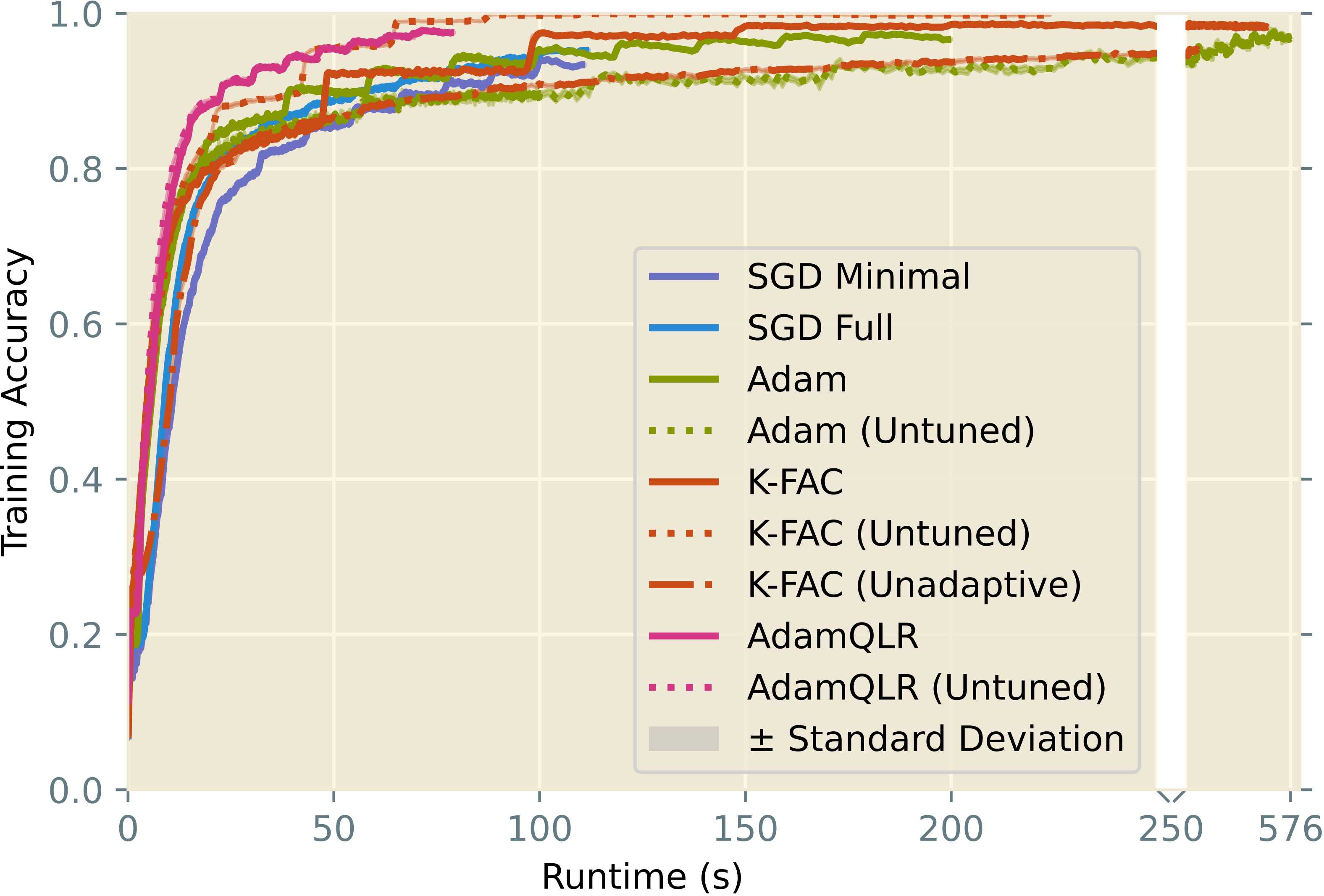}
        & \includegraphics[align=c,width=\linewidth]{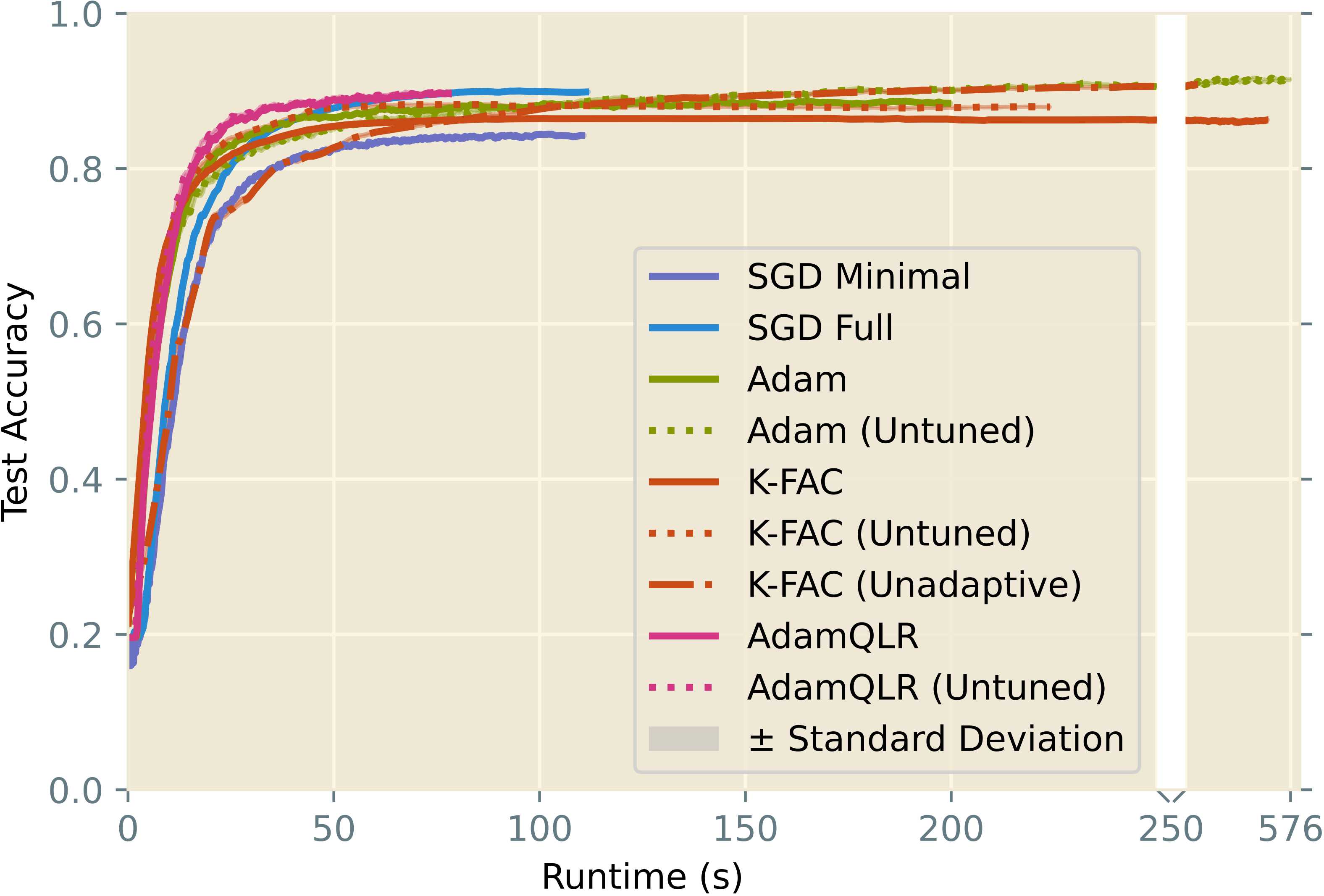} \vfill
    \end{tabularx}
    \caption{Reprise of Figure~\ref{fig:AlgorithmMixedResults}, with additional results on a \emph{K-FAC (Unadaptive)} setting replacing adaptive heuristics with fixed tuned values.}
    \label{fig:AblationKFACHeuristics}
\end{figure*}

In addition to extending Adam with heuristics from K-FAC, we briefly study the impact of \emph{removing} these heuristics from K-FAC. To this end, we introduce a \emph{K-FAC (Unadaptive)} baseline, which replaces the adaptive damping, learning rate and momentum of the \emph{K-FAC} setting with fixed tuned values. The results of this additional comparison are shown in Figure~\ref{fig:AblationKFACHeuristics}, which shows the \emph{Unadaptive} variant makes much slower progress than vanilla \emph{K-FAC}, and in some cases the lack of adaptivity completely destabilises K-FAC. This gives convincing evidence that the adaptive heuristics make a significant difference to \emph{K-FAC}, and thus reinforces the central hypothesis of this work.

We also note that correctly implementing K-FAC is an extremely subtle and complex task, with many libraries excluding some aspects of the algorithm presented by \citet{martens_optimizing_2015}. For this reason, we feel many results in the literature are not entirely fair comparisons between K-FAC and other methods, which may lead to contradictory conclusions about the effectiveness of K-FAC.

\subsubsection{Learning Rate Schedules}
\begin{figure*}[p]
    \centering
    \newcommand{\rotatecaption}[1]{%
        \rotatebox[origin=c]{90}{\begin{minipage}{3cm}#1\end{minipage}} }
    \begin{tabularx}{0.82\linewidth}{p{2ex}XX}
        \rotatecaption{\subcaption{UCI Energy}}
        & \includegraphics[align=c,width=\linewidth]{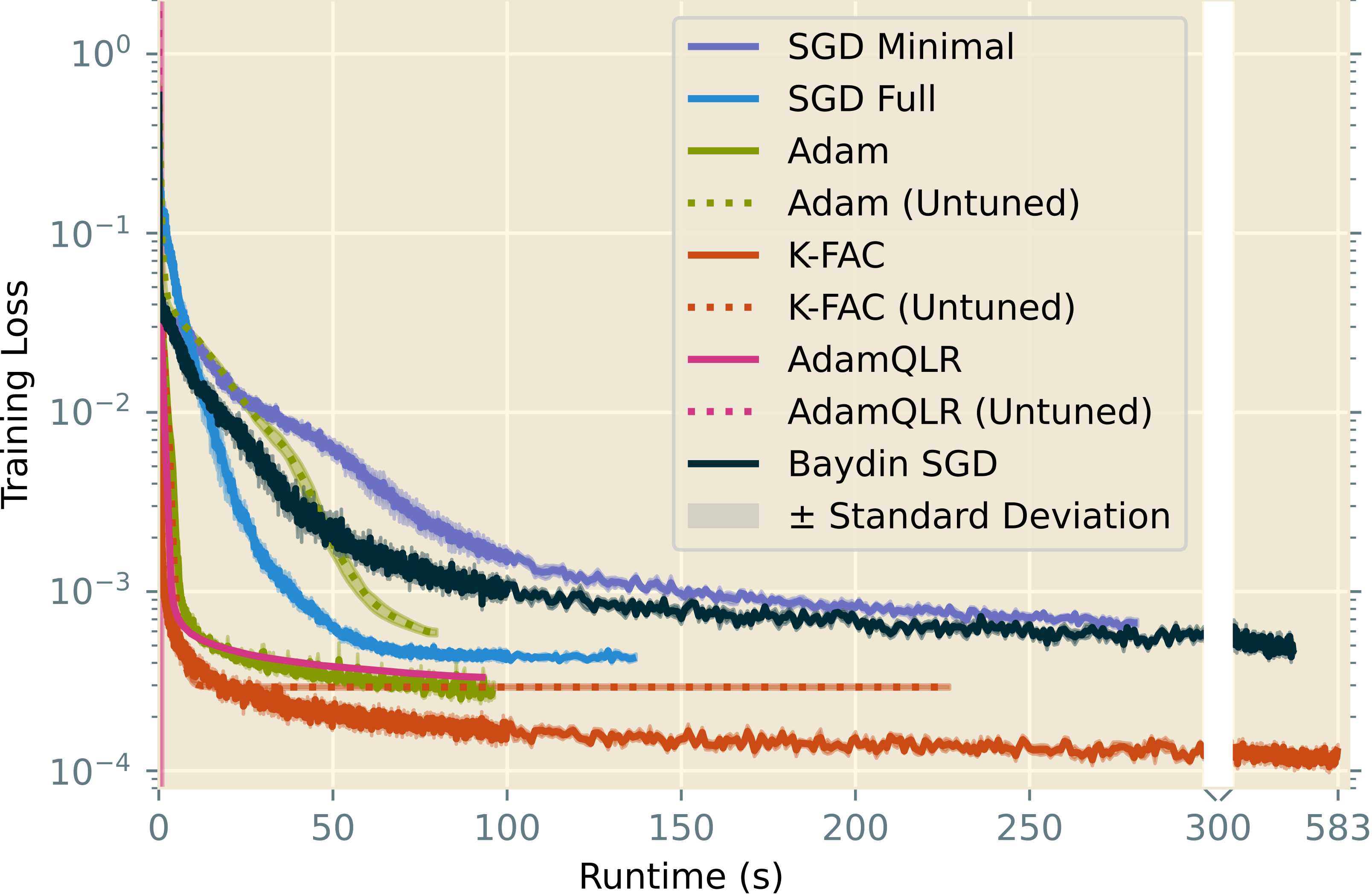}
        & \includegraphics[align=c,width=\linewidth]{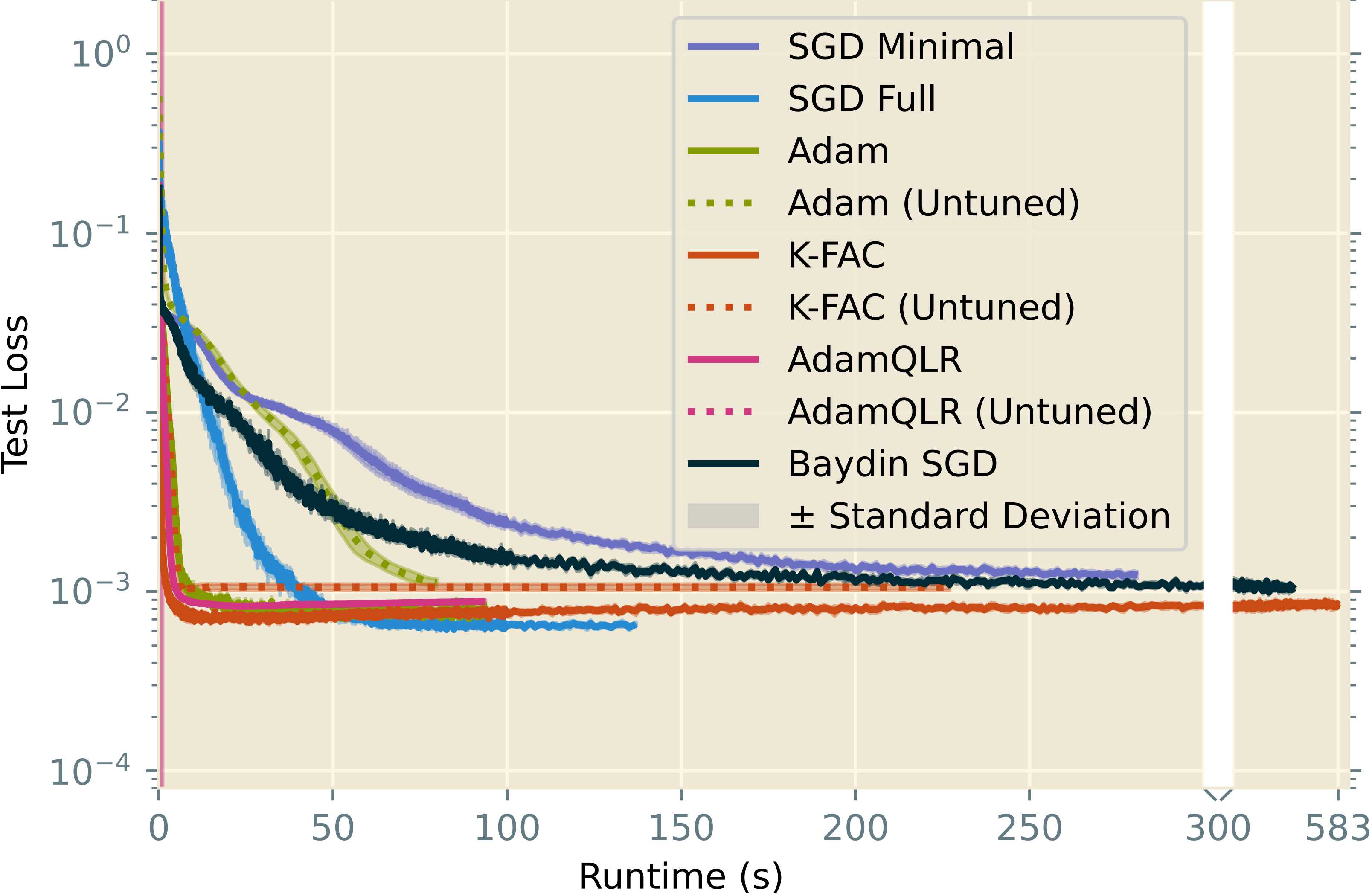} \vfill\\
        
        \rotatecaption{\subcaption{UCI Protein}}
        & \includegraphics[align=c,width=\linewidth]{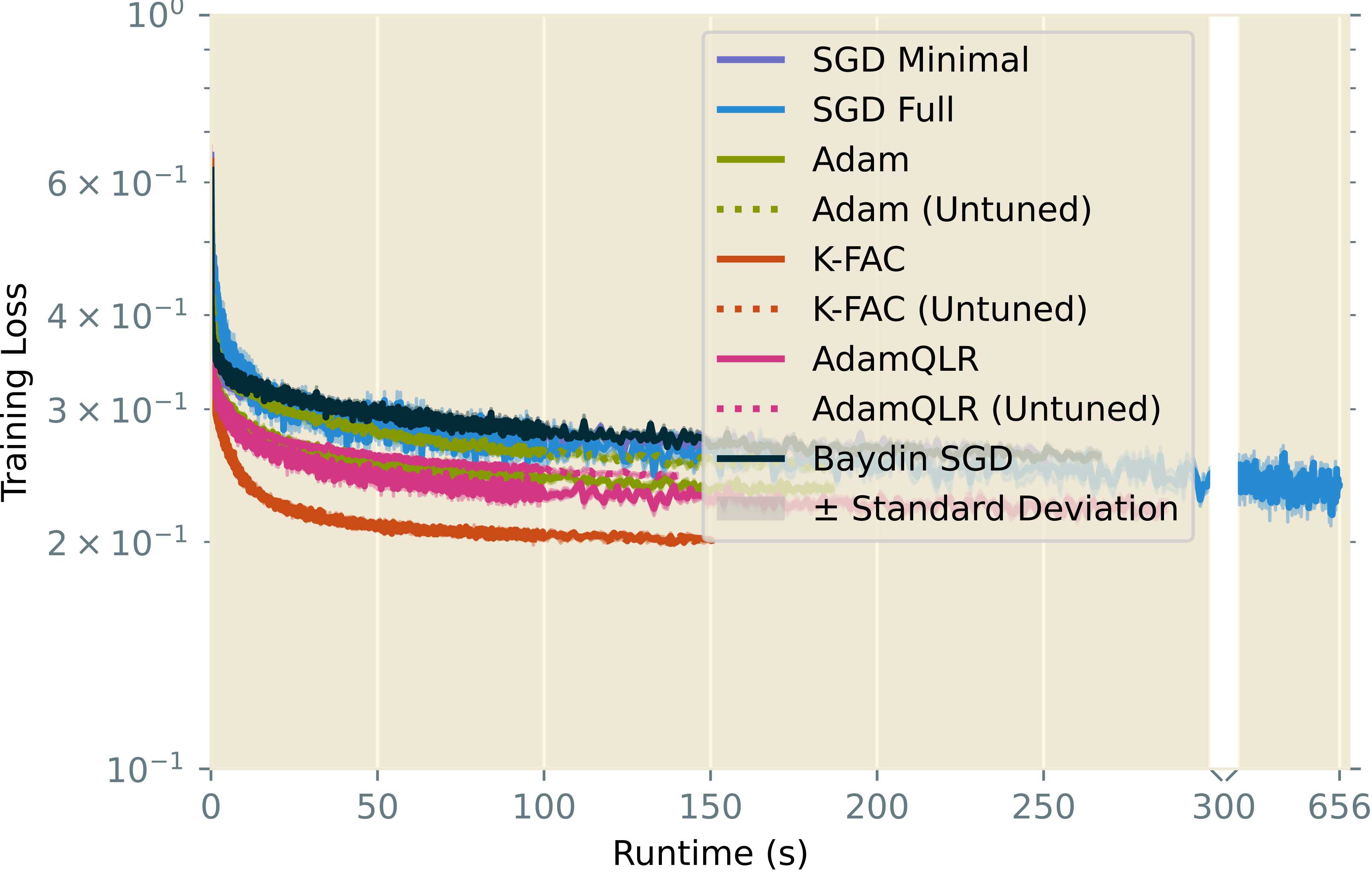}
        & \includegraphics[align=c,width=\linewidth]{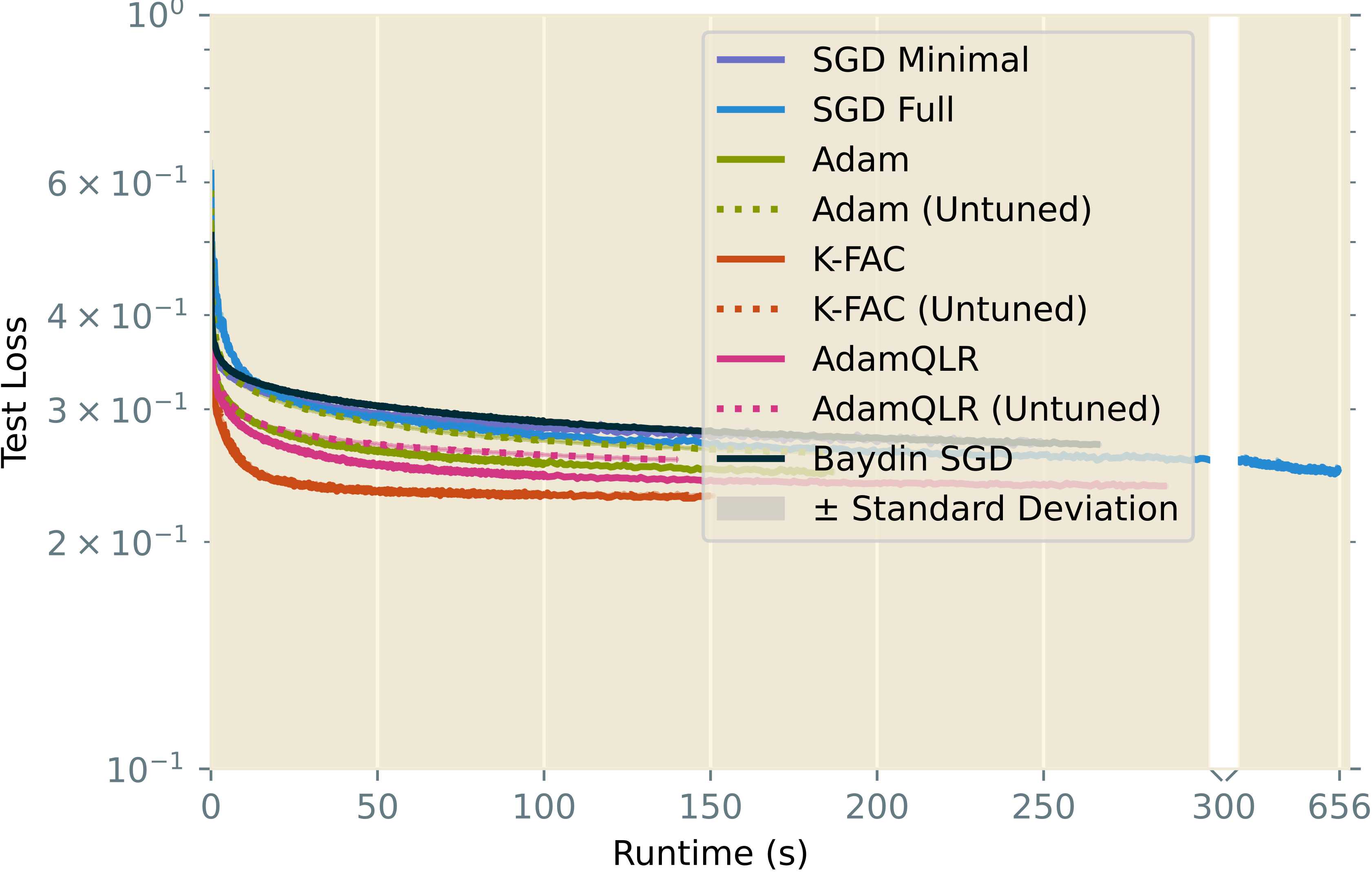} \vfill\\
        
        \rotatecaption{\subcaption{Fashion-MNIST}}
        & \includegraphics[align=c,width=\linewidth]{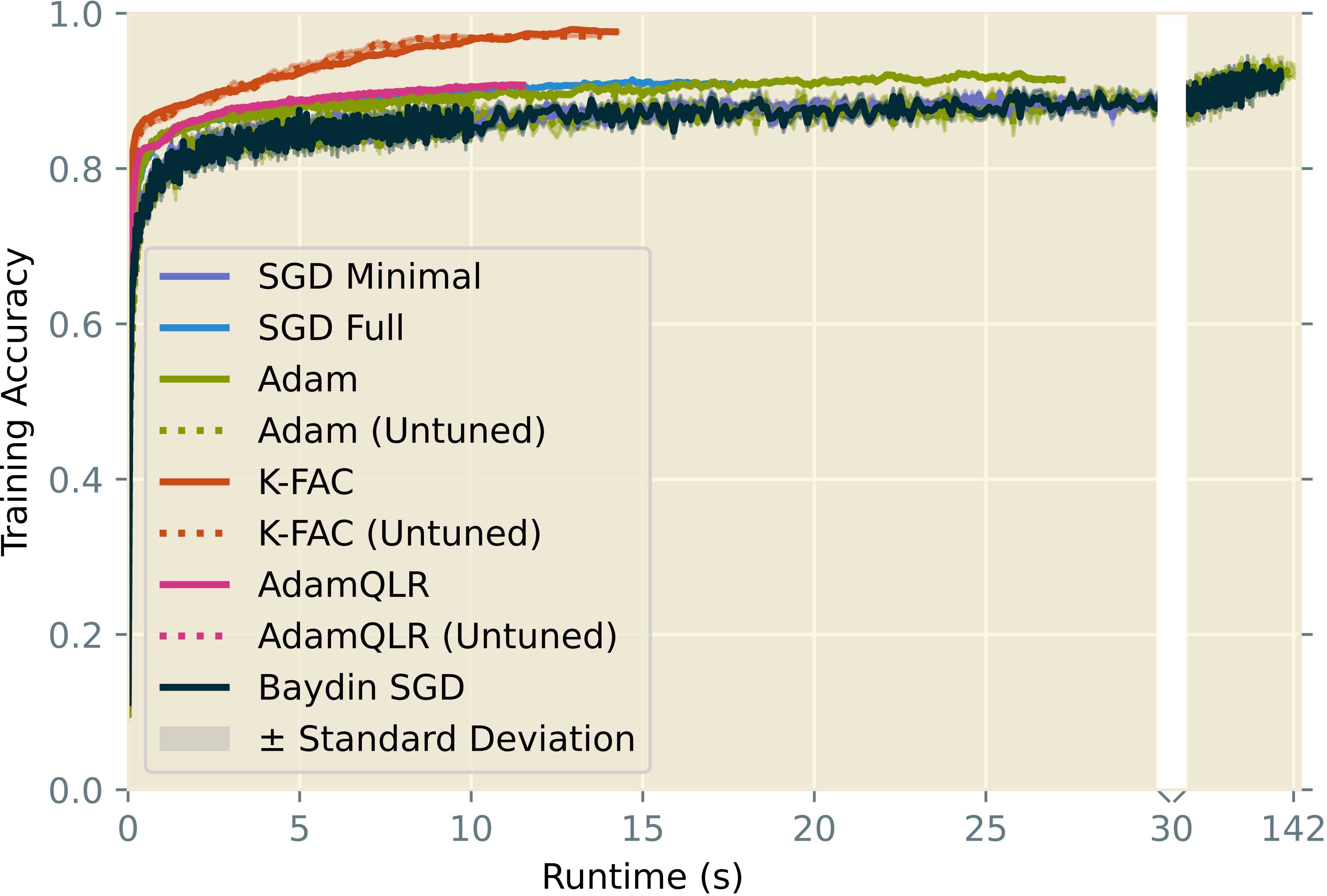}
        & \includegraphics[align=c,width=\linewidth]{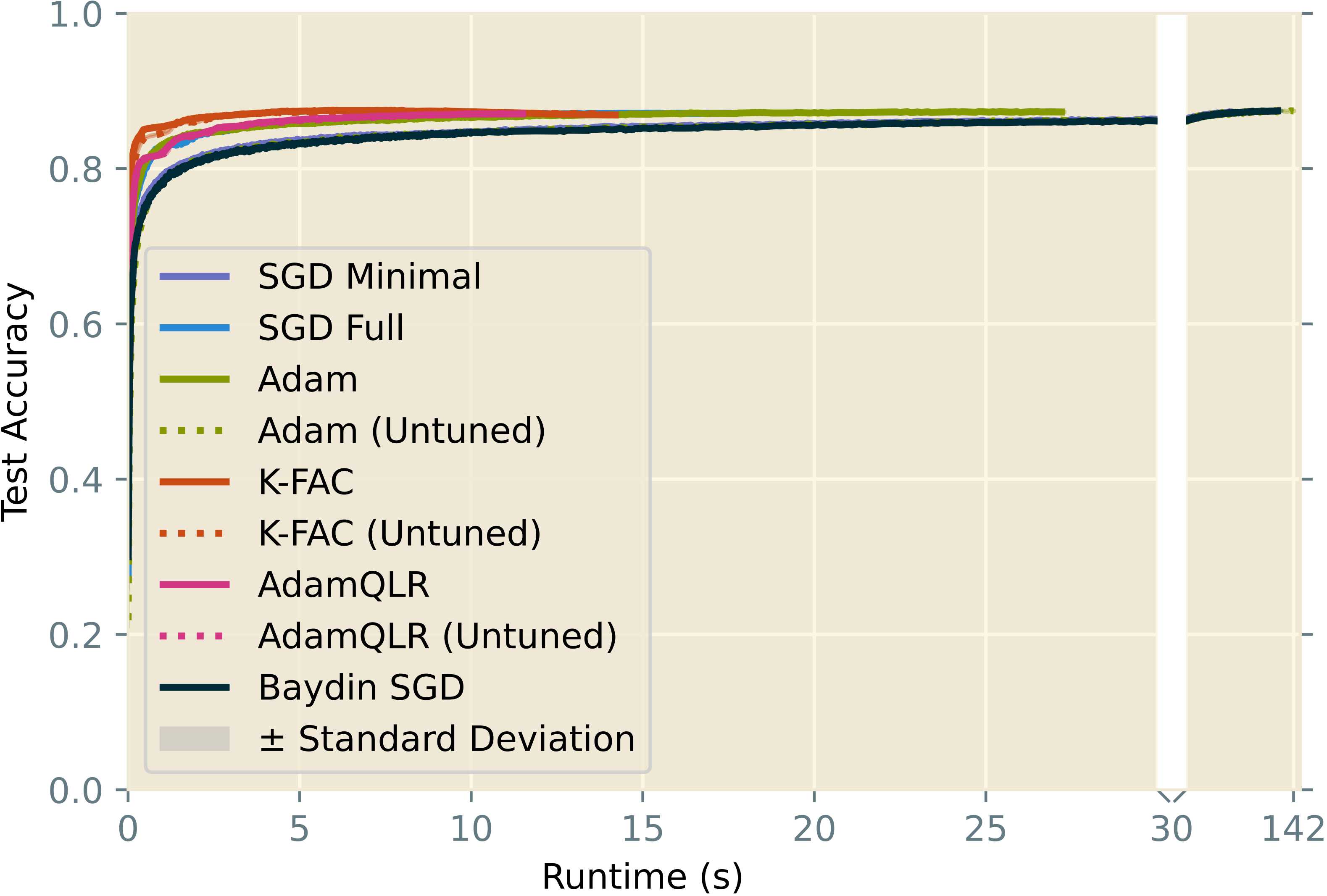} \vfill\\
        
        \rotatecaption{\subcaption{SVHN}}
        & \includegraphics[align=c,width=\linewidth]{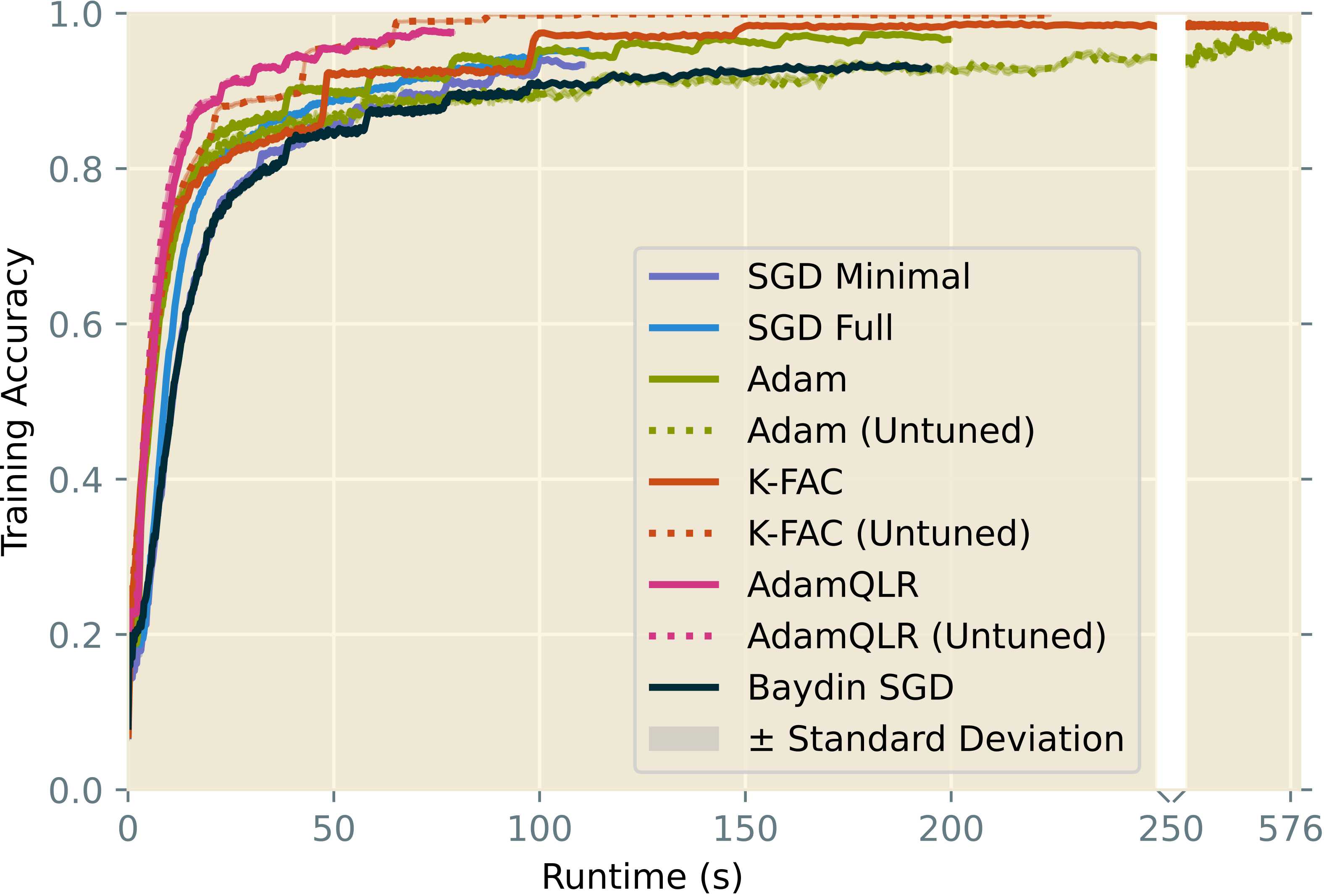}
        & \includegraphics[align=c,width=\linewidth]{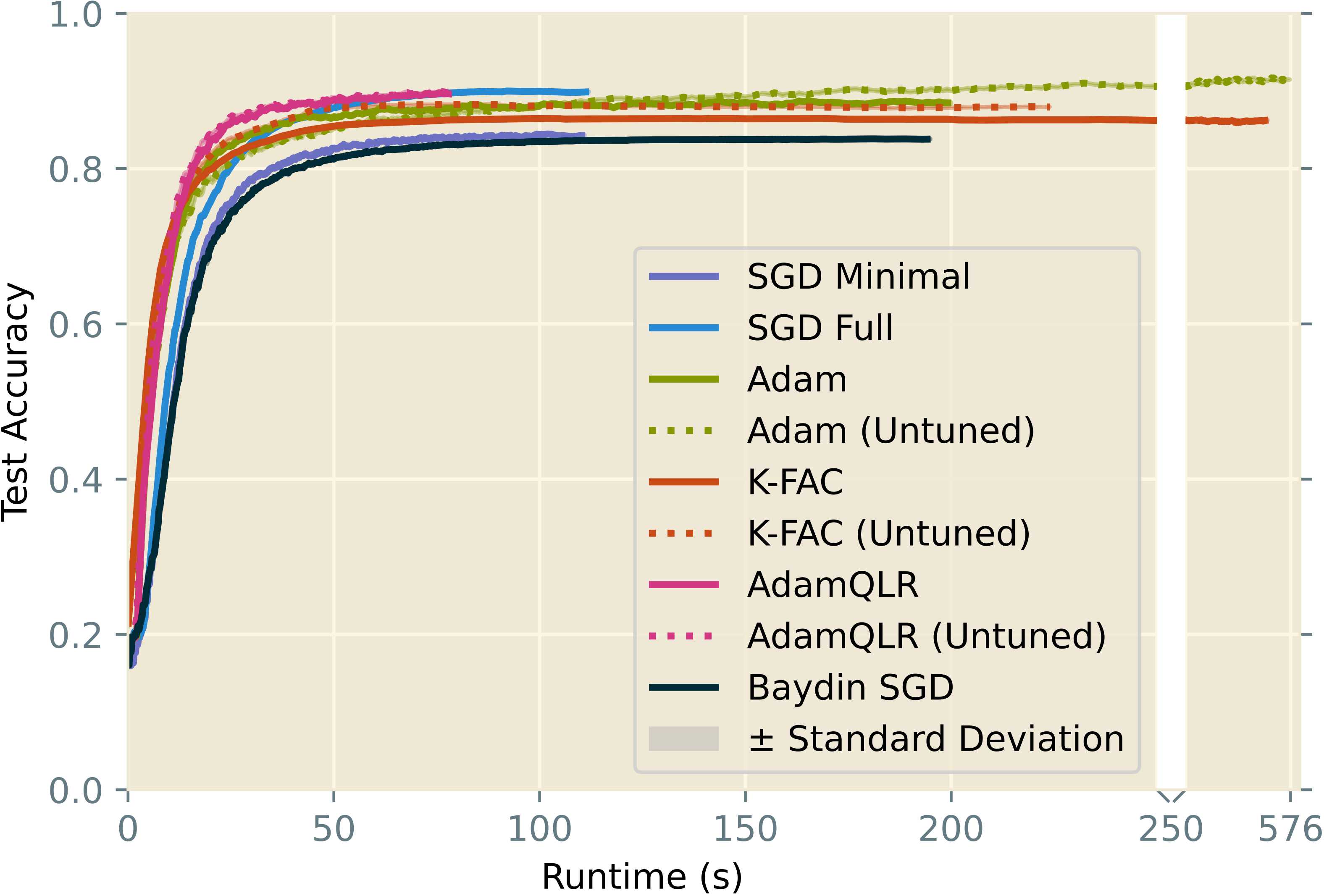} \vfill
    \end{tabularx}
    \caption{Reprise of Figure~\ref{fig:AlgorithmMixedResults}, with additional results on a \emph{Baydin SGD} setting examining the effect of a simple adaptive learning rate strategy.}
    \label{fig:AblationLearningRateSchedules}
\end{figure*}

Many ML training approaches use learning rates which follow a predefined schedule in order to improve performance. We have chosen not to focus on learning rate scheduling in this work, partly due to the myriad schedules available, and partly because this can easily be incorporated into any method we consider, including AdamQLR. In support of this position, we introduce a new \emph{Baydin SGD} baseline, which combines the adaptive learning rate method of \citet{baydin_online_2018} with our \emph{SGD Minimal} setting, and compare to this method in Figure~\ref{fig:AblationLearningRateSchedules}.

These results show the adaptive learning rate --- even when its initialisation and hyper-learning rate are tuned --- does not perform significantly differently to the tuned fixed learning rate of \emph{SGD Minimal}. This concurs with the general belief that well-optimised learning rates tend to outperform adaptive learning rate methods in practice.

\section{Curvature Matrices: Hessian and Fisher}
\label{sec:CurvatureMatrices}
In this section we discuss in more detail the two main candidates for the curvature matrix $\vec{C}$ in our algorithm. Recall from Section~\ref{sec:AdamQLR} that throughout we consider an arbitrary function $f(\vec{\theta})$ representing the loss function of some network parameterised by $\vec{\theta}$.

\subsection{Hessian Matrix}
In this setting, the Hessian curvature matrix follows naturally from the definition of the objective function. A first derivative with respect to $\vec{\theta}$ yields the gradient vector $\vec{g} = (\nabla_{\vec{\theta}} f)(\vec{\theta})$, and repeating the derivative yields the Hessian $\vec{H} = (\nabla_\vec{\theta} (\nabla_\vec{\theta} f)\trans) (\vec{\theta})$.

\subsection{Fisher Information Matrix}
To draw a connection with the Fisher matrix, we must restate our problem in a probabilistic form. We shall separate the loss function from the neural network, naming the latter $\vec{w}_\vec{\theta} (\cdot)$, and consider input-output data pairs $(\vec{x}, \vec{y})$. Let the input data have some ground truth distribution $p(\vec{x})$, and suppose we choose to interpret the output of the network as a probabilistic relationship, such that $\vec{w}_\vec{\theta}(\vec{x}) = \log p(\vec{y}|\vec{x})$.

For this model $\vec{w}$, the \emph{Fisher Information Matrix} (FIM, or ``the Fisher'') is defined as:
\begin{equation}
    \label{eq:Fisher}
    \vec{F} = \mathbb{E}_{\vec{x} \sim p(\vec{x})} \mathbb{E}_{\vec{y} \sim p(\vec{y} | \vec{x})} \left[ \pderiv{\log p(\vec{y} | \vec{x}) }{\vec{\theta}} \pderiv{\log p(\vec{y} | \vec{x}) }{\vec{\theta}} \trans \right].
\end{equation}
In its exact form, the Fisher bears many favourable properties for use in optimisation: it is positive semi-definite by construction (so represents a convex space), it is amenable to efficient computation in the form of a matrix=vector product, and provides a parameterisation-independent view of the problem (as in the Natural Gradient Descent \citep{amari_natural_1998} family of methods).

Since $\pderiv{\log p(\vec{y} | \vec{x}) }{\vec{\theta}}$ is the Jacobian of the network output $\vec{w}_{\vec{\theta}}$ with respect to the parameters $\vec{\theta}$, the outer product of derivatives is readily available as part of our standard training regime. Although $p(\vec{x})$ is unknown, in the mini-batched training setting it is commonly approximated by the empirical distribution $\widehat{p}(\vec{x})$ implied by our training dataset. It is important to stress that the expectation of $\vec{y}$ is taken with respect to the output distribution of the network, \emph{not} with respect to any ground-truth or empirical distribution $\widehat{p}(\vec{y} | \vec{x})$ given by the training data. However, some previous work uses the latter distribution as an approximation, resulting in the \emph{empirical} Fisher matrix, which is known to be inferior to the true Fisher.

\subsection{Gauss-Newton Matrix}
We briefly note the Gauss-Newton matrix, which is also commonly used in second-order optimisation algorithms. We chose not to focus on this matrix because it is motivated as a lossy approximation to the Hessian, which seemed undesirable when we could access the exact Hessian and Fisher matrices through Jacobian-vector products. While it has been used in a derivation of K-FAC, we note also the equivalence of the Fisher and Generalised Gauss-Newton matrices when applying negative log-likelihood loss to the natural parameters of an exponential family \citep{botev_practical_2017}, and to our knowledge there is only a marginal performance difference between Fisher- and Gauss-Newton-based K-FAC.

\subsection{Adam and Fisher Matrix}
\label{sec:AdamAndFisherMatrix}

While Adam is described by its authors as representing an approximation to the Fisher matrix \citep{kingma_adam_2015}, we seek here to make the connection more explicit.

The matrix computed inside the expectation of Equation~\ref{eq:Fisher} has as its diagonal the elementwise square of $\pderiv{\log p(\vec{y} | \vec{x}) }{\vec{\theta}}$. This is connected to the quantity $\vec{g}_t = \nabla_\vec\theta f(\vec{\theta}_{t-1})$ computed by Adam; by the chain rule, $\vec{g}_t$ is precisely the product of $\pderiv{\log p(\vec{y} | \vec{x}) }{\vec{\theta}}$ and the derivative of the loss function with respect to the model output. Neglecting the effect of the latter allows us to view Adam's second-moment buffer $\widehat{\vec{v}}_t$ as an approximation to the diagonal of the outer product in Equation~\ref{eq:Fisher}.

Further, because $\vec{g}_t$ is averaged over a mini-batch of input data, we are automatically taking approximate expectations over $\widehat{p}(\vec{x})$ and $\widehat{p}(\vec{y} | \vec{x})$. The approximation arises because the underlying Fisher matrix is not constant, so the contributions from each mini-batch relate to different underlying curvatures. However, the argument motivates the idea that Adam develops an approximation to the diagonal of the empirical Fisher matrix in its buffer $\widehat{\vec{v}}_t$.

From this perspective, Adam's elementwise division by the reciprocal of $\widehat{\vec{v}}_t$ is simply multiplication by the inverse (approximate) empirical Fisher, and we may interpret $\epsilon$ as a fixed damping term. This picture is slightly corrupted by the square root of $\widehat{\vec{v}}_t$ being the quantity actually used by Adam; this operation brings the eigenvalues of the approximate empirical Fisher closer to one, in particular increasing problematic near-zero eigenvalues to more stable values, thus justifying \citeauthor{kingma_adam_2015}'s statement that the square root permits more ``conservative'' preconditioning.

\end{document}